\documentclass[final,1p,twocolumn]{elsarticle}

\usepackage[cmex10]{amsmath}
\usepackage{amsmath}
\usepackage{amsxtra}
\usepackage{amstext}
\usepackage{amssymb}
\usepackage{latexsym}
\usepackage{dsfont}
\usepackage{graphicx}
\usepackage{caption}
\usepackage{subfig}
\usepackage{epsfig}
\usepackage{epstopdf}
\usepackage{fontenc}
\graphicspath{{./Images/}}

\journal{Pattern Recognition}

\begin{document}

\begin{frontmatter}



\title{Perceptually Motivated Shape Context Which Uses Shape Interiors}


\author{Vittal Premachandran\corref{cor1}}
\ead{vittalp@pmail.ntu.edu.sg}
\author{Ramakrishna Kakarala}
\ead{ramakrishna@ntu.edu.sg}
\address{50, Nanyang Avenue, School of Computer Engineering, Nanyang Technological University, Singapore-639798}
\cortext[cor1]{Corresponding author. Ph: +65 67906124, Fax: +65 67900375}
%

\begin{abstract}
In this paper, we identify some of the limitations of current-day shape matching techniques. We provide examples of how contour-based shape matching techniques cannot provide a good match for certain visually similar shapes. To overcome this limitation, we propose a perceptually motivated variant of the well-known shape context descriptor. We identify that the interior properties of the shape play an important role in object recognition and develop a descriptor that captures these interior properties. We show that our method can easily be augmented with any other shape matching algorithm. We also show from our experiments that the use of our descriptor can significantly improve the retrieval rates.
\end{abstract}


\end{frontmatter}


\section{Introduction}
\label{}

Accurately measuring the similarity between two objects is a fundamental problem in many computer vision applications, and is still a largely unsolved problem. Many applications such as shape-matching, shape-retrieval, and shape-based object detection, rely on a strong and robust similarity measure. However, coming up with such a similarity measure has proven to be a difficult task since the definition of similarity itself is rather subjective. Given two cars, one might say that their similarity should be measured based on their colour, while others might argue that the make, and model, are better metrics for measuring the similarity. Given two shapes, one might justify their similarity based on the number of parts in the shape, while another might feel that the symmetry of the objects is an important criterion.

Most pattern recognition problems are required to overcome this apparent vagueness in the definition of similarity and come up with a quantitative similarity (or dissimilarity) measure between objects. Restricting ourselves to the identification of dissimilarity between shapes, given two shapes, $S_1$ and $S_2$, dissimilarity measures try to identify the cost of transforming the shape $S_1$ into the shape $S_2$. The more similar the shapes are to each other, the easier it is to transform one into another, and thus, lower is the cost of matching.

The challenge that still remains is to come up with a good measure, which can give reasonable costs between two shapes. The metric should ideally be invariant to translation, rotation, and scaling; the metric should also be able to account for non-rigid shapes i.e., it should be invariant to articulations, the metric should be robust enough to ignore noise in the shape boundary, it should be able to handle deformations, and, hopefully, it would permit partial matching of shapes. Most current-day shape matching techniques cannot handle such diversity.

The task of matching two shapes is currently being thought of as a task of matching their respective contours. A 2-D shape, $S$, is modeled as a surface residing in $\mathds{R}^2$, which has a well-defined boundary, $\mathcal{B}$. Most algorithms sample this boundary and define features at the sampled locations. A correspondence problem is then solved, and the total cost of matching the two sets of features is considered as the cost of matching the two boundaries, and therefore, the cost of matching the two shapes (Section \ref{sec:sscdesc} gives a detailed description of the method).

\begin{figure}
  \centering
  \vspace{-10ex}
  {\shortstack{\includegraphics[width=3cm, height=3cm]{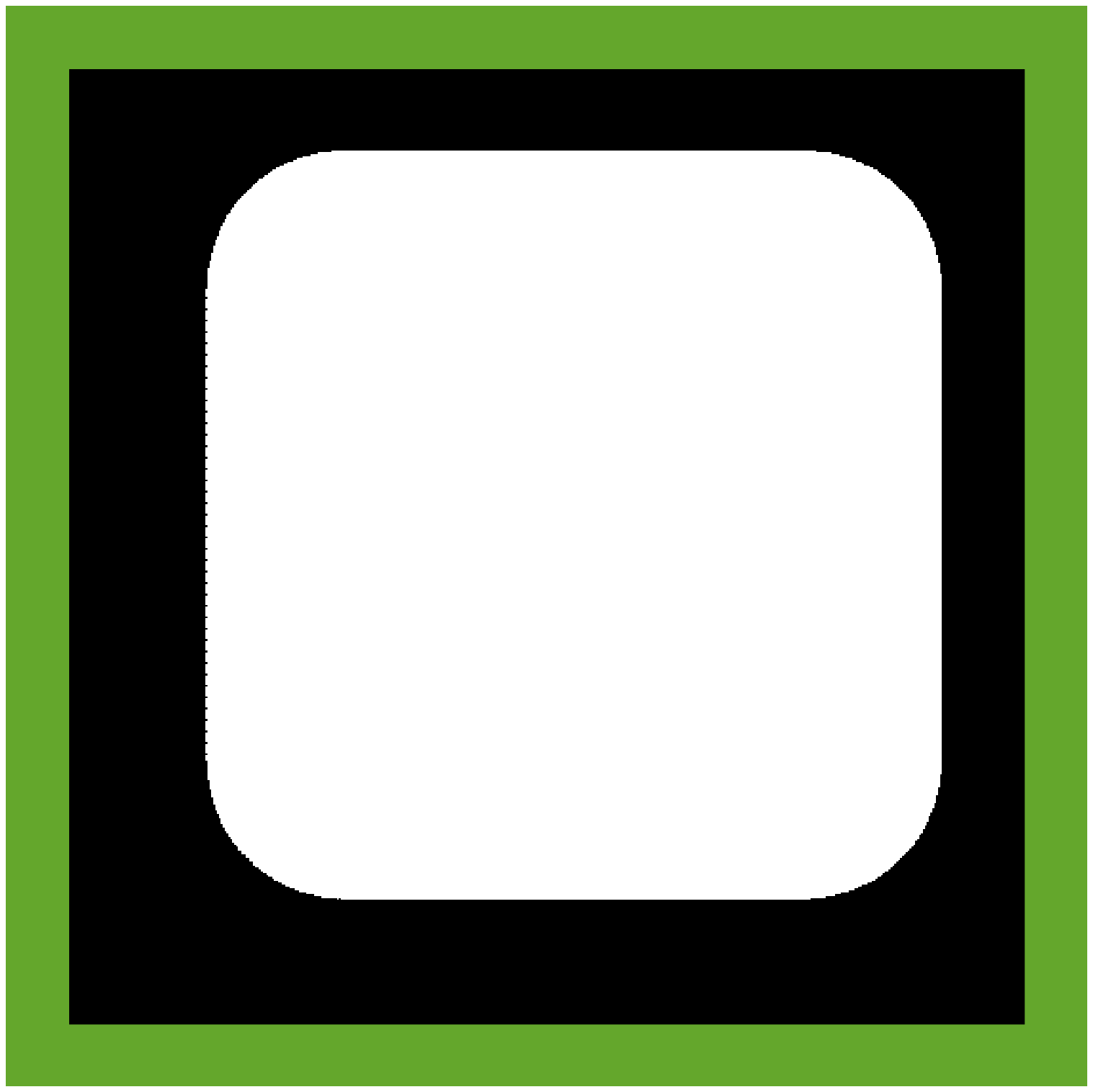}
\includegraphics[width=3cm, height=3cm]{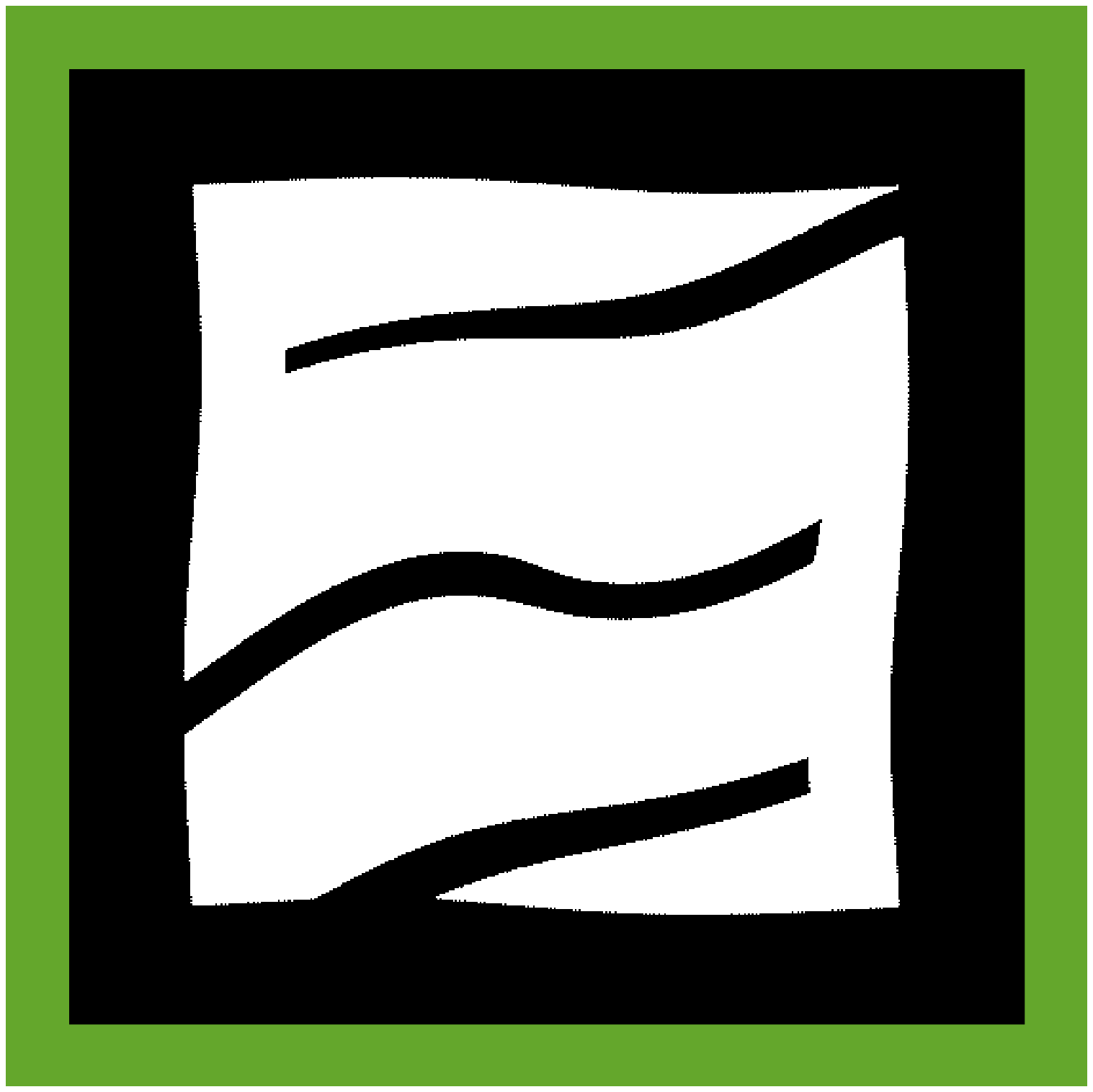}
\includegraphics[width=3cm, height=3cm]{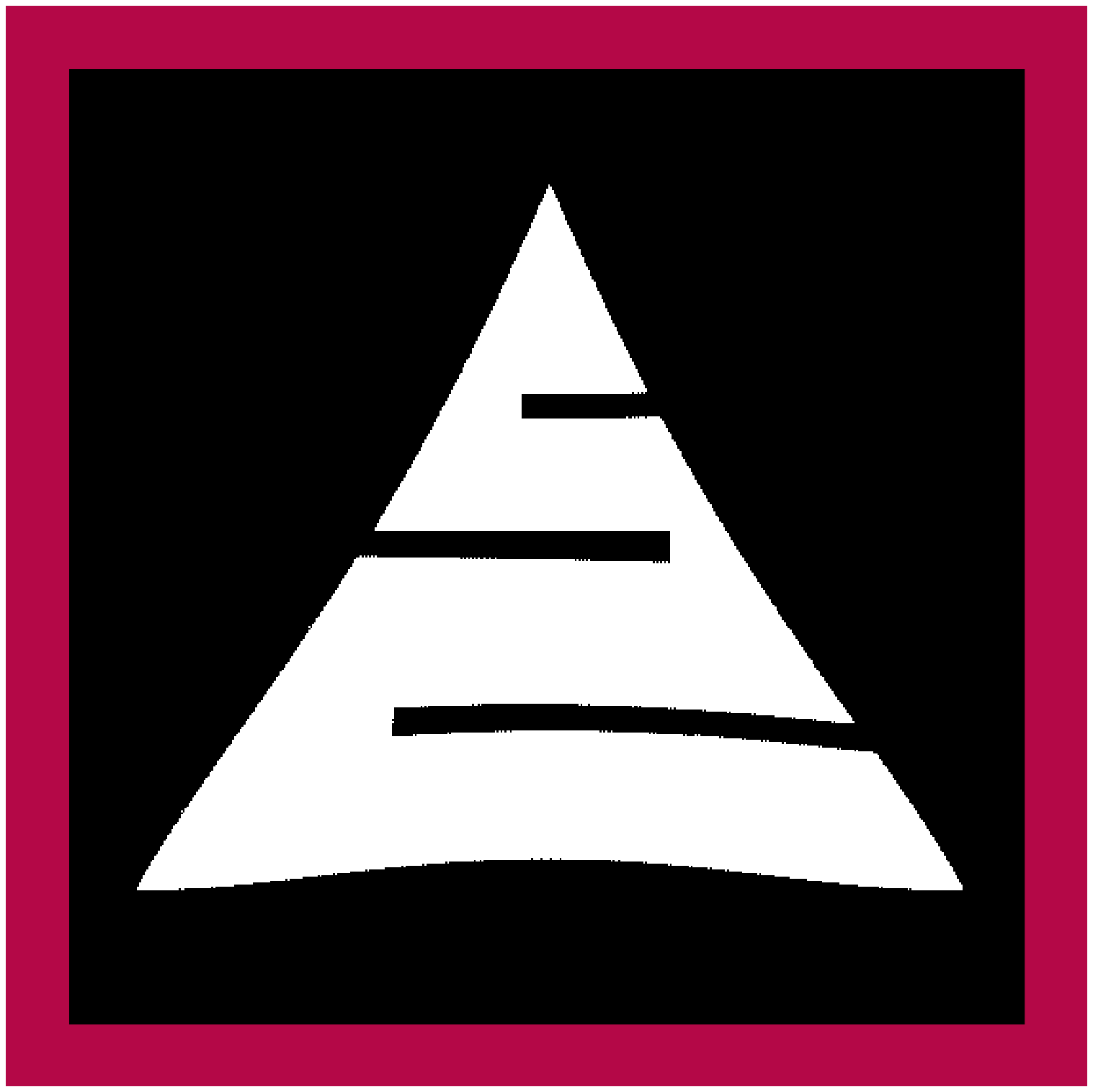}
\includegraphics[width=3cm, height=3cm]{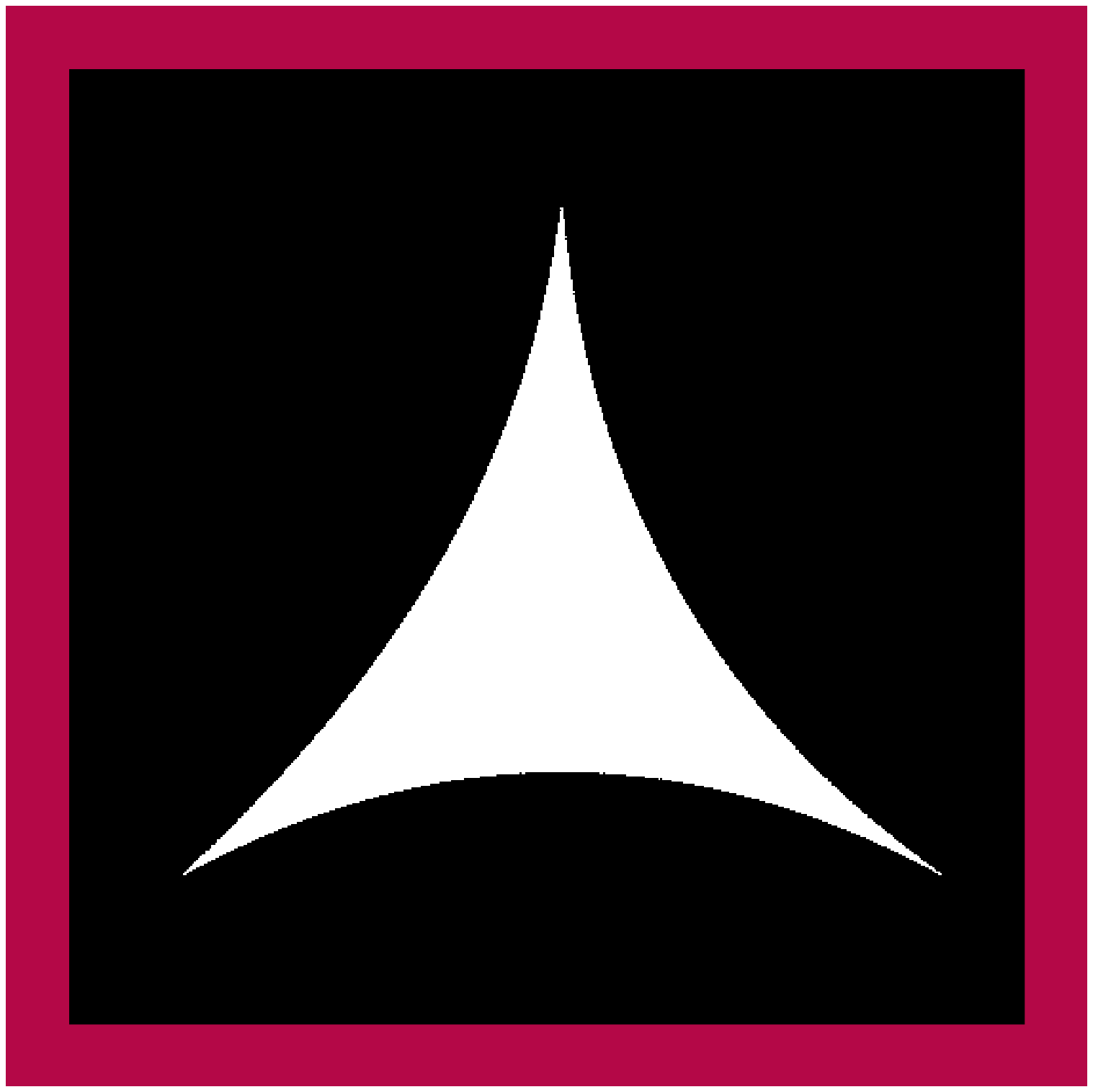}\\
\includegraphics[width=3cm, height=3cm]{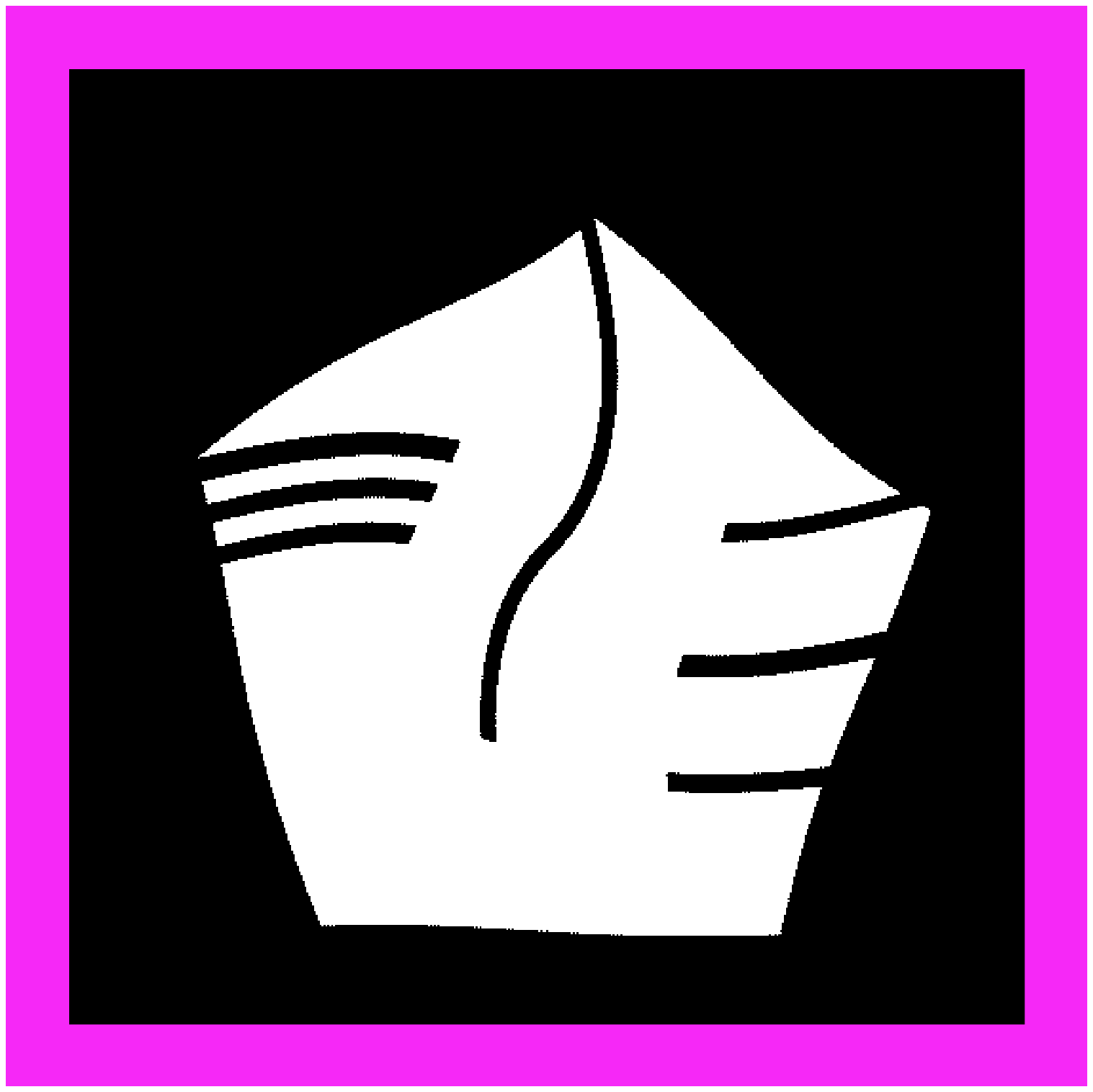}
\includegraphics[width=3cm, height=3cm]{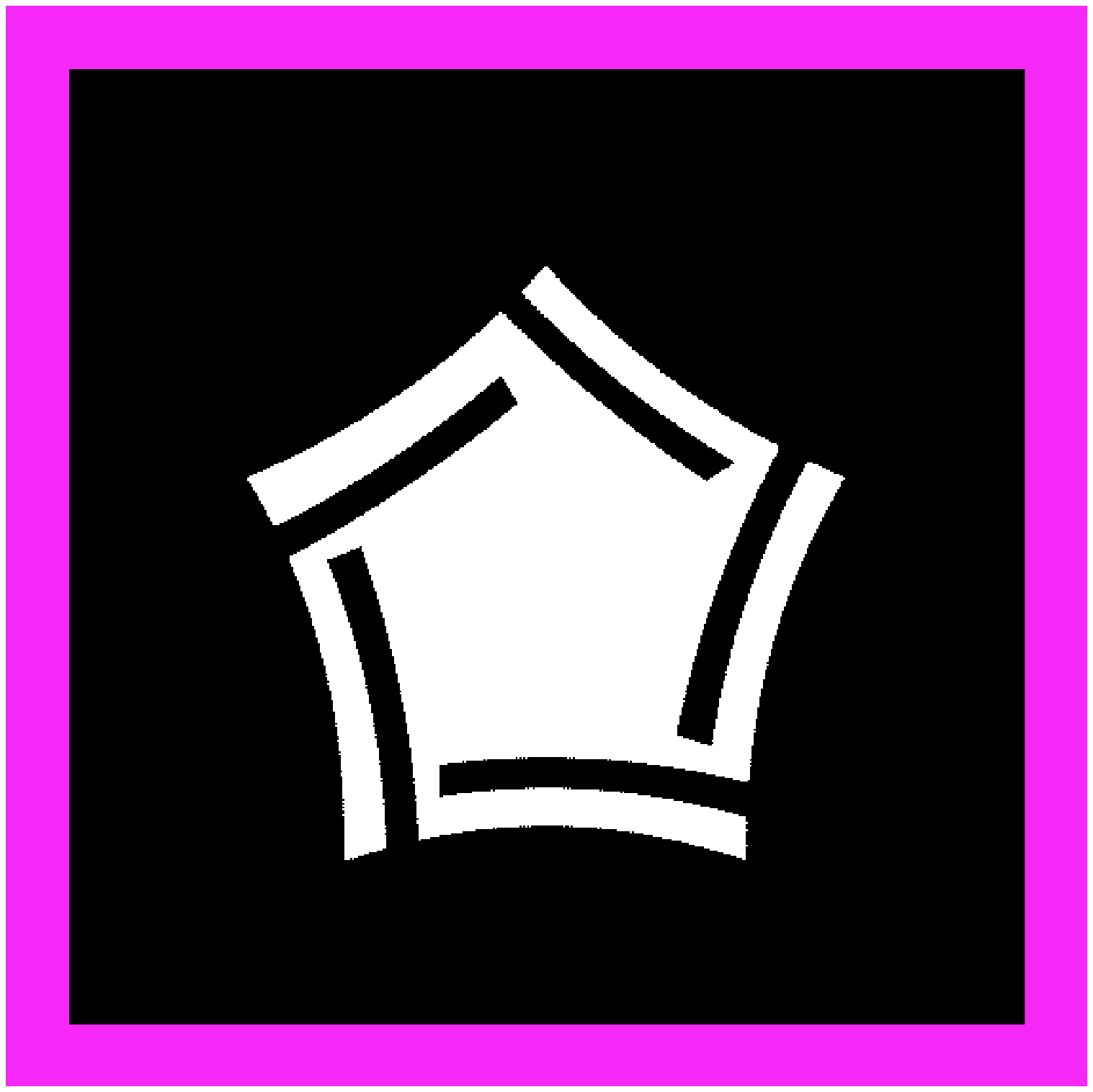}
  \includegraphics[width=3cm, height=3cm]{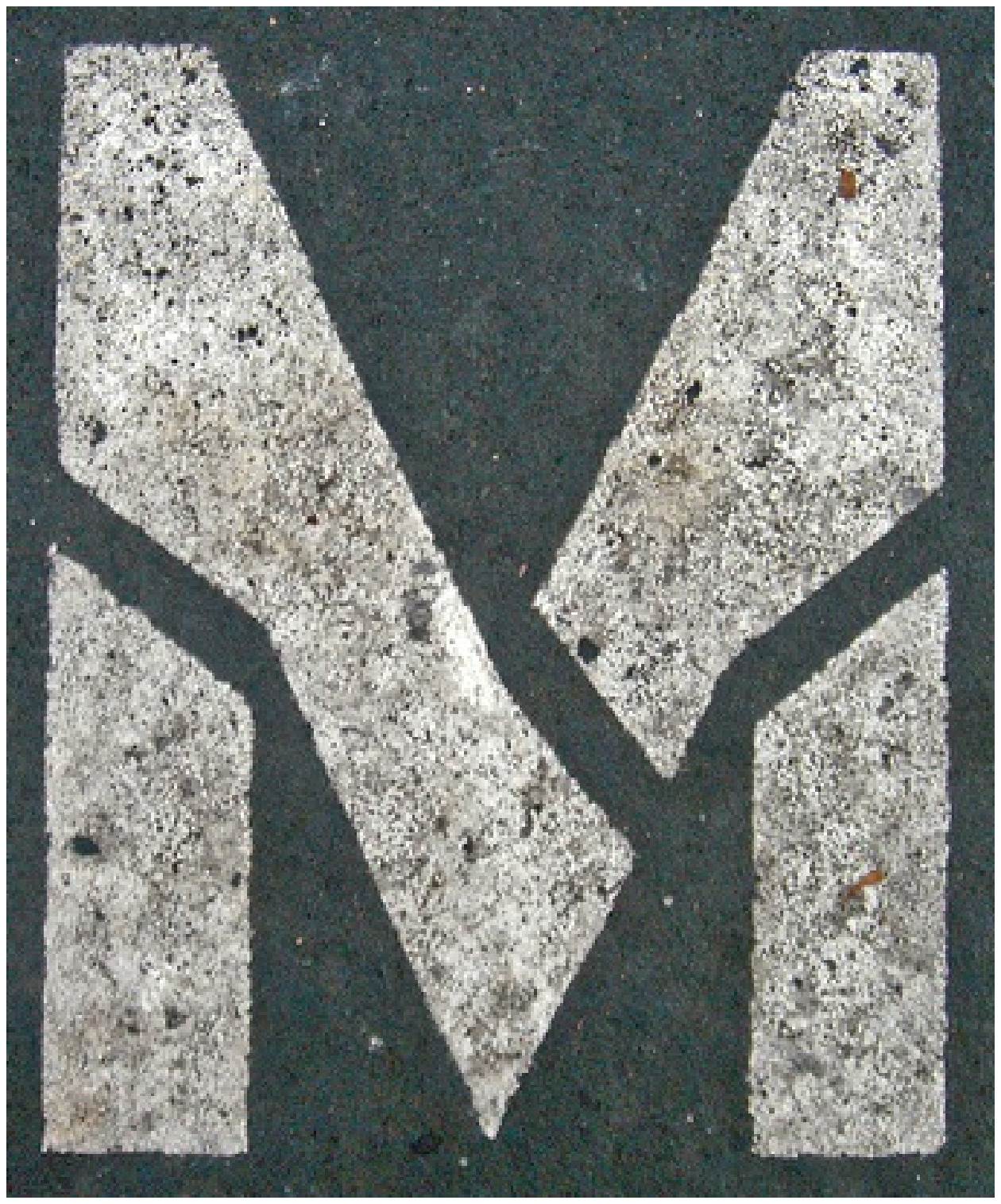}
  \includegraphics[width=3cm, height=3cm]{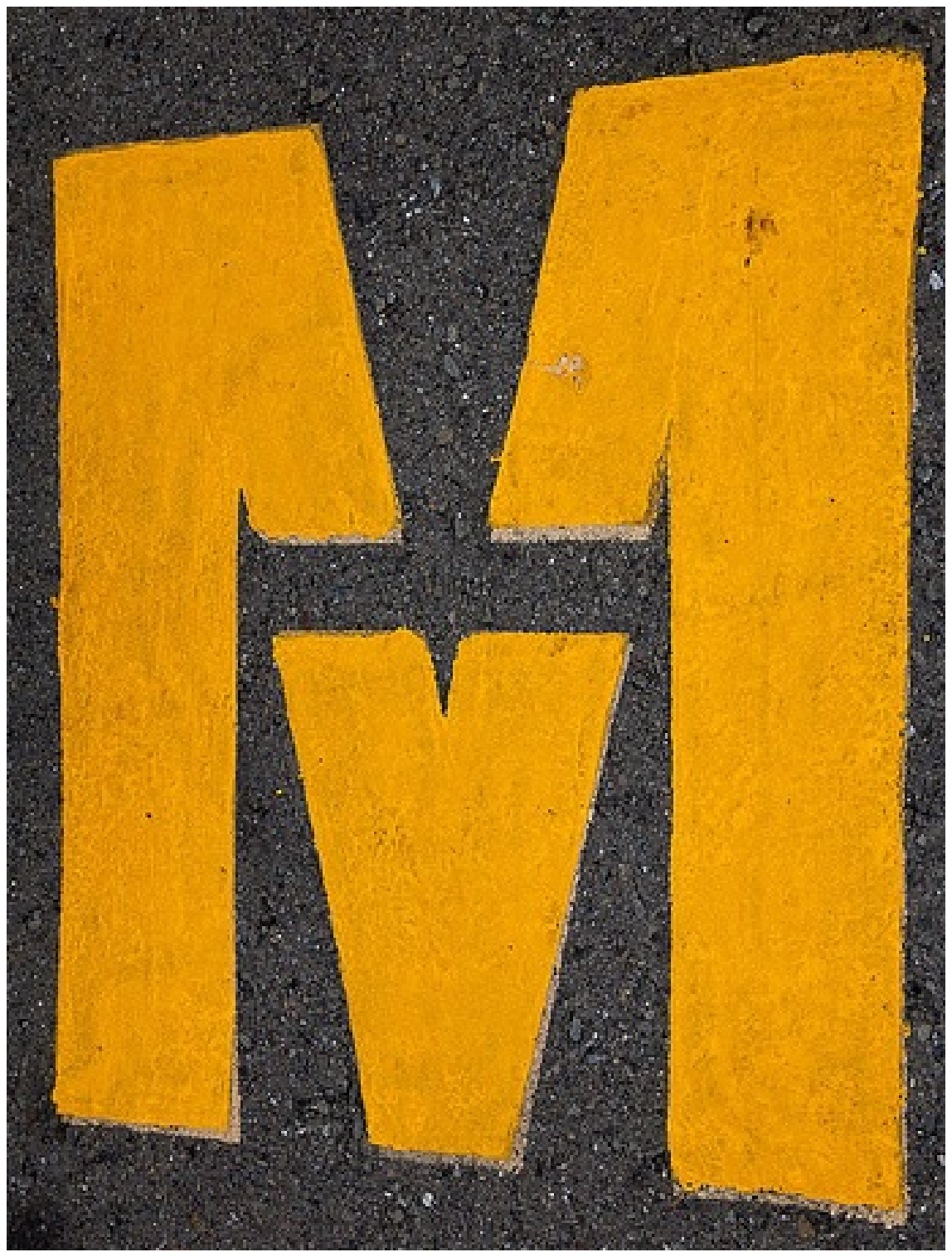}}}
  \caption{[Best Viewed in Colour] Figure shows examples of objects that are visually similar to each other even though they have multiple indentations (and even breaks) within their contours. Algorithms that perform contour-based matching of shapes cannot be used while matching such objects. Visually similar objects are appropriately color-coded using their bounding boxes.}
  \label{fig:similar}
\end{figure}

While most of the shape information can usually be extracted from just the object's contour, it is not true in cases where the objects have a strong base structure. In such cases, indentations in their boundaries have minimal effect on the human visual system. Figure \ref{fig:similar} shows examples of objects that are visually similar to each other even though some of them have multiple indentations in their contours. People tend to neglect these minor (or even major) indentations while perceiving the object's shape. This is in accordance with  Gestalt psychology, which maintains that the human eye sees objects in their entirety before perceiving their individual parts. The gestalt effect is the form-generating capability of our senses, particularly with respect to the visual recognition of figures, and whole forms, instead of just a collection of simple lines and curves.

Due to the Gestalt effect, approaches that perform shape-matching based on part decomposition, or curve matching, will not perform well on objects such as those shown in Figure \ref{fig:similar}. Somehow, there is a need for the development of matching techniques that can some how capture this Gestalt effect \cite{desolneux2007gestalt}.

In this paper, we propose a novel way of extracting the shape properties that capture the object's shape in its entirety. We show how this can help in improving the retrieval rates by testing on the well-known MPEG7 shape database. We also show improvements in performance over other recently proposed perceptually motivated techniques.

The rest of the paper is organized as follows. Section \ref{sec:related} discusses some of the previous work on shape-based object detection. We also identify some of the problems that the recent techniques face. In Section \ref{sec:ssc}, we explain our method in detail and show how it can be used to tackle some of the problems mentioned in Section \ref{sec:related}. In Section \ref{sec:exp}, we provide results from our experiments, which shows an improvement over some of the recently proposed techniques. Finally, in Section \ref{sec:conclusion}, we conclude the paper, with directions for future work.

\section{Related Work}
\label{sec:related}
Shape matching has been recognized as an important area of computer vision and has been actively pursued in the recent past. Some notable advances that have been made in this area over the past decade are discussed below. A typical approach to measure shape similarity is through non-rigid shape deformation \cite{sebastian2003aligning, felzenszwalb2007hierarchical}. Such methods measure the difficulty in transforming one shape into another. Geometrically, one can think of a shape $S$ as a point on some low-dimensional manifold $\mathbb{M}$, residing in some high-dimensional space. The energy required to transform a shape $S_1$ into a shape $S_2$ can be thought of as the geodesic distance\footnote{In this paper, we will use the terms cost, distance, and energy, interchangeably}  of the shortest path between the two points lying on the manifold.

Most approaches equate the task of shape-matching to the matching of the respective object boundaries. The shape boundaries are discretized into a set of $n$ landmark points, $S = \{p_1, p_2, ... p_n\}$, for easier representation and matching. Belongie et al. \cite{belongie2002shape} showed that these points could be located at any place on the object boundary and that they need not be restricted to extrema points on the curve. They also proposed to describe the shape using shape contexts at each of these sampled points. The shape context at each sampled point is given by the relative distribution of the rest of the $n-1$ points, which is represented as a 2-D histogram of distances and angles.

The shape context (SC) can be made invariant to translation, rotation and scale. However, while SC matching performs well on rigid objects, it is susceptible to articulations. This is because the SC histogram is composed of Euclidean distance and angle, which cannot handle articulations. To overcome this problem, Ling et al. \cite{ling2007shape} proposed a variant of SC, namely, Inner Distance Shape Context (IDSC). The IDSC uses inner distance (the length of the shortest path connecting the two points, such that the path lies completely within the shape) and inner angle, instead of Euclidean distance and angle, to generate the histograms at the sampled points. The use of this changed metric makes the descriptor invariant to articulations. Also, as suggested by Thayananthan et al. \cite{thayananthan2003shape}, they make use of the figural continuity constraints and perform the context matching using a dynamic programming scheme. Though IDSC looks at distances between points such that path connecting them lies completely within the shape's boundary, it still cannot capture the interior density of the shape. It relies completely on the distances between points that lie on the shape's boundary. We show how this important interior property can be captured in Section \ref{sec:dense}, and show how it can be used to construct meaningful shape descriptors in Section \ref{sec:sscdesc}.

Bronstein et al. \cite{bronstein2009partial} tackle the problem of partial similarity and show how objects that have large similar parts (but not completely similar) can be matched. They present a novel approach, which shows how partiality can be quantified using the notion of Pareto optimality. They use inner distance in order to handle non-rigid objects \cite{bronstein2008analysis}. The notion of Pareto optimality has since been applied by other authors for measuring partiality of shapes \cite{donoser2009efficient}.

Gopalan et al. \cite{gopalan2010articulation} identified that though the use of inner distance provided invariance to articulations, it could not be directly applied to ``non-ideal" 2-D projections of 3-D objects. If the projection took place using a weak perspective, then not all parts of the 3-D model would get accurately projected onto the 2-D plane. In order to overcome this problem, they modeled an articulating object as a combination of approximate convex parts and performed affine normalization of these parts. They then use inner distance to perform shape matching on the normalized shapes. Their near-convex decomposition algorithm takes as input the contour of the object and splits the object into multiple convex parts. However, such an approach cannot be followed for shapes such as those shown in Figure \ref{fig:similar}, since the algorithm would split the object into multiple parts, yielding undesirable results.

\begin{figure}
  \centering
  \subfloat{\includegraphics[width=2.5cm, height=2.5cm]{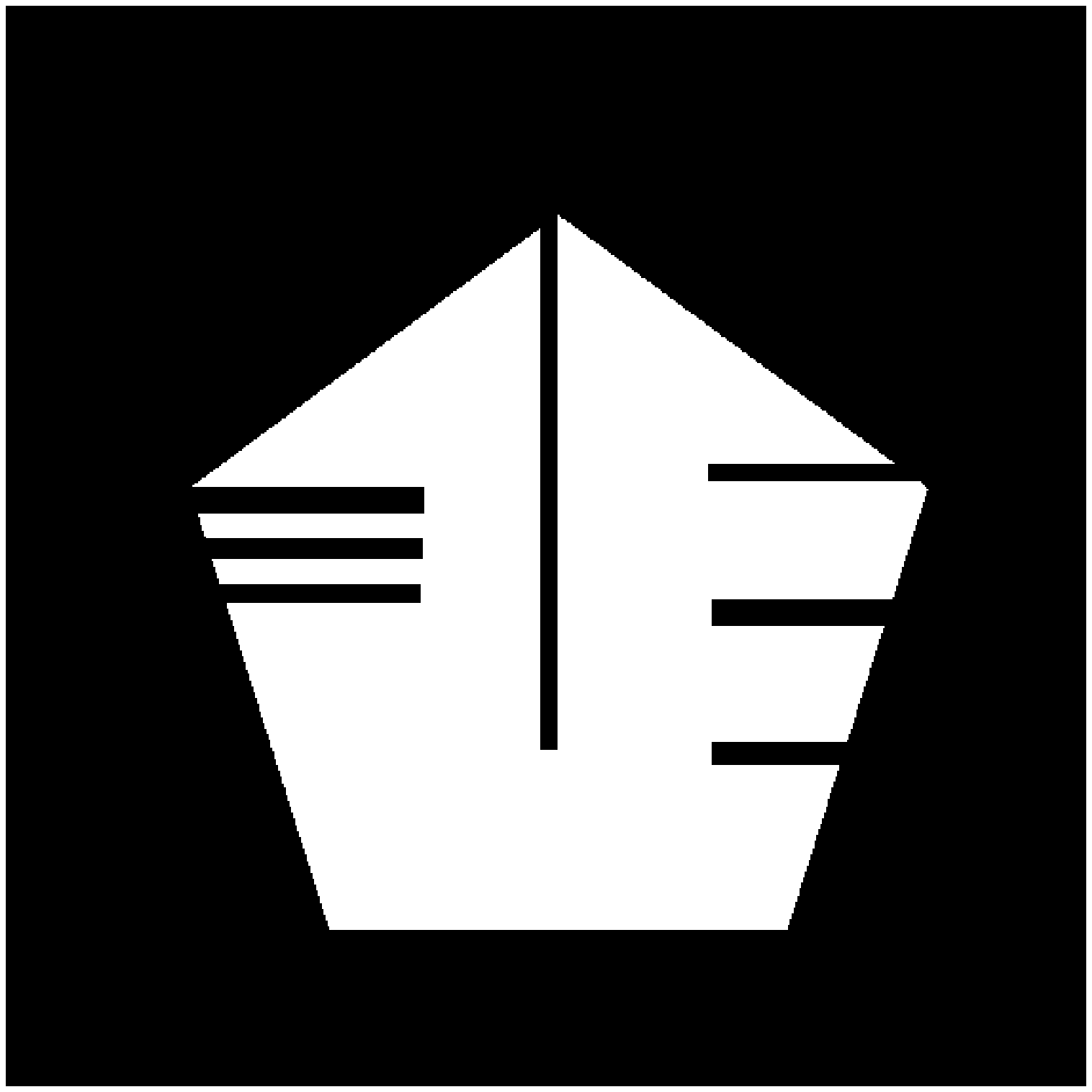}}
  \subfloat{\includegraphics[width=2.5cm, height=2.5cm]{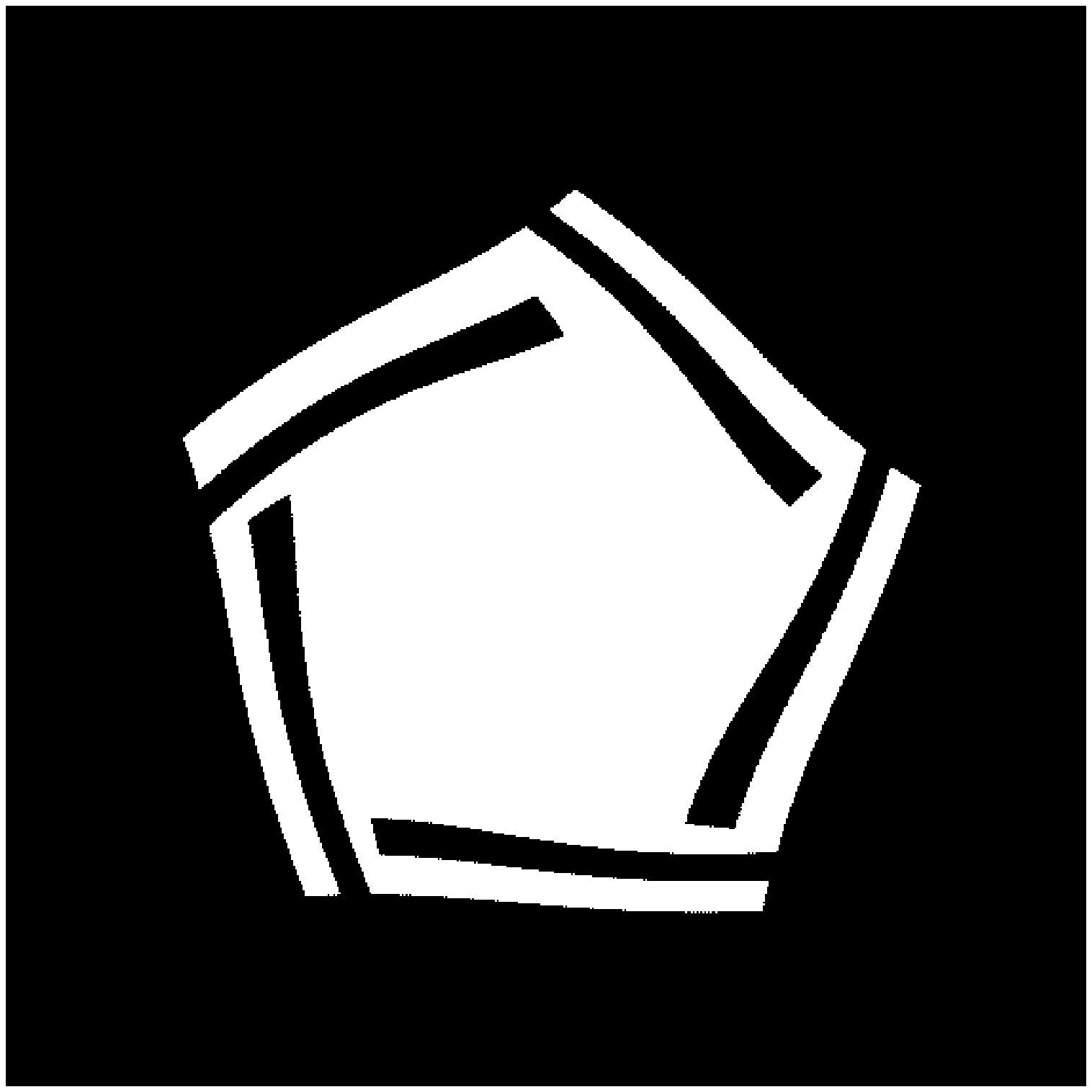}}
  \subfloat{\includegraphics[width=2.5cm, height=2.5cm]{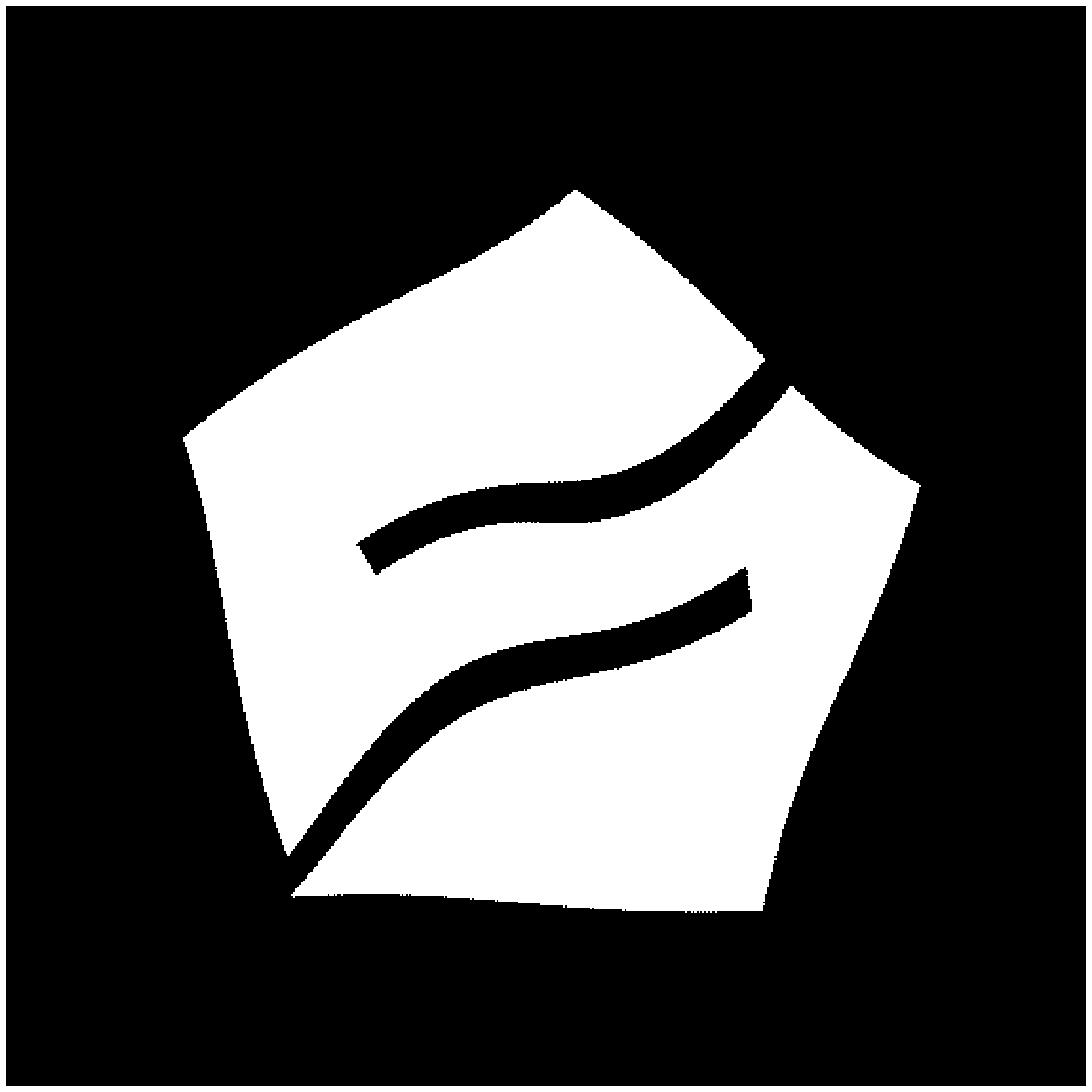}}
  \subfloat{\includegraphics[width=2.5cm, height=2.5cm]{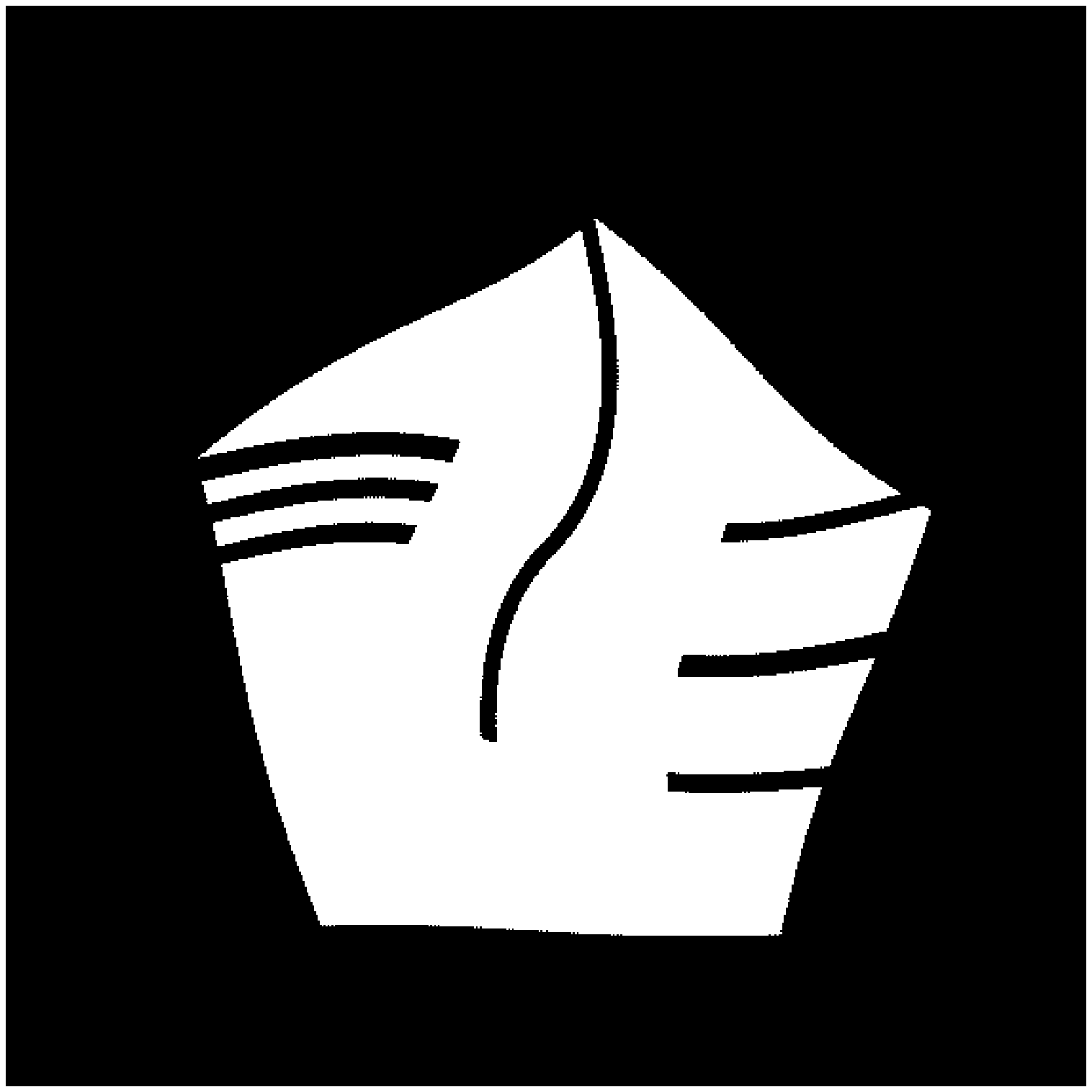}}
  \subfloat{\includegraphics[width=2.5cm, height=2.5cm]{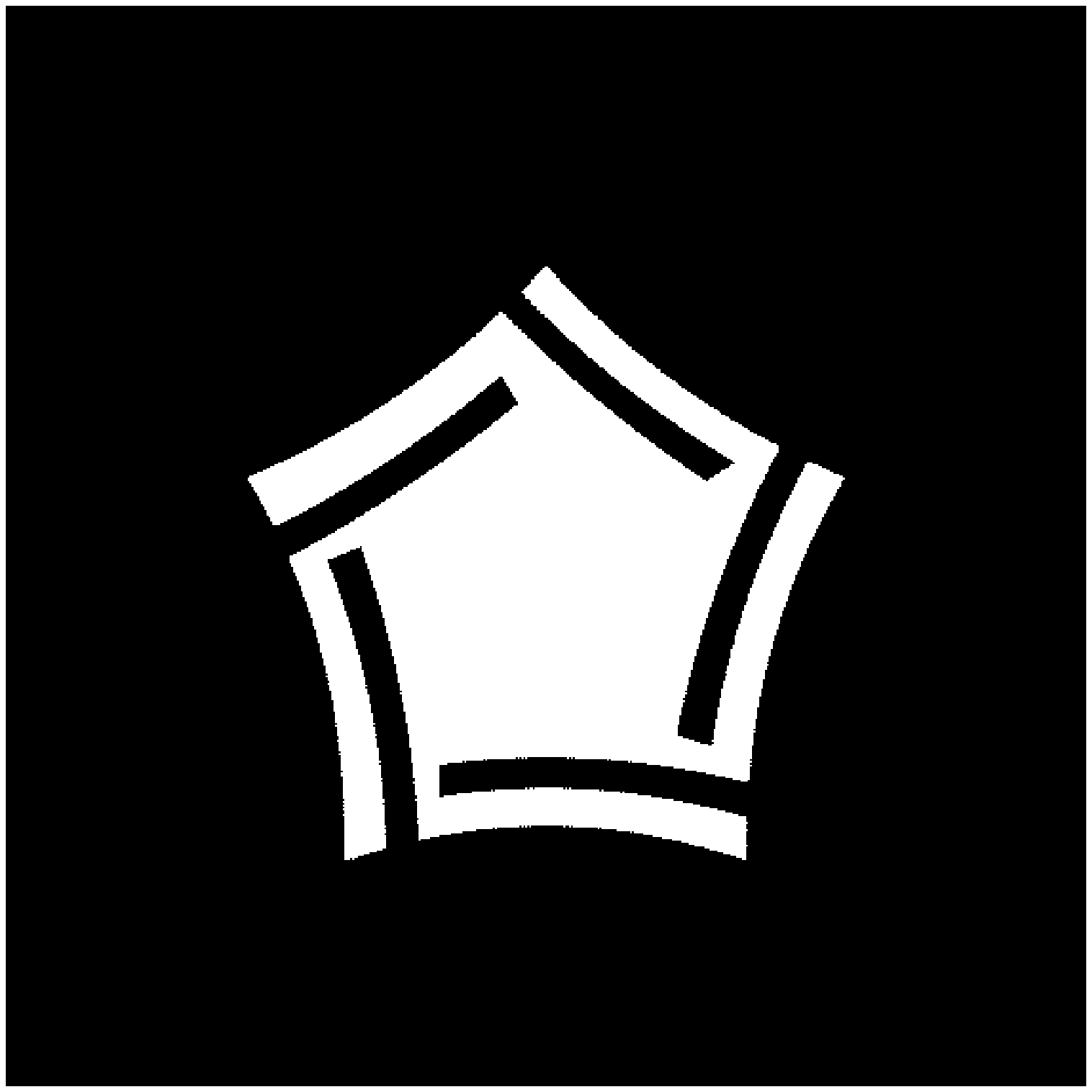}}\\

  \subfloat{\includegraphics[width=2.5cm, height=2.5cm]{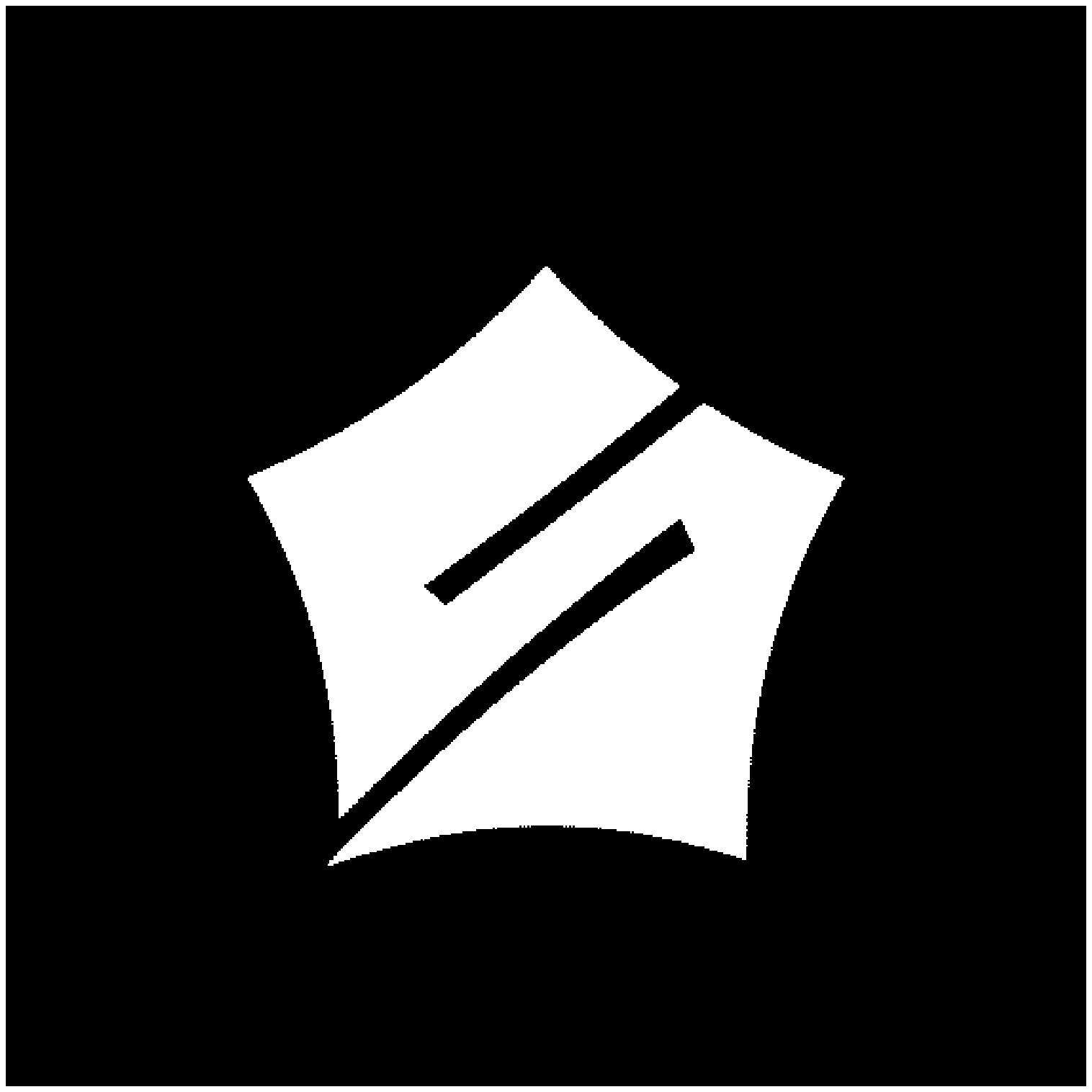}}
  \subfloat{\includegraphics[width=2.5cm, height=2.5cm]{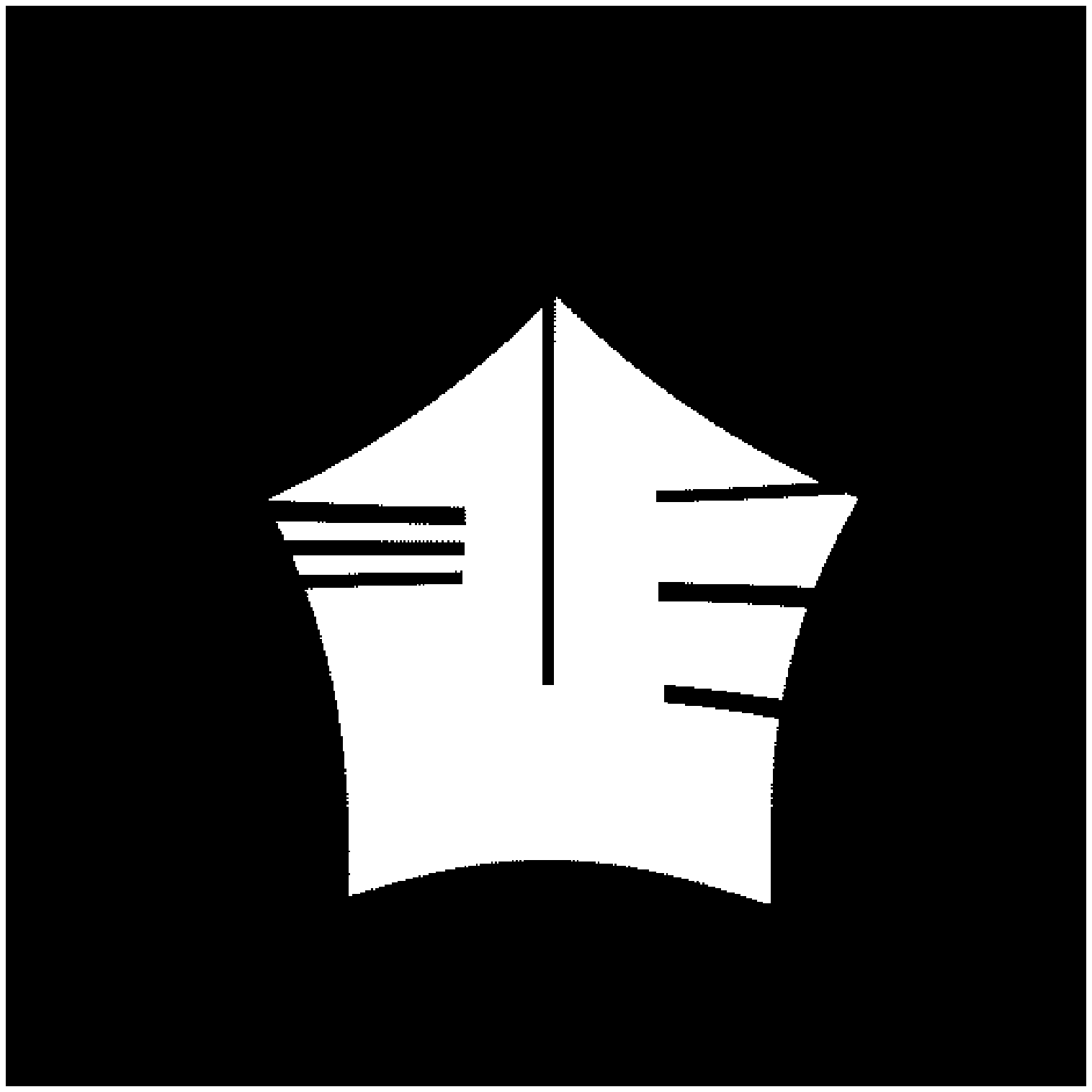}}
  \subfloat{\includegraphics[width=2.5cm, height=2.5cm]{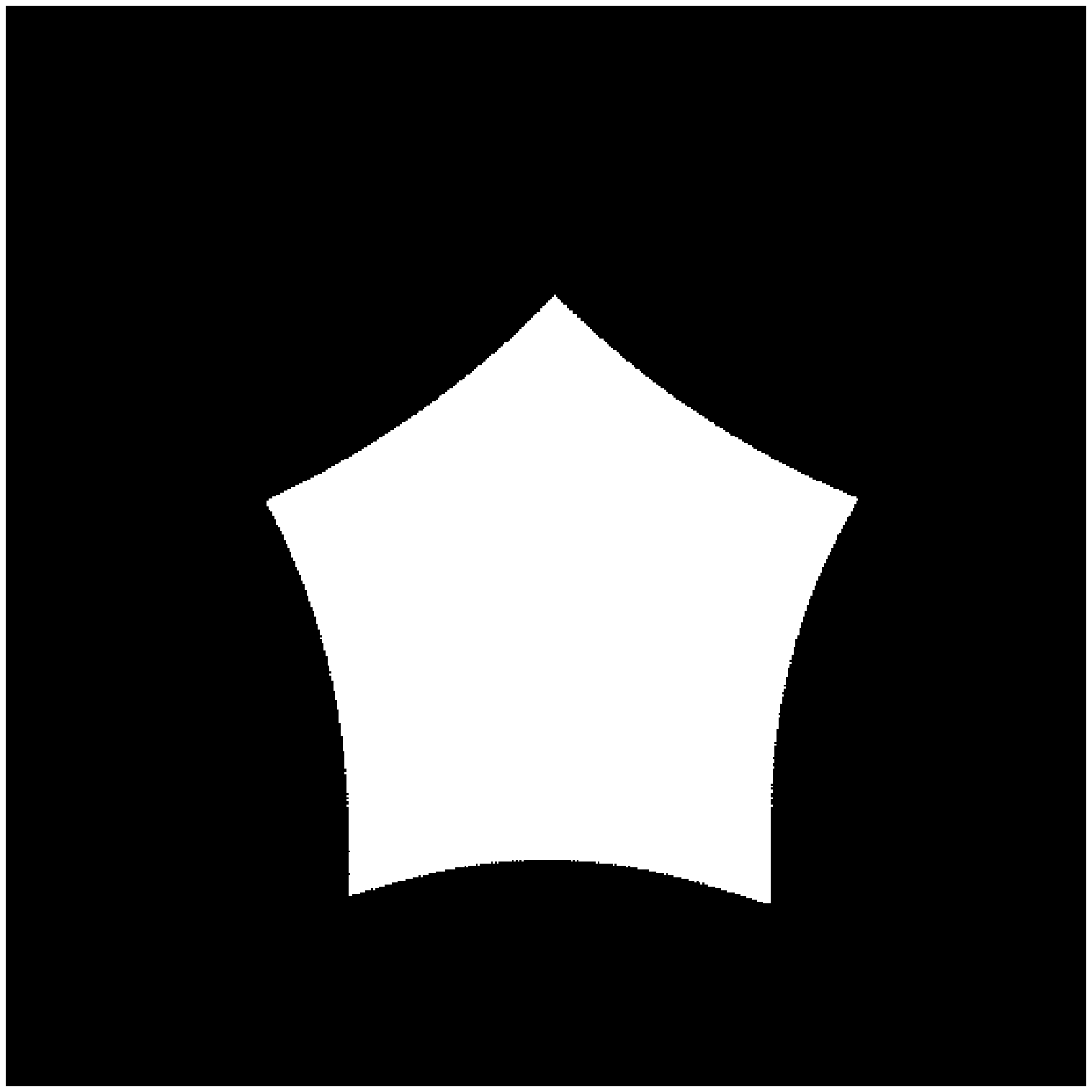}}
  \subfloat{\includegraphics[width=2.5cm, height=2.5cm]{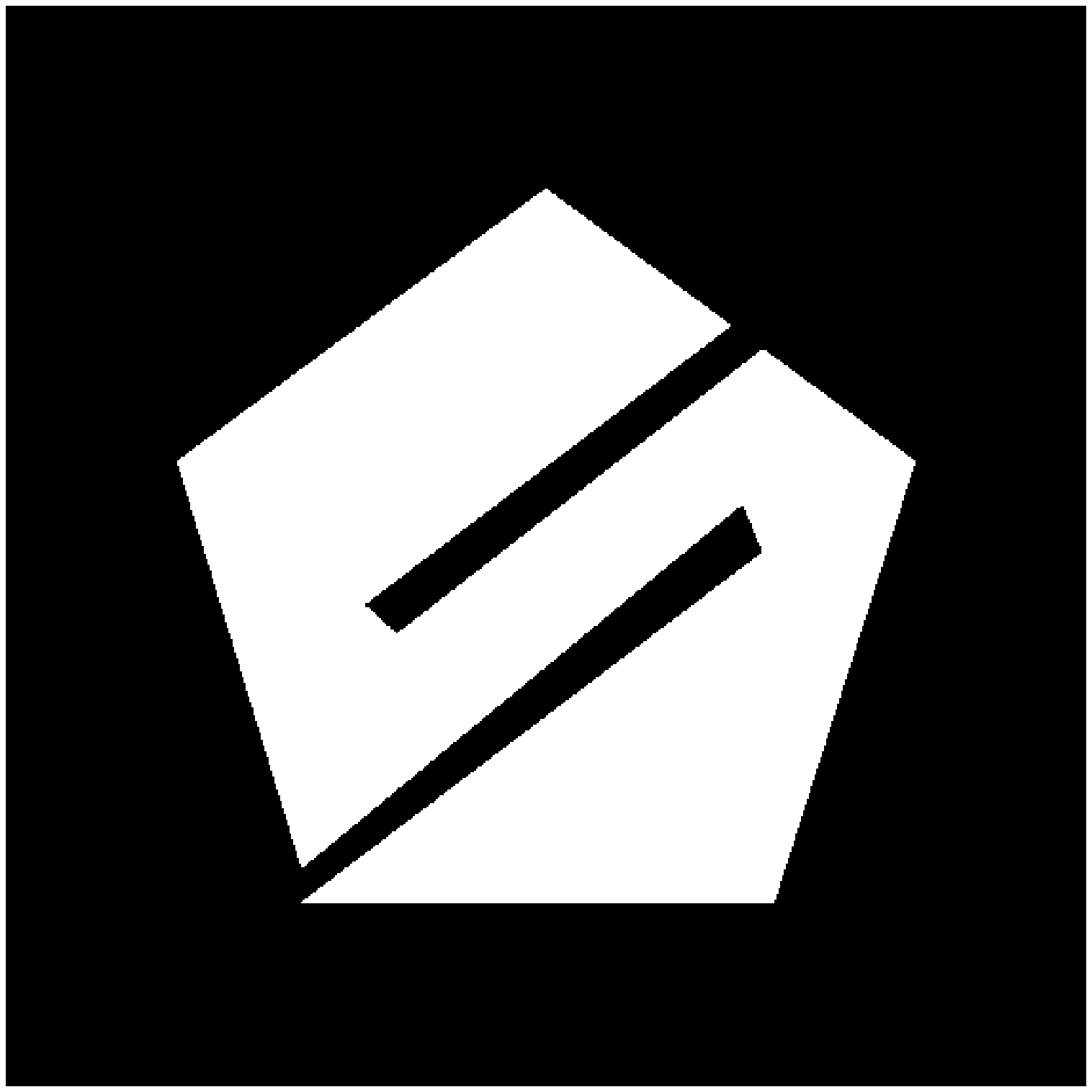}}
  \subfloat{\includegraphics[width=2.5cm, height=2.5cm]{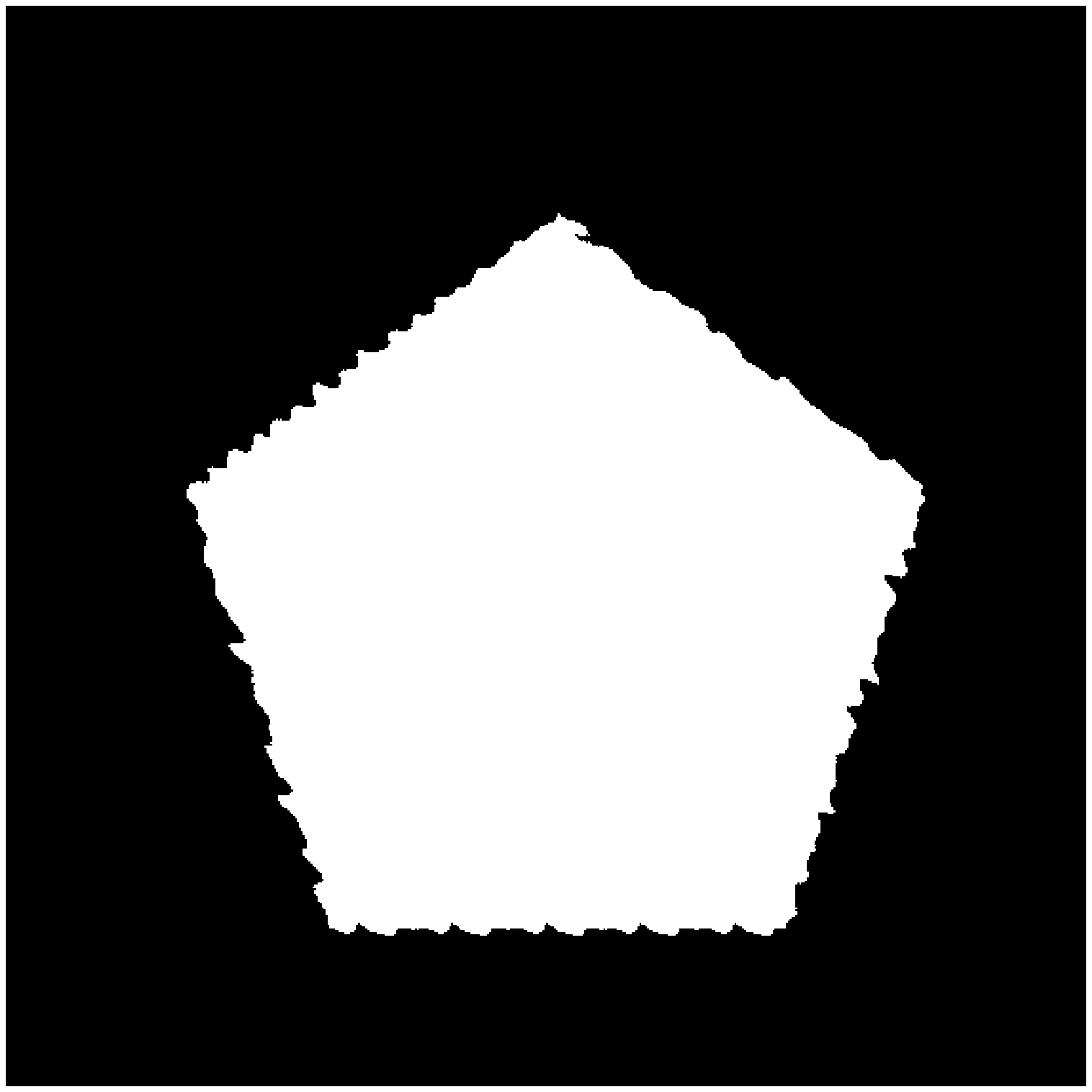}}
  \caption{Figure shows more examples of a particular class of objects from the MPEG7 database. All of the above objects have different contour properties. However, their overall visual similarity is still that of a pentagon.}
  \label{fig:device6}
\end{figure}

The Medial Axis Transform (MAT) and its variant, shock graphs, have been used by certain authors for matching shapes \cite{siddiqi1999shock,sebastian2004recognition}. The medial axis, or skeleton, is the locus of the centers of all maximally inscribed circles of the object. While the MAT captures the interior properties of the shape to a large extent, by definition, the generation of a skeleton depends on the boundary of the object. Therefore, the objects shown in Figure \ref{fig:device6} will all have vastly different skeletons. Xie et al.\cite{xie2008shape}, proposed to model shapes using skeletal contexts. Their contexts are calculated at the skeleton endings and the bins are populated by the non-uniformly sampled points from the boundary. Relying on the skeleton, and the boundary points, makes their method susceptible to indentations in the contour. We show, in Section \ref{sec:ssc}, how our method does not fall prey to such boundary perturbations.

Due to the diversity involved in shape-matching, it has become difficult to come up with a single measure that incorporates all the requirements. While the use of Euclidean distance is beneficial for identifying certain classes of objects, the use of inner distance favours some others. As a result, researchers have started to fuse two or more techniques while calculating the distance between two shapes. Ling et al. \cite{ling2010balancing} identified that the use of inner distance was ``overkill" for certain classes of objects and proposed a technique to balance deformability and discriminability. They calculate the cost between two shapes with the help of various distance measures, parameterised by an aspect weight, and retain the ``best" cost. However, they still use points sampled from the contour and their algorithm would therefore be susceptible to objects with strong base structures that have indentations in their contours.

Recently, some effort as gone into the development of perceptually motivated techniques \cite{temlyakov2010two, Hu20123222}. These techniques tackle cases, such as those shown in Figure \ref{fig:device6}. To the human visual system, all objects in the figure appear to belong to the same class. However, measures that rely on the contour to obtain the object's shape properties cannot fathom this similarity.

Temlyakov et al. \cite{temlyakov2010two} propose to split the object into a base structure and multiple strand structures. They define strands as structures that are thin and long, relatively small in size, and attached to a base structure. The strands may be made of inward or outward strands. When comparing two shapes, they compare the base structures and strand structures separately. They use IDSC for comparing base structures, and for the strands, they just check if the two objects have similar number of strands, without giving much importance to the detailed geometry. Secondly, they also identify objects with a single axis of symmetry and normalize the aspect ratios of the two shapes before comparison. Fusing these two strategies along with IDSC helps them achieve better retrieval rates. Such an approach will work well if the object has a strong base structure. However, in many cases, the objects do not have a well-defined base. Even in the case of a strong base structure, multiple parameters, such as area, length, and width, have to be set to identify the strands.

More recently, Hu et al. \cite{Hu20123222} proposed a morphological approach to model human perceptions. To ``close" the objects, they perform morphological closing on the shapes. They compare the shapes using IDSC before and after performing the morphological operation, and retain the better of the two costs. They perform the morphological operations over multiple scales. This calls for an additional scale parameter to be set. Secondly, selecting the structuring element for performing the closing operation is also a difficult task. In their experiments, they try using structuring elements of different sizes and report results from all sizes. In the next section, we explain our novel method of capturing the shape properties in their entirety, and in Section \ref{sec:exp}, we show that our method can help generate better retrieval results than \cite{temlyakov2010two} and \cite{Hu20123222}.

All the techniques described above were directed towards the development of a good distance measure between pairs of images, where the similarity of an object was influenced by just one other object. However, recent works have shown that an improvement in the retrieval performance can be achieved if other similar shapes are allowed to influence the pair-wise scores. For a given similarity measure, a new similarity measure is learned through graph transduction \cite{bai2010learning}. Many methods that focus on improving the transduction algorithms have been proposed in the recent past \cite{kontschieder2010beyond, yang2012affinity,yang2008improving}.

Starting the diffusion with a good similarity matrix will lead us to obtain better similarities at the end. A good similarity matrix is one in which similar shapes have high affinity. We show that our method helps in generating a better similarity matrix after the diffusion process. We use the Locally Constrained Diffusion Process (LCDP) \cite{yang2009locally} to learn the manifold structure of the shapes and show, in Section \ref{sec:exp}, that our matrix is able to generate highly competitive retrieval rates.

\section{Solid Shape Context}
\label{sec:ssc}
In the previous section, we reviewed past work in the area of shape matching and pointed to the fact that more research needs to be done in the development of perceptually motivated techniques. In this section, we introduce one such perceptually motivated technique, which can capture the shape properties in their entirety.

To motivate our work, let us go back to the examples in Figures \ref{fig:similar} and \ref{fig:device6}. We identify that the human visual system not only recognises shapes by their external contour, but also by their ``density". We perceive a solid disc as a different object compared to a ring, though both have a circle as their outer contour. From this example, we can see that the interior solidity plays an important role in the identification of an object. We propose to utilize this important interior property of a shape by coming up with a descriptor, which is a variant of the well-known Shape Context descriptor.

To capture the interior properties of a shape, we propose to sample a set of uniformly-spaced \emph{Dense Points} that lie within the object's body (the reader can see this as the blue points in Figure \ref{fig:dense}). We then sample a much smaller set of points, called \emph{Sparse Points}, where we compute the object's features (crosses in Figure \ref{fig:sparseCH}, sampled along the convex hull). The computed features are our modifications of the shape context, and are described using the previously sampled \emph{Dense Points}. Given two shapes, the contexts at their respective \emph{Sparse Points} are used for comparison. In the following subsections, we describe in detail how we sample the dense and sparse points, and how our Solid Shape Context (SSC) is computed.

\subsection{Dense Points}
\label{sec:dense}
Motivated by the sampling techniques used to approximate probability density functions, we propose to approximate the interior shape of an object by sampling points lying within the object's boundary. Each part of the object is equally important in understanding the shape properties. Therefore, we use an uniform sampling scheme to sample points that lie uniformly within the shape.

The issues that we face while sampling from an arbitrary shape are similar to the issues that we face while sampling from an arbitrary distribution. Uniformly sampling from a well-known and simple shape, such as a square, rectangle, circle, or a triangle, is relatively straightforward. However, uniformly sampling a fixed set of points from a random shape is not that simple.

\begin{figure}
  \centering
  \includegraphics[trim=0.8cm 0.8cm 0.8cm 0.8cm, clip=true, width=7cm, height=6cm]{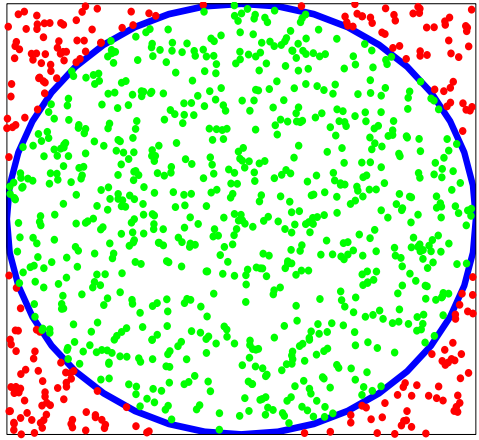}
  \caption{In order to sample points from within a circle, we uniformly sample points from a bounding square. We retain the points that fall within the circle (green) and reject the points that fall outside the circle (red).}\label{fig:rejection}
\end{figure}

One common technique that could be adopted is the rejection sampling technique. We can encompass the arbitrary shape using a well known, and simple shape (say, circle or a square), and uniformly sample points from within it. We can then retain only those points that fall within the shape boundary and reject the rest that lie outside the shape. Figure \ref{fig:rejection} gives an illustration of the rejection sampling technique.

While rejection sampling is a very simple method, there are a some issues that we encounter. It is difficult to efficiently
sample a fixed number of points lying inside the shape boundary without
wasting samples. For shapes with elongated parts, such as the tentacles
of an octopus, the accept/reject method wastes a number of samples, which is
proportional to the ratio in areas between the bounding rectangle and the
object; this ratio can be quite high for objects with parts spread
over a large region, such as horseshoes, octopi, or insects.  Even for
simple shapes such as the circle in Figure \ref{fig:rejection}, a large number of points
shown in red are wasted. Our method, which is described below, does not
waste any samples, and is therefore able to maintain a constant complexity
regardless of the shape of object.


The above problems are encountered because of two reasons:
\begin{enumerate}
    \item We are trying to sample points from arbitrary shapes, for which there are no elegant sampling techniques.
    \item We are not restricting ourselves to the interior of the shape, before sampling.
\end{enumerate}

We wish to overcome these problems by making use of the object's boundary constraints. Firstly, we restrict our sampling area such that it lies totally within the object's boundary. Secondly, we ensure that the area we are sampling from is a simple shape, to ensure easy sampling. Below, we explain in detail how we propose to sample a fixed number of \emph{Dense Points} without wasting any samples.

\begin{figure*}
  \centering
  \subfloat[]{\includegraphics[width=4.5cm, height=3.5cm]{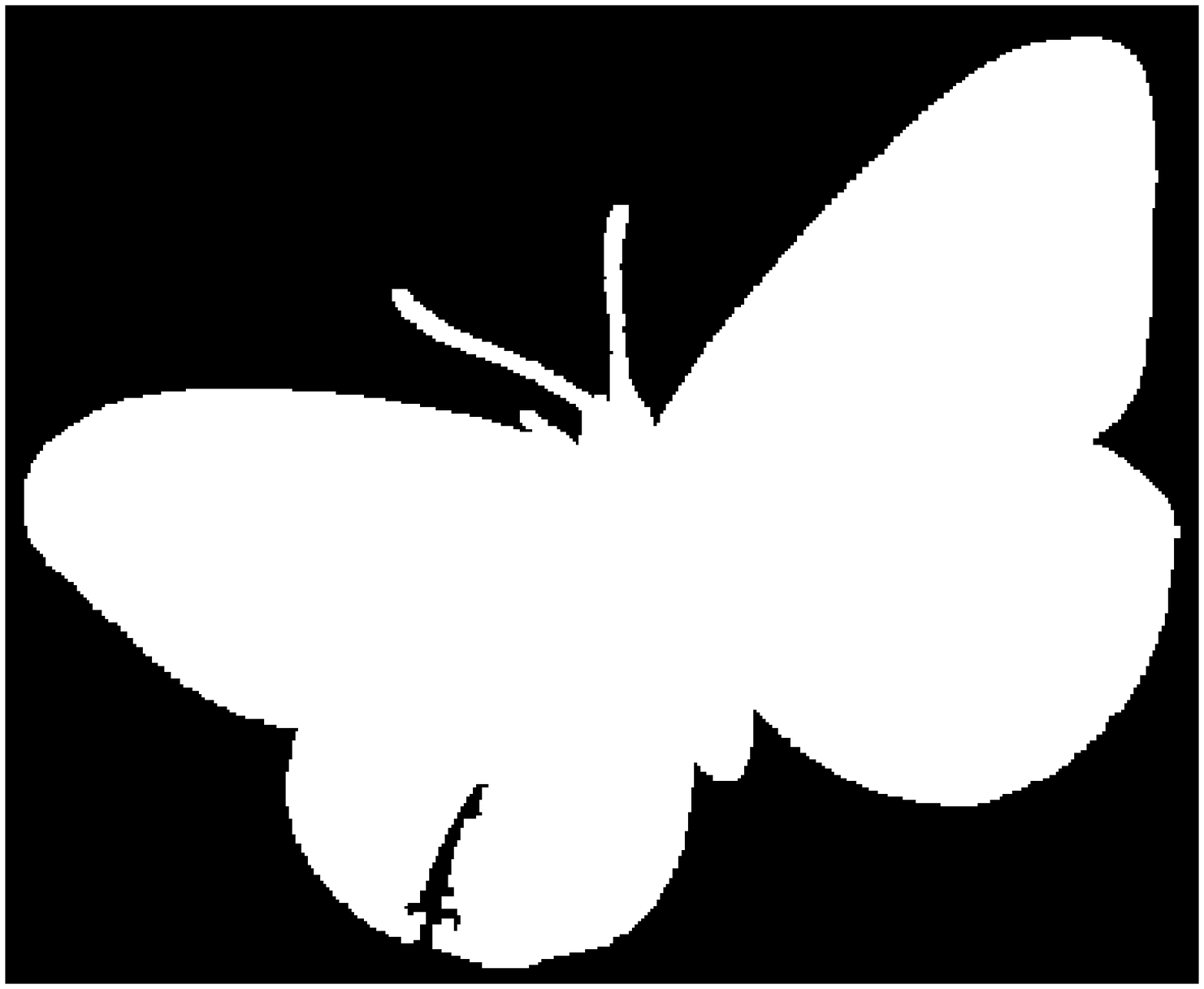}}
  \subfloat[]{\label{fig:contSample}\includegraphics[trim=2cm 2cm 1.25cm 0.8cm, clip=true, width=4.5cm, height=3.5cm]{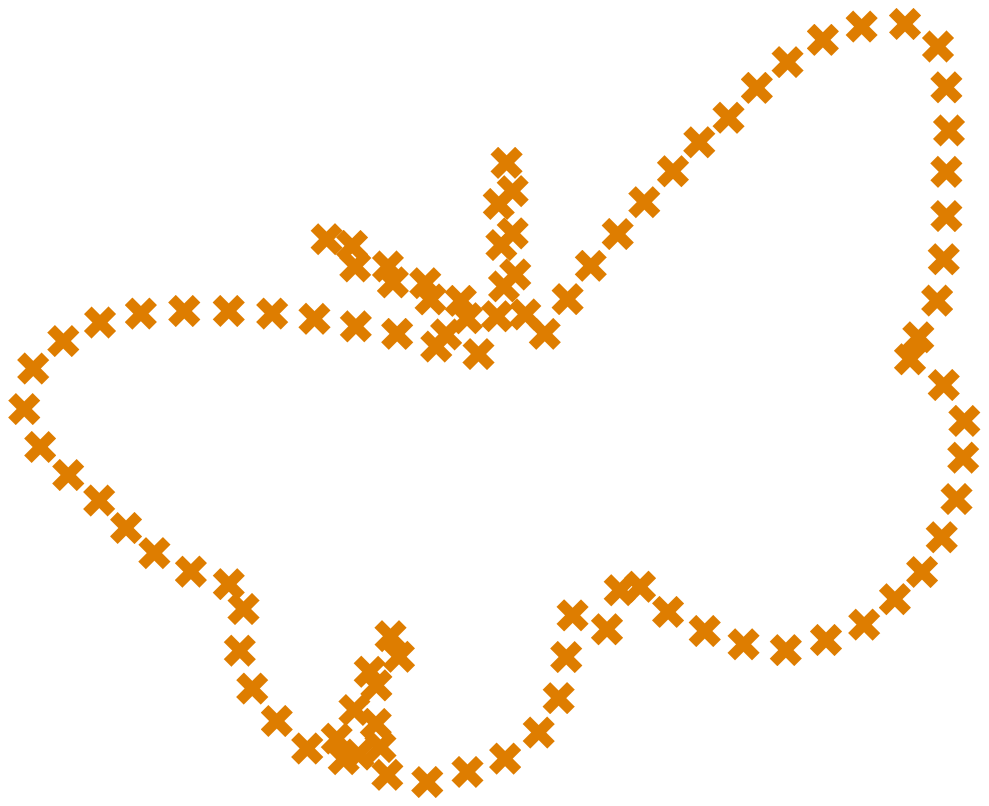}}
  \subfloat[]{\label{fig:triang}\includegraphics[trim=1.75cm 1cm 1.25cm 0.75cm, clip=true, width=4.5cm, height=3.5cm]{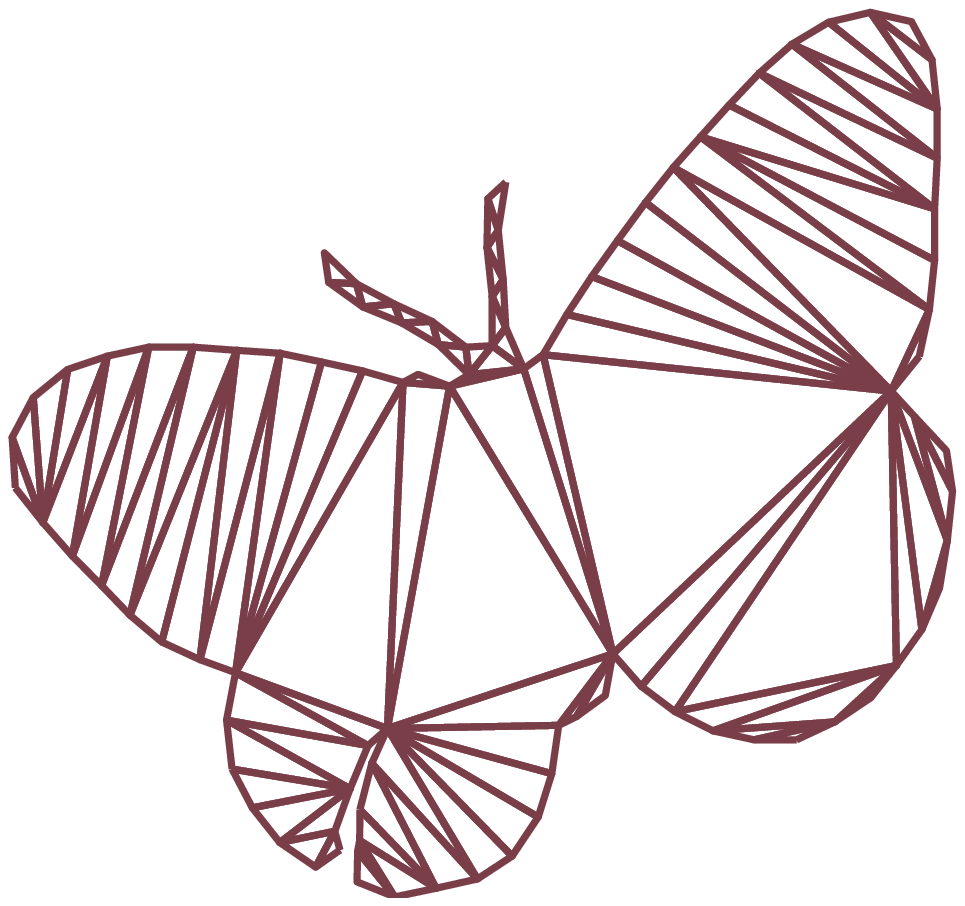}}\\
  \subfloat[]{\label{fig:dense}\includegraphics[trim=7cm 1.8cm 4cm 1cm, clip=true, width=4.5cm, height=3.5cm]{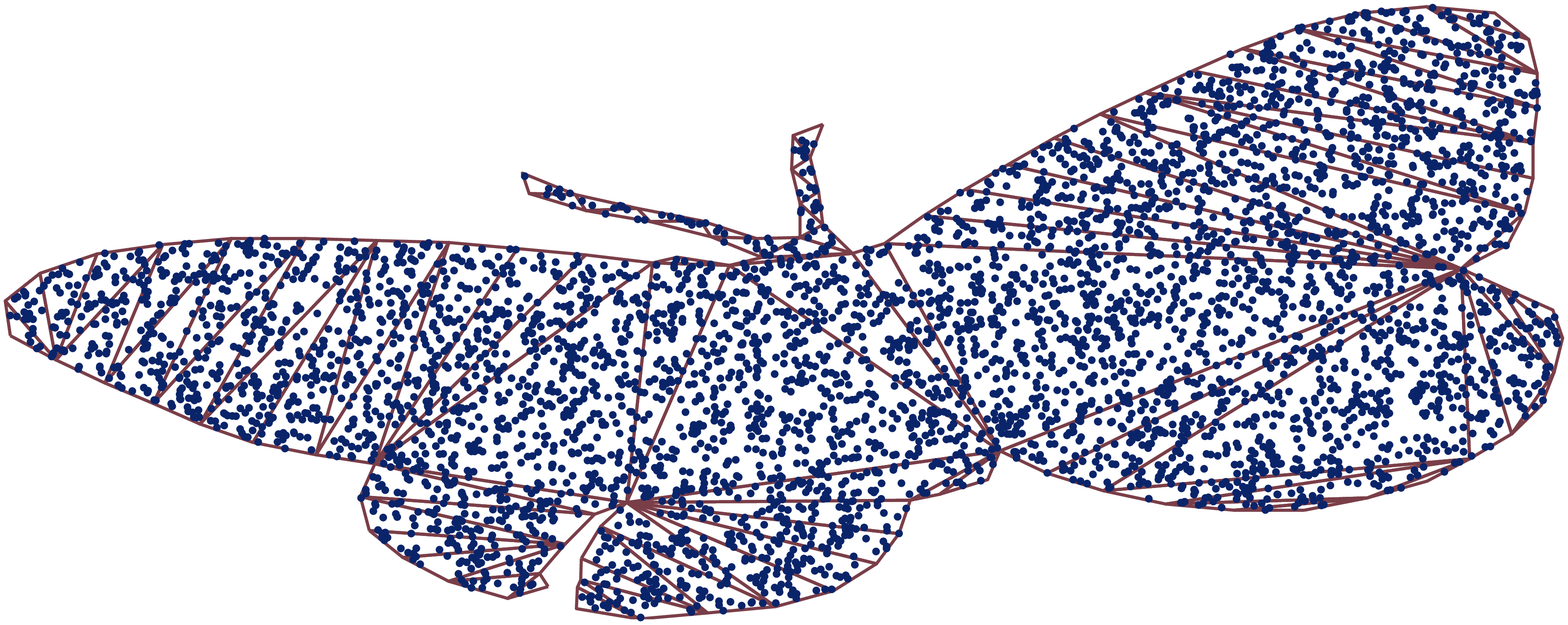}}
  \subfloat[]{\label{fig:sparseCH}\includegraphics[trim=6cm 2cm 22cm 11cm, clip=true, width=4.5cm, height=3.5cm]{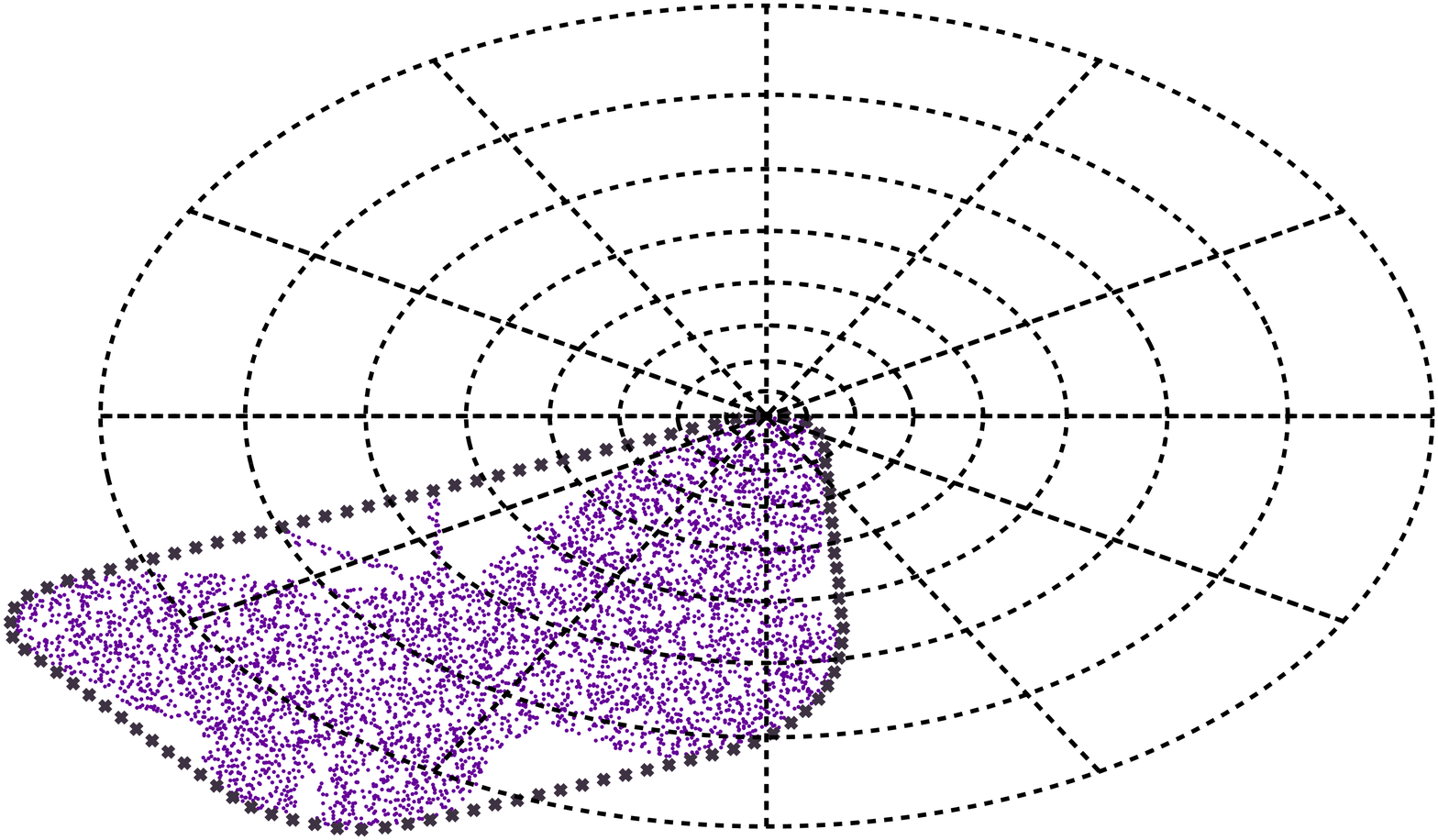}}
  \subfloat[]{\label{fig:sscHist}\includegraphics[trim=0cm 0cm 0cm 0cm, clip=true, width=4.5cm, height=3.5cm]{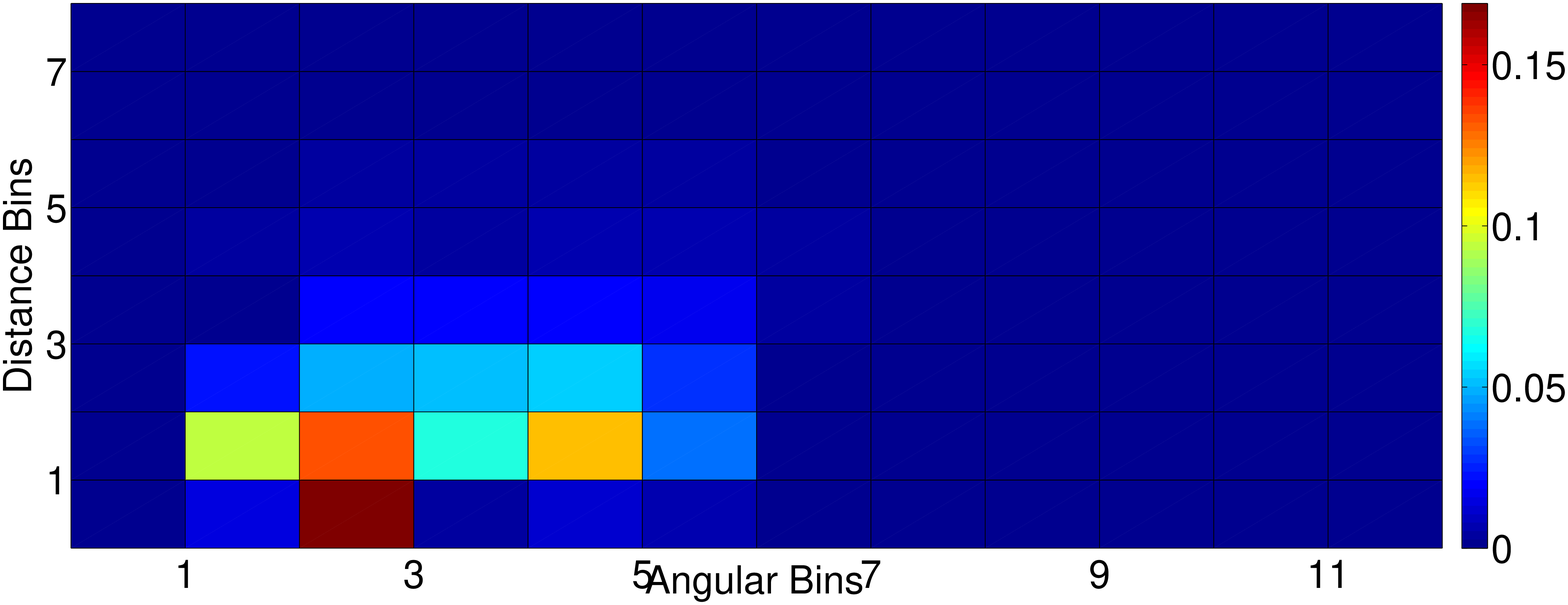}}
  \caption{[Best Viewed in Color] (a) The figure shows the silhouette of a butterfly with a noisy indent in the contour. (b) Uniformly sampled boundary points, $\mathcal{B}^S_P$, from the contour. (c) Output of the Constrained Delaunay Triangulation. The constraint is a simple neighbourhood continuity constraint of the sampled contour points. All the triangles now lie within the object's boundary. (d) \emph{Dense Points} sampled from inside each triangle according to Equations \ref{eq:triangle} and \ref{eq:proportional}. (e) \emph{Sparse Points}, represented by crosses (zoom into the figure), are sampled from the boundary of object's convex hull. Solid Shape Context histogram is computed using log-polar bins at each \emph{Sparse Point}. (f) A visualization of the Solid Shape Context (SSC) histogram.}\label{fig:fullProcedure}
\end{figure*}

Given a shape $S$, we can easily extract its boundary, $\mathcal{B}^S$. We then sample a set of uniformly spaced points, $\mathcal{B}^S_P = \{ \mathcal{B}^S_{p_1}, \mathcal{B}^S_{p_2}, ..., \mathcal{B}^S_{p_{|\mathcal{B}^S_P|}}\}$, that lie on the boundary of the object (Figure \ref{fig:contSample}). A point $\mathcal{B}^S_{p_{i}}$ neighbours just two other points $\mathcal{B}^S_{p_{i-1}}$ and $\mathcal{B}^S_{p_{i+1}}$ (the indices are taken modulo $|\mathcal{B}^S_P|$). We make use of this neighbourbood constraint and perform a Constrained Delaunay Triangulation (CDT) of these points. A CDT ensures that the edges specified as the constraints are retained in the triangulation process \cite{paul1989constrained}. The constraints that we specify are the neighbourhood constraints i.e., our constraint ensures that a point $\mathcal{B}^S_{p_{i}}$ has an edge to its two neighbours $\mathcal{B}^S_{p_{i-1}}$ and $\mathcal{B}^S_{p_{i+1}}$. Once the triangulation is performed, we remove the triangles that lie in the concavities and holes of a shape \cite{shewchuk1996triangle}. This guarantees that the triangles generated from the triangulation lie totally within the object's boundary. For a given set of $N_\mathcal{B}$ points on the boundary, such a Constrained Delaunay Triangulation produces $N_{\mathcal{B}}-2$ triangles, $Tri^S = \{Tri^S_{1}, Tri^S_{2}, ..., Tri^S_{{N_{\mathcal{B}}-2}}\}$. Figure \ref{fig:triang} shows the output of the constrained triangulation. Notice that all the triangles now lie within the object's boundary, especially at the bottom left of the butterfly where there is a noisy indentation.

With the above formulation, it becomes easy to sample a fixed number of points such that they lie completely within the object's boundary. For any triangle, $Tri_{i}^{S}$, with vertices $X$, $Y$, and $Z$, a random point $p$, lying inside the triangle, can be generated using
\begin{equation}\label{eq:triangle}
  p = (1-\sqrt{r_1})X+\sqrt{r_1}(1-r_2)Y+\sqrt{r_1}r_2Z,
\end{equation}
where $r_1$, $r_2 \in [0,1]$ are two random numbers, independent of each other \cite{osada2002shape}. $X$, $Y$, and $Z$, are 2-D vectors containing the $x$ and $y$ coordinates of the three vertices. The set of points lying inside $Tri^S_i$ is given by $\mathcal{P}^{Tri^S_i} = \{p^{Tri^S_i}_{1}, p^{Tri^S_i}_{2}, ..., p^{Tri^S_i}_{|\mathcal{P}^{Tri^S_i}|}\}$. In order to generate $|\mathcal{P}^{Tri^S_i}|$ points lying inside the triangle $Tri^S_i$, all we have to do is generate $|\mathcal{P}^{Tri^S_i}|$ pairs of random numbers $(r_1, r_2)$, from a uniform distribution, and use Equation \ref{eq:triangle} to generate the points.

In order to generate $N_{DP}$ number of uniformly distributed \emph{Dense Points}, lying within the object's boundary (Figure \ref{fig:dense}), we generate $|\mathcal{P}^{Tri^S_i}|$ uniformly distributed points from within each triangle $Tri^S_i, \forall i \in \{1,2, ...,N_{\mathcal{B}}-2\}$, such that $|\mathcal{P}^{Tri^S_i}|$ is proportional to the area of triangle $Tri^S_i$, as given in Equation \ref{eq:proportional}.

\begin{equation}\label{eq:proportional}
  |\mathcal{P}^{Tri^S_i}| = \frac{A_{Tri^S_i}}{\sum_{j=1}^{N_{\mathcal{B}}-2} A_{Tri^S_j}}N_{DP}
\end{equation}

$A_{Tri^S_i}$ is the area of triangle $Tri^S_i$. Therefore,

\begin{equation}\label{eq:sumndp}
N_{DP} =  \sum_{i=1}^{N_{\mathcal{B}}-2} |\mathcal{P}^{Tri^S_i}|
\end{equation}

We have shown how we can generate a fixed number of points from within any random shape, easily. Our method overcomes the two problems that were previously listed in this section. Firstly, we restrict the sampling area to lie within the shape, thus preventing any sampled points from being wasted. Secondly, we sample from a very simple polygon, a triangle, thus making the sampling of uniformly spaced random points quick and easy.

These densely sampled points approximate the interior density of a shape. Our shape descriptor models the shape in its entirety by making use of these \emph{Dense Points}. The SSC shape descriptors are generated at each \emph{Sparse Point} location. A discussion of how to select the location of these \emph{Sparse Points} is given in the following subsection.

\subsection{Sparse Points}
\label{sec:sparse}
A shape $S$ is described using SSC, at $N_{SP}$ locations, where $N_{SP} \ll N_{DP}$. Due to the fact that they are relatively less in number, compared to the \emph{Dense Points}, we call them \emph{Sparse Points}. It is usually enough if we generate the shape descriptors at these sparse set of locations, instead of generating them at each dense point.

The next question that arises is, how and where on the object to localise these sparse points. Ideally, we would like these feature locations to be uniformly spread across the object. We want the descriptors to describe the shape from a varied number of vantage points. One way to do this would be to generate a minimal enclosing rectangle for the object, and uniformly divide the rectangle into $N_{SP}$ number of cells, and mark the centers of these cells as the locations of the \emph{Sparse Points}. However, doing so would not enable us to make use of the continuity constraints while comparing the descriptors between two shapes.

Another approach could be to make use of the uniformly sampled points on the boundary, $\mathcal{B}^S_P$, as the \emph{Sparse Points}, similar to the boundary sampling used in \cite{belongie2002shape} and \cite{ling2007shape}. While this would enable us to make use of the continuity constraints that occur naturally, it would lead us to obtain certain erroneous matches, resulting in increased costs of matching two shapes. These erroneous matches would occur in cases where there are strong indentations in the boundary of the object, such as the examples shown in Figure \ref{fig:similar}. All the descriptors at the landmark points that lie on the indentations will have a vastly different representation of the shape compared to the descriptors that are extracted from a shape without similar (or, any) indentations. Thus, selecting the set of points, $\mathcal{B}^S_P$, as the landmark points does not seem to be a good idea.

To retain the advantage of the continuity constraints and still have \emph{Sparse Points} that are independent of the indentations in the contour,  we propose to sample the feature point locations along the boundary of the convex hull of the shape. Sampling landmark points along the convex hull gives us many advantages. Since the convex hull encloses the object completely, we retain the advantage of having the descriptors describe the object from various vantage points. Secondly, sampling from the convex hull gives us larger insensitivity to boundary perturbations. Along with the densely sampled points, which help in handling noisy indentations, sampling along the convex hull also prevents such indentations from unnecessarily affecting the landmark selection. Thirdly, sampling along the convex hull gives a better rotation invariance to the descriptor. Rotation invariance is usually added to the descriptor by tangent angle normalization. Calculating the tangent angle on the boundary of the convex hull gives better invariance to rotation than when the normalization angle is calculated using the tangent on a noisy contour. Such unwanted perturbations in the boundary would randomly skew the tangents along the boundary, thus causing large amounts of noise to be added during the angle normalization step. Finally, using the convex hull can be an advantage even when the shapes are highly concave. Since our sampling procedure ensures that the sampled points always lie inside the shape boundary, the absence of dense points in the concavities of the shape help capture the concave properties of the shape. Ex: The characteristic property of a horseshoe is its concavity, and this property is captured in our shape descriptor by means of zero height bins (see the following subsection).

Due to the above mentioned advantages, similar to $\mathcal{B}^S_P$, we obtain the set of \emph{Sparse Points}, $\mathcal{SP}^S = \{\mathcal{SP}^S_1, \mathcal{SP}^S_2, ..., \mathcal{SP}^S_{N_{SP}}\}$, for shape $S$, sampled along the convex hull of the shape, and compute the SSC descriptor at each of these points. Figure \ref{fig:sparseCH} shows how the sparse points are sampled from the object's convex hull. Notice the insensitivity of the \emph{Sparse Points} to the indentations in the butterfly's boundary.

Now that we have a set of \emph{Dense Points} that can be used to model the interior of the shape, and a set of \emph{Sparse Points} to represent the shape, we go on to describe how we generate our SSC descriptor using both these sets of points.

\subsection{Solid Shape Context Descriptor}
\label{sec:sscdesc}

At each sparse point $\mathcal{SP}^S_i$, we generate a 2-D histogram
\begin{equation}
\label{eq:hist}
  \mathcal{H}^S_i(k) = \#\{\mathcal{DP}^S_j : \mathcal{DP}^S_j \in bin(k)\},
\end{equation}
where, $\mathcal{DP}^S_j$ is the $j$-th \emph{Dense Point}, $k$ is the bin number, $i \in \{1,2,...,N_{SP}\}$, and $j \in \{1,2,...,N_{DP}\}$. Similar to \cite{ling2007shape}, we use $8$ distance bins and $12$ angular binsp to generate the log-polar histogram. We use the Euclidean distance and Euclidean angle (similar to \cite{belongie2002shape}) to calculate the distance, and angle, between a \emph{Sparse Point} and a \emph{Dense Point}. A given shape $S$ can now be described by a set of histograms, $\mathcal{SSC}^{S} = \{\mathcal{H}^S_1,\mathcal{H}^S_2, ...,\mathcal{H}^S_{N_{SP}}\}$. Similar to SC and IDSC, SSC is inherently invariant to translation. It can be made invariant to rotations, and scale, by tangent normalization, and mean distance normalization, respectively. Figure \ref{fig:sscHist} gives a visualization of the SSC histogram for one of the sparse locations.

Given two shapes $S_1$ and $S_2$, matching them now boils down to matching their respective histogram sets, $\mathcal{SSC}^{S_1}$ and $\mathcal{SSC}^{S_2}$. The goal of the matching stage is to find a mapping function $\phi$, which minimizes the cost of mapping the histogram $\mathcal{H}^{S_1}_{i}$ to $\mathcal{H}^{S_2}_{\phi(i)}$. The total cost of matching shape $S_1$ to shape $S_2$ is given by

\begin{equation}\label{eq:cost}
  \Psi_{SSC}(S_1,S_2) = \sum_{i=1}^{N_{SP}^{S_1}} \psi(\mathcal{H}^{S_1}_{i}, \mathcal{H}^{S_2}_{\phi(i)}).
\end{equation}

The distance between two histograms, $\psi(\mathcal{H}^{S_1}_{i}, \mathcal{H}^{S_2}_{\phi(i)})$, is defined by the $\chi^2$ test statistic. If the distance between the two histograms is greater than an acceptable threshold $\tau (=0.6)$, we set the distance to equal $\tau$, and set $\phi(i)$ to $0$, which means to say that we were not able to find a suitable match for $\mathcal{H}^{S_1}_{i}$, in shape $S_2$. Similar to \cite{ling2007shape}, we use a dynamic programming scheme to match the two sets of histograms.

Finally, the true cost between the two shapes $S_1$ and $S_2$ can be computed as
\begin{equation}\label{eq:fusion}
  \Psi(S_1,S_2) = \min (\Psi_{IDSC}, \alpha\Psi_{SSC}),
\end{equation}
where $\Psi_{IDSC}$ is the cost of matching the two shapes using the standard IDSC method \cite{ling2007shape}, $\Psi_{SSC}$ is given by Equation \ref{eq:cost}, and $\alpha$ is a normalization constant, which is used to normalize the two costs. Fusing two or more costs to obtain the smallest cost has become popular in the recent past and is used in \cite{ling2010balancing}, \cite{temlyakov2010two} and \cite{Hu20123222}.

Figure \ref{fig:fullProcedure} illustrates all the steps involved in the generation of the SSC shape descriptor. In the next section, we demonstrate the effectiveness of our SSC descriptor using the results obtained from our experiments.

\section{Experiments and Results}
\label{sec:exp}

\begin{figure}[!th]
\centering
\subfloat{\includegraphics[trim=0.1cm 0cm 0.12cm 0cm, clip=true, width=1.3cm, height=1.5cm]{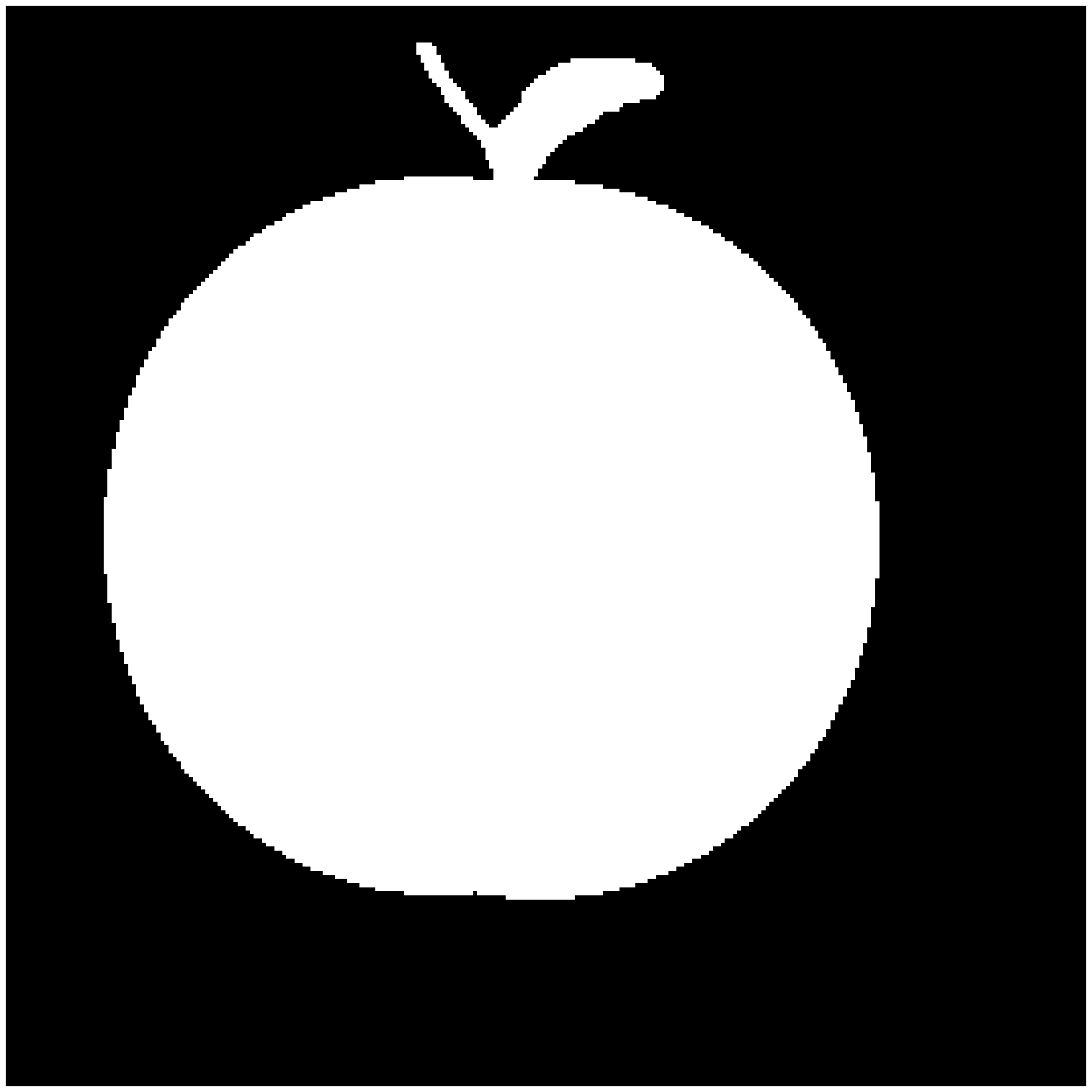}}
\subfloat{\includegraphics[trim=0.1cm 0cm 0.12cm 0cm, clip=true, width=1.3cm, height=1.5cm]{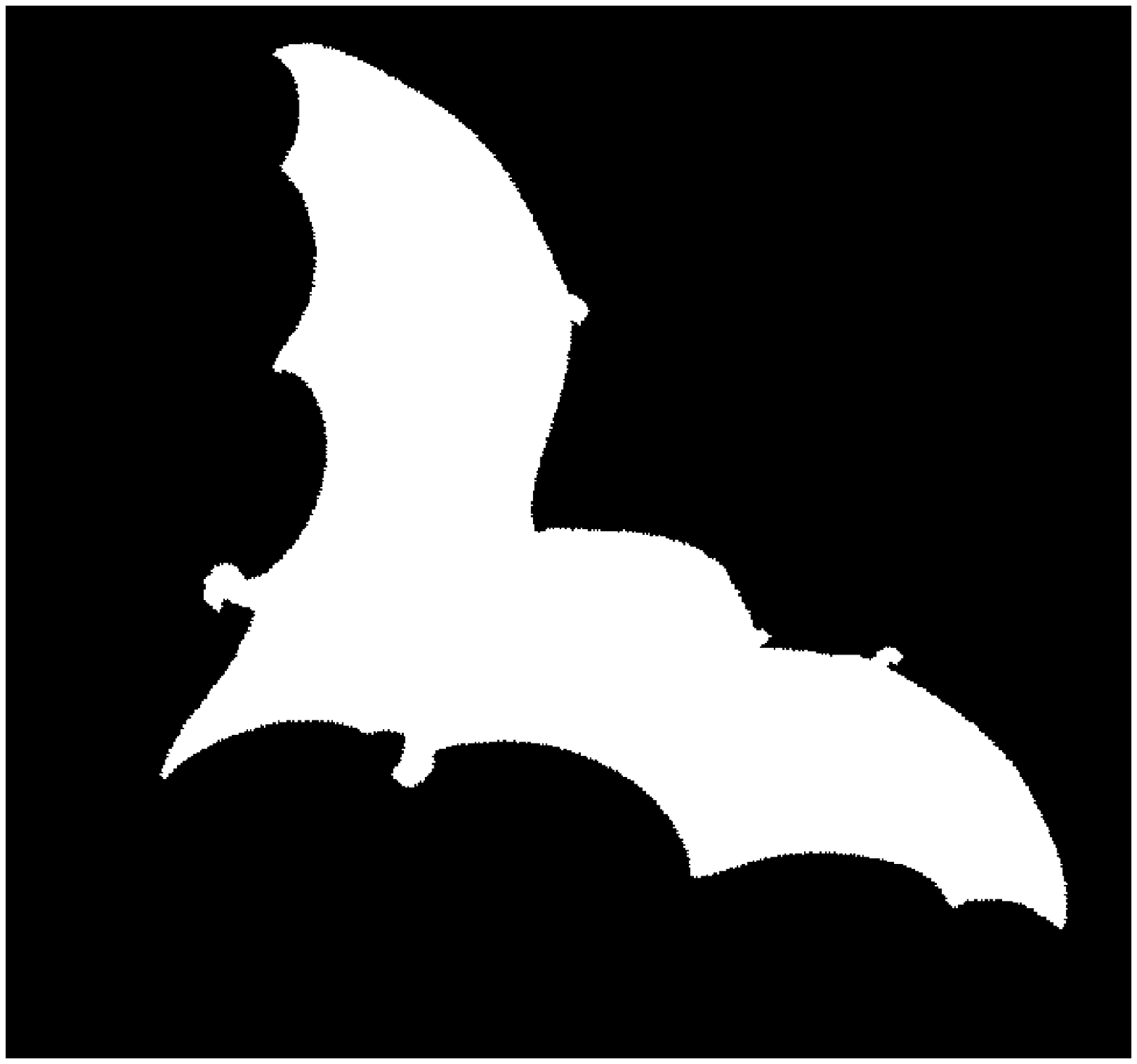}}
\subfloat{\includegraphics[trim=0.1cm 0cm 0.12cm 0cm, clip=true, width=1.3cm, height=1.5cm]{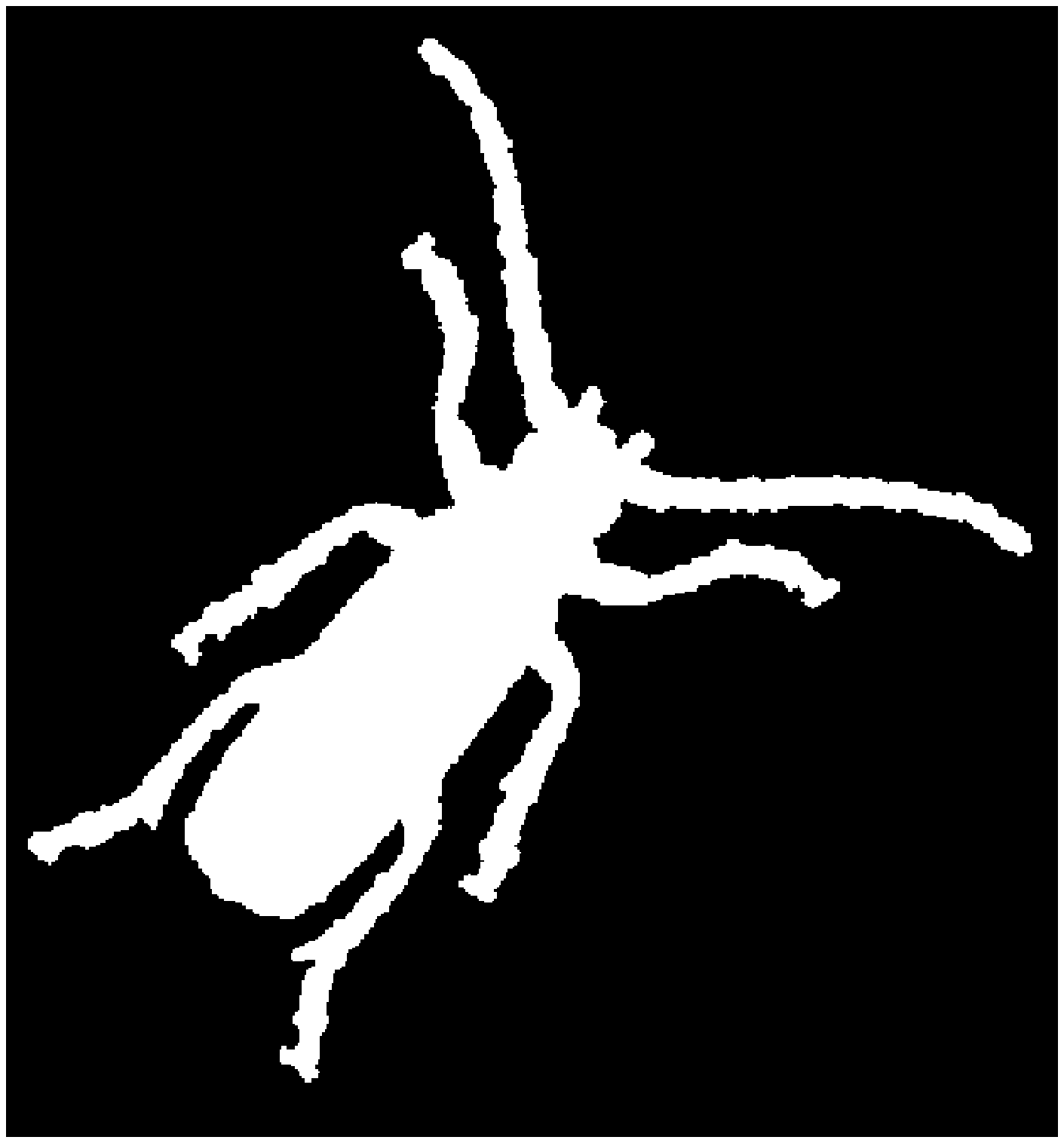}}
\subfloat{\includegraphics[trim=0.1cm 0cm 0.12cm 0cm, clip=true, width=1.3cm, height=1.5cm]{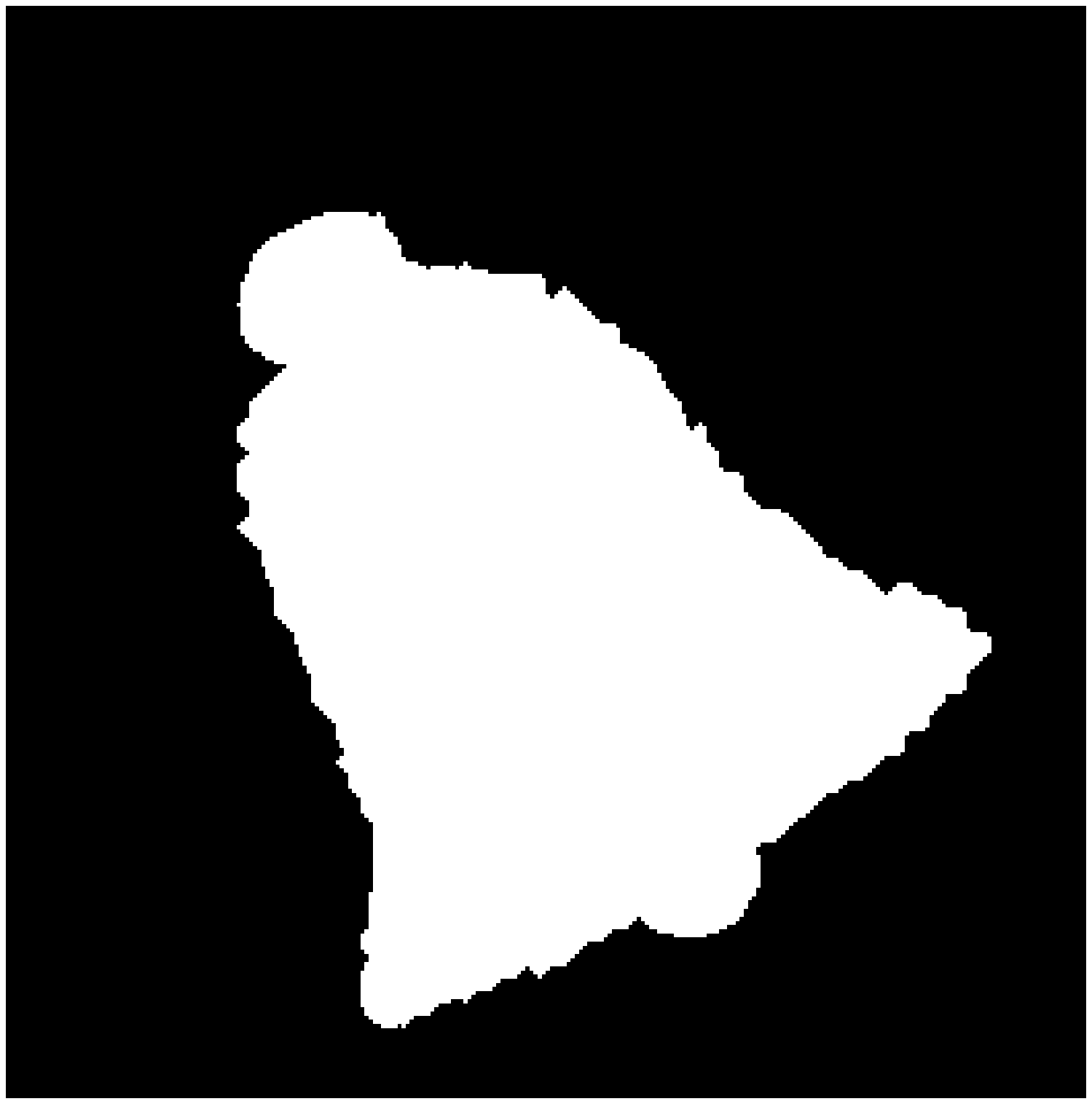}}
\subfloat{\includegraphics[trim=0.1cm 0cm 0.12cm 0cm, clip=true, width=1.3cm, height=1.5cm]{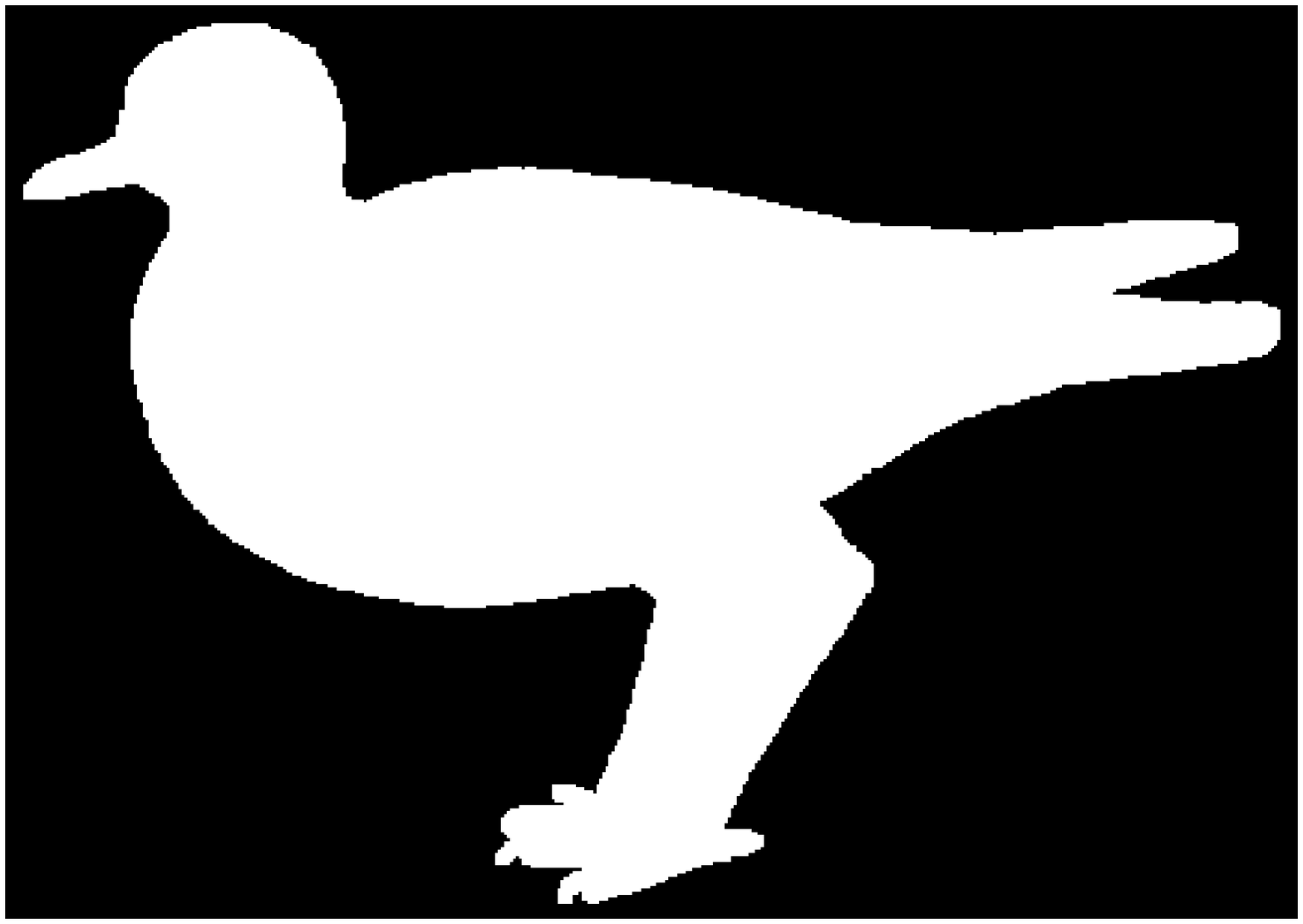}}
\subfloat{\includegraphics[trim=0.1cm 0cm 0.12cm 0cm, clip=true, width=1.3cm, height=1.5cm]{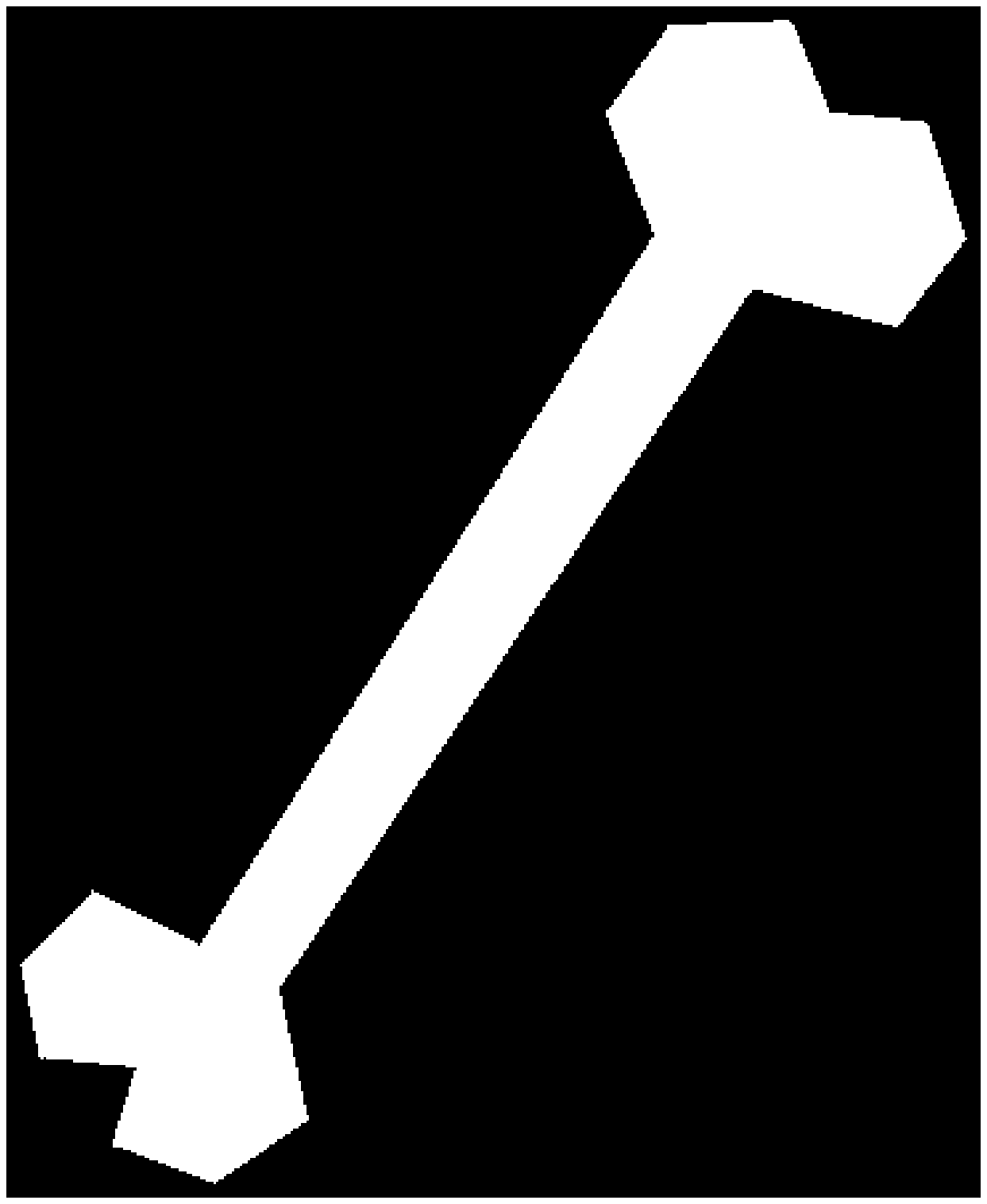}}
\subfloat{\includegraphics[trim=0.1cm 0cm 0.12cm 0cm, clip=true, width=1.3cm, height=1.5cm]{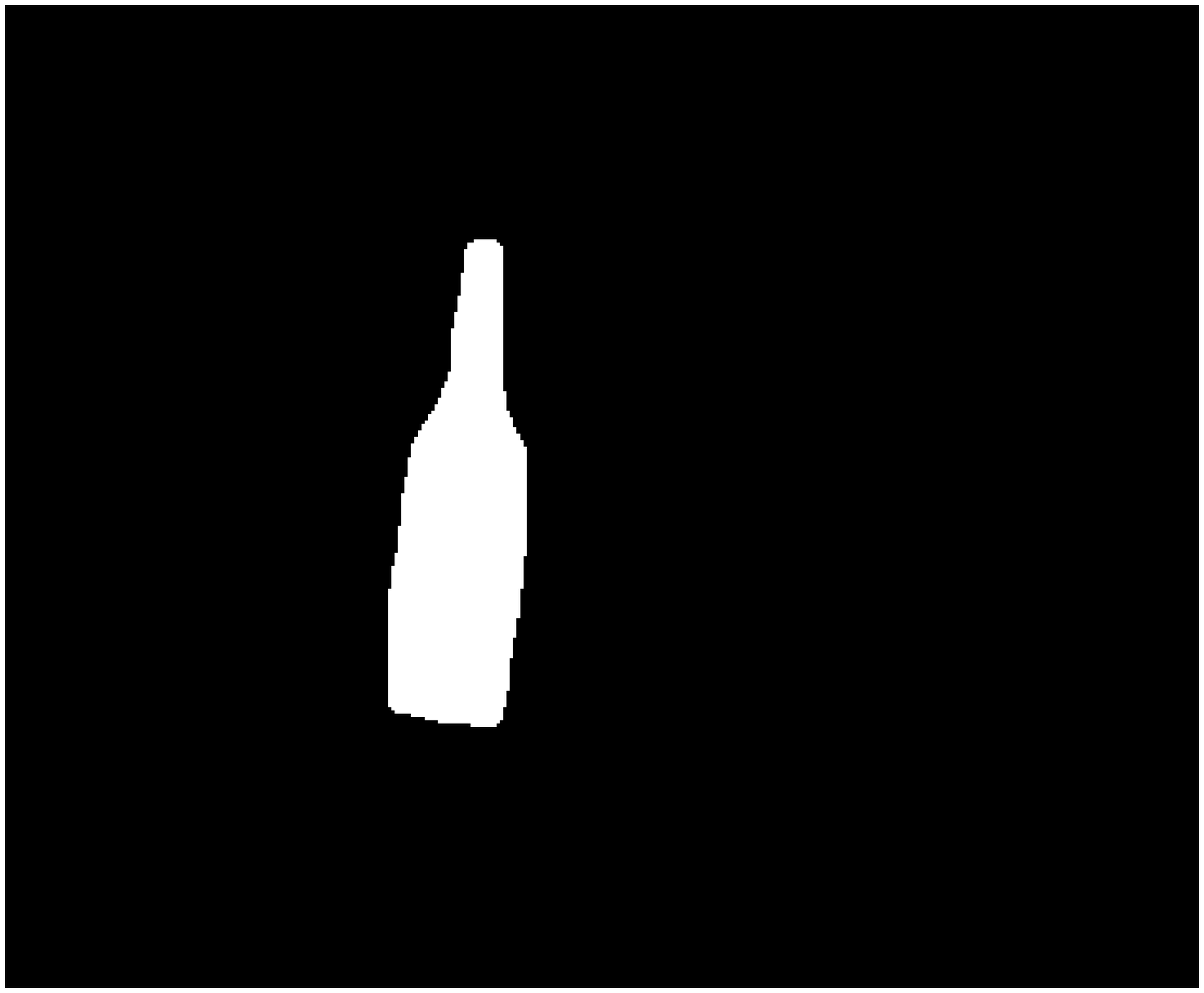}}
\subfloat{\includegraphics[trim=0.1cm 0cm 0.12cm 0cm, clip=true, width=1.3cm, height=1.5cm]{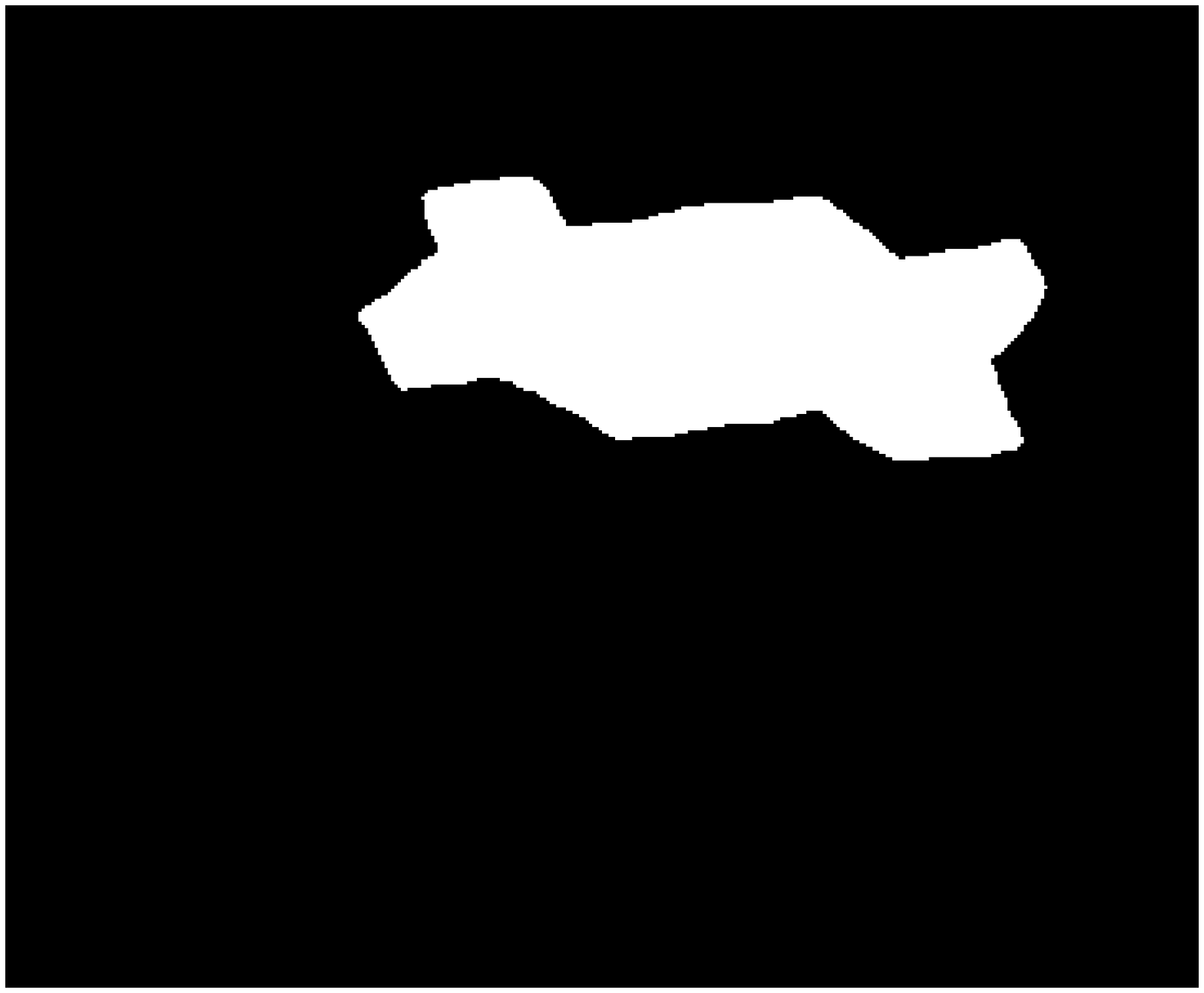}}
\subfloat{\includegraphics[trim=0.1cm 0cm 0.12cm 0cm, clip=true, width=1.3cm, height=1.5cm]{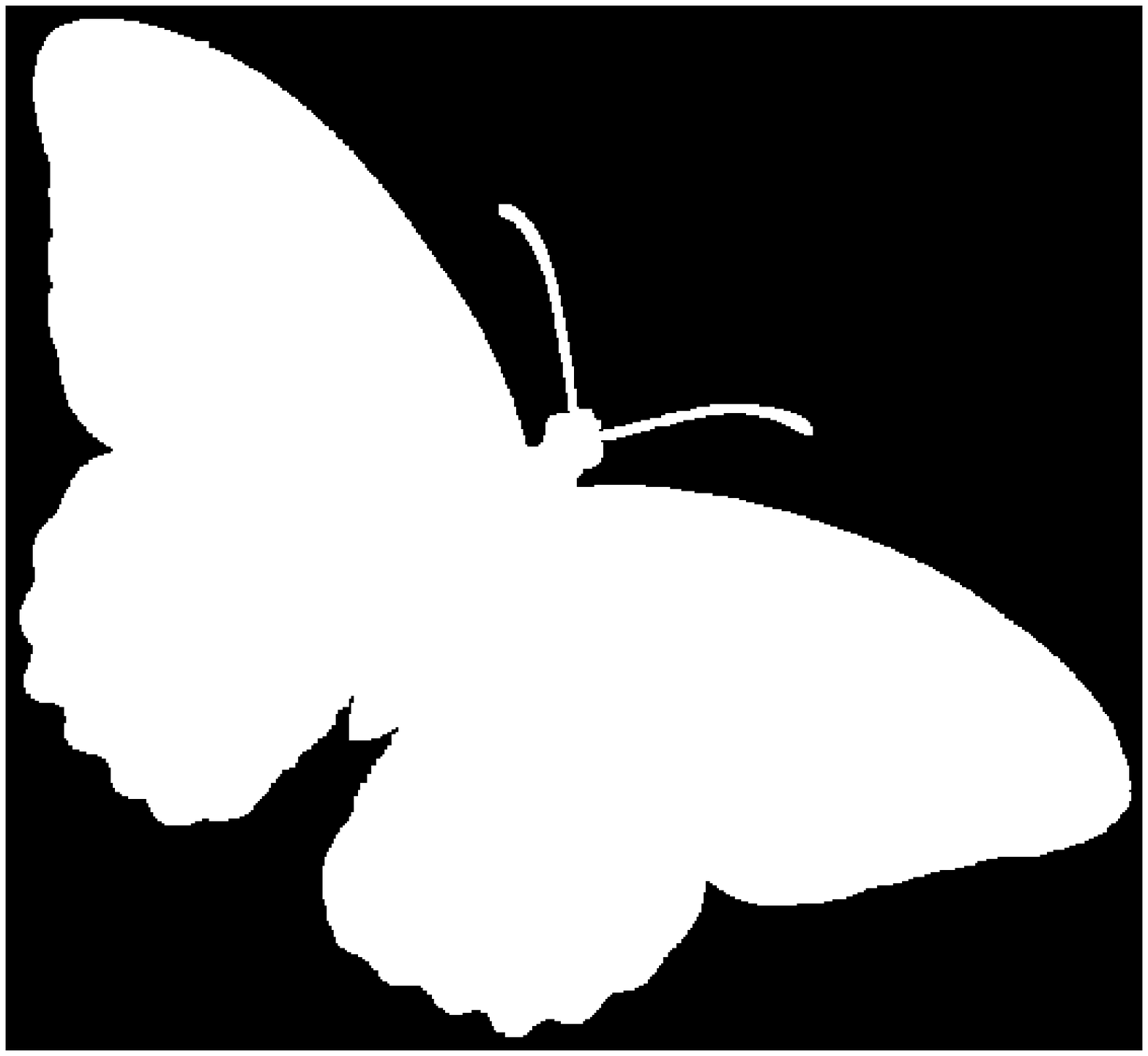}}
\subfloat{\includegraphics[trim=0.1cm 0cm 0.12cm 0cm, clip=true, width=1.3cm, height=1.5cm]{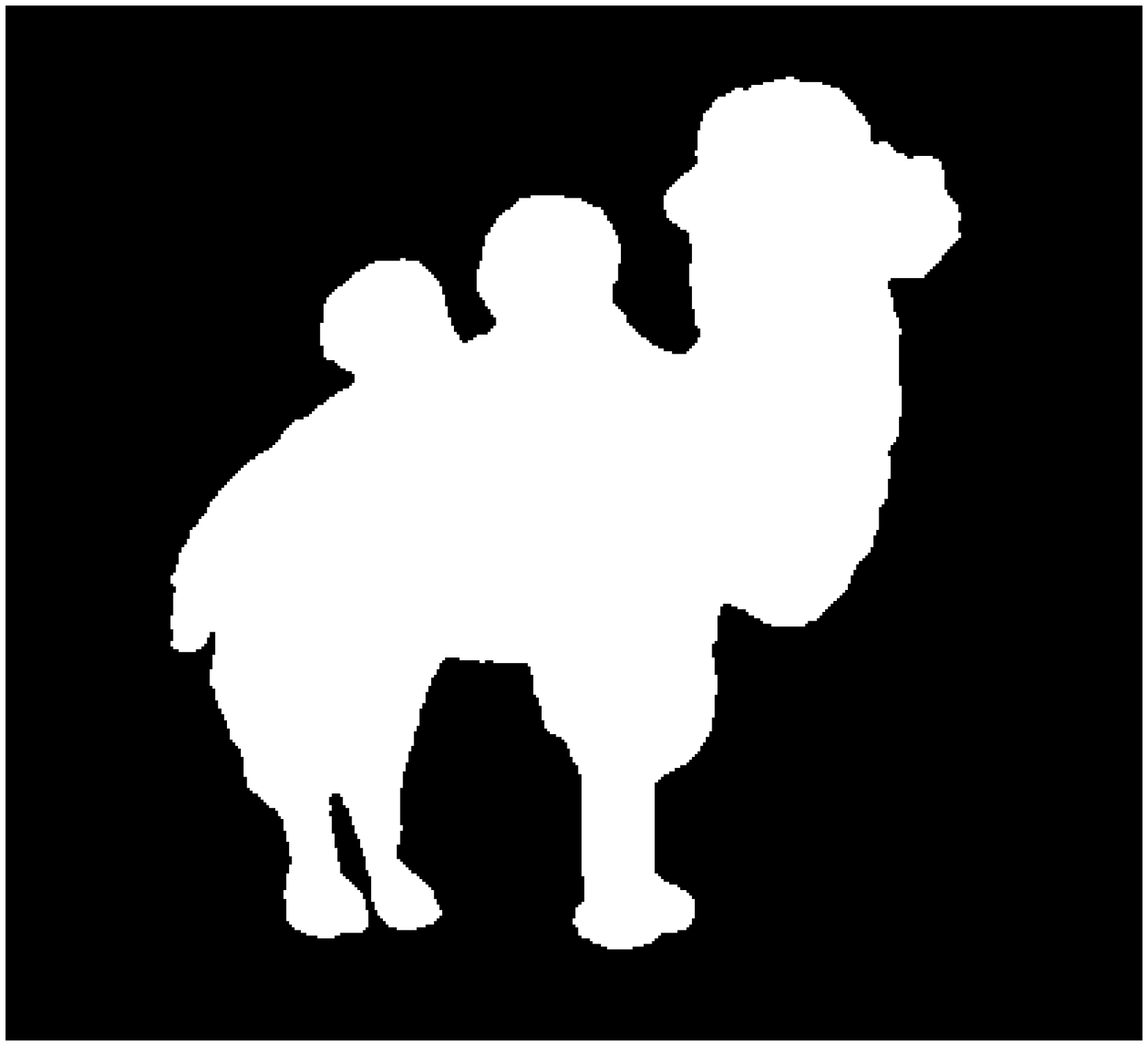}}\\

\subfloat{\includegraphics[trim=0.1cm 0cm 0.12cm 0cm, clip=true, width=1.3cm, height=1.5cm]{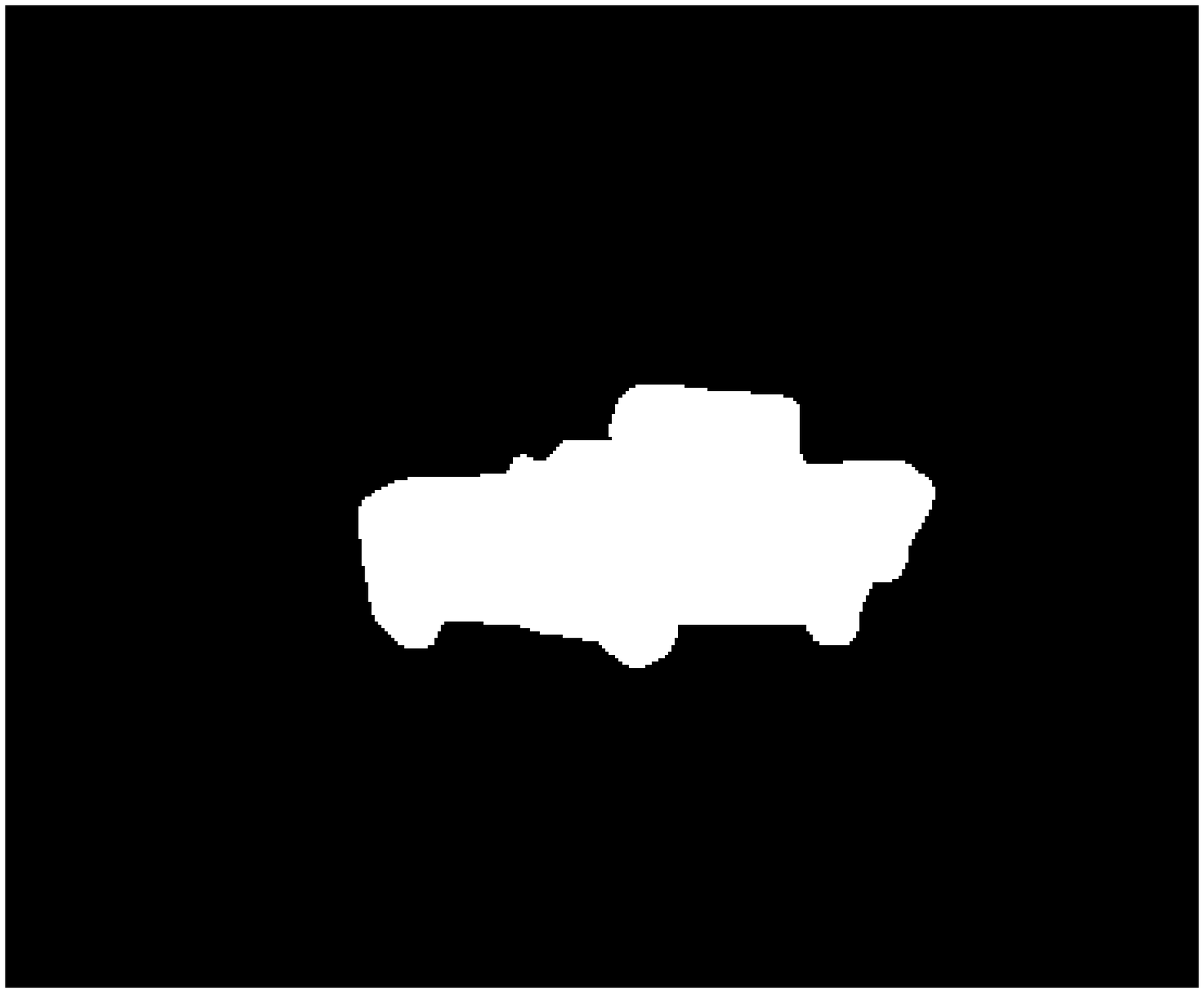}}
\subfloat{\includegraphics[trim=0.1cm 0cm 0.12cm 0cm, clip=true, width=1.3cm, height=1.5cm]{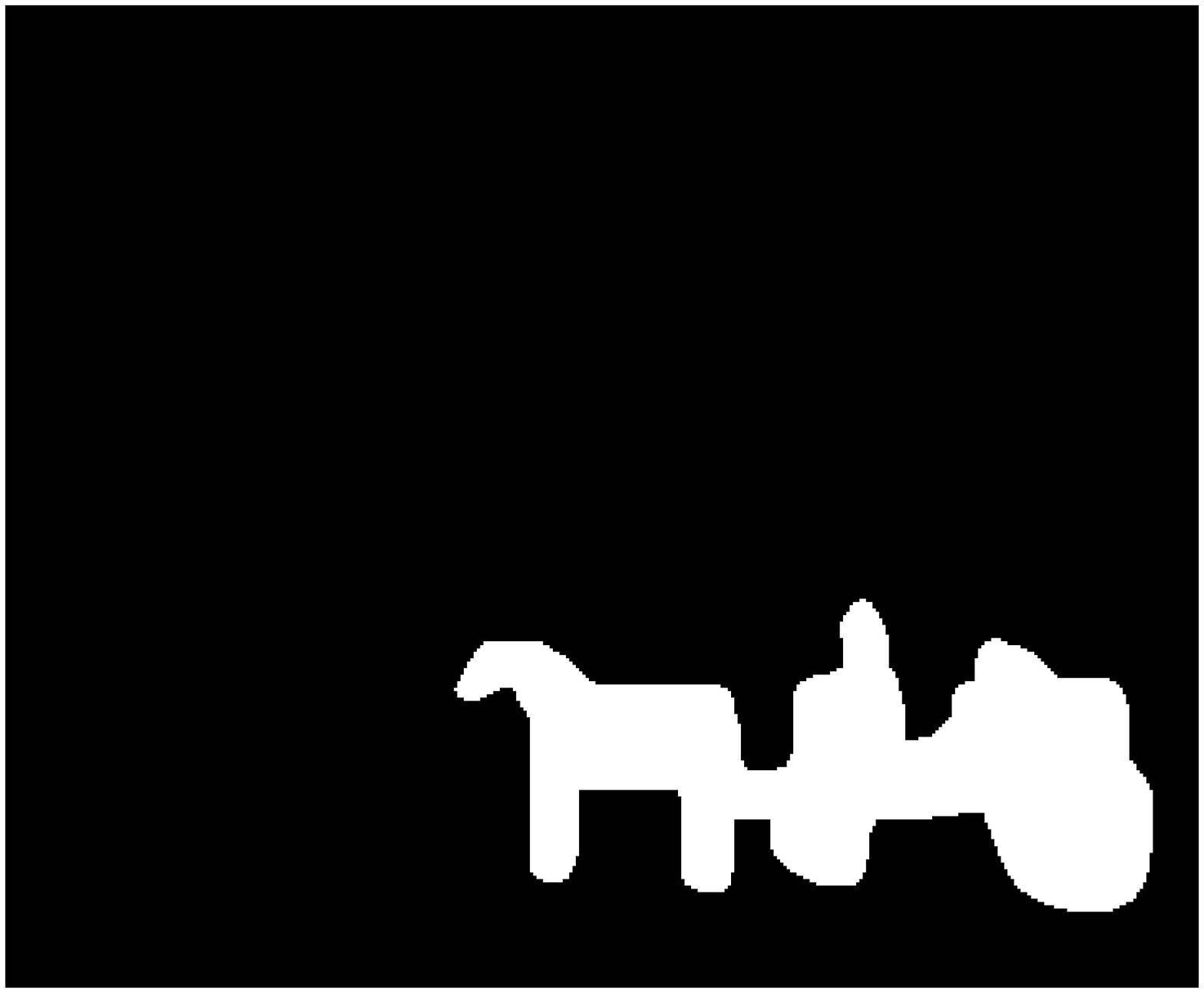}}
\subfloat{\includegraphics[trim=0.1cm 0cm 0.12cm 0cm, clip=true, width=1.3cm, height=1.5cm]{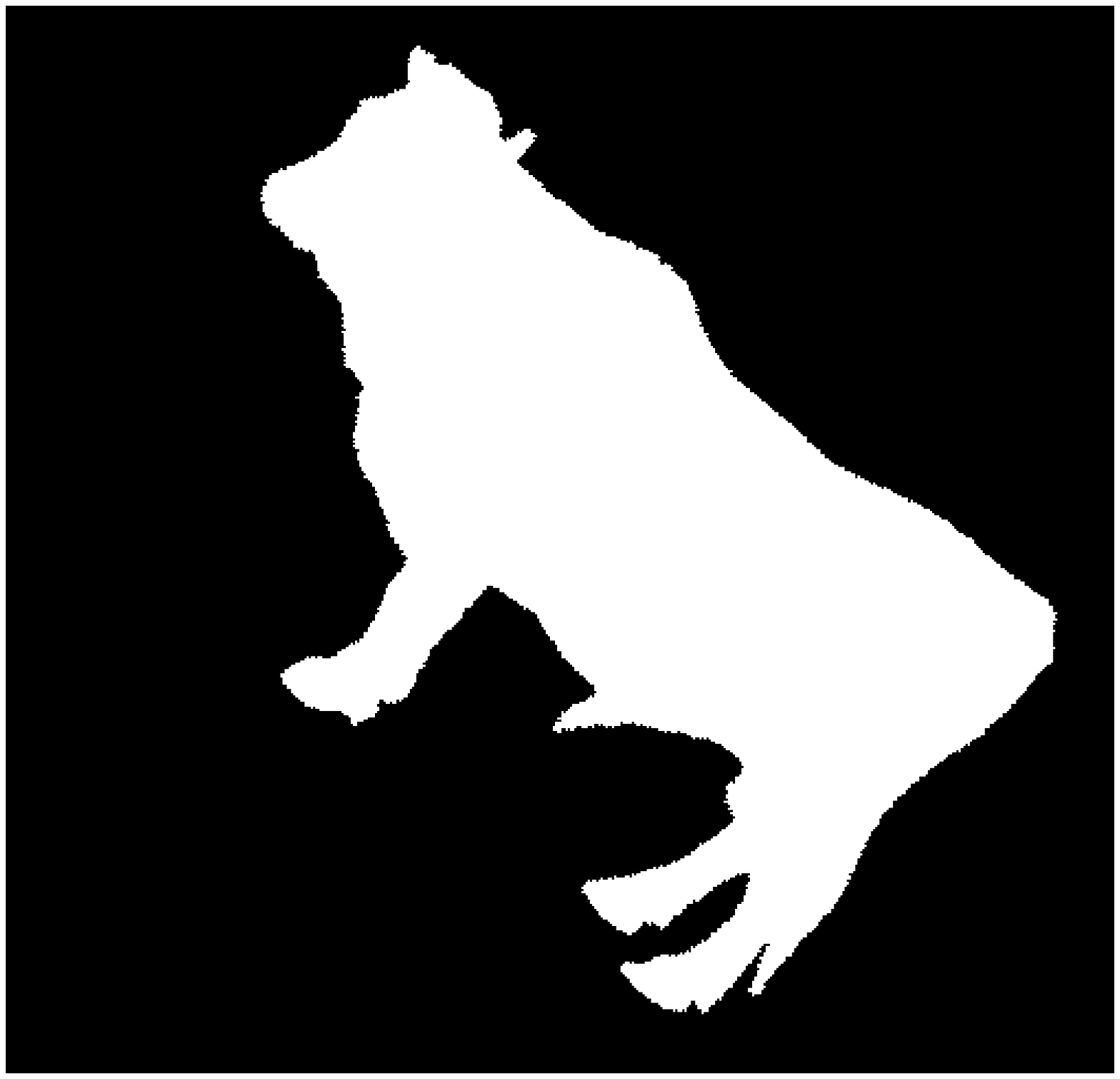}}
\subfloat{\includegraphics[trim=0.1cm 0cm 0.12cm 0cm, clip=true, width=1.3cm, height=1.5cm]{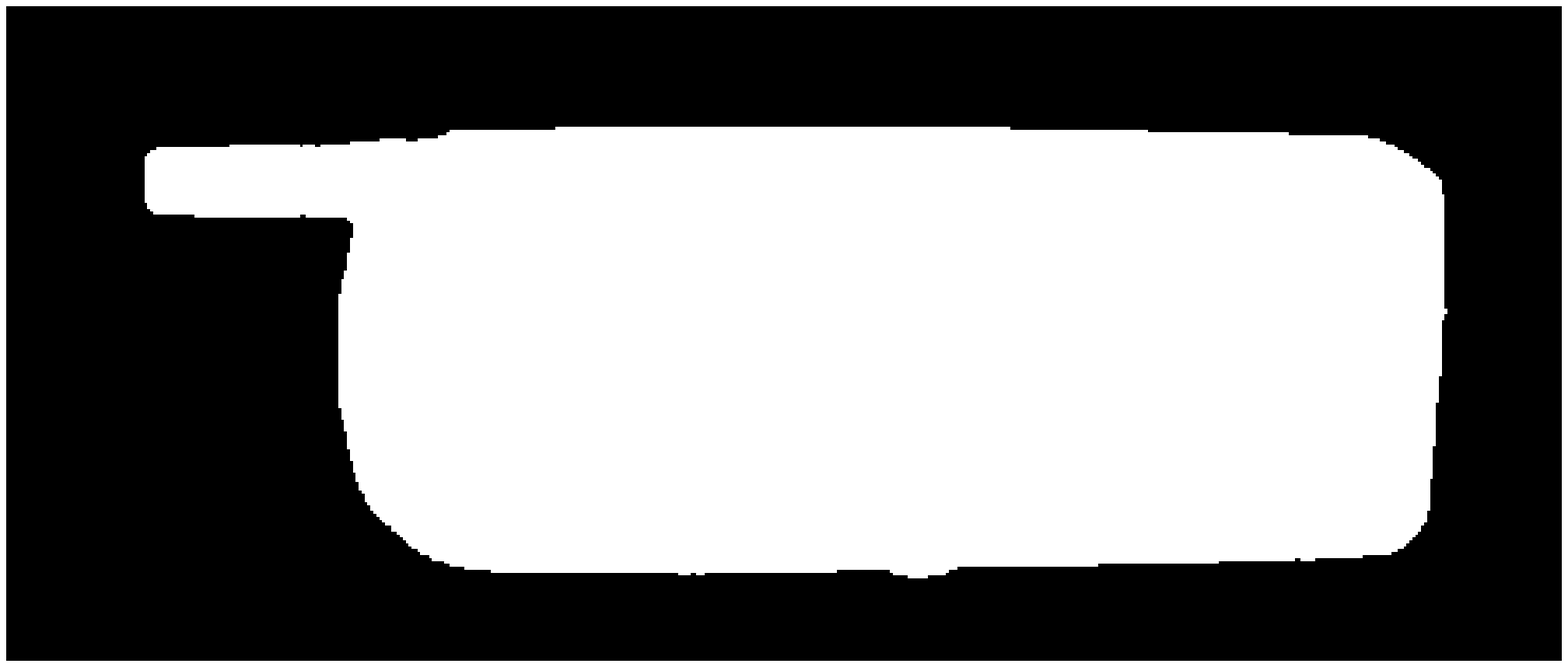}}
\subfloat{\includegraphics[trim=0.1cm 0cm 0.12cm 0cm, clip=true, width=1.3cm, height=1.5cm]{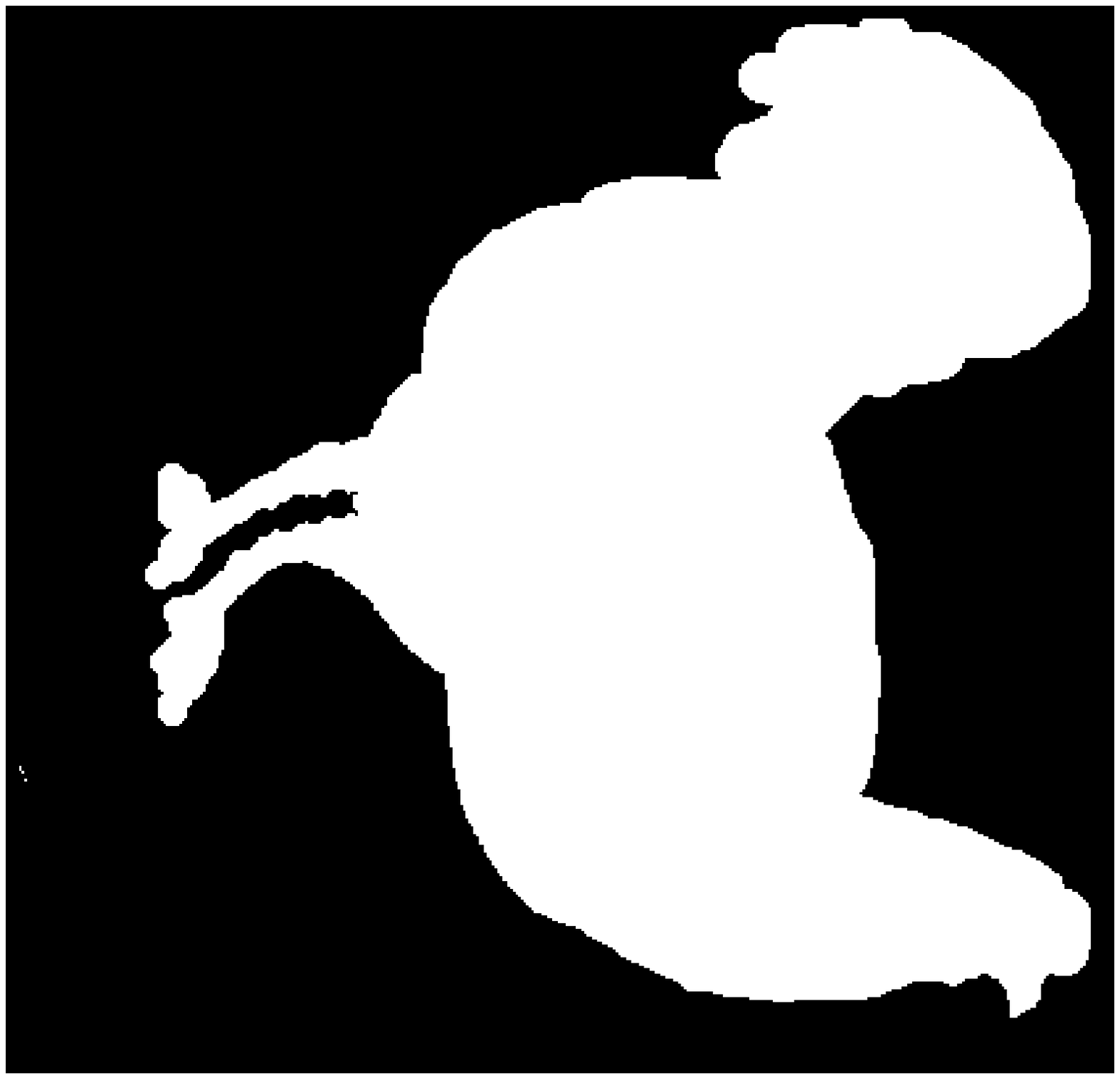}}
\subfloat{\includegraphics[trim=0.1cm 0cm 0.12cm 0cm, clip=true, width=1.3cm, height=1.5cm]{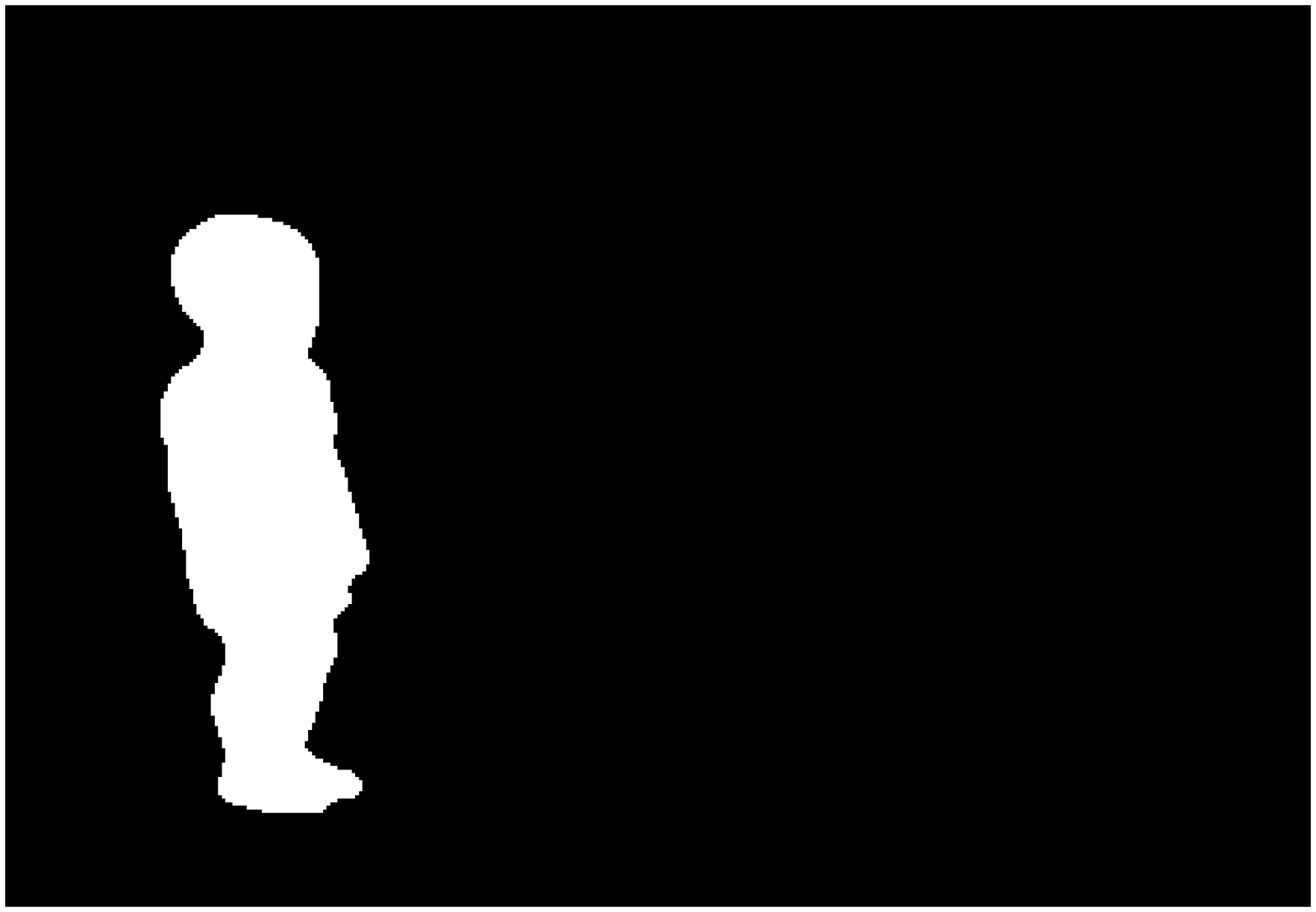}}
\subfloat{\includegraphics[trim=0.1cm 0cm 0.12cm 0cm, clip=true, width=1.3cm, height=1.5cm]{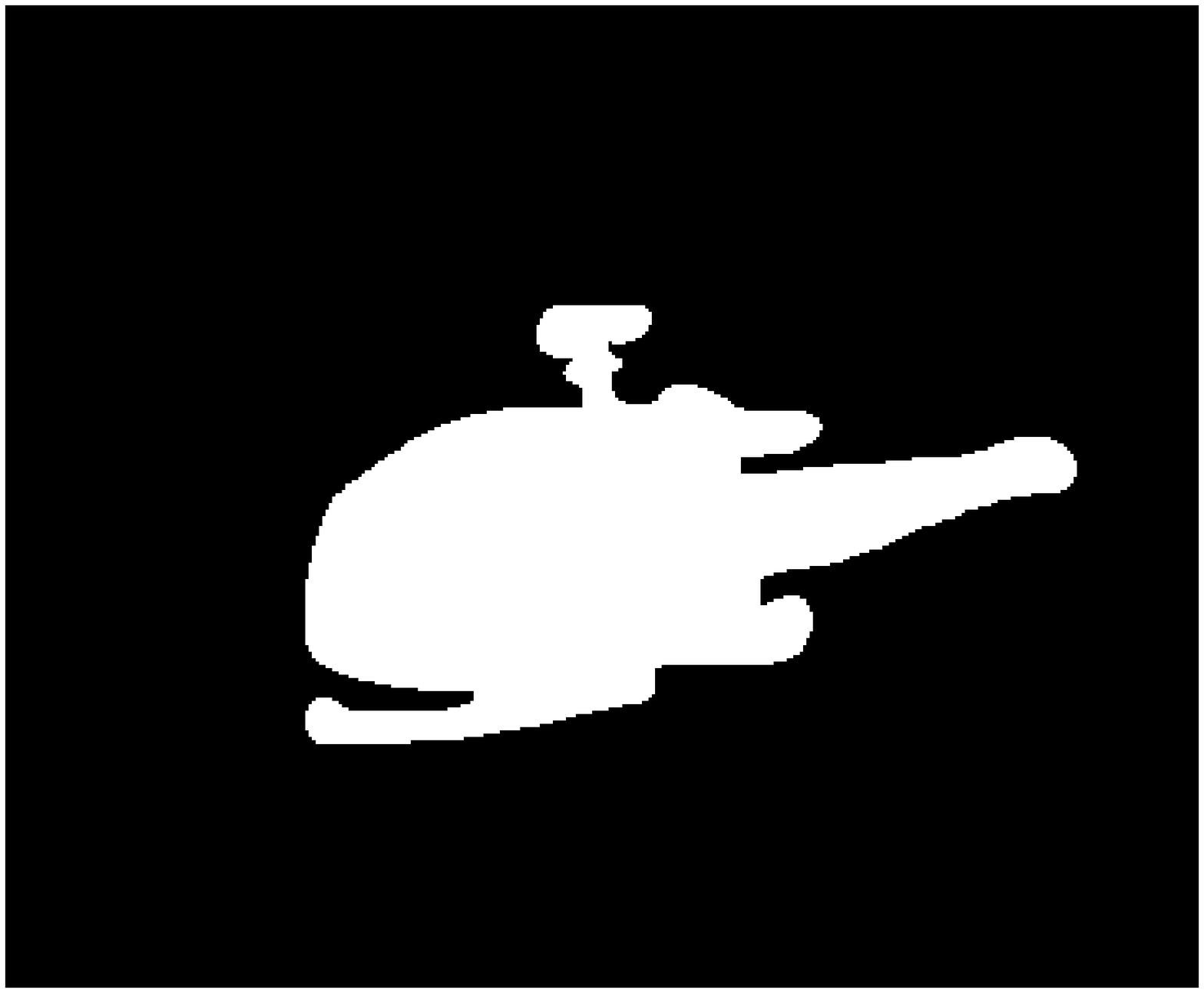}}
\subfloat{\includegraphics[trim=0.1cm 0cm 0.12cm 0cm, clip=true, width=1.3cm, height=1.5cm]{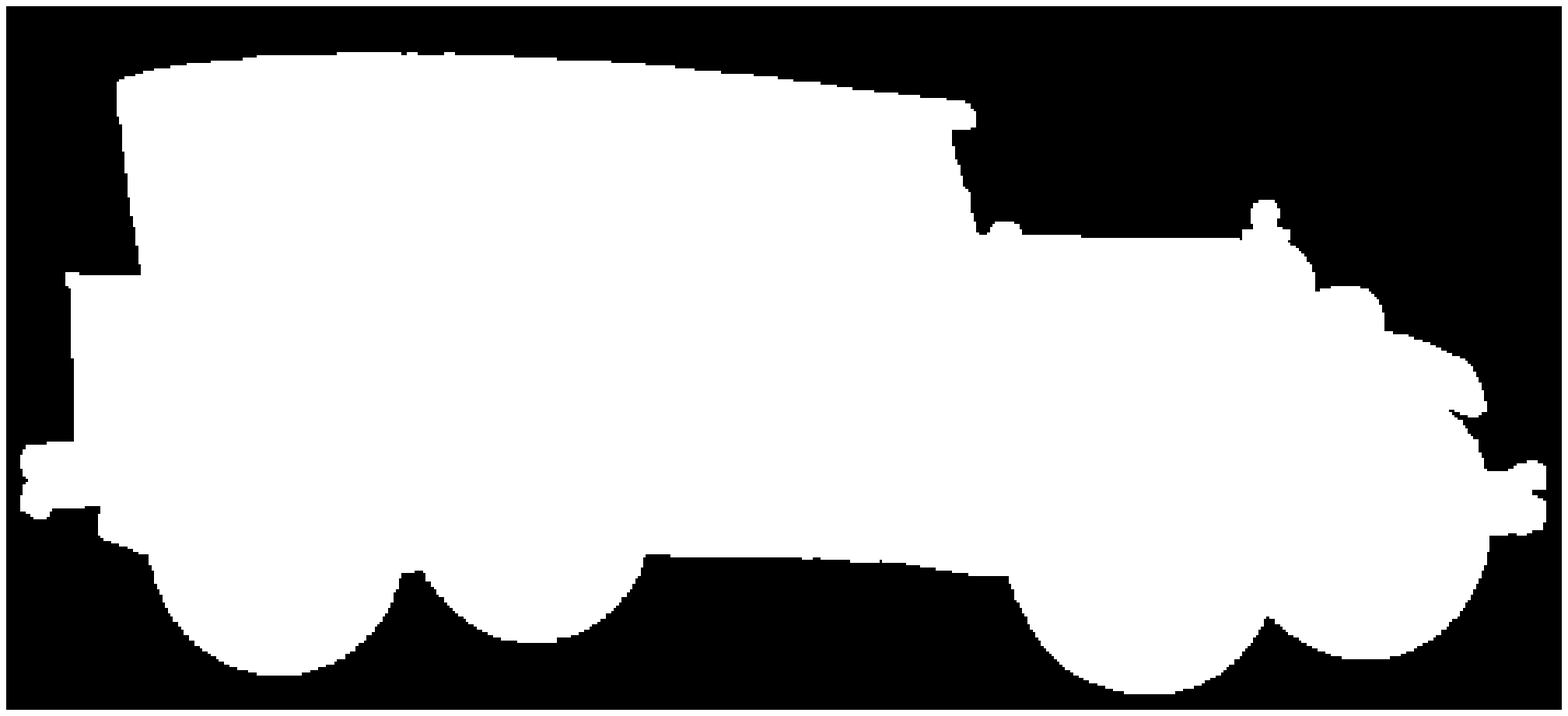}}
\subfloat{\includegraphics[trim=0.1cm 0cm 0.12cm 0cm, clip=true, width=1.3cm, height=1.5cm]{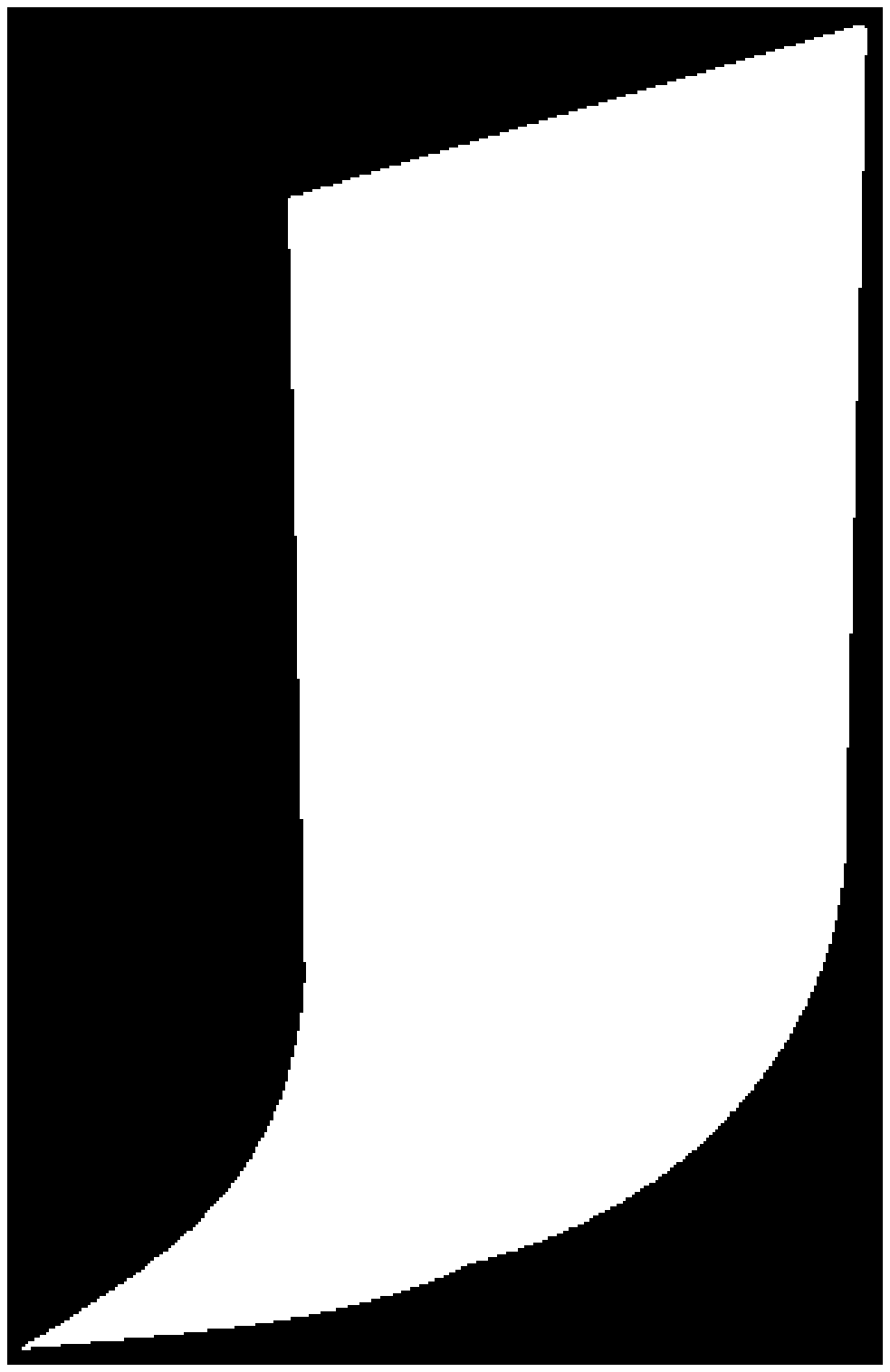}}
\subfloat{\includegraphics[trim=0.1cm 0cm 0.12cm 0cm, clip=true, width=1.3cm, height=1.5cm]{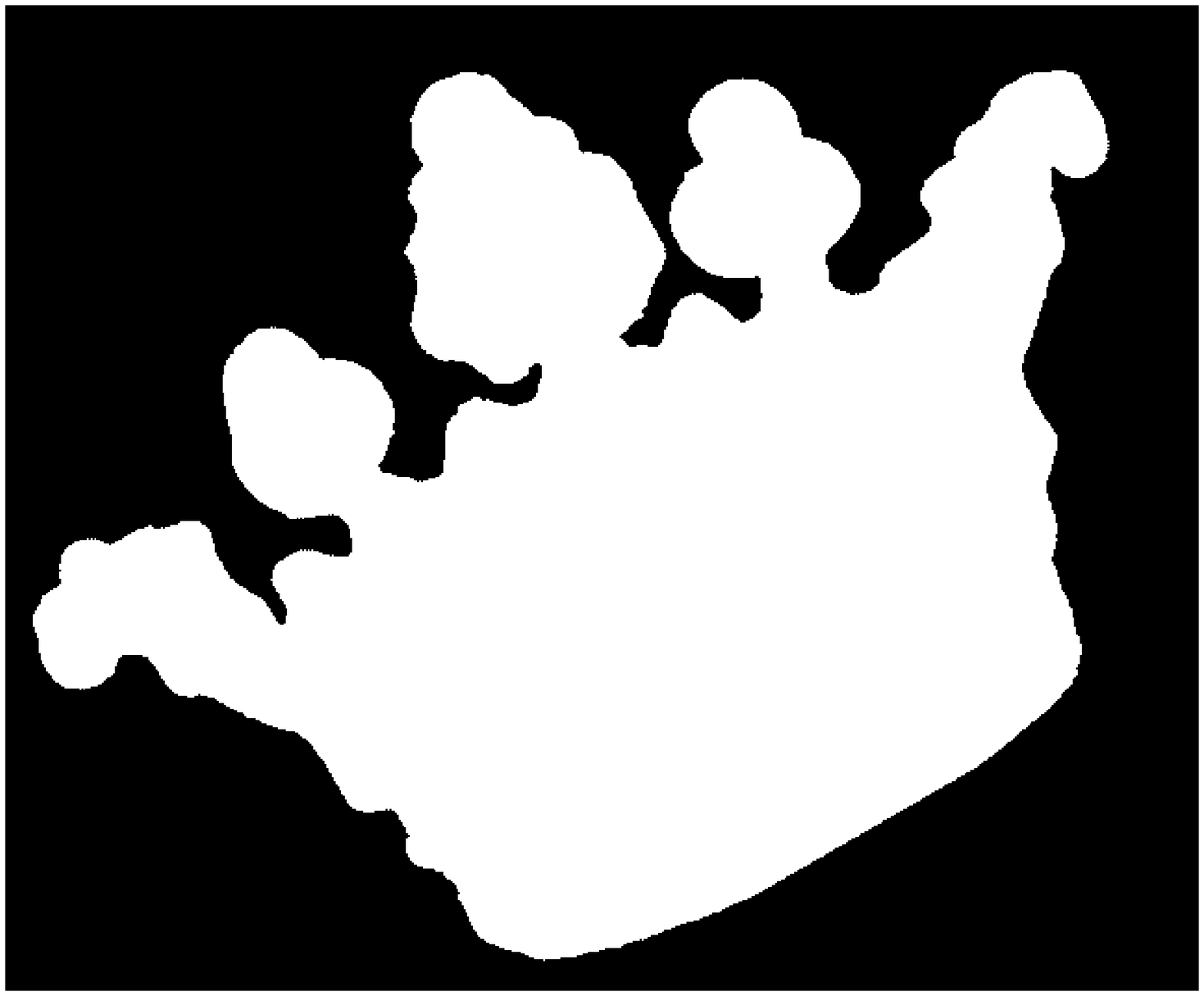}}\\

\subfloat{\includegraphics[trim=0.1cm 0cm 0.12cm 0cm, clip=true, width=1.3cm, height=1.5cm]{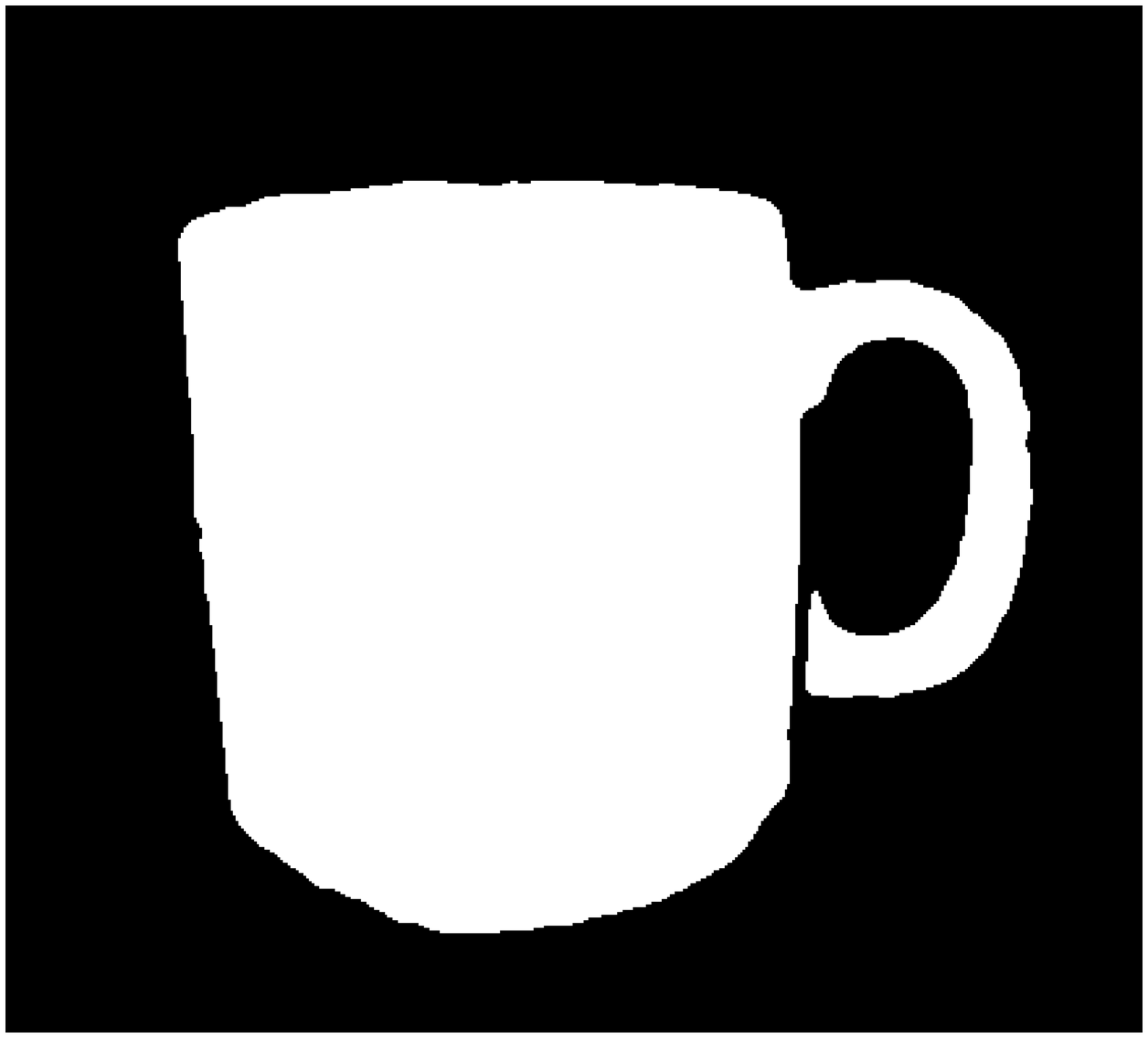}}
\subfloat{\includegraphics[trim=0.1cm 0cm 0.12cm 0cm, clip=true, width=1.3cm, height=1.5cm]{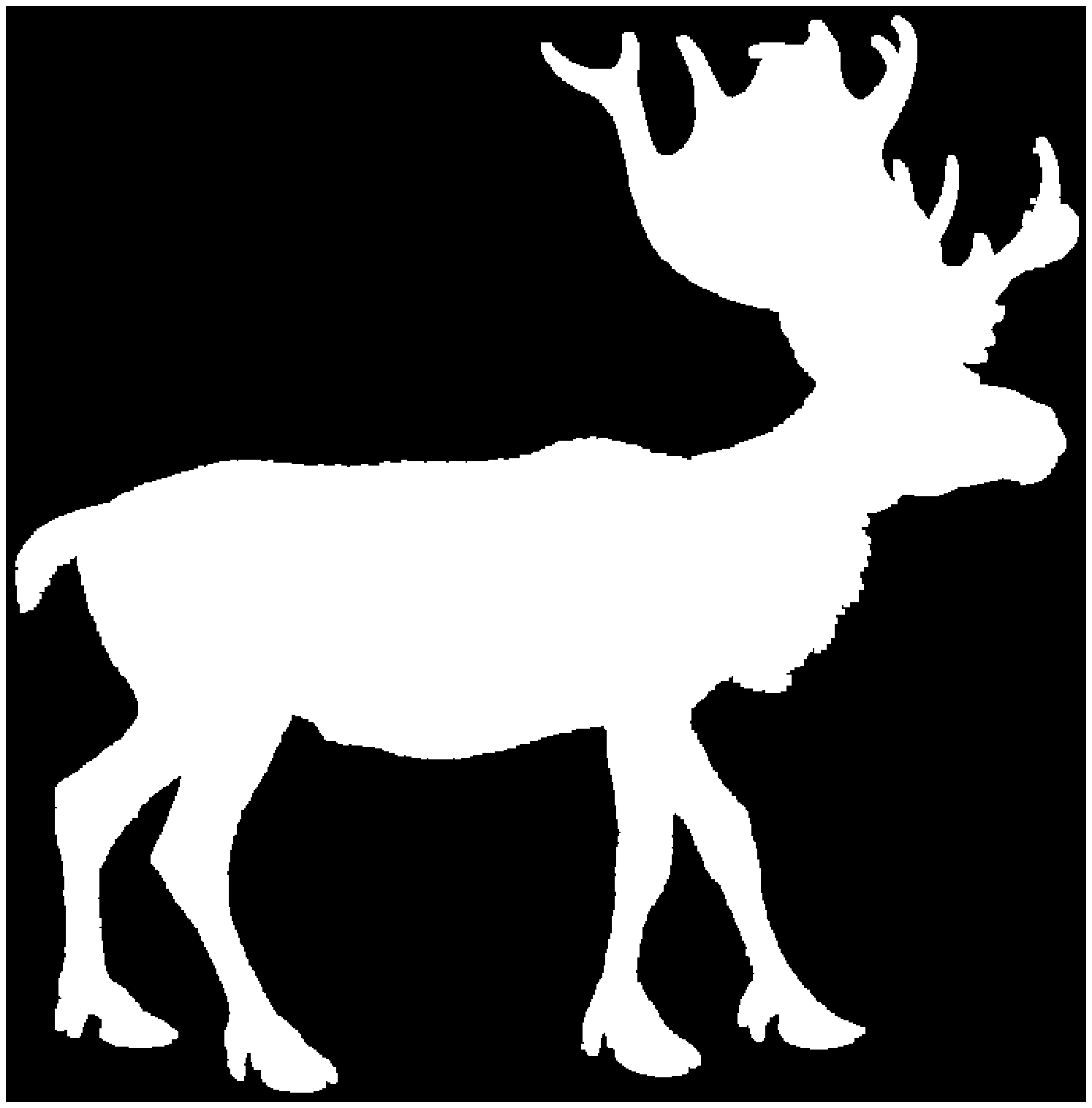}}
\subfloat{\includegraphics[trim=0.1cm 0cm 0.12cm 0cm, clip=true, width=1.3cm, height=1.5cm]{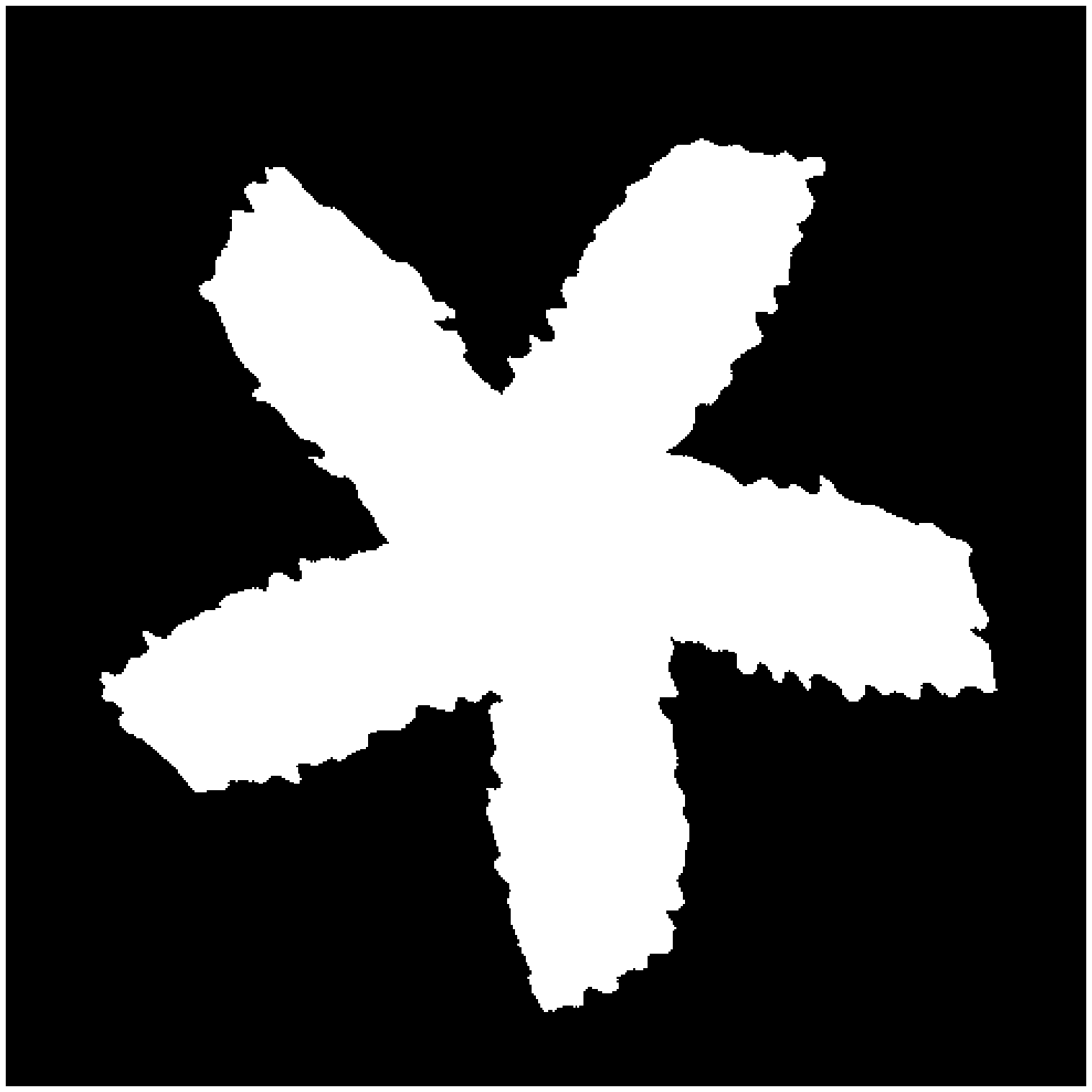}}
\subfloat{\includegraphics[trim=0.1cm 0cm 0.12cm 0cm, clip=true, width=1.3cm, height=1.5cm]{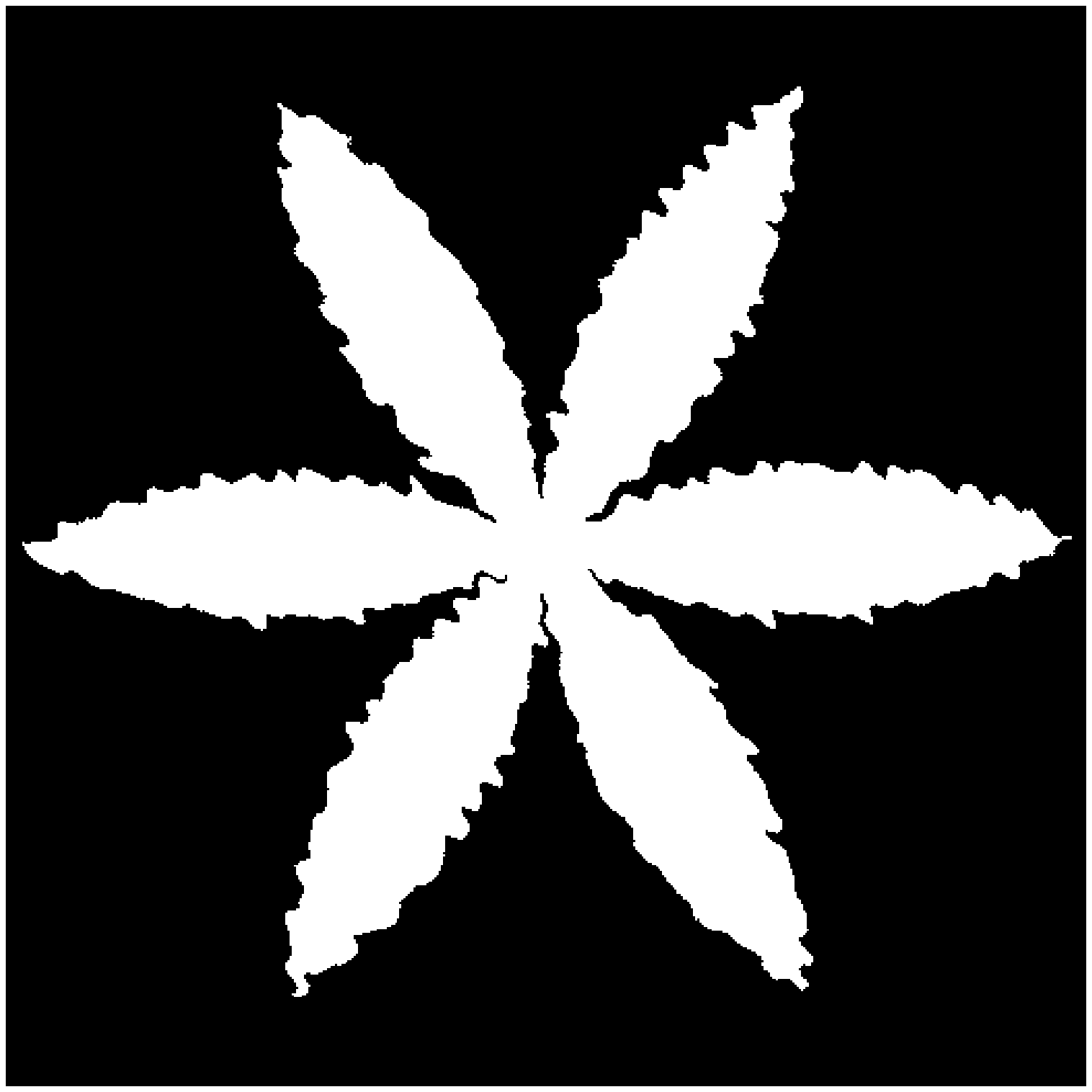}}
\subfloat{\includegraphics[trim=0.1cm 0cm 0.12cm 0cm, clip=true, width=1.3cm, height=1.5cm]{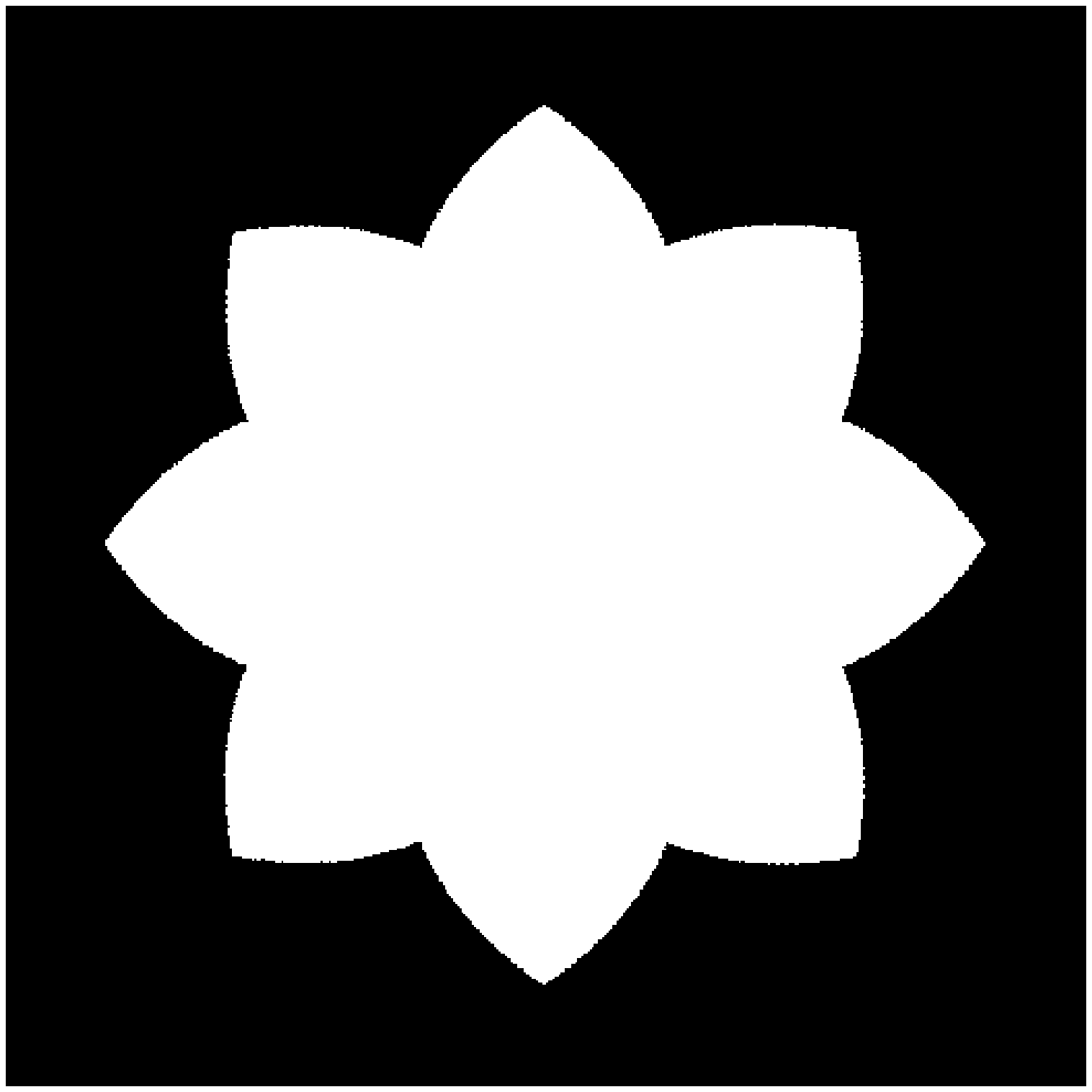}}
\subfloat{\includegraphics[trim=0.1cm 0cm 0.12cm 0cm, clip=true, width=1.3cm, height=1.5cm]{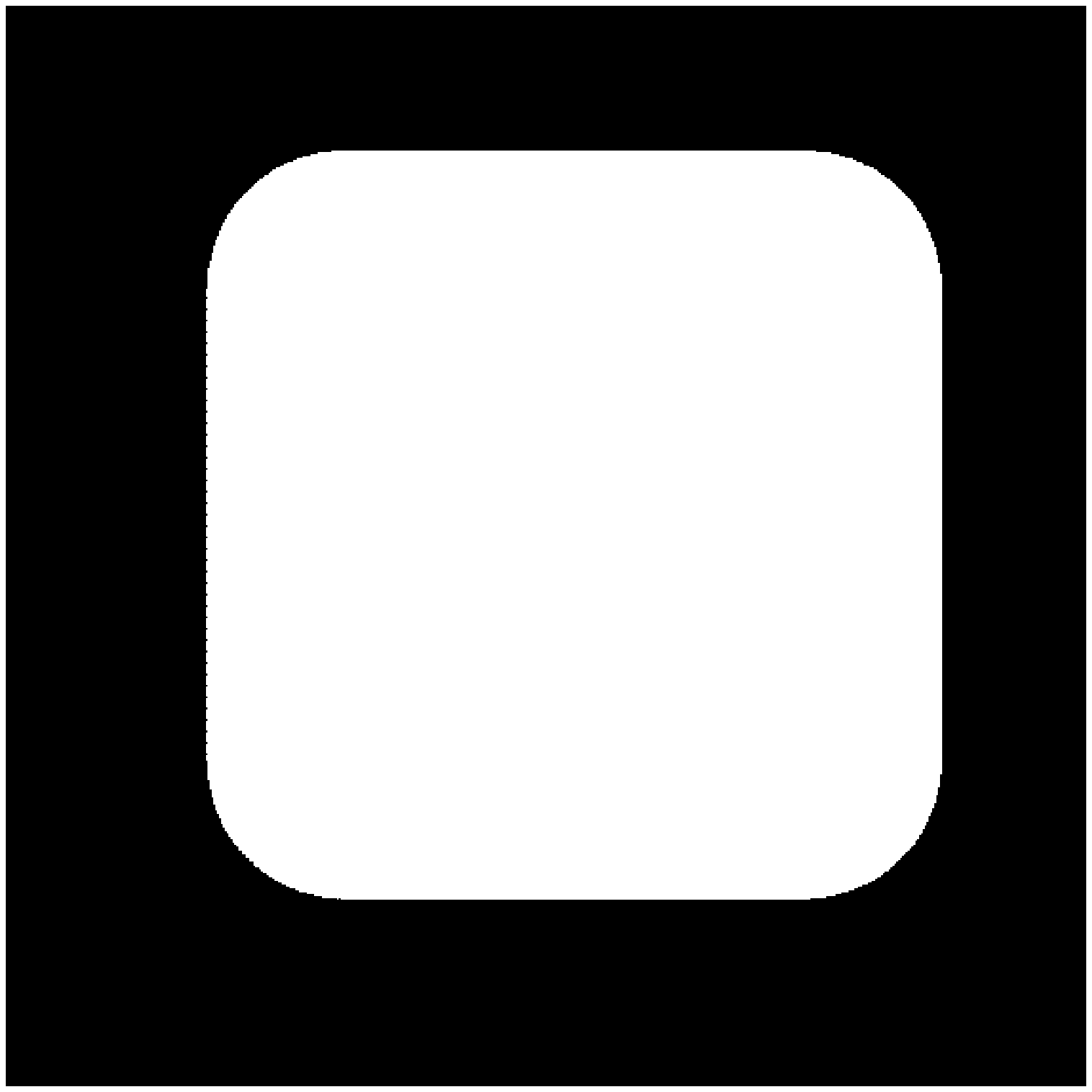}}
\subfloat{\includegraphics[trim=0.1cm 0cm 0.12cm 0cm, clip=true, width=1.3cm, height=1.5cm]{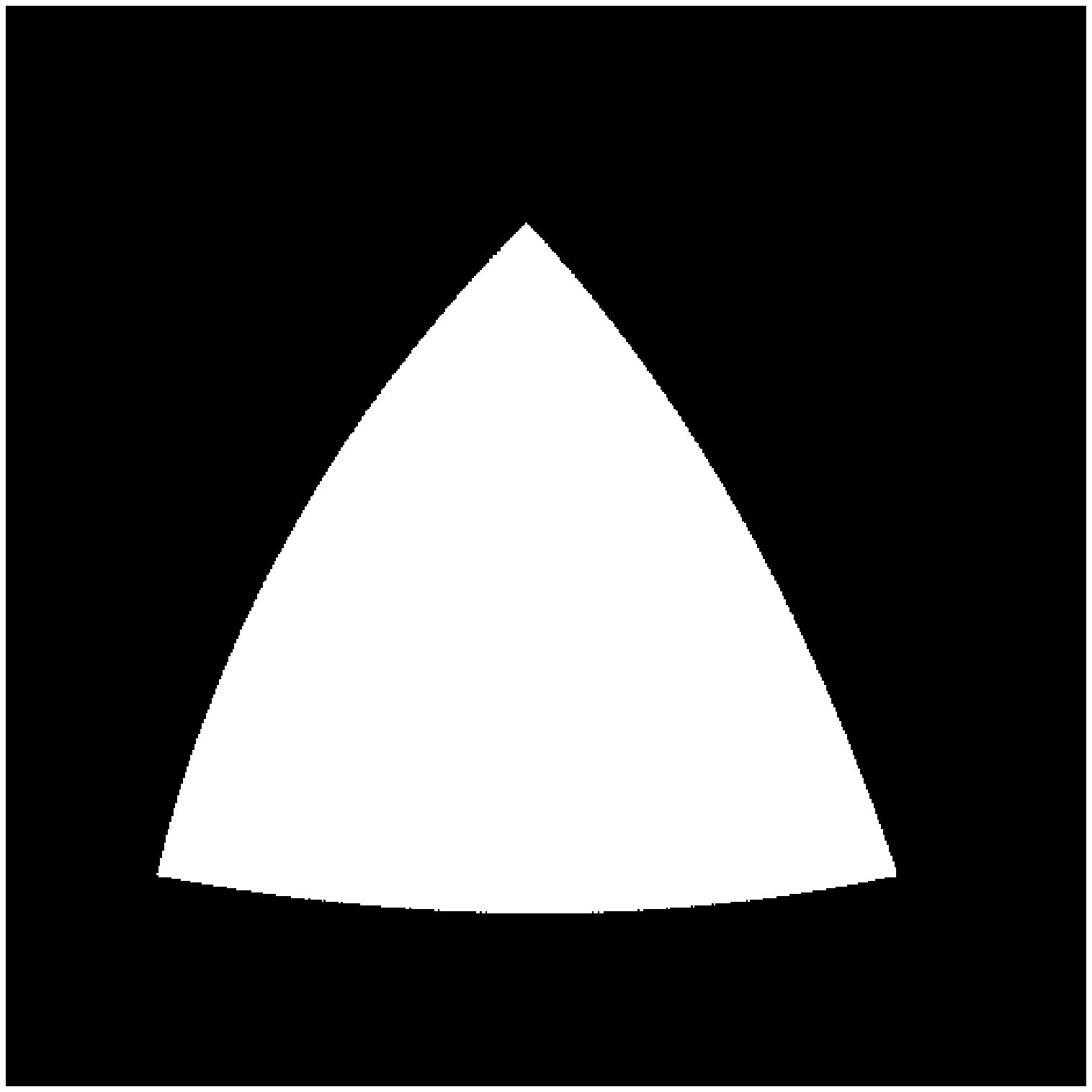}}
\subfloat{\includegraphics[trim=0.1cm 0cm 0.12cm 0cm, clip=true, width=1.3cm, height=1.5cm]{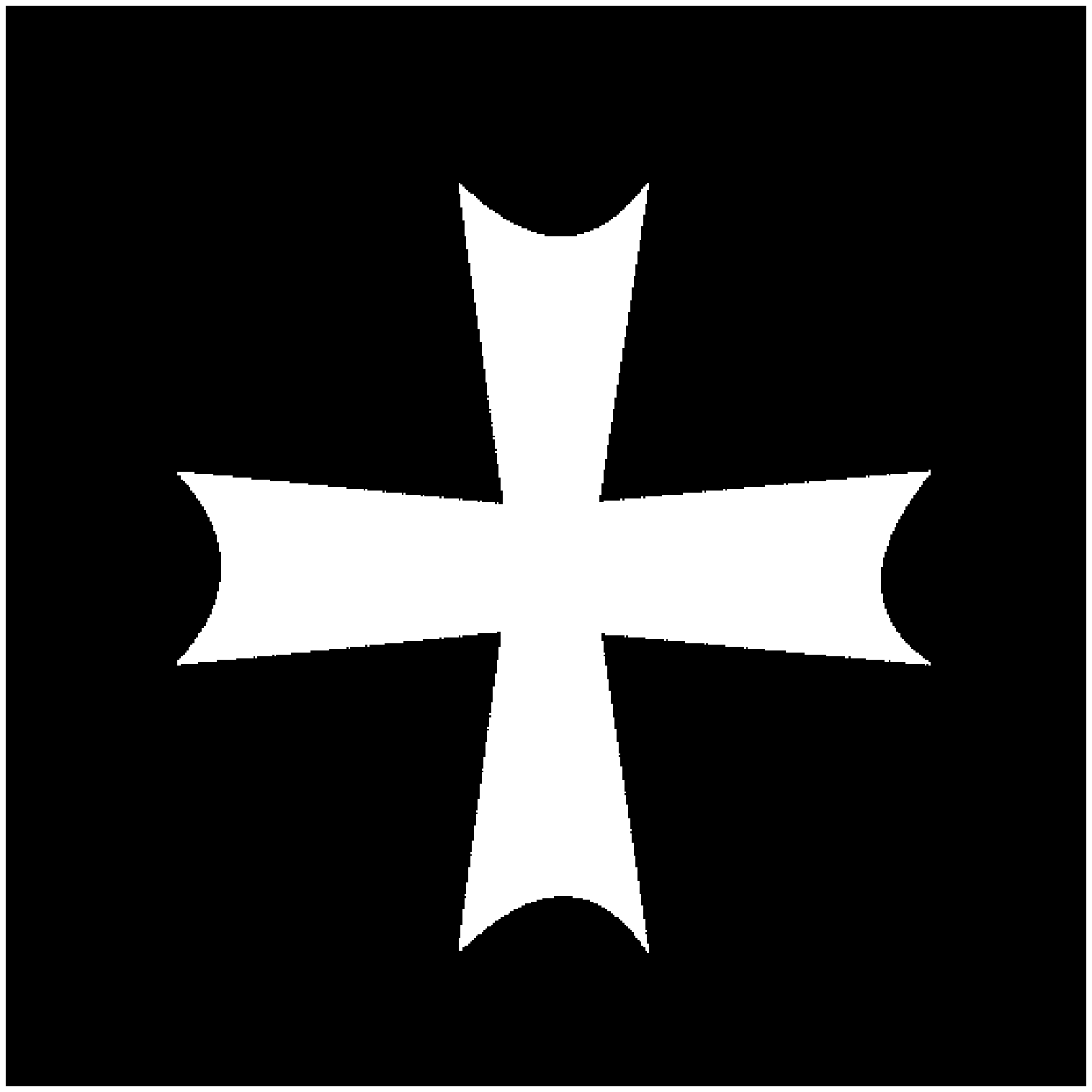}}
\subfloat{\includegraphics[trim=0.1cm 0cm 0.12cm 0cm, clip=true, width=1.3cm, height=1.5cm]{device6-1.eps}}
\subfloat{\includegraphics[trim=0.1cm 0cm 0.12cm 0cm, clip=true, width=1.3cm, height=1.5cm]{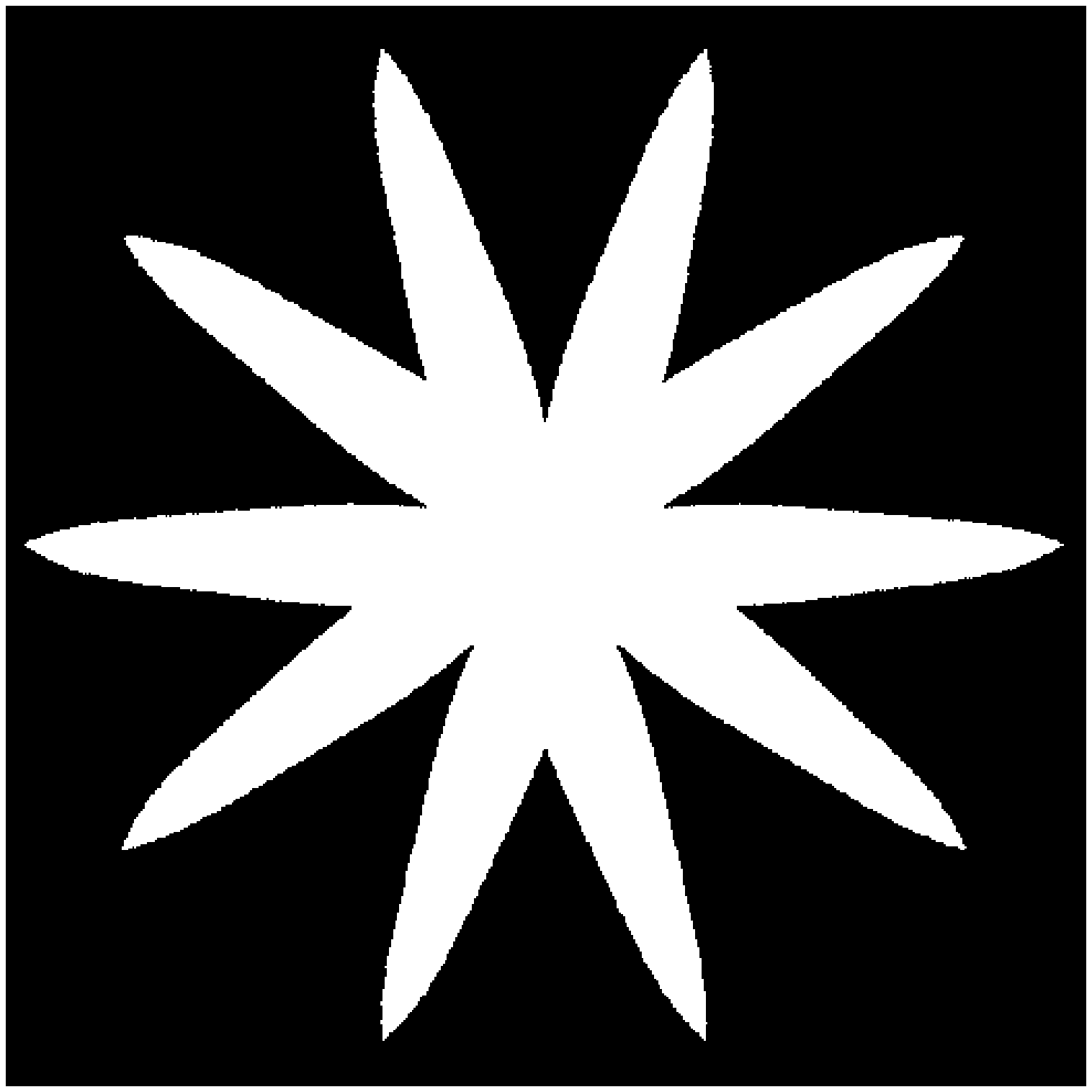}}\\

\subfloat{\includegraphics[trim=0.1cm 0cm 0.12cm 0cm, clip=true, width=1.3cm, height=1.5cm]{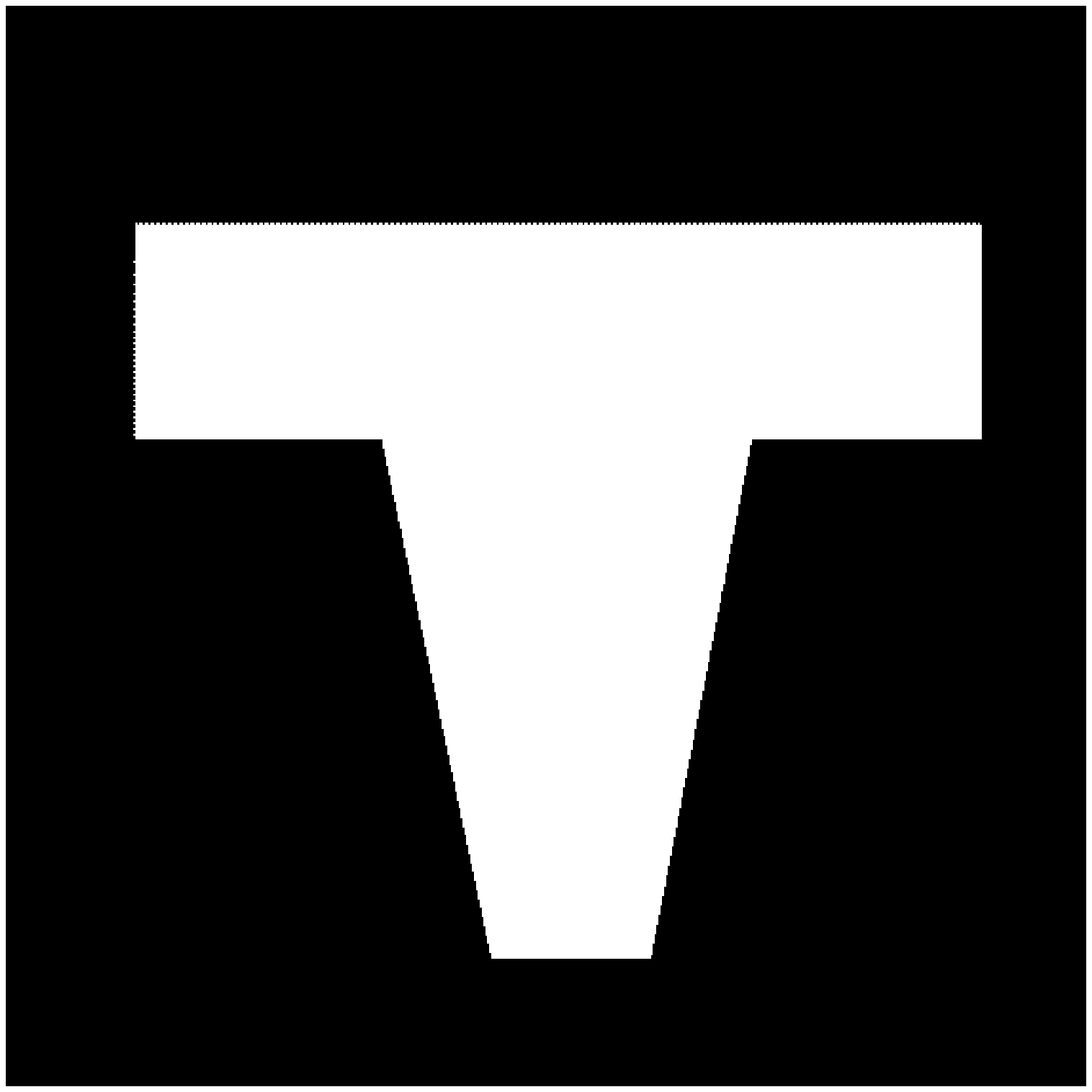}}
\subfloat{\includegraphics[trim=0.1cm 0cm 0.12cm 0cm, clip=true, width=1.3cm, height=1.5cm]{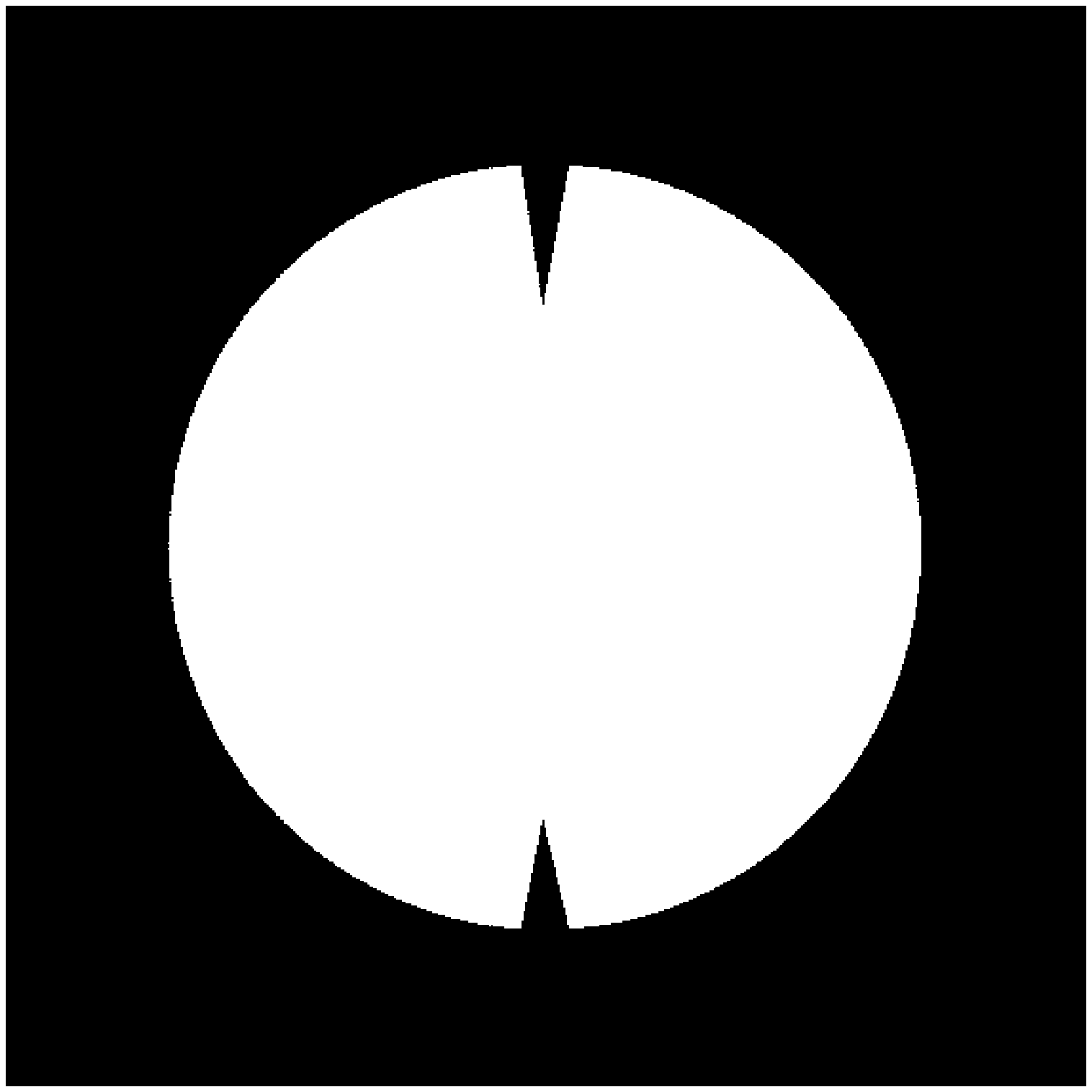}}
\subfloat{\includegraphics[trim=0.1cm 0cm 0.12cm 0cm, clip=true, width=1.3cm, height=1.5cm]{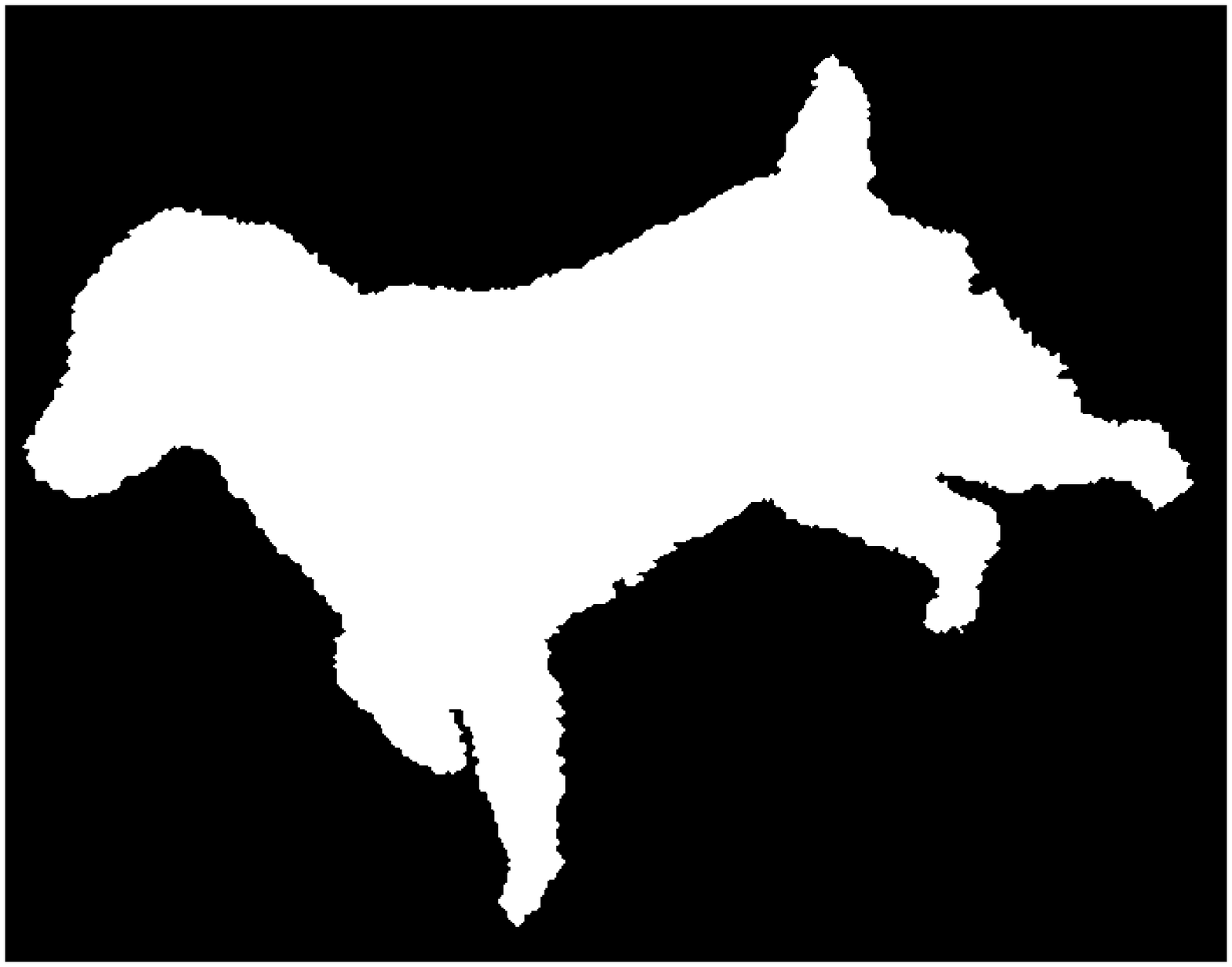}}
\subfloat{\includegraphics[trim=0.1cm 0cm 0.12cm 0cm, clip=true, width=1.3cm, height=1.5cm]{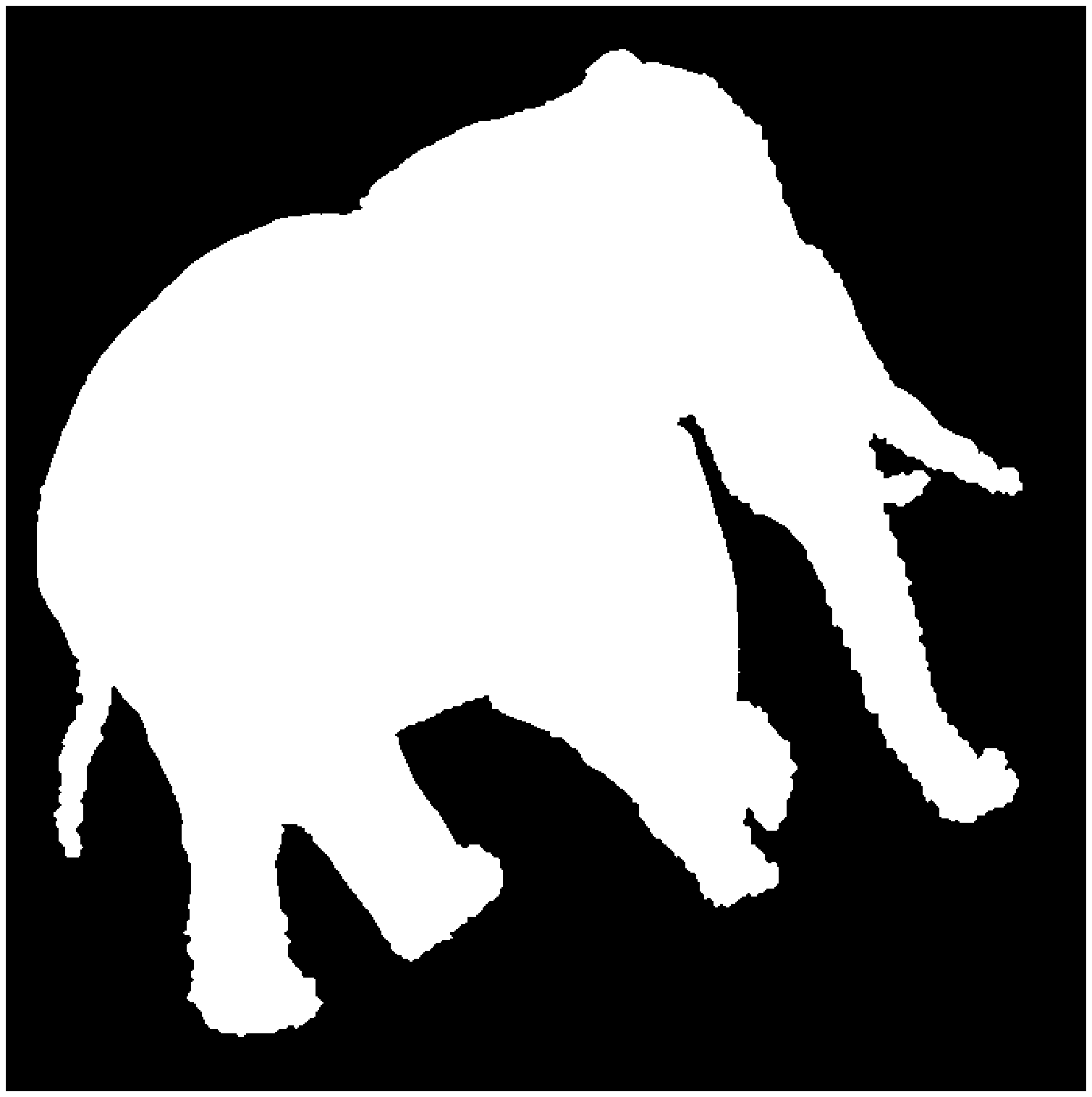}}
\subfloat{\includegraphics[trim=0.1cm 0cm 0.12cm 0cm, clip=true, width=1.3cm, height=1.5cm]{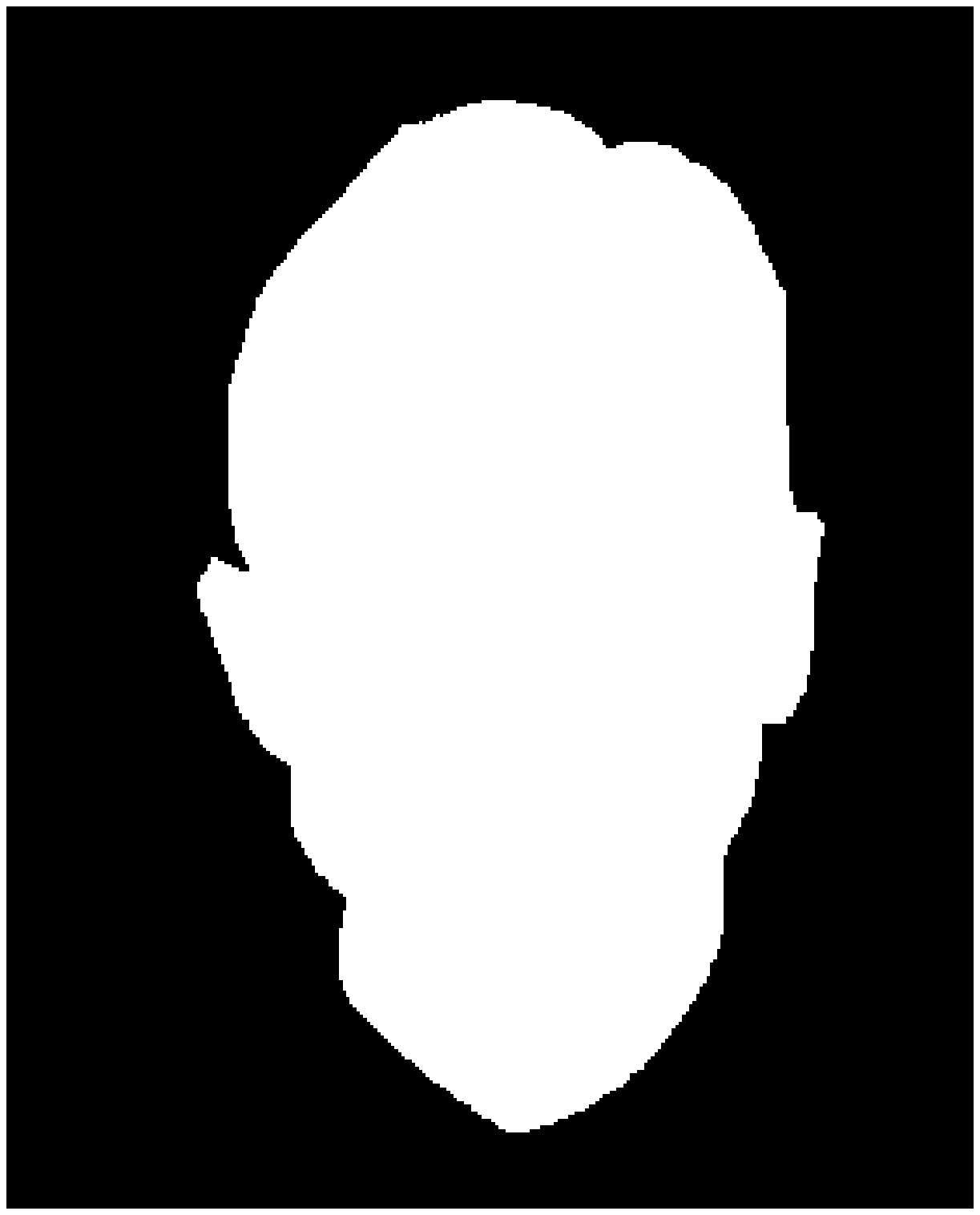}}
\subfloat{\includegraphics[trim=0.1cm 0cm 0.12cm 0cm, clip=true, width=1.3cm, height=1.5cm]{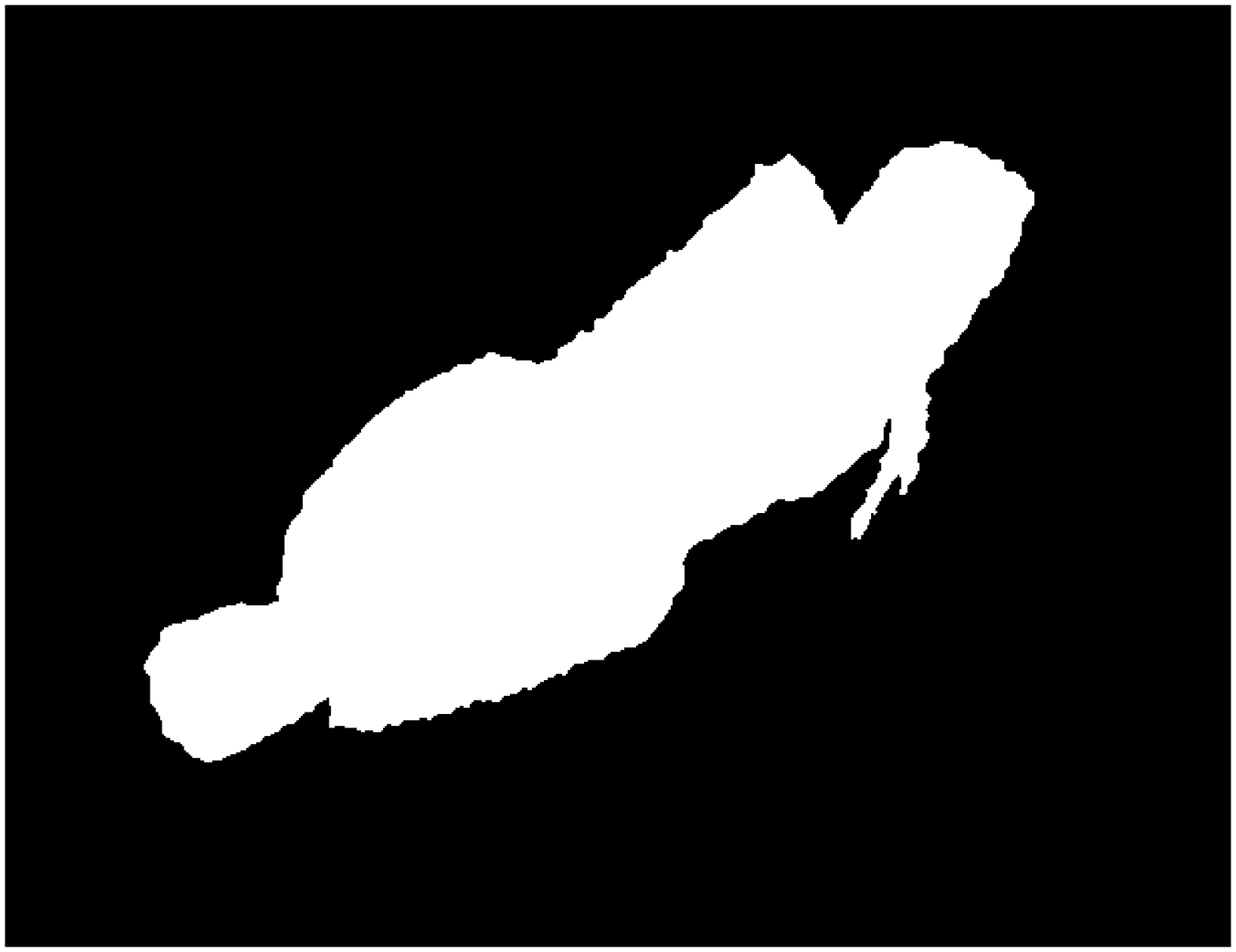}}
\subfloat{\includegraphics[trim=0.1cm 0cm 0.12cm 0cm, clip=true, width=1.3cm, height=1.5cm]{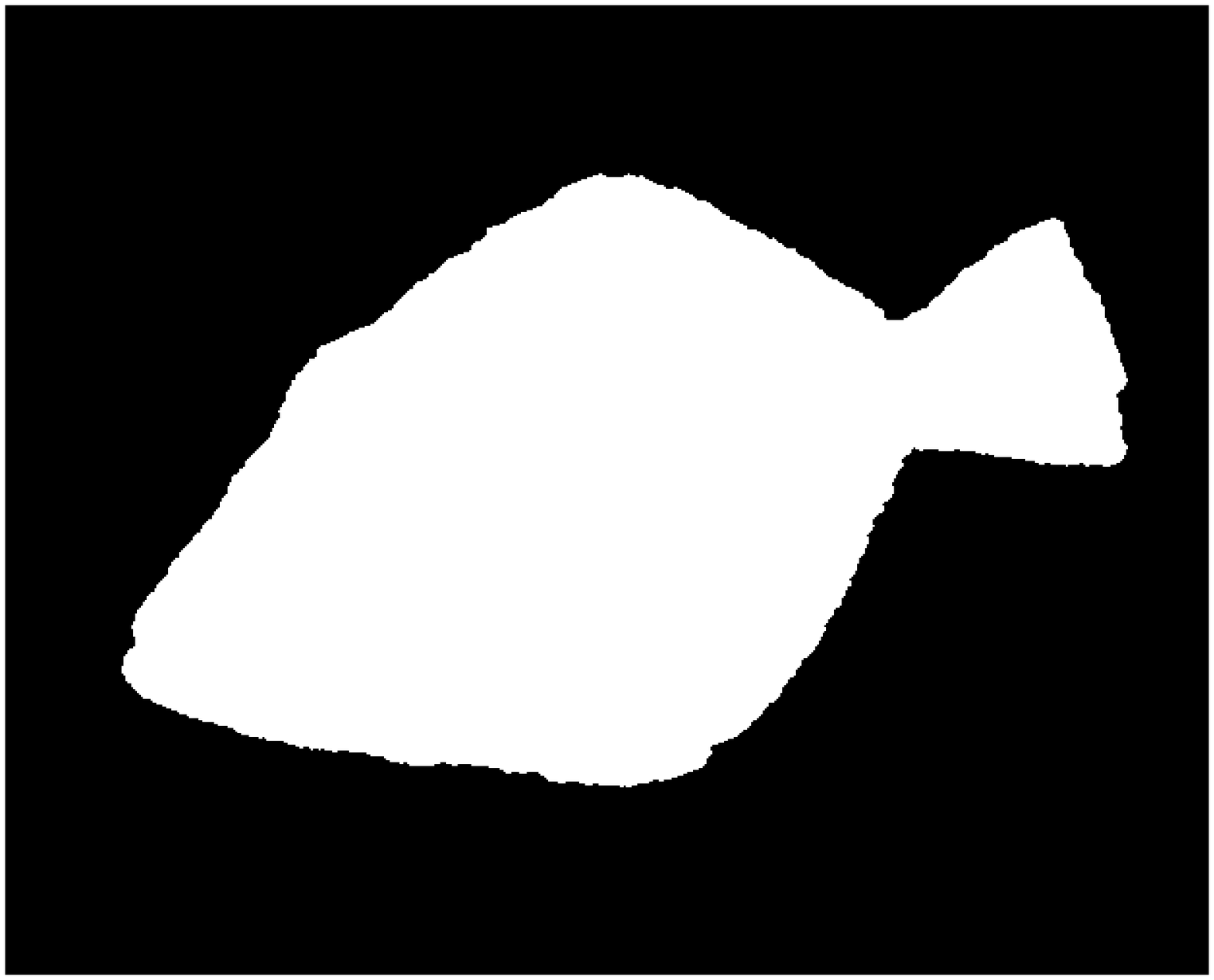}}
\subfloat{\includegraphics[trim=0.1cm 0cm 0.12cm 0cm, clip=true, width=1.3cm, height=1.5cm]{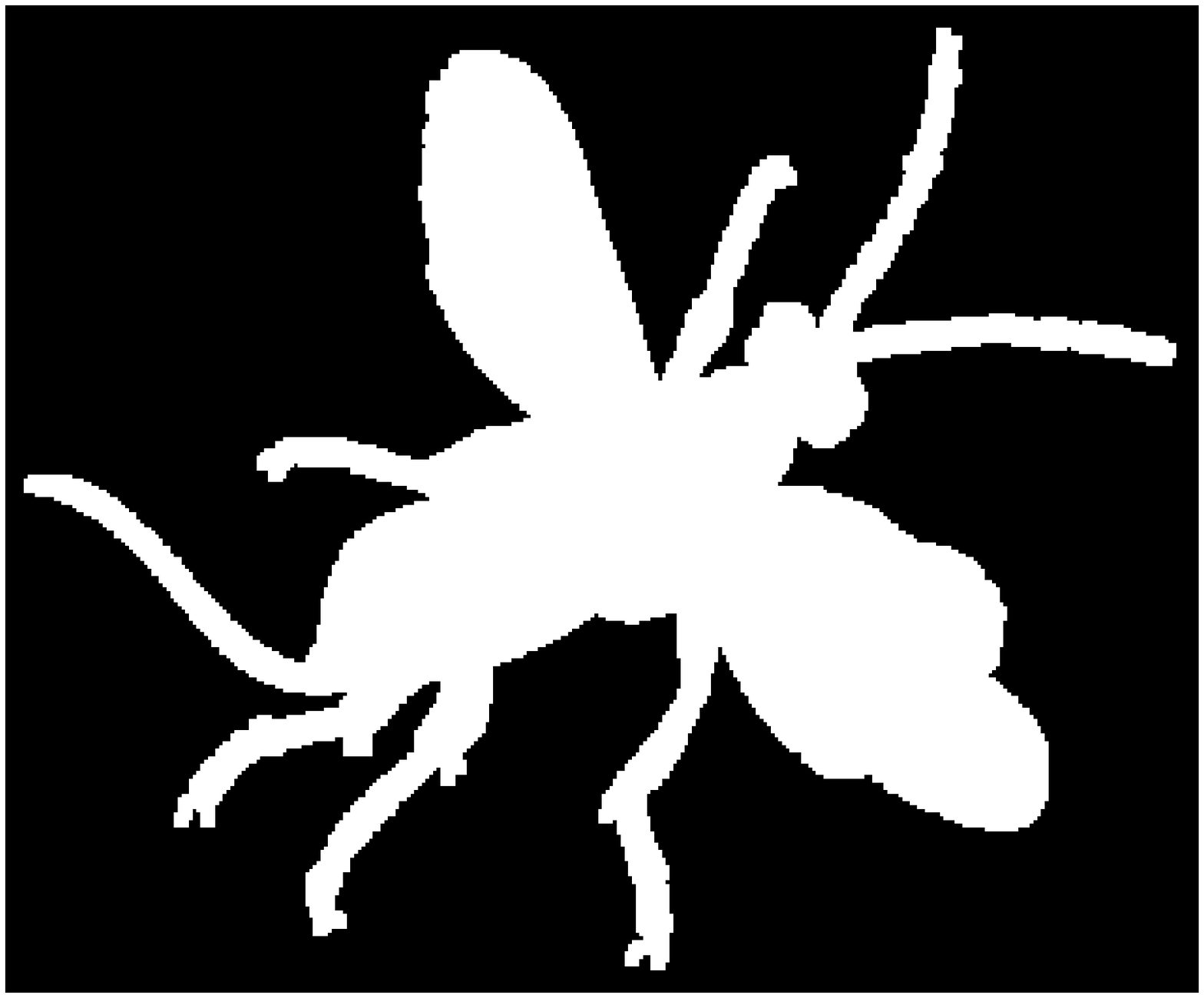}}
\subfloat{\includegraphics[trim=0.1cm 0cm 0.12cm 0cm, clip=true, width=1.3cm, height=1.5cm]{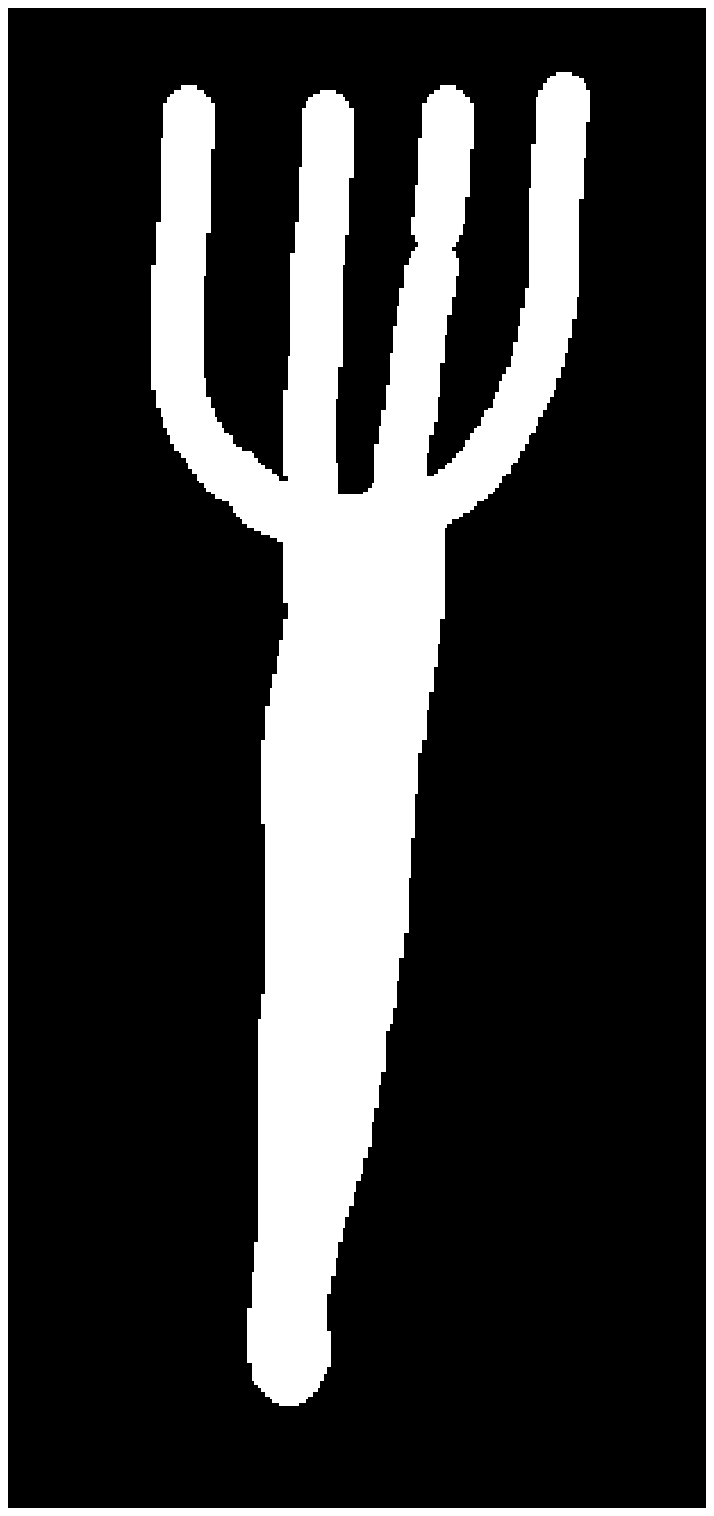}}
\subfloat{\includegraphics[trim=0.1cm 0cm 0.12cm 0cm, clip=true, width=1.3cm, height=1.5cm]{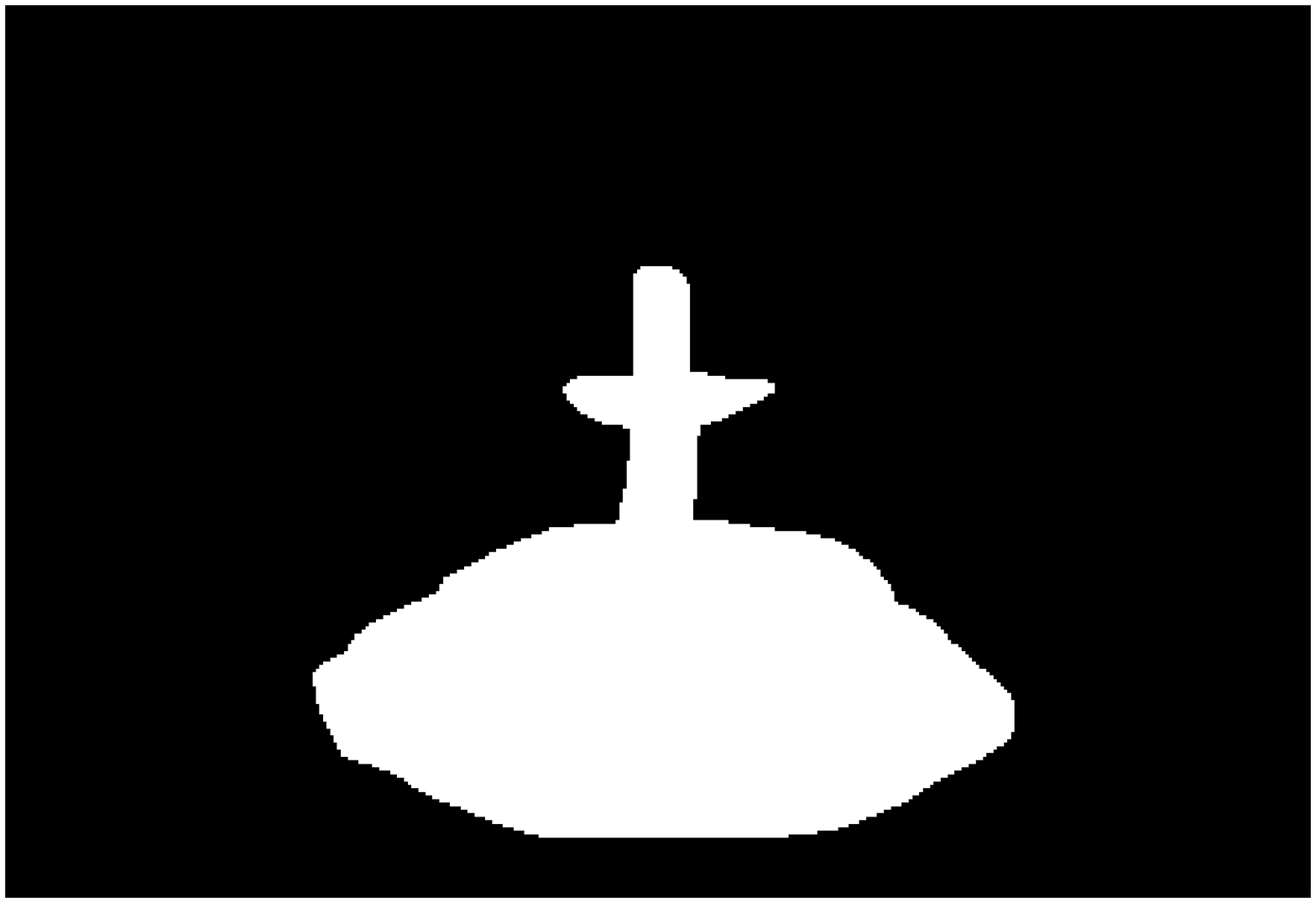}}\\

\subfloat{\includegraphics[trim=0.1cm 0cm 0.12cm 0cm, clip=true, width=1.3cm, height=1.5cm]{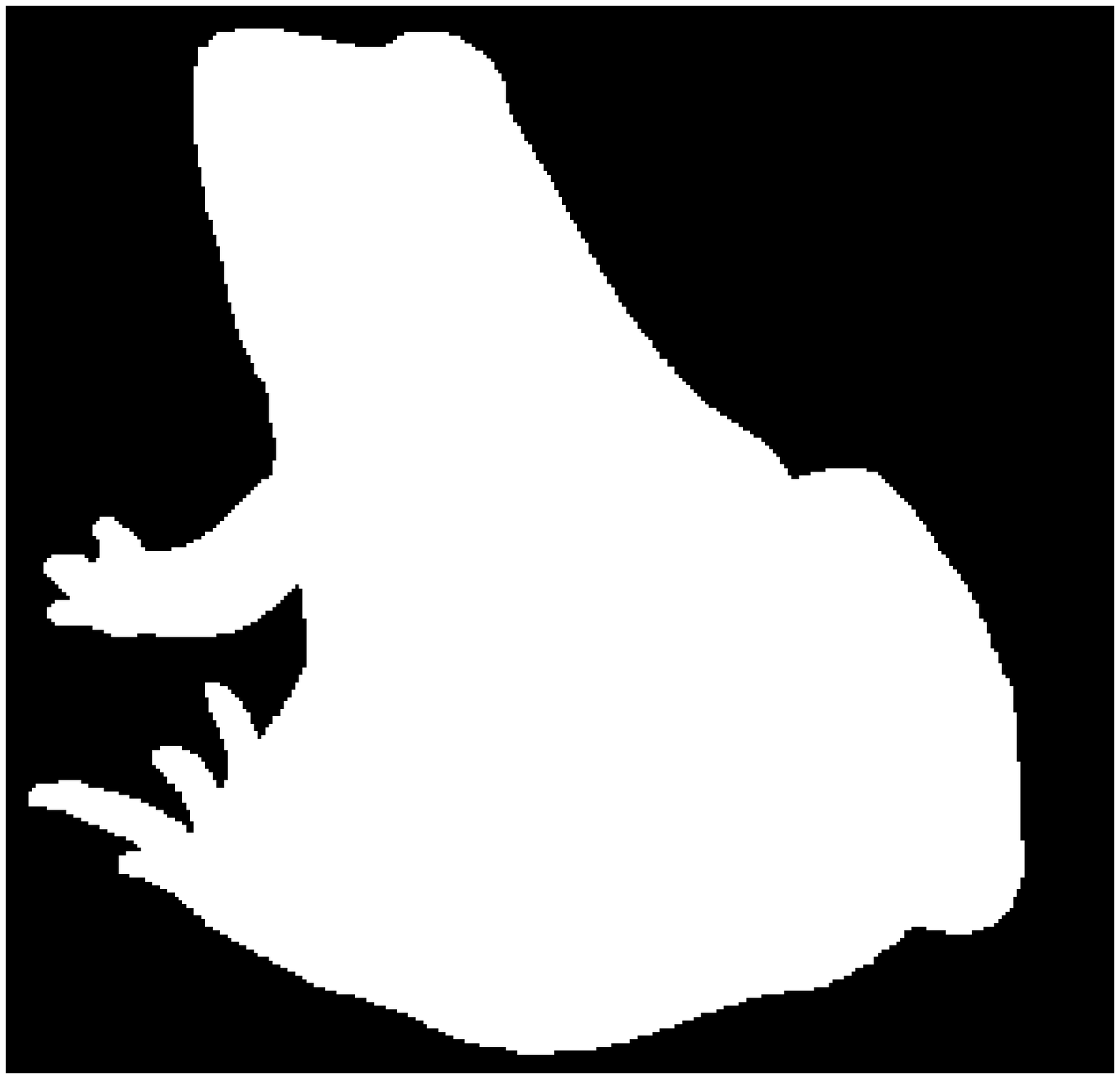}}
\subfloat{\includegraphics[trim=0.1cm 0cm 0.12cm 0cm, clip=true, width=1.3cm, height=1.5cm]{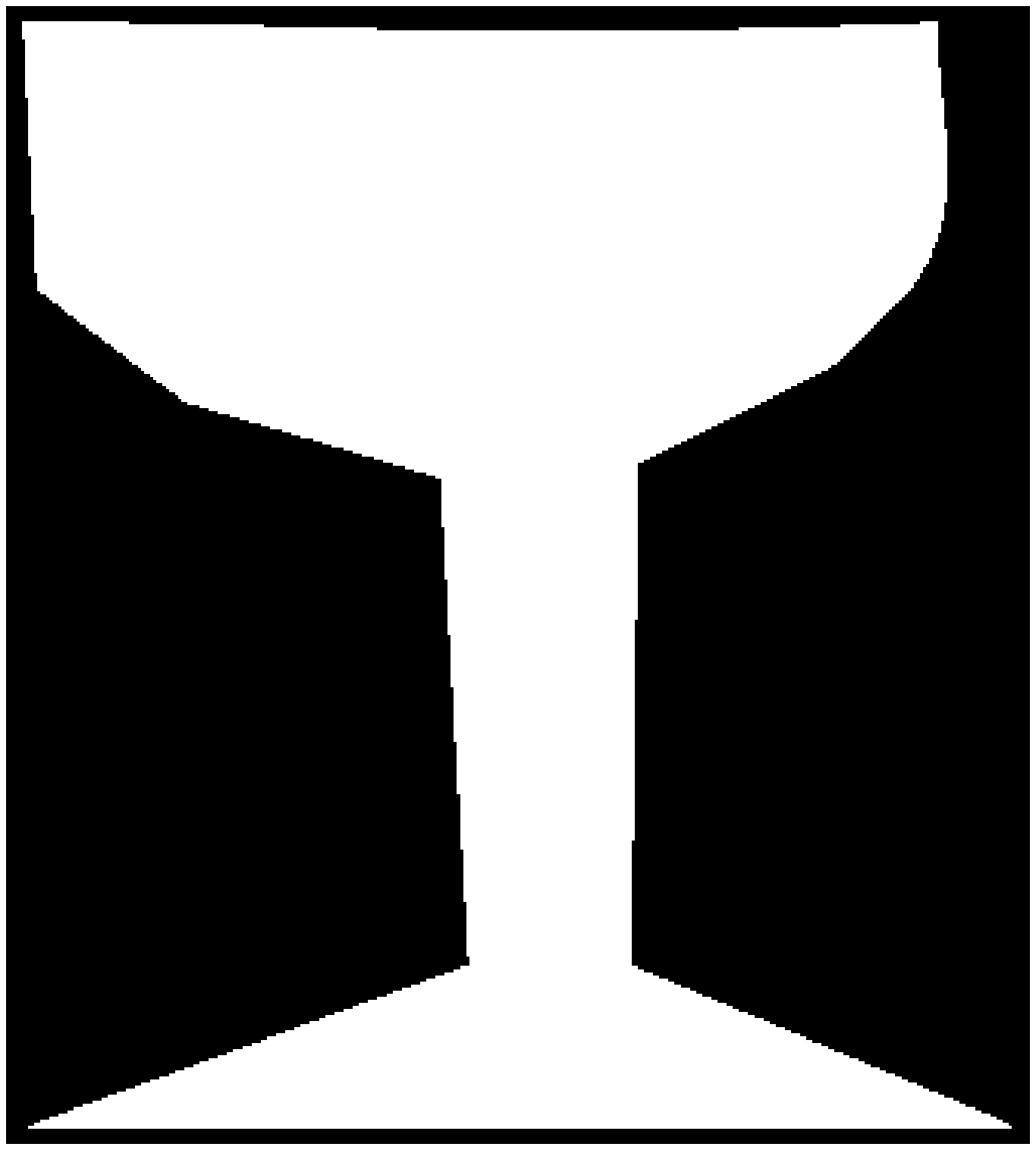}}
\subfloat{\includegraphics[trim=0.1cm 0cm 0.12cm 0cm, clip=true, width=1.3cm, height=1.5cm]{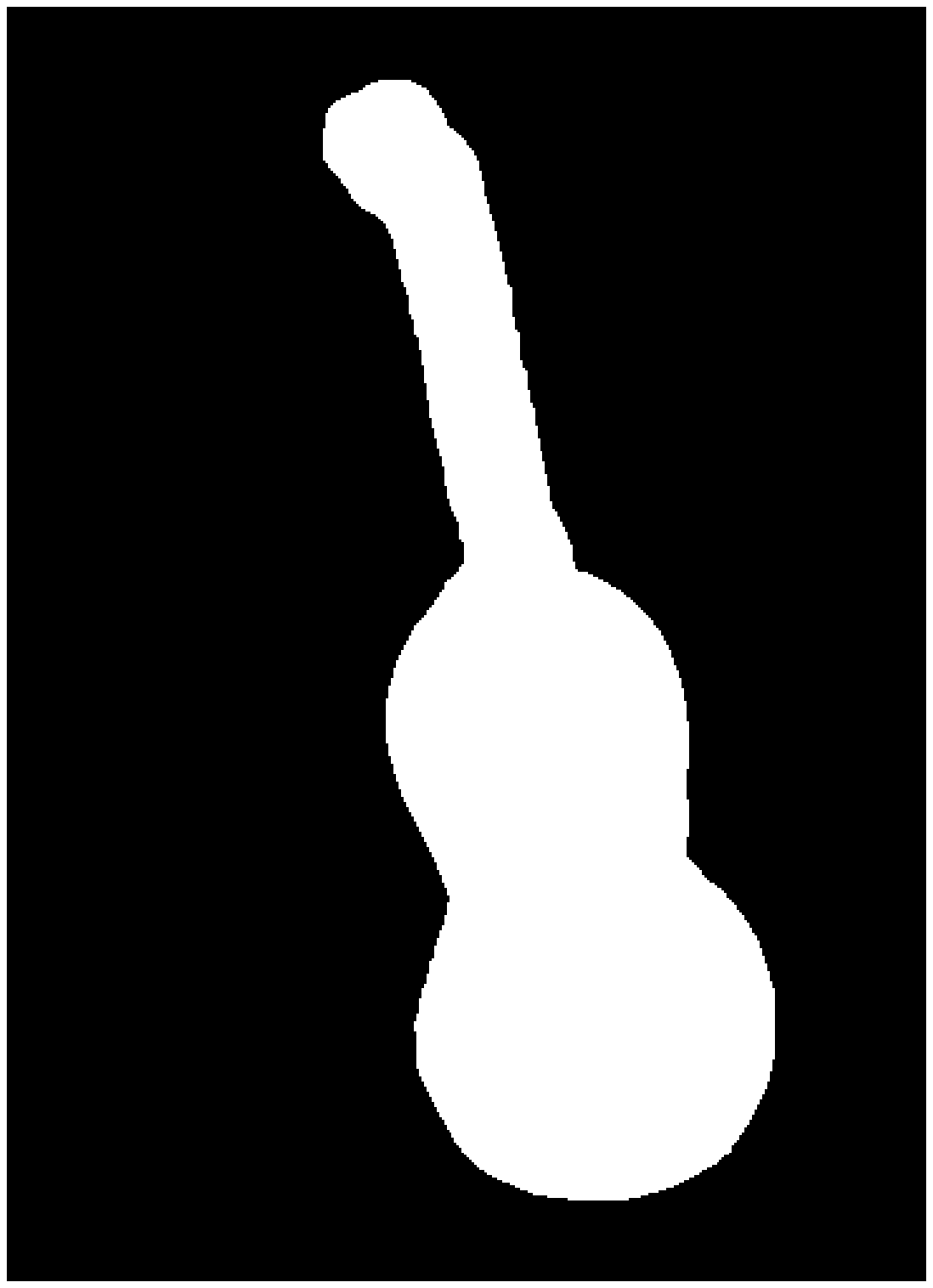}}
\subfloat{\includegraphics[trim=0.1cm 0cm 0.12cm 0cm, clip=true, width=1.3cm, height=1.5cm]{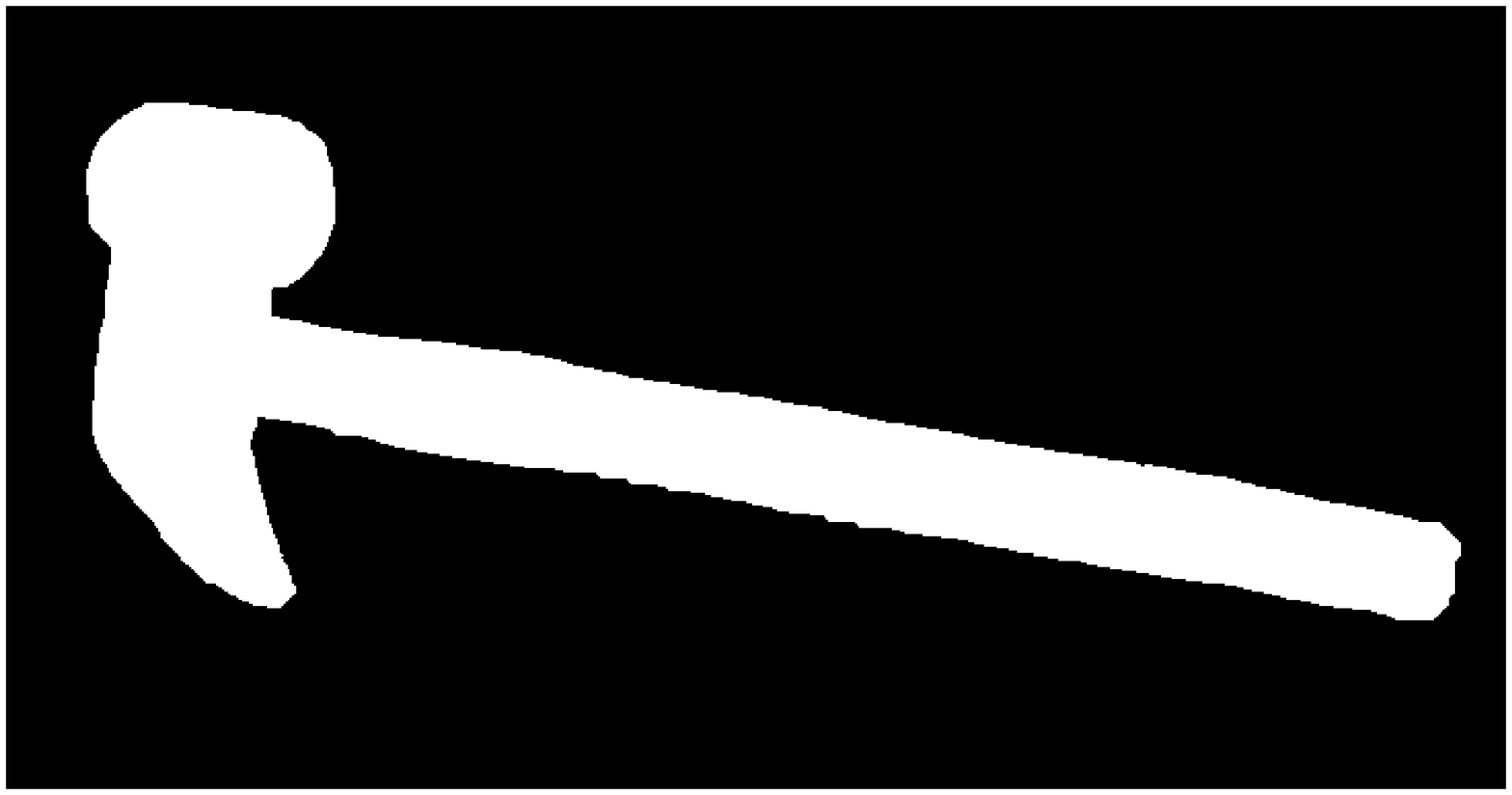}}
\subfloat{\includegraphics[trim=0.1cm 0cm 0.12cm 0cm, clip=true, width=1.3cm, height=1.5cm]{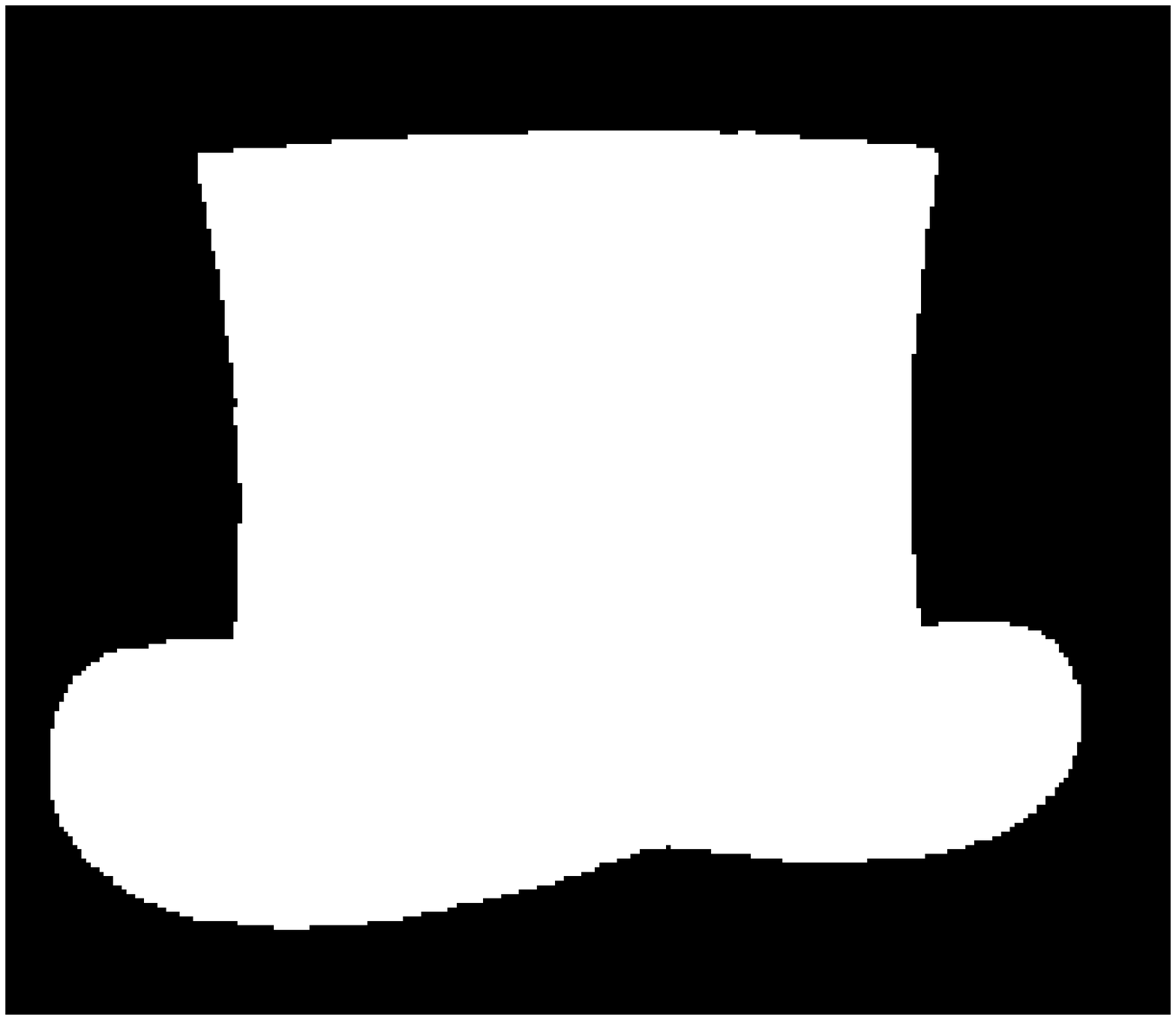}}
\subfloat{\includegraphics[trim=0.1cm 0cm 0.12cm 0cm, clip=true, width=1.3cm, height=1.5cm]{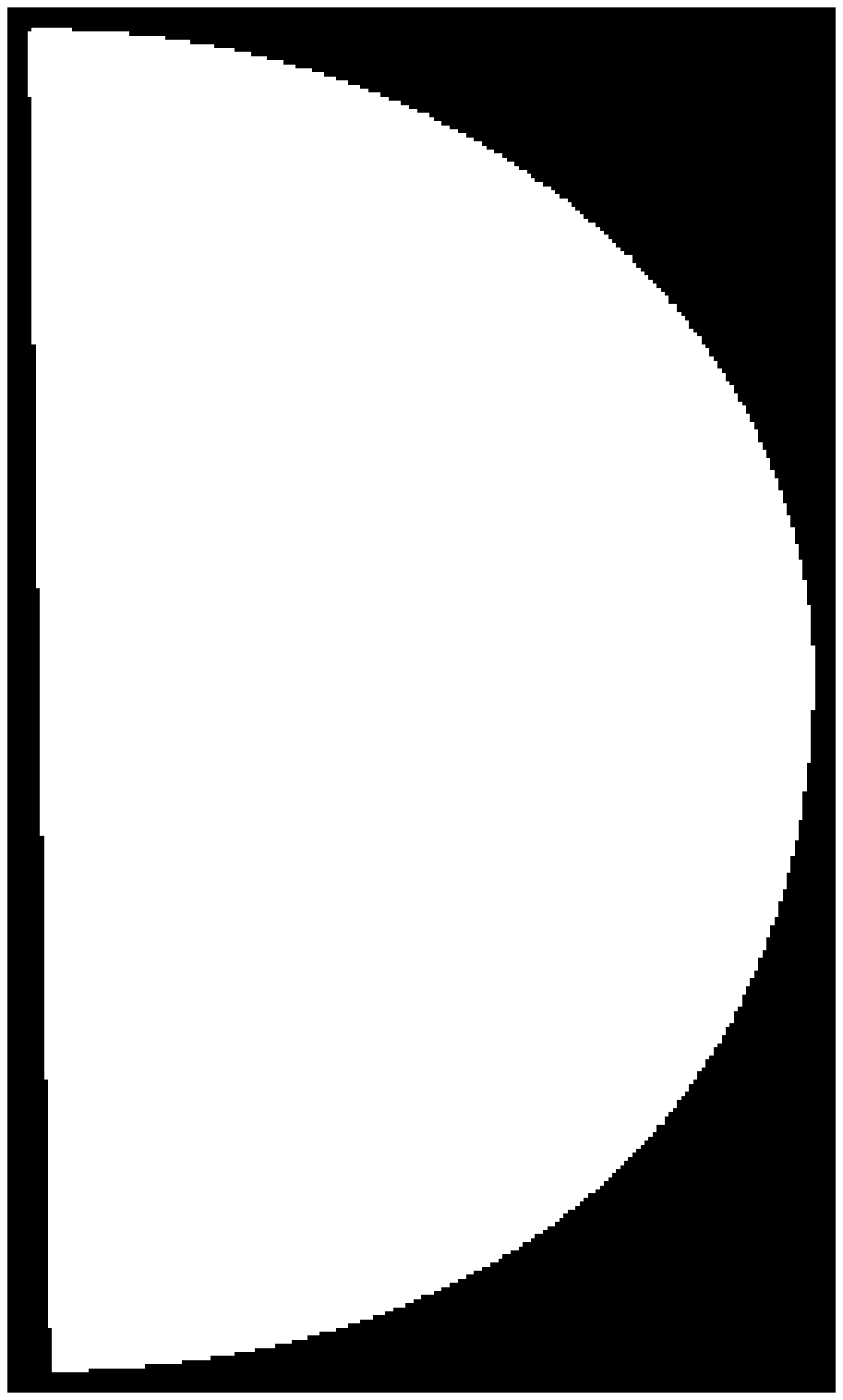}}
\subfloat{\includegraphics[trim=0.1cm 0cm 0.12cm 0cm, clip=true, width=1.3cm, height=1.5cm]{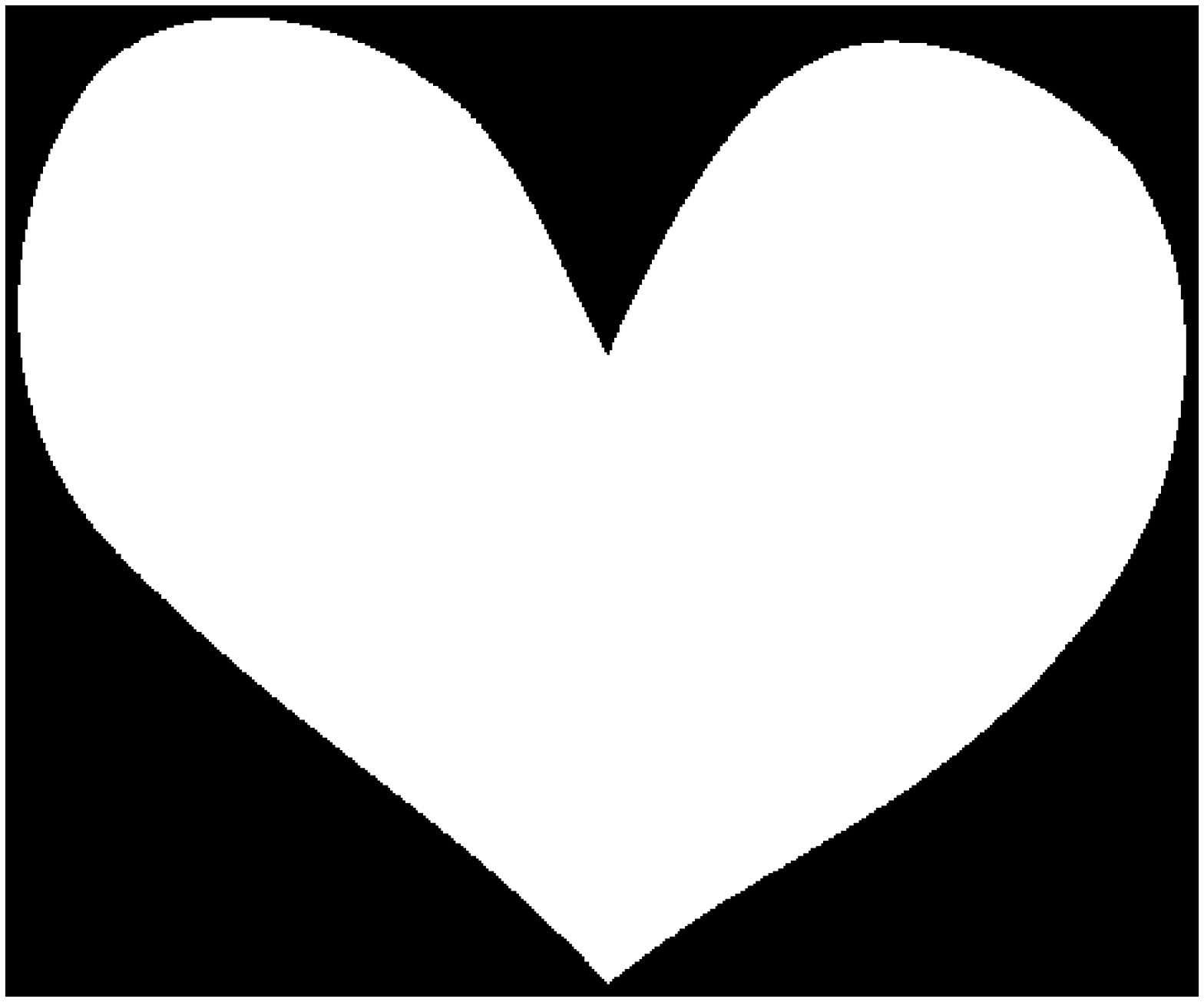}}
\subfloat{\includegraphics[trim=0.1cm 0cm 0.12cm 0cm, clip=true, width=1.3cm, height=1.5cm]{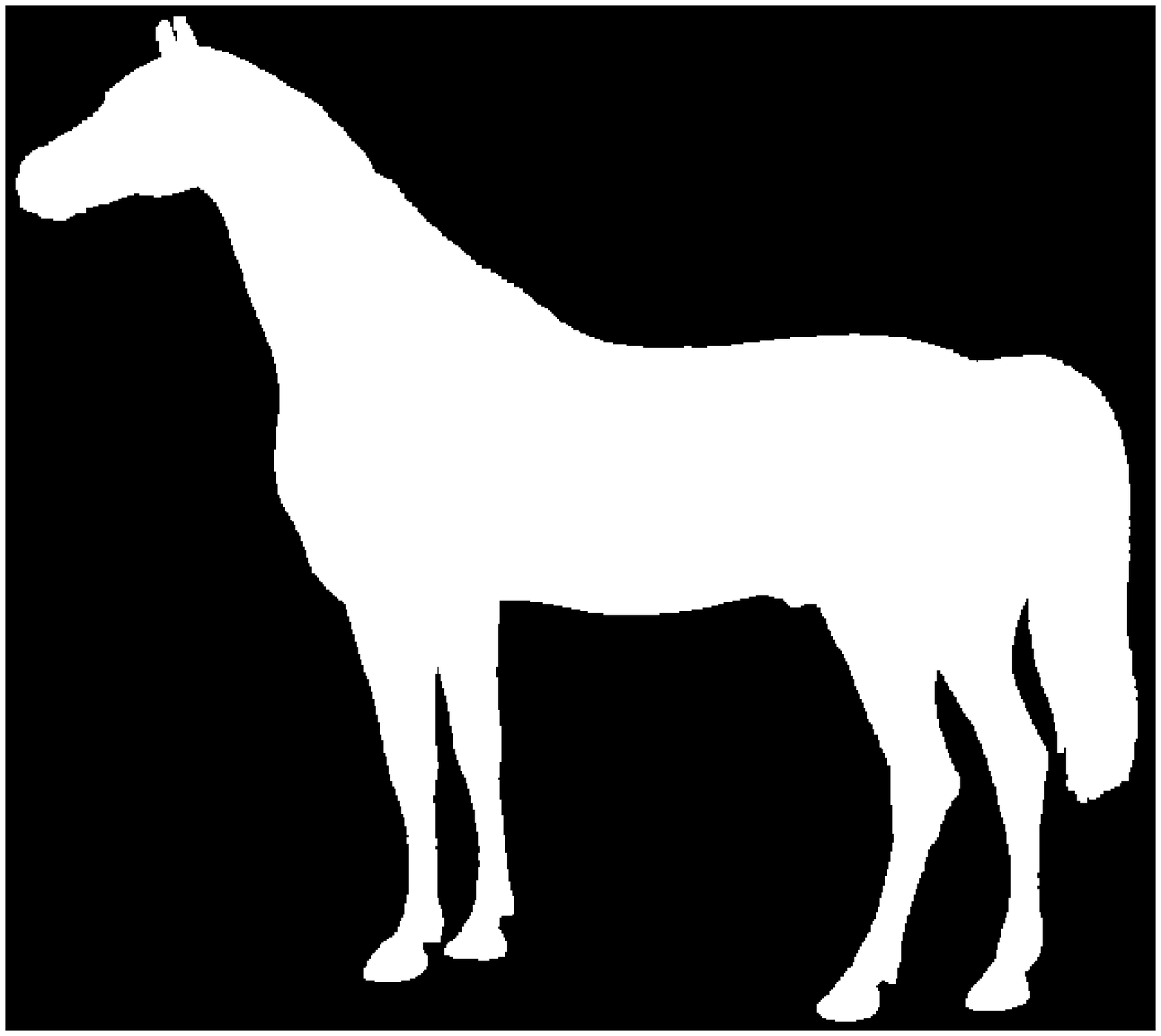}}
\subfloat{\includegraphics[trim=0.1cm 0cm 0.12cm 0cm, clip=true, width=1.3cm, height=1.5cm]{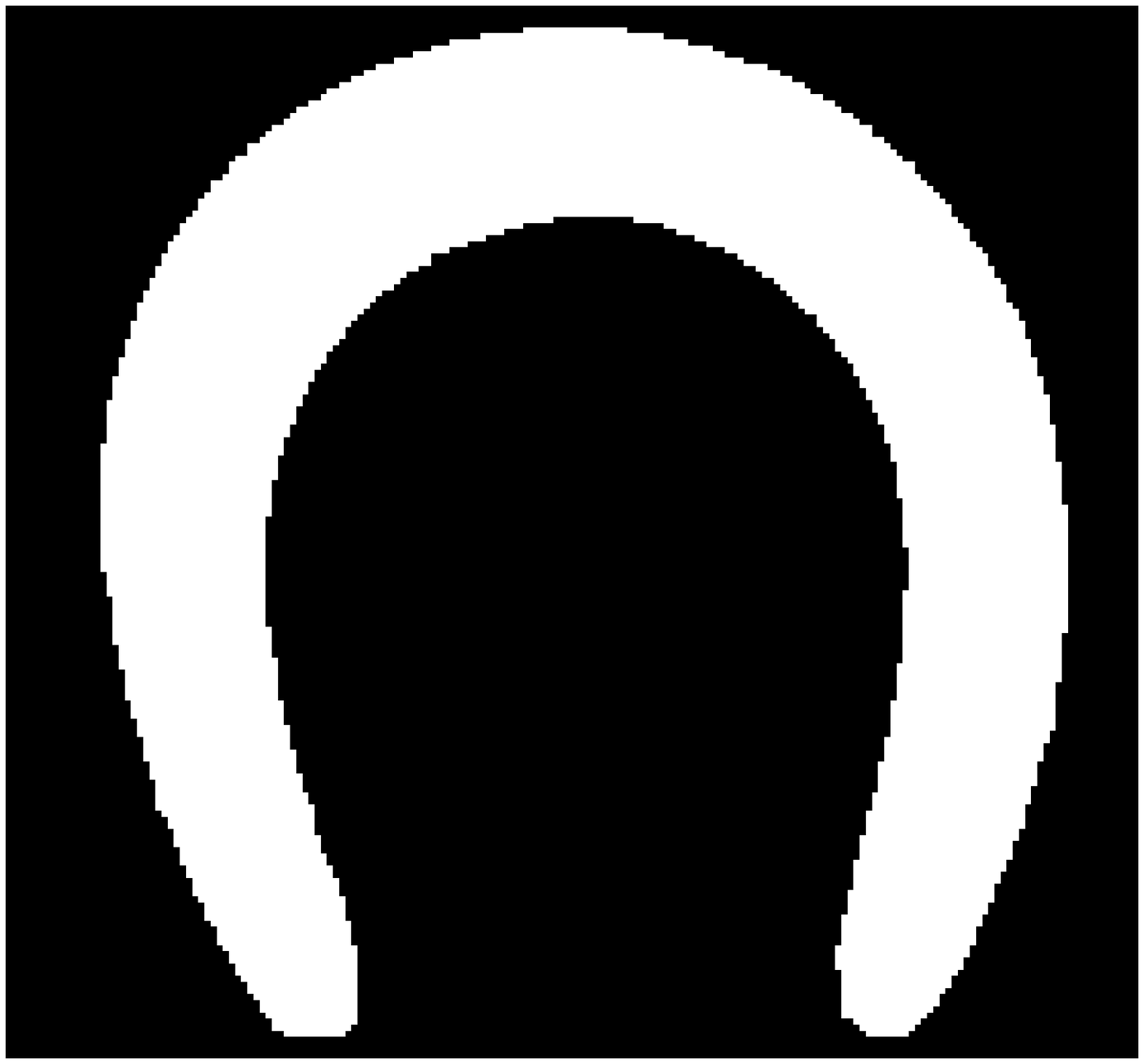}}
\subfloat{\includegraphics[trim=0.1cm 0cm 0.12cm 0cm, clip=true, width=1.3cm, height=1.5cm]{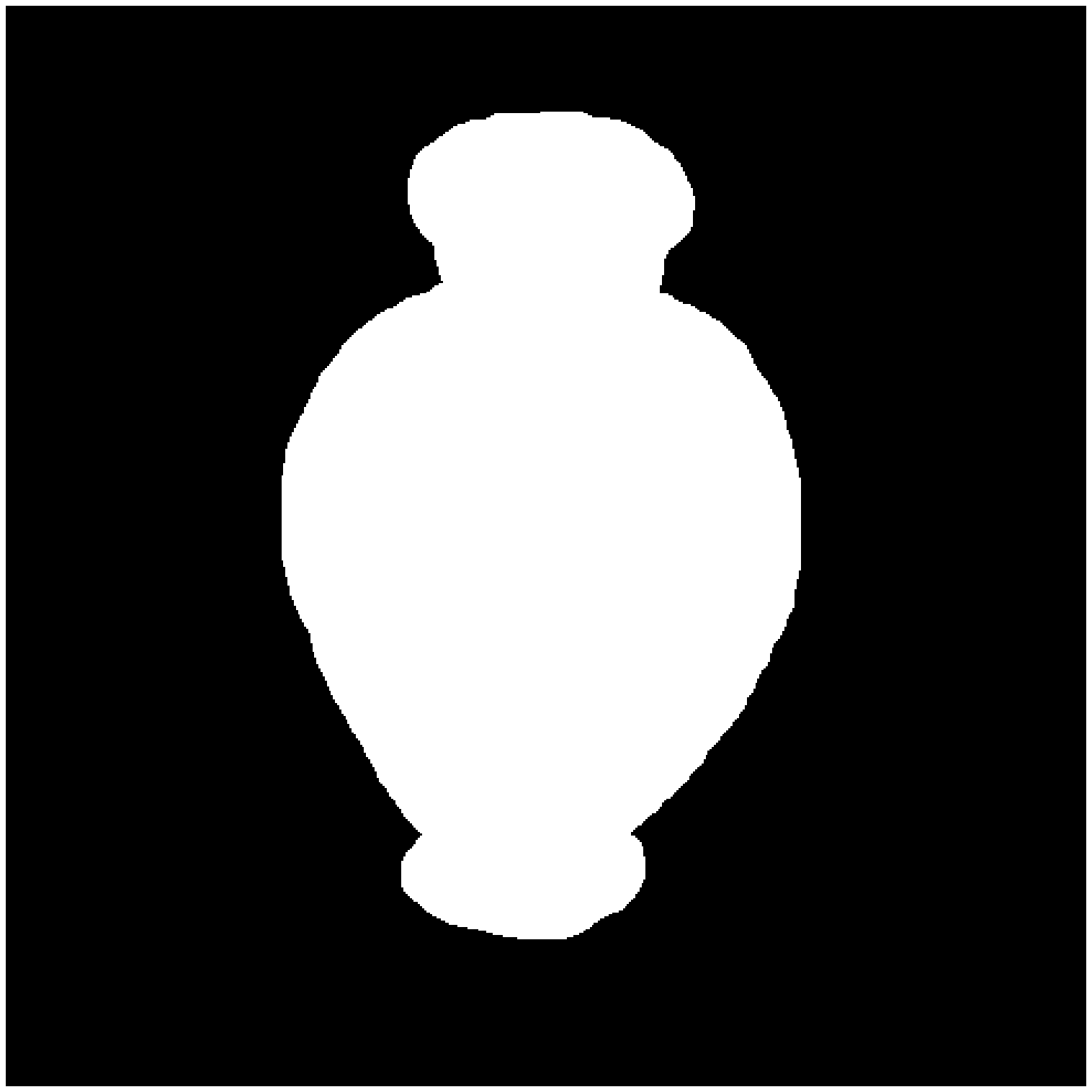}}\\

\subfloat{\includegraphics[trim=0.1cm 0cm 0.12cm 0cm, clip=true, width=1.3cm, height=1.5cm]{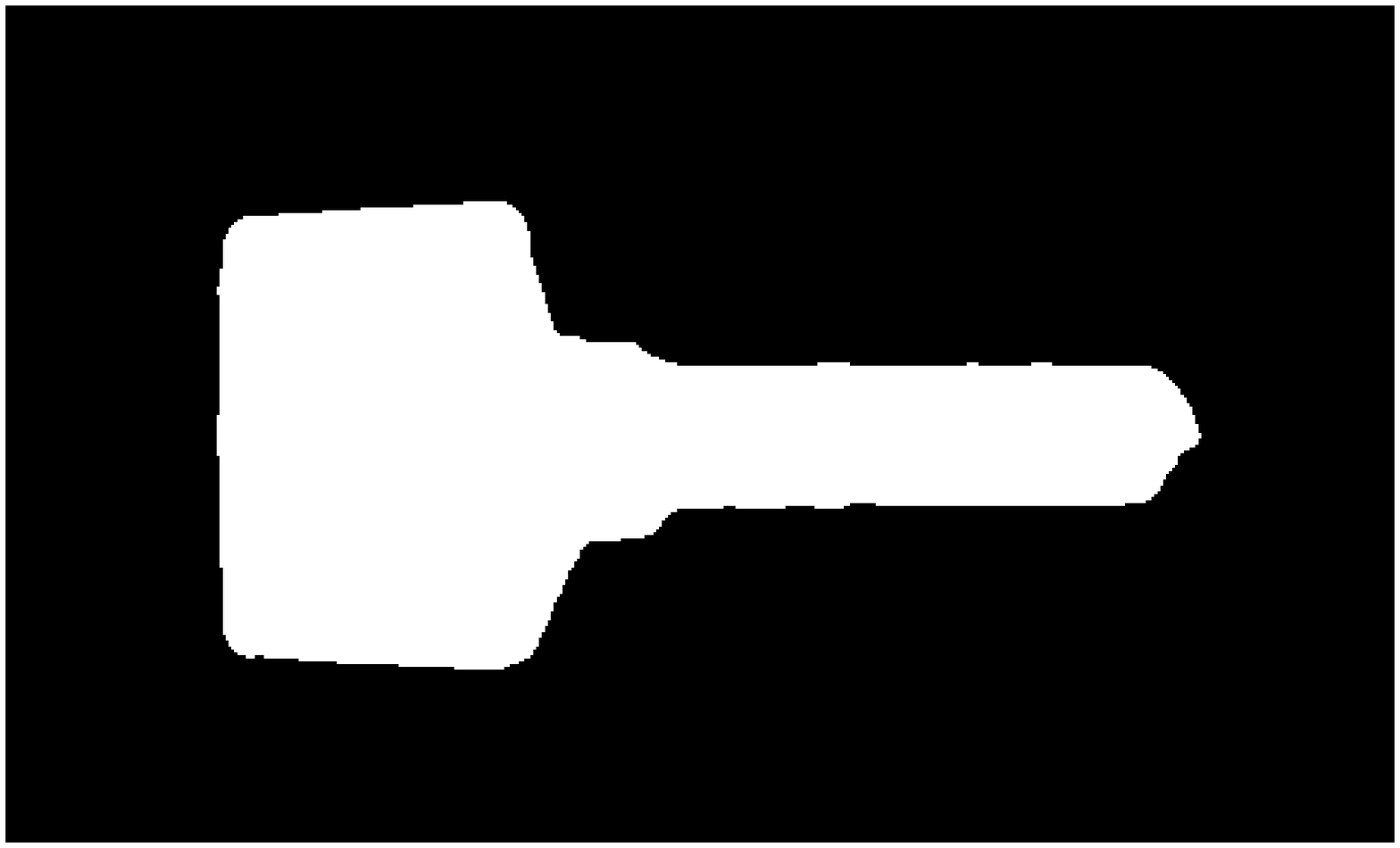}}
\subfloat{\includegraphics[trim=0.1cm 0cm 0.12cm 0cm, clip=true, width=1.3cm, height=1.5cm]{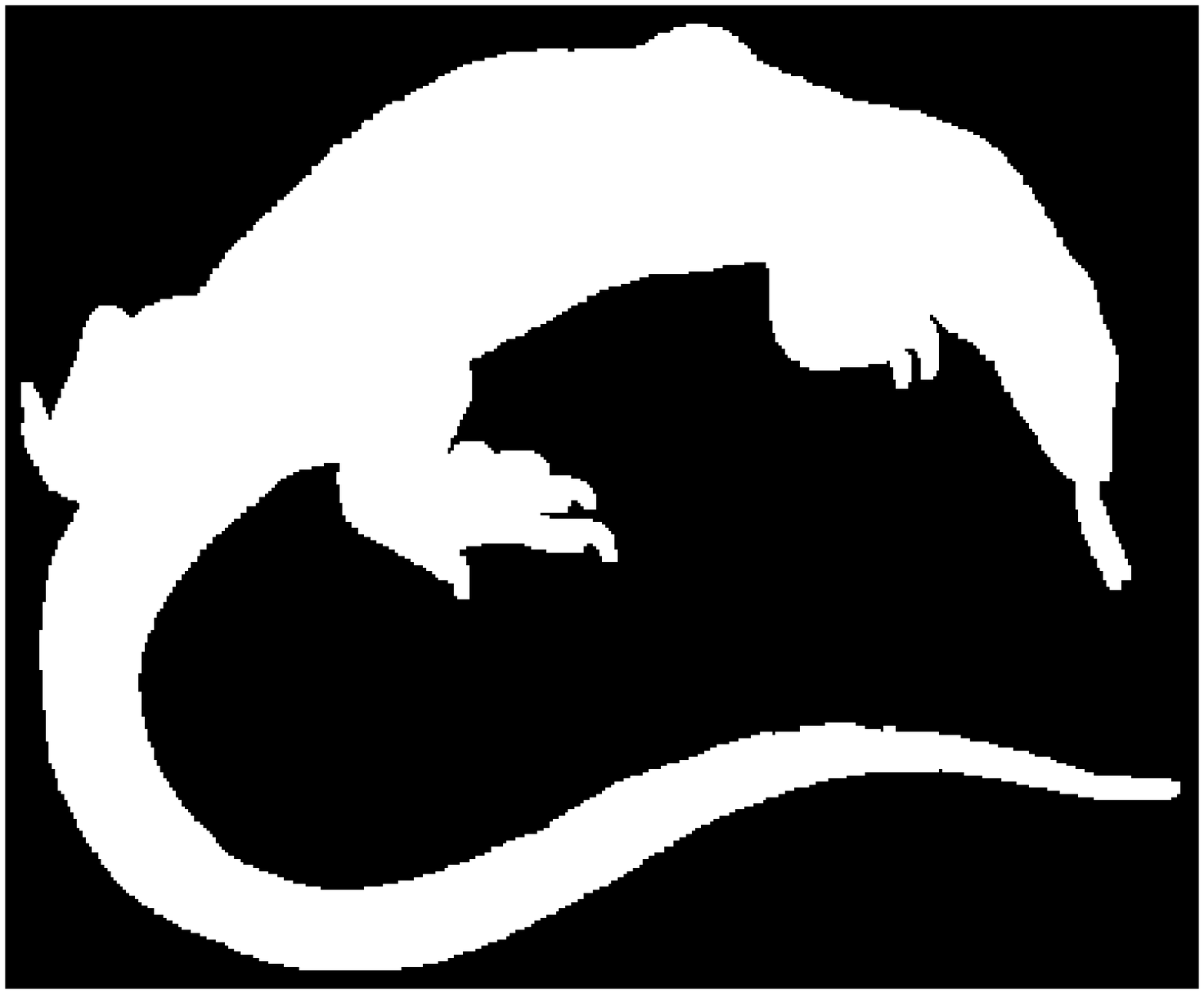}}
\subfloat{\includegraphics[trim=0.1cm 0cm 0.12cm 0cm, clip=true, width=1.3cm, height=1.5cm]{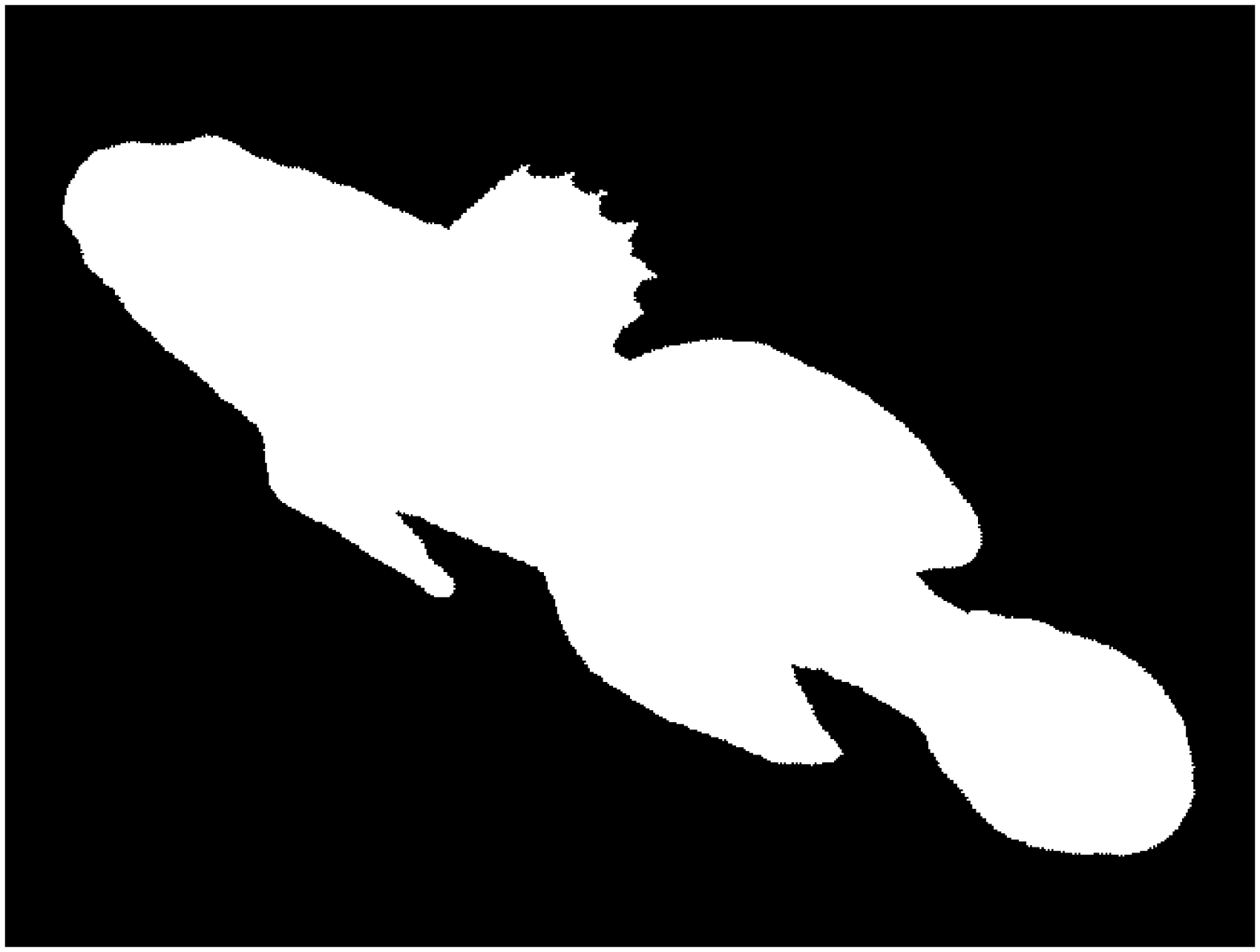}}
\subfloat{\includegraphics[trim=0.1cm 0cm 0.12cm 0cm, clip=true, width=1.3cm, height=1.5cm]{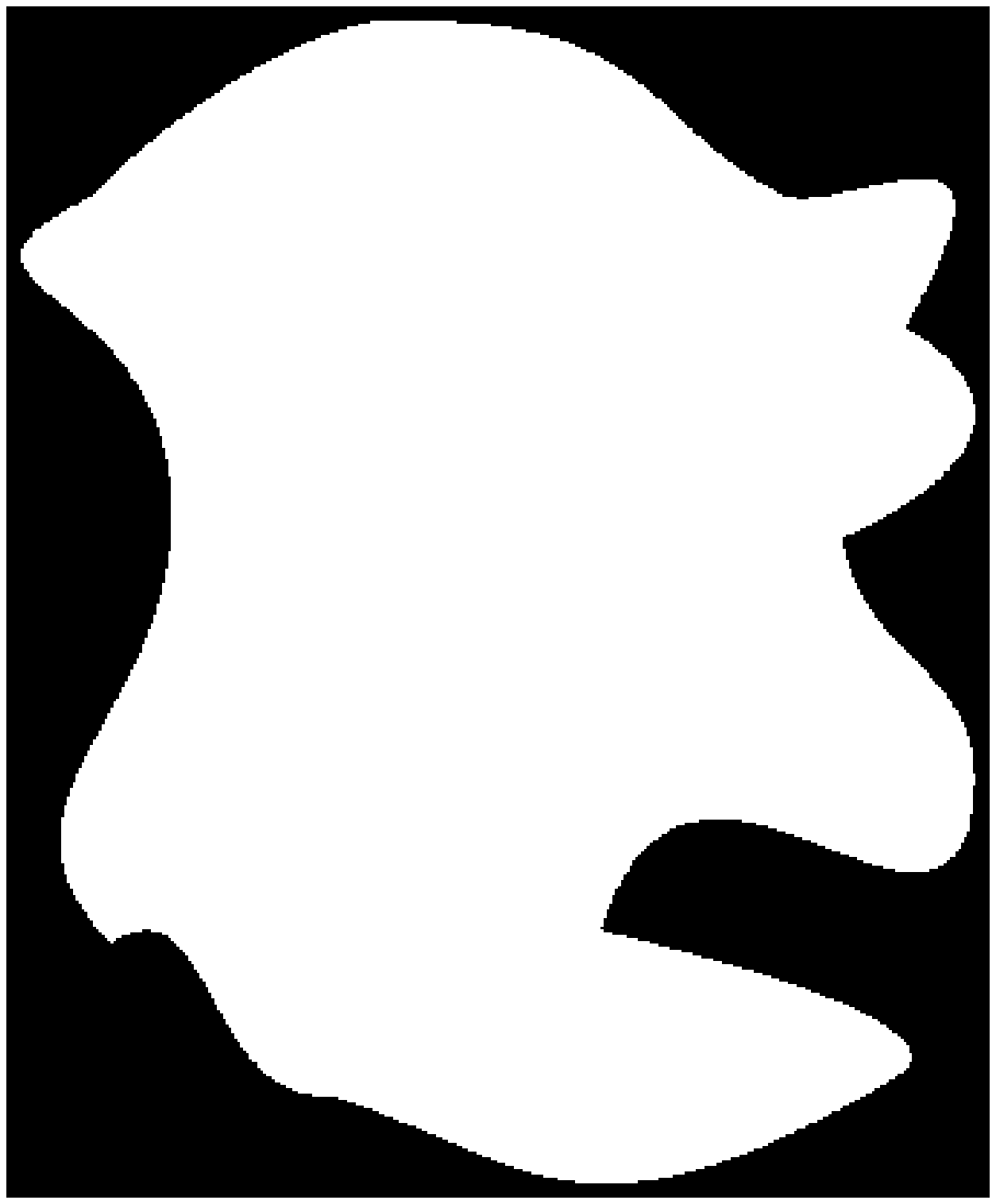}}
\subfloat{\includegraphics[trim=0.1cm 0cm 0.12cm 0cm, clip=true, width=1.3cm, height=1.5cm]{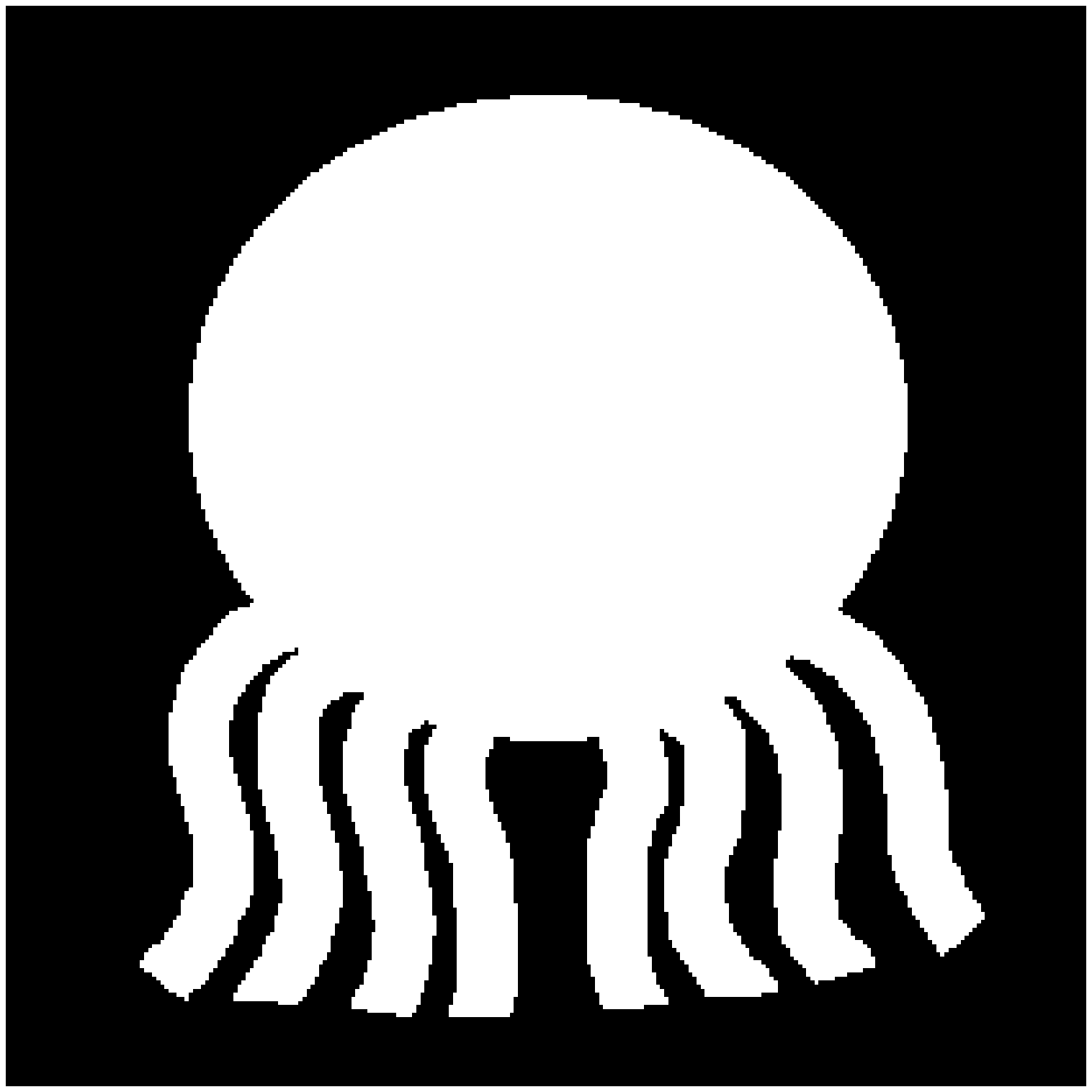}}
\subfloat{\includegraphics[trim=0.1cm 0cm 0.12cm 0cm, clip=true, width=1.3cm, height=1.5cm]{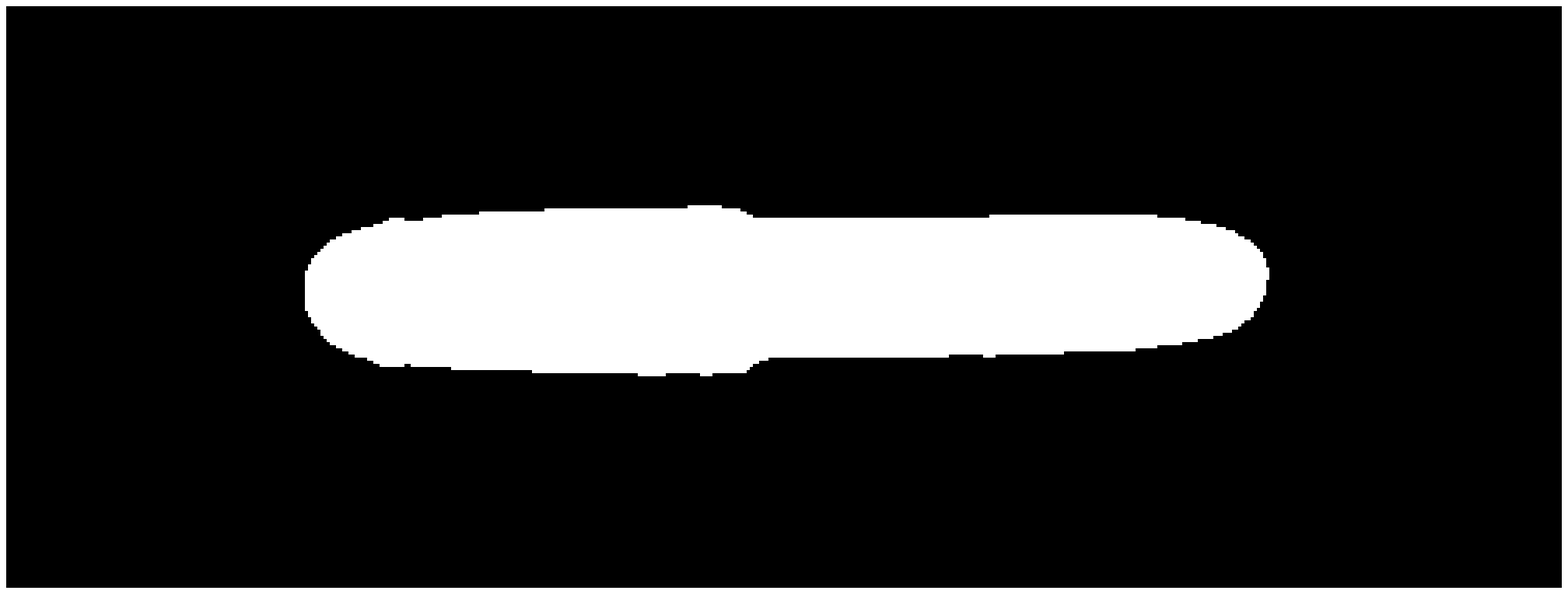}}
\subfloat{\includegraphics[trim=0.1cm 0cm 0.12cm 0cm, clip=true, width=1.3cm, height=1.5cm]{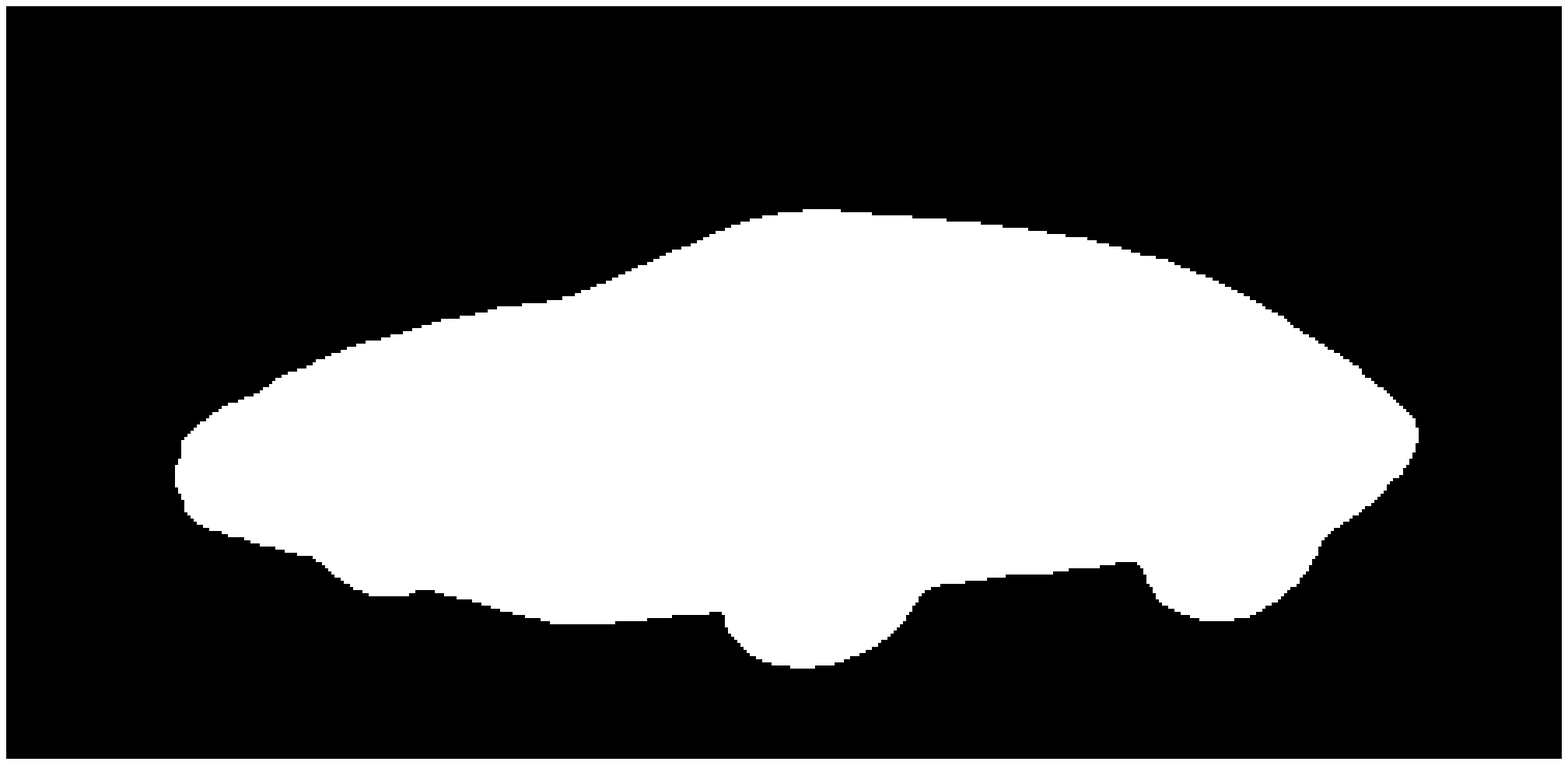}}
\subfloat{\includegraphics[trim=0.1cm 0cm 0.12cm 0cm, clip=true, width=1.3cm, height=1.5cm]{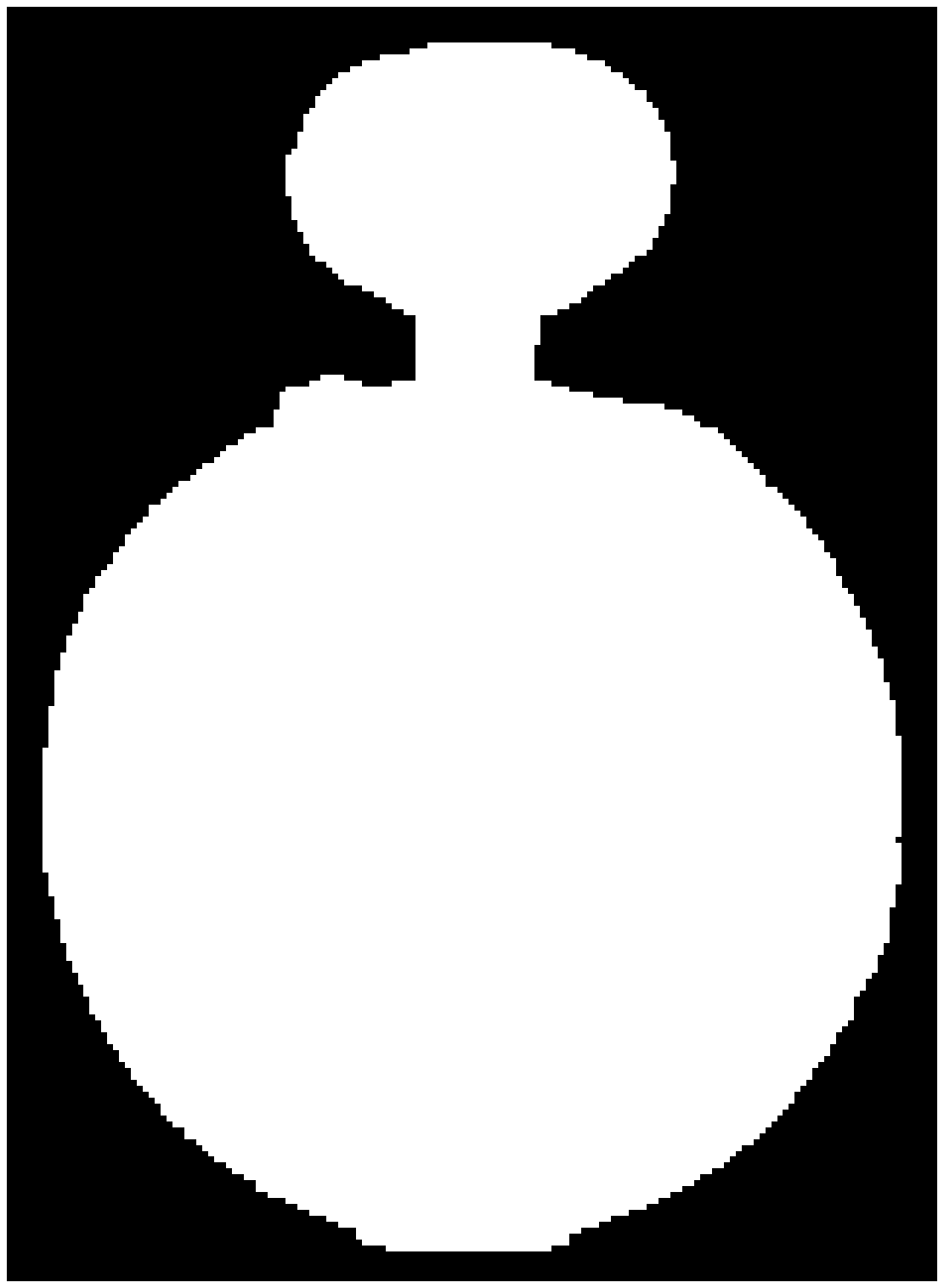}}
\subfloat{\includegraphics[trim=0.1cm 0cm 0.12cm 0cm, clip=true, width=1.3cm, height=1.5cm]{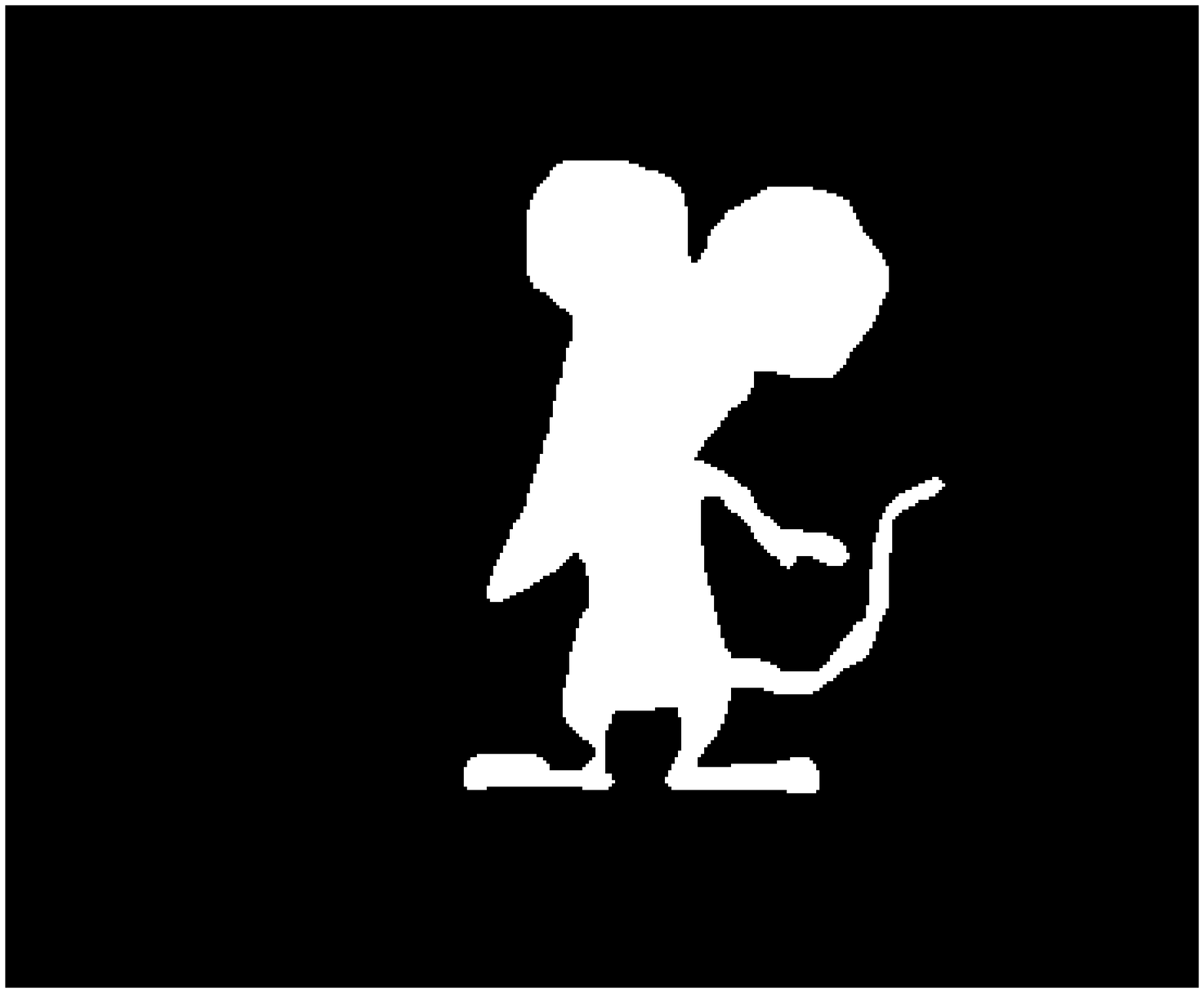}}
\subfloat{\includegraphics[trim=0.1cm 0cm 0.12cm 0cm, clip=true, width=1.3cm, height=1.5cm]{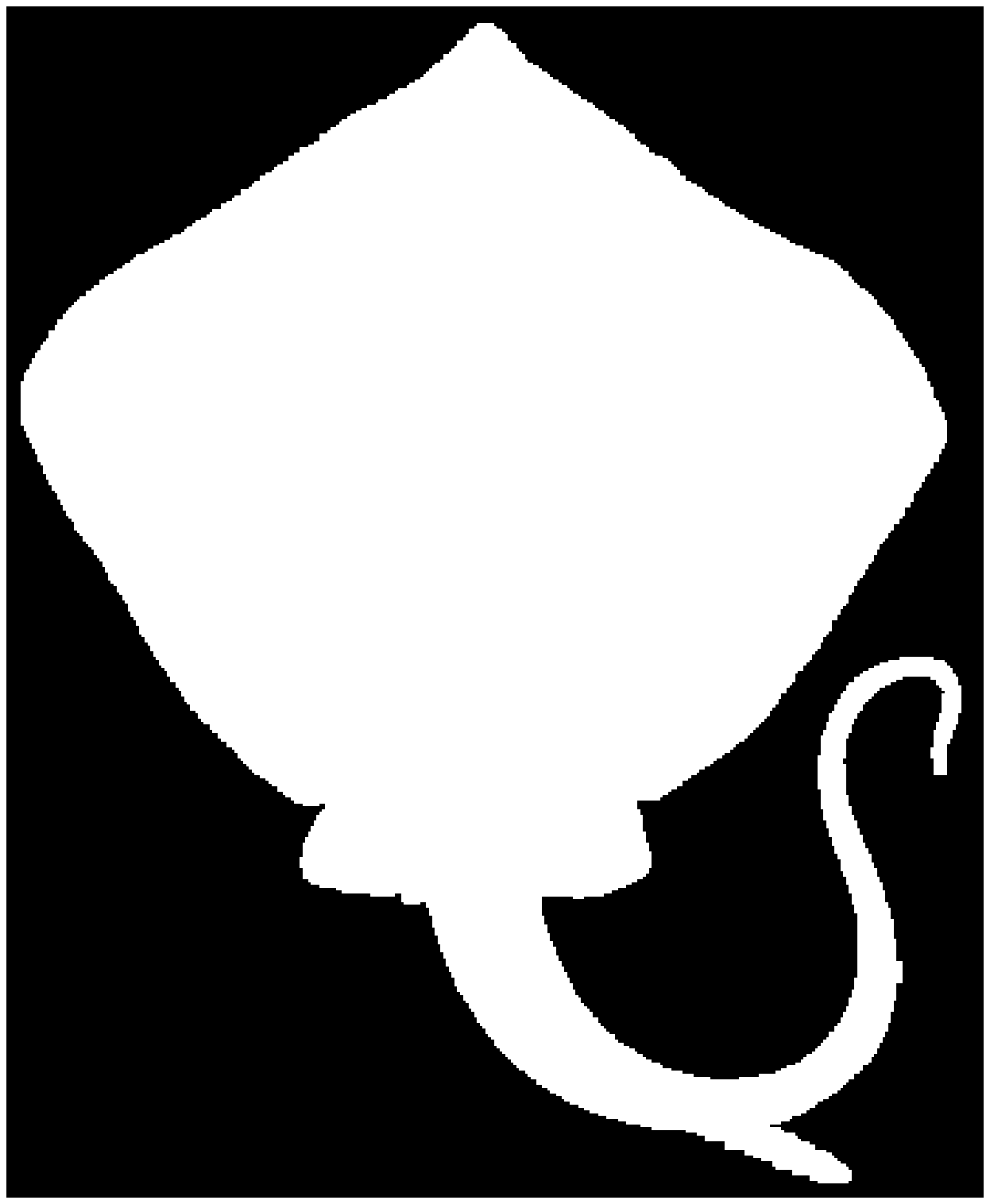}}\\

\subfloat{\includegraphics[trim=0.1cm 0cm 0.12cm 0cm, clip=true, width=1.3cm, height=1.5cm]{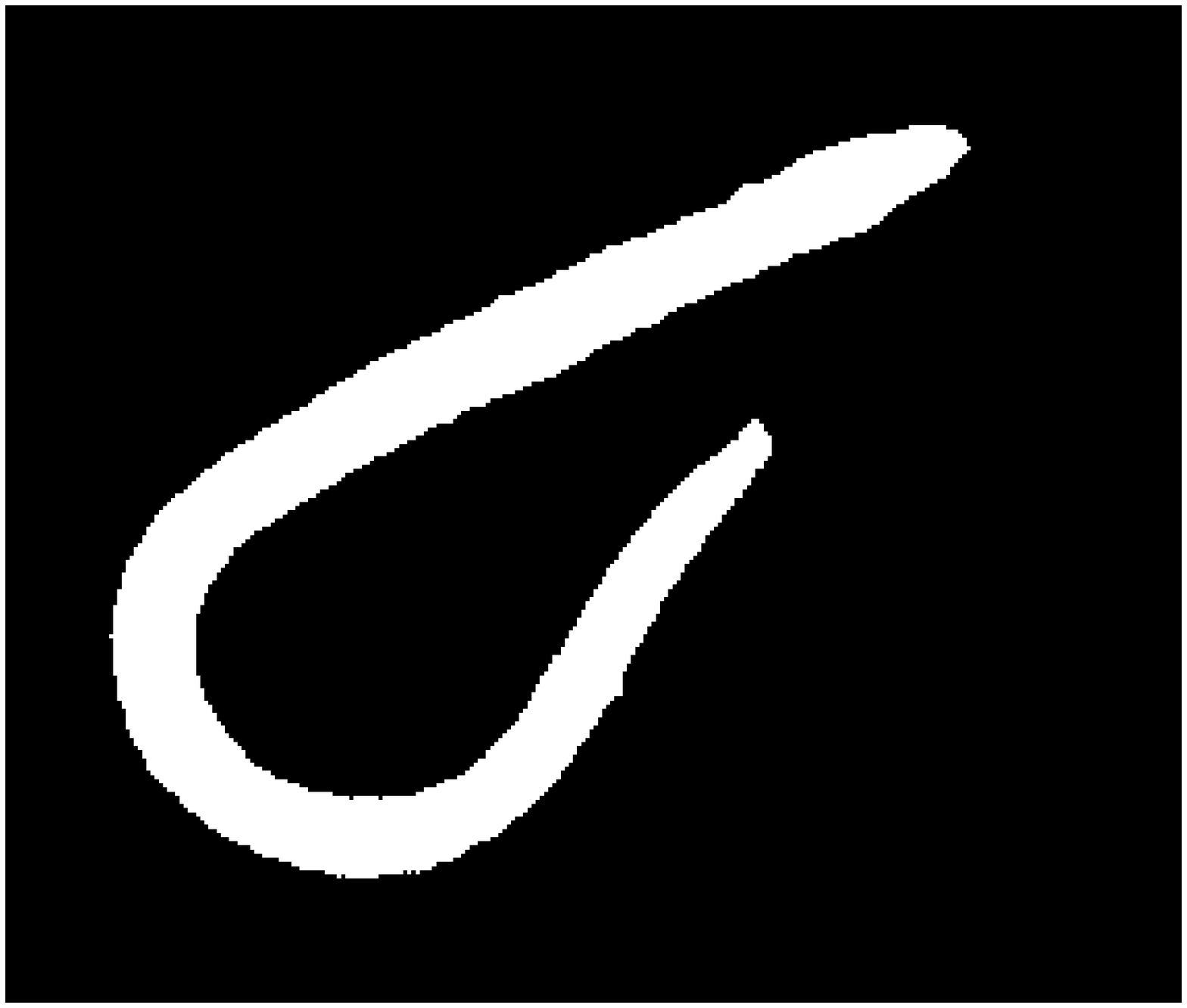}}
\subfloat{\includegraphics[trim=0.1cm 0cm 0.12cm 0cm, clip=true, width=1.3cm, height=1.5cm]{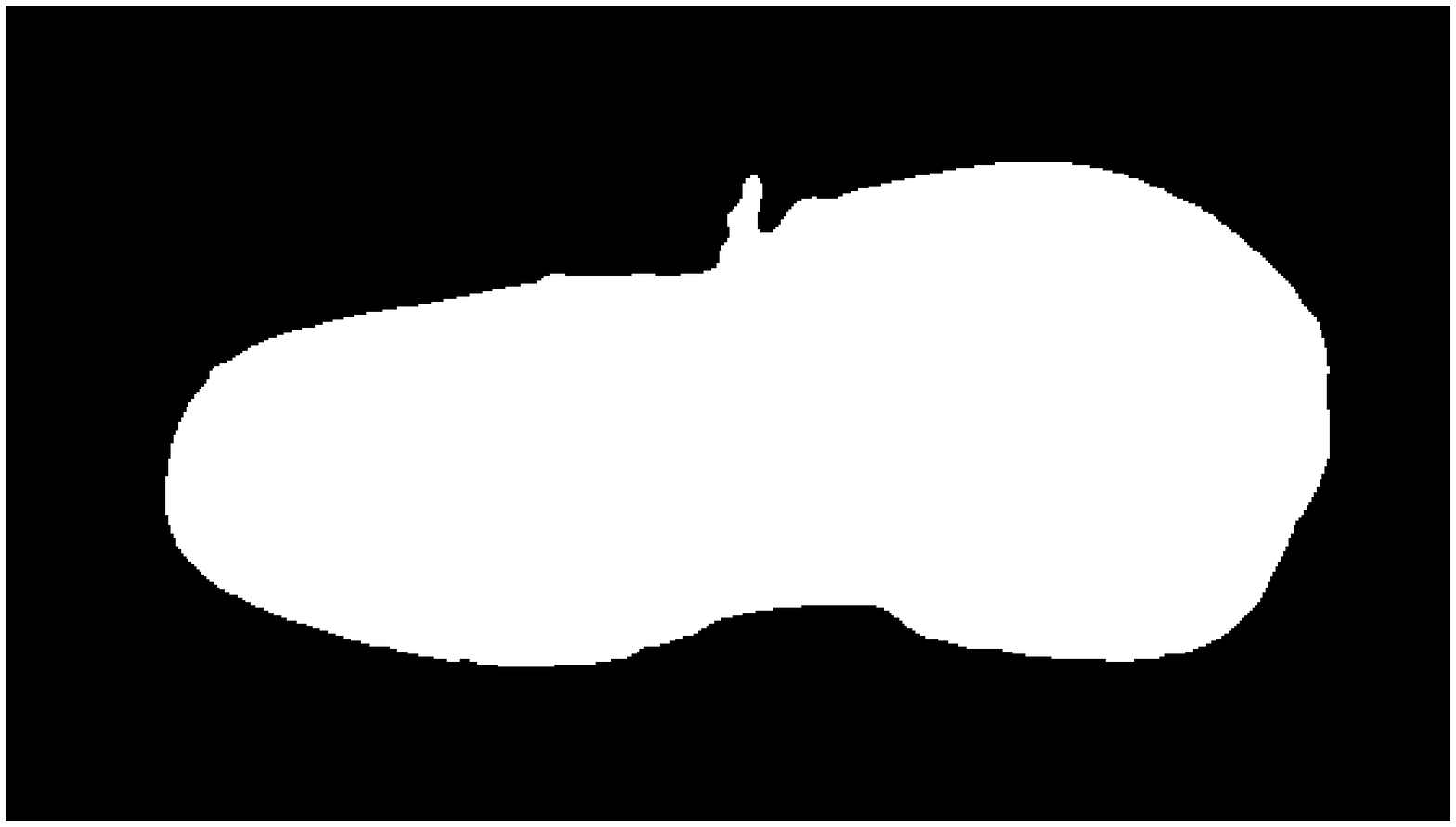}}
\subfloat{\includegraphics[trim=0.1cm 0cm 0.12cm 0cm, clip=true, width=1.3cm, height=1.5cm]{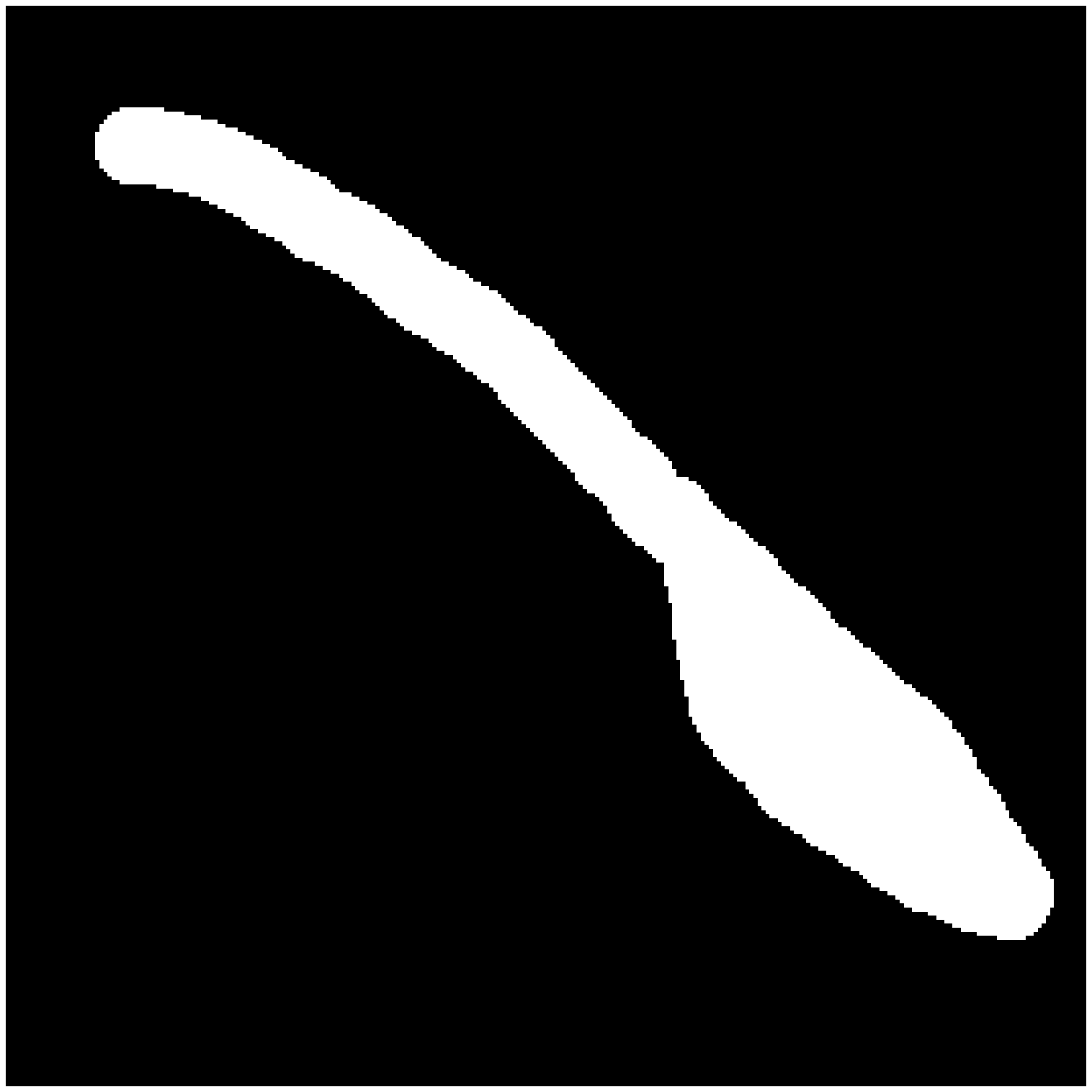}}
\subfloat{\includegraphics[trim=0.1cm 0cm 0.12cm 0cm, clip=true, width=1.3cm, height=1.5cm]{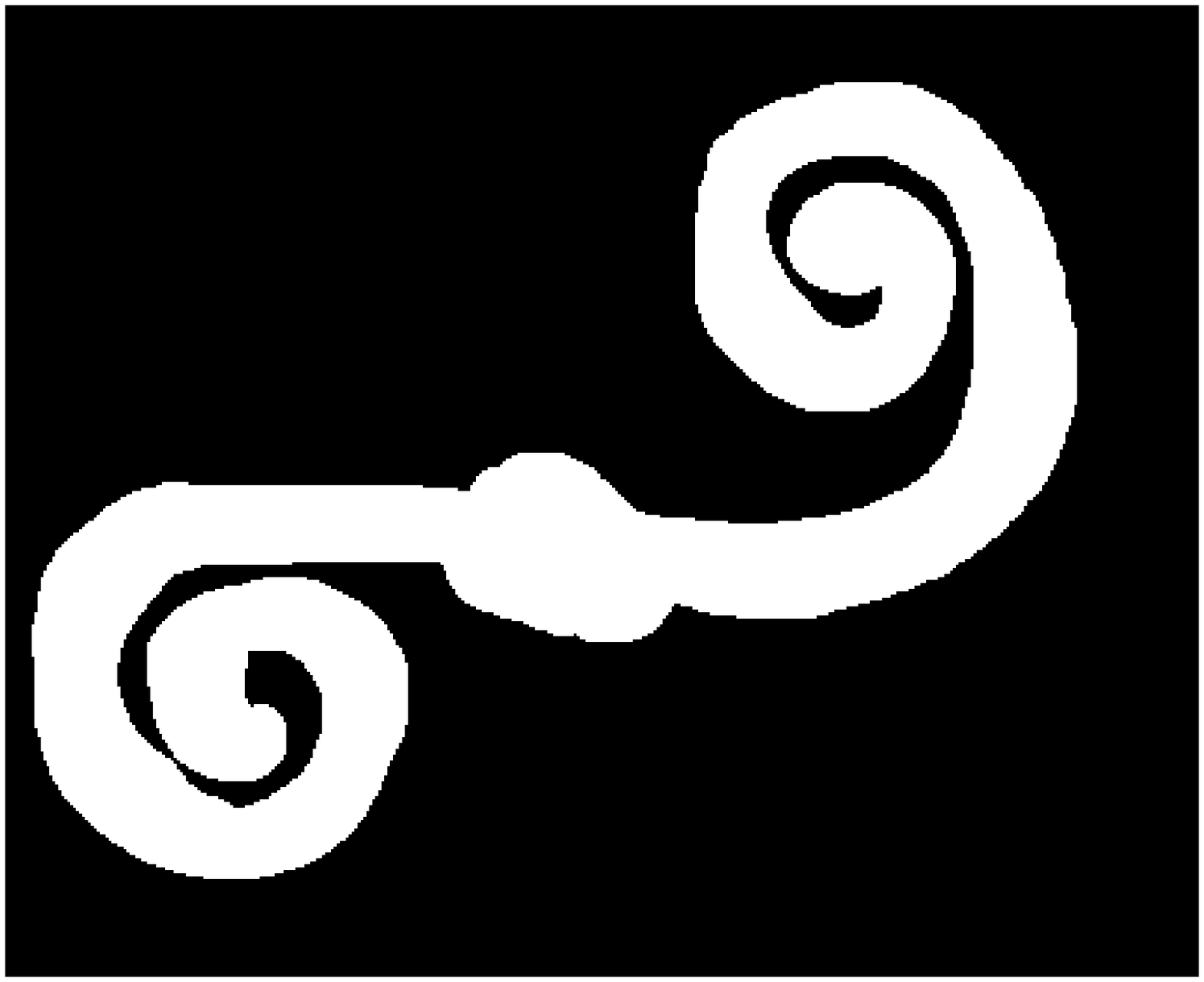}}
\subfloat{\includegraphics[trim=0.1cm 0cm 0.12cm 0cm, clip=true, width=1.3cm, height=1.5cm]{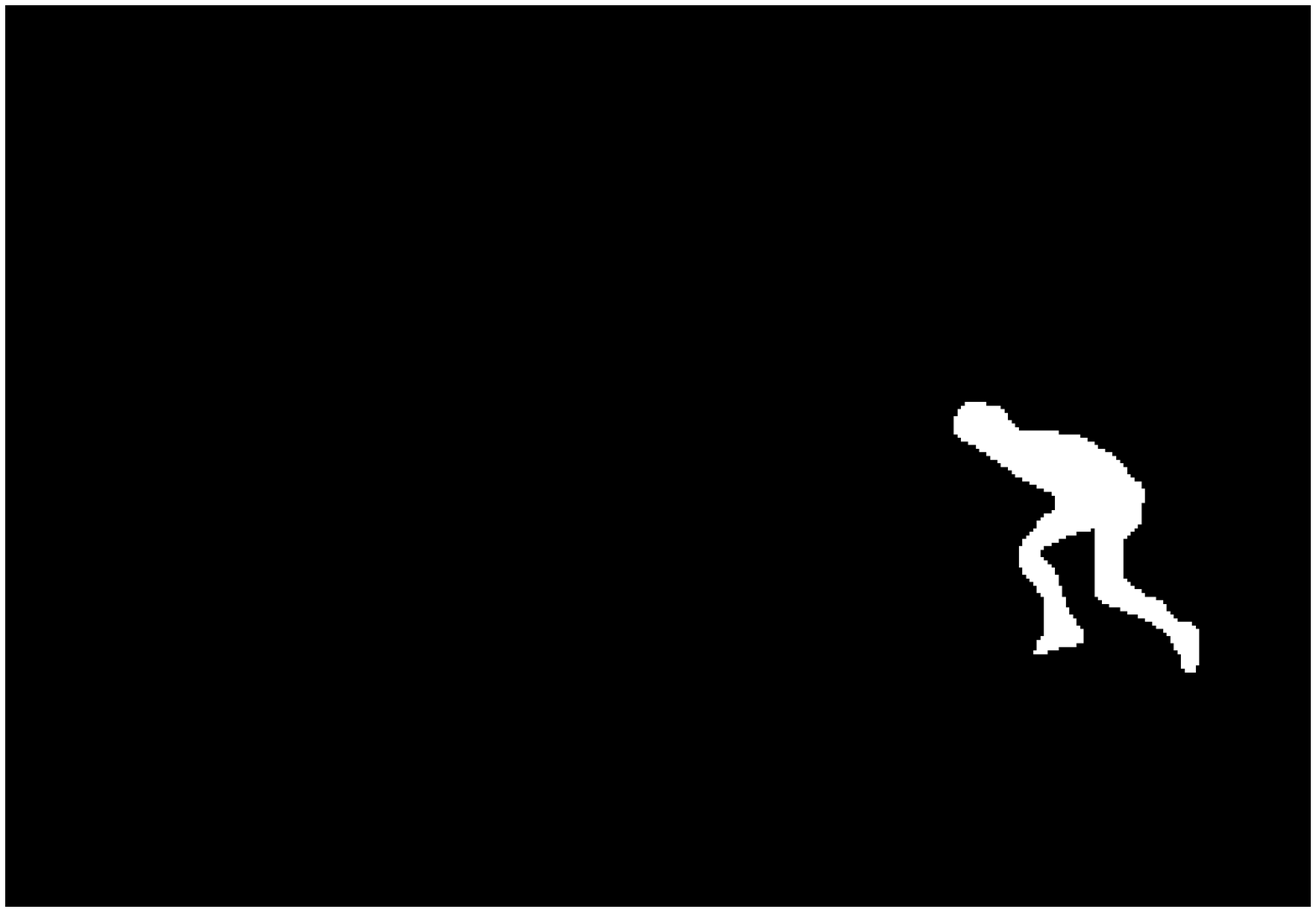}}
\subfloat{\includegraphics[trim=0.1cm 0cm 0.12cm 0cm, clip=true, width=1.3cm, height=1.5cm]{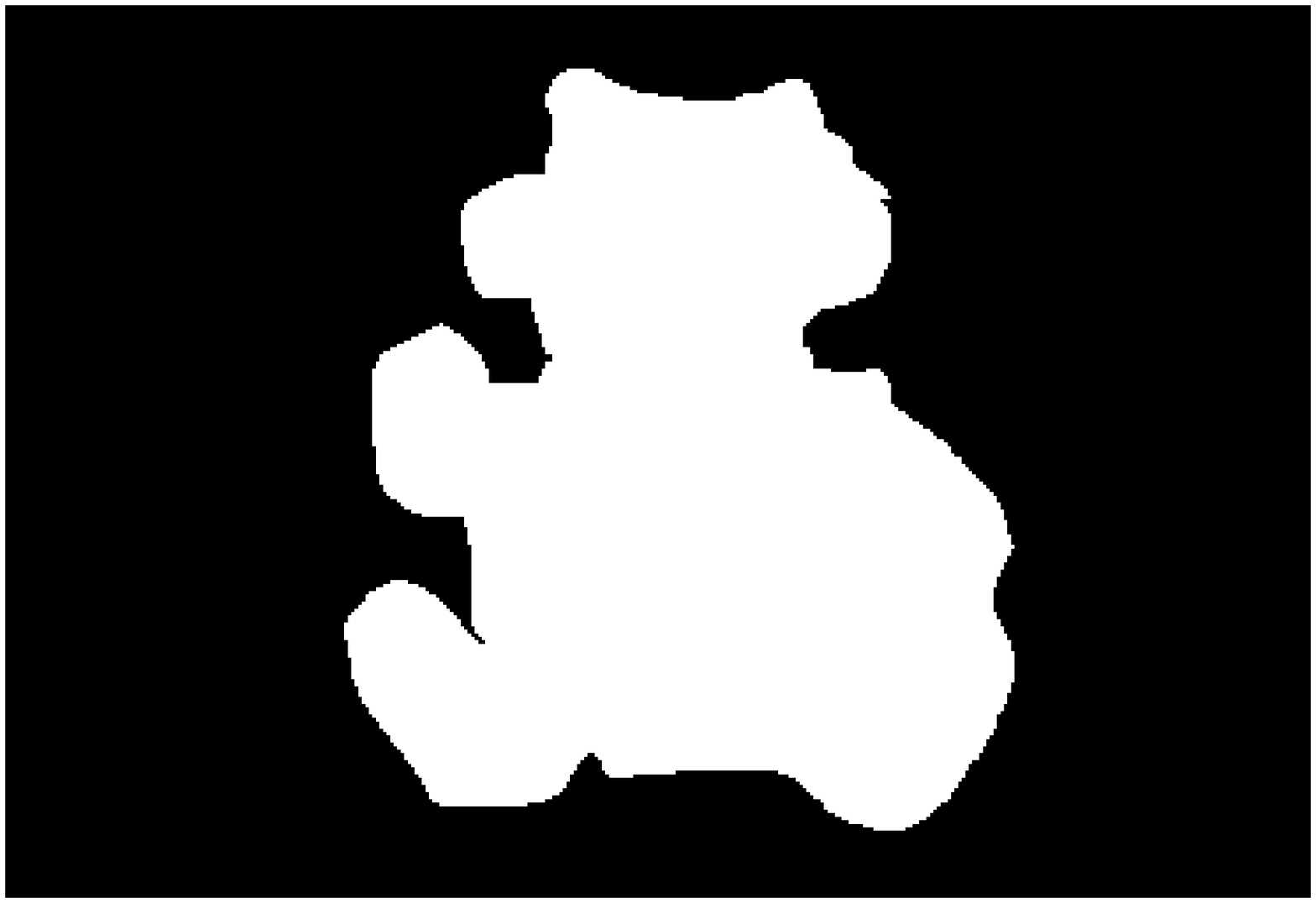}}
\subfloat{\includegraphics[trim=0.1cm 0cm 0.12cm 0cm, clip=true, width=1.3cm, height=1.5cm]{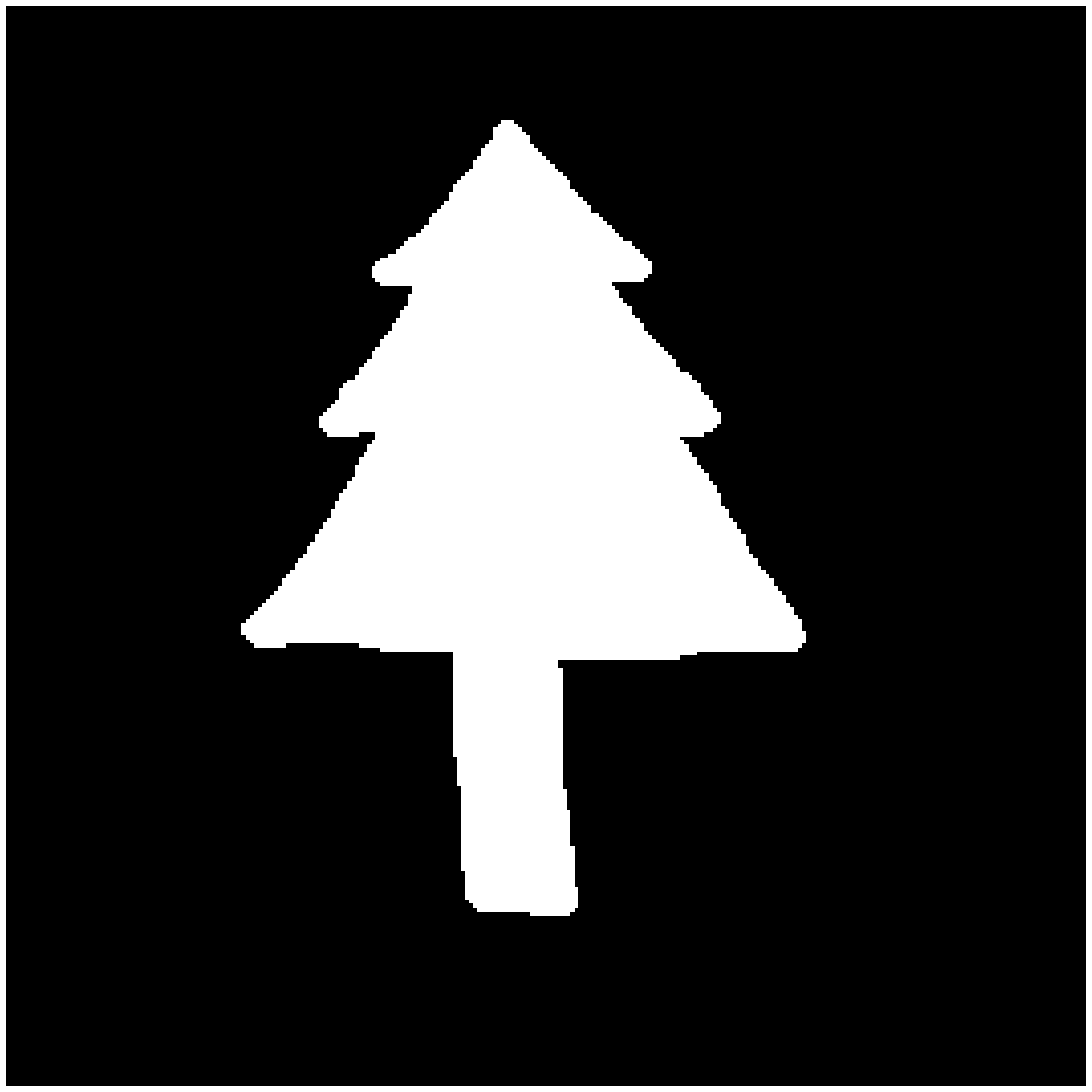}}
\subfloat{\includegraphics[trim=0.1cm 0cm 0.12cm 0cm, clip=true, width=1.3cm, height=1.5cm]{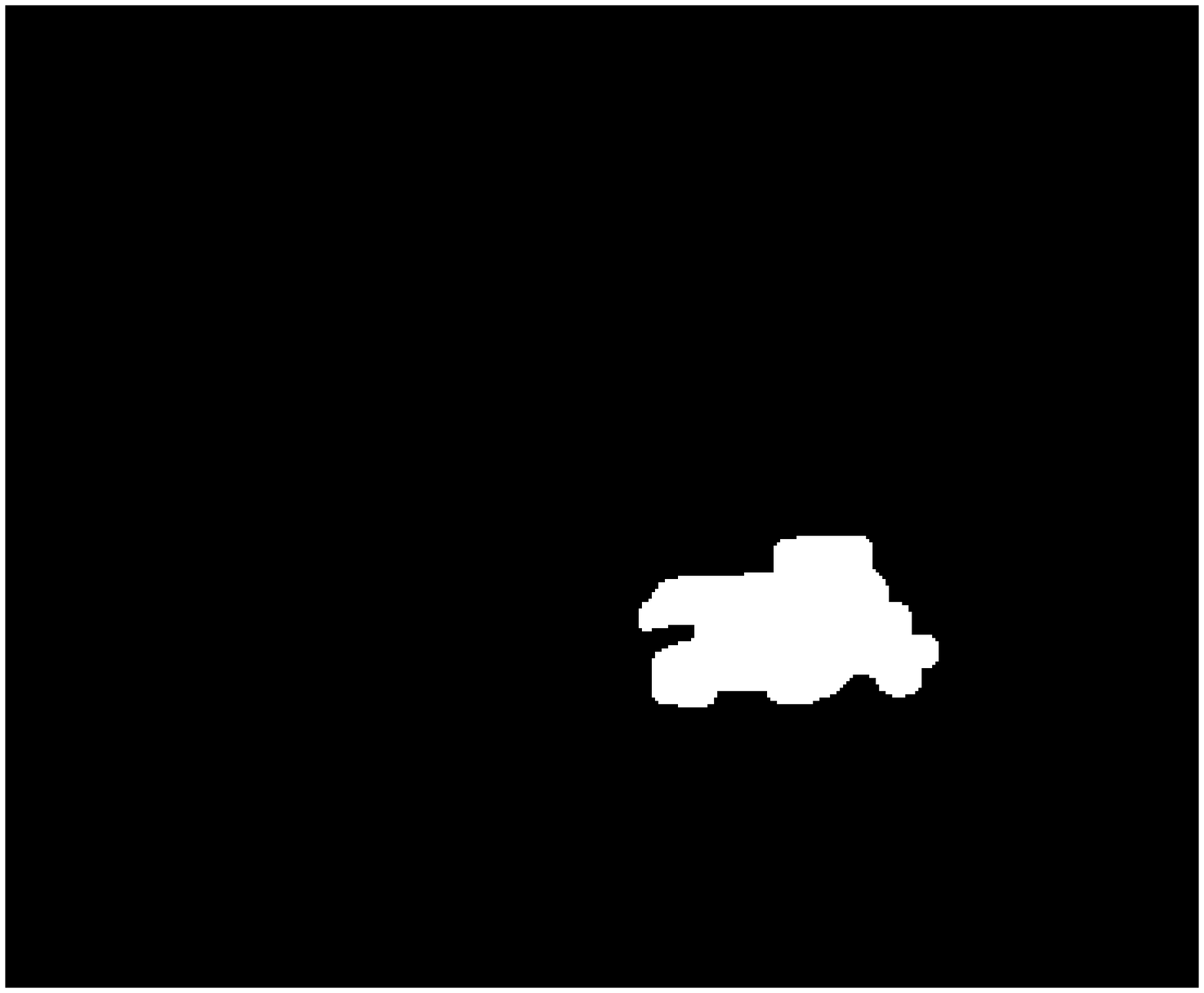}}
\subfloat{\includegraphics[trim=0.1cm 0cm 0.12cm 0cm, clip=true, width=1.3cm, height=1.5cm]{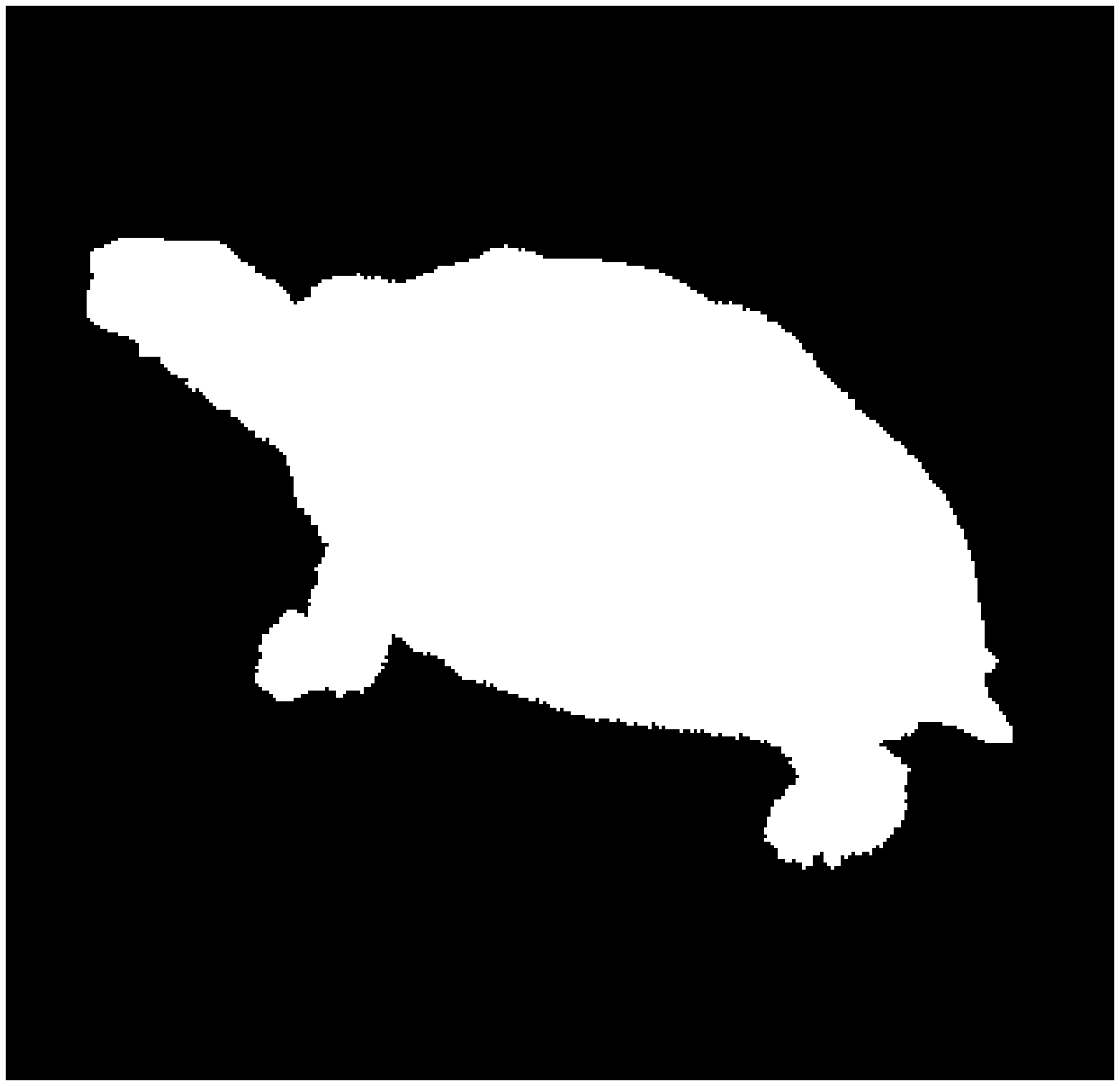}}
\subfloat{\includegraphics[trim=0.1cm 0cm 0.12cm 0cm, clip=true, width=1.3cm, height=1.5cm]{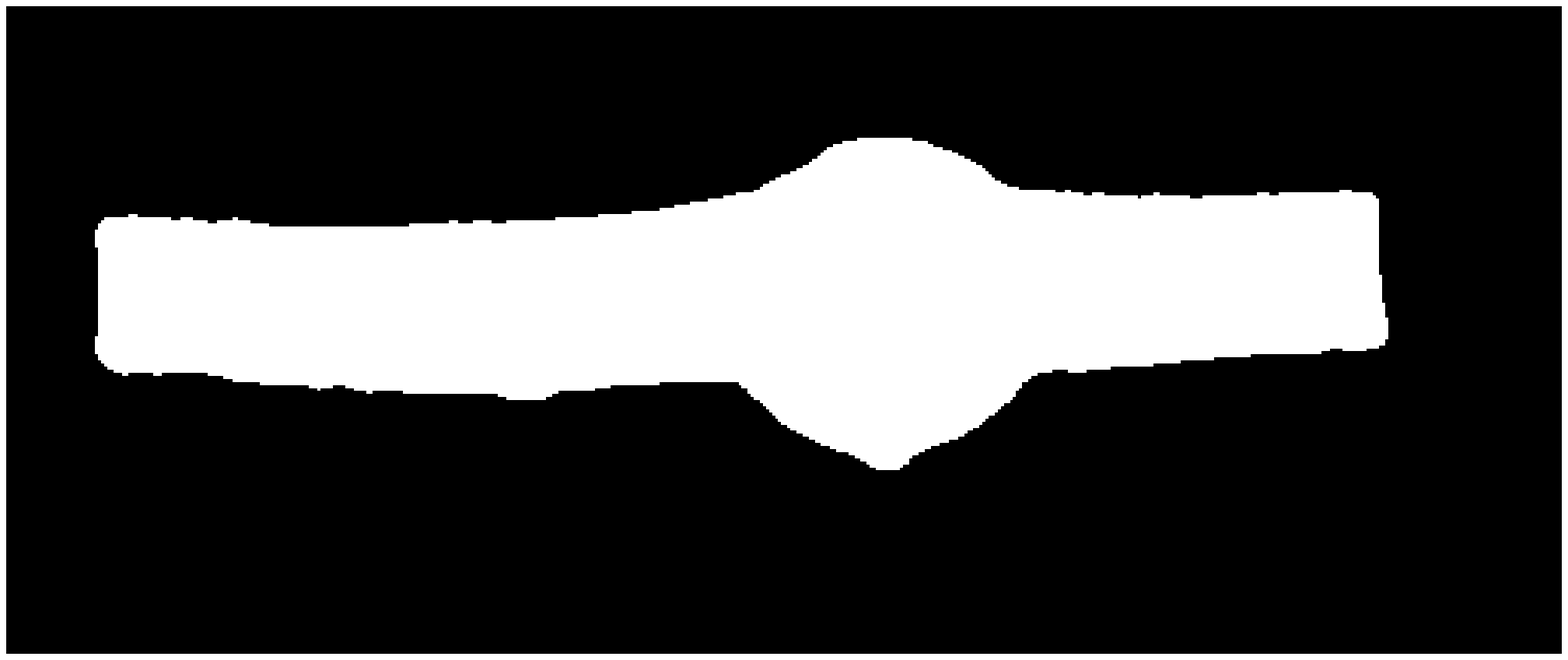}}\\
   \caption{The figure shows example images from the MPEG7 Database. Shown above is an example from each class. As can be seen, the database consists of images from both rigid and non-rigid objects. }
\label{fig:mpeg7}
\end{figure}

We use the well-known, and widely used, MPEG7 CE-Shape-1 Part B dataset for testing our algorithm. The database consists of silhouettes of $1400$ images with a wide variety among them. The database is split into $70$ classes, with each class containing $20$ example images. The database consists of both rigid and non-rigid objects. The objects in the database have varied levels of translations, rotations, scales, articulations, deformations and occlusions. The objects belonging to a particular class are not only similar by the contour properties, but also by their overall visual similarity. The database is considered as a challenging database as there are many instances where the inter-class object similarity is more than intra-class object similarity. Figure \ref{fig:mpeg7} shows an example object from each of the $70$ classes.

The performance of the algorithm on the database is measured by the Bullseye score. To calculate the Bullseye score, each image is compared to every other image in the database. The top $40$ best-matching images are retained, of which at most $20$ images can belong to the same class. Of the top $40$ best matches, the number of objects belonging to the same class as the template image are counted. This number is divided by $20$ to get the Bullseye score for the template image under consideration. The average Bullseye score over all the images in the database gives the Bullseye score for the complete database.

In our experiments, we set $N_\mathcal{B} = 100$ for triangulating the shape, $N_{SP}$ to $300$, $N_{DP}$ to $2000$, and $\alpha$ to 4. The mean of the costs obtained from IDSC was about four times the mean of all the SSC costs. The range of values was also smaller than the range of values from IDSC. Hence, the choice of $\alpha = 4$.
We set $N_{SP}$ to 300 as previous works \cite{ling2010balancing} have used 300 feature points for shape comparison. Minor improvements were seen in the Bullseye score for $N_{SP} > 300$.
No major decrease in performance was seen for $N_{SP} < 300$.
A coarse representation of the shape is sufficient for interior sampling. With $N_\mathcal{B} = 50$, the overall shape boundary was not decipherable for some highly convoluted shapes. Thus, we increased it to 100. With further increases such as \{200, 300, 400\}, we did not see any major improvement in the overall results.
We also tried experimenting with larger values of $N_{DP}$ ($2500$, $3000$, $3500$ and $4000$), but did not find any significant improvement in the Bullseye score. Table \ref{tab:bullseye} lists our Bullseye score along with the Bullseye scores for various algorithms.

\begin{table}
  \centering
    \vspace{-6ex}
  \begin{tabular}{c|c}
  \hline
  \hline
  Algorithm & Bullseye Score \\
  \hline
  Visual Parts \cite{latecki2000shape} & 76.45\% \\
  SC+TPS \cite{belongie2002shape} & 76.51\% \\
  Generative Model \cite{tu2004shape} & 80.03\% \\
  Curvature Scale Space \cite{mokhtarian2003curvature} & 81.12\% \\
  SSC & 82.39\%\\
  Polygonal Multiresolution \cite{attalla2005robust} & 84.33\%\\
  Multiscale Representation \cite{adamek2004multiscale} & 84.93\% \\
  IDSC \cite{ling2007shape} & 85.40\% \\
  Symbolic Representation \cite{daliri2008robust} & 85.92\%\\
  Hierarchical Procrustes Matching \cite{mcneill2006hierarchical} & 86.35\% \\
  IDSC(EMD) \cite{ling2007efficient} & 86.53\% \\
  Triangle Area \cite{alajlan2008geometry} & 87.23\% \\
  Shape Tree \cite{felzenszwalb2007hierarchical} & 87.70\% \\
  ASC \cite{ling2010balancing} & 88.30\% \\
  IDSC+AspectNorm.+StrandRemoval \cite{temlyakov2010two} & 88.39\% \\
  Contour Flexibility \cite{xu20092d} & 89.31\% \\
  IDSC+PMMS \cite{Hu20123222} & 90.18\% \\
  IDSC+LP \cite{yang2008improving} & 91.00\% \\
  \textbf{IDSC+SSC} & \textbf{91.65\%} \\
  \textbf{IDSC+AspectNorm.+SSC} & \textbf{91.83\%} \\
  IDSC+LCDP \cite{yang2009locally} & 92.36\% \\
  IDSC+Affine Normalization \cite{gopalan2010articulation} & 93.67\% \\
  IDSC+AspectNorm.+StrandRemoval+LCDP \cite{temlyakov2010two}\cite{yang2009locally} & 95.60\% \\
  ASC+LCDP \cite{ling2010balancing}\cite{yang2009locally} & 95.96\% \\
  IDSC+PMMS+LCDP \cite{Hu20123222}\cite{yang2009locally} & 98.56\% \\
  \textbf{IDSC+SSC+LCDP} & \textbf{98.85\%}\\
  IDSC+Affine Normalization+TPG \cite{yang2012affinity} & 99.99\% \\
  \hline
\end{tabular}
  \vspace{-2ex}
  \caption{{\small The table gives a comprehensive list of shape-matching techniques proposed in the literature, along with their respective Bullseye scores. We can see that our method helps in significantly improving the Bullseye score when fused with IDSC. Diffusion techniques, such as LCDP, further improve our Bullseye score.}}\label{tab:bullseye}
\end{table}

As can be seen from Table \ref{tab:bullseye}, quite a lot of work has been done in the area of shape matching. We fuse the costs from our algorithm with the costs from IDSC. Doing so significantly improves the Bullseye score from $85.40\%$, to $91.65\%$. The objects in the MPEG7 database have different aspect ratios as well. Performing aspect normalization of shapes, as in \cite{temlyakov2010two}, helps improve the Bullseye score further, to $91.83\%$. Temlyakov et al. \cite{temlyakov2010two} perform a similar fusion, and their algorithm helps improve the Bullseye score to $88.39\%$, while Hu et al.'s \cite{Hu20123222} method improves the score to $90.18\%$. We specifically compare our algorithm to these two methods as they are also perceptually motivated techniques. We would like to mention that the method in \cite{temlyakov2010two} requires the setting of threshold parameters for the identification of strand structures. Also, the method in \cite{Hu20123222} requires the selection of an appropriate structuring element and the identification of a proper scale at which to perform the morphological closing operation. Our method can help achieve a better Bullseye score without the requirement of such additional parameters.

\begin{figure}
  \centering
  \subfloat[]{\label{fig:individual}\includegraphics[trim=3.5cm 0.3cm 0.8cm 0.5cm, clip=true, width=15cm, height=5cm]{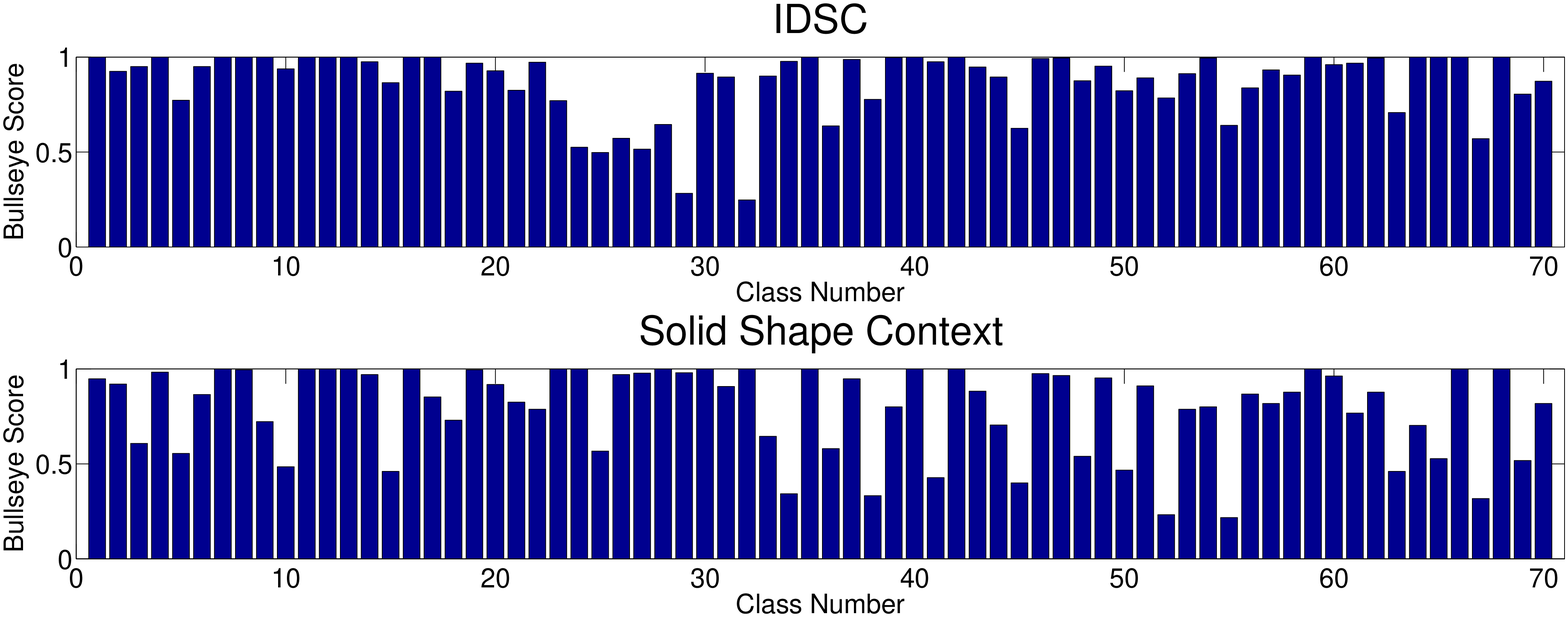}}

  \subfloat[]{\label{fig:diffHist}\includegraphics[trim=3.5cm 0.3cm 0.8cm 1cm, clip=true, width=15cm, height=3.5cm]{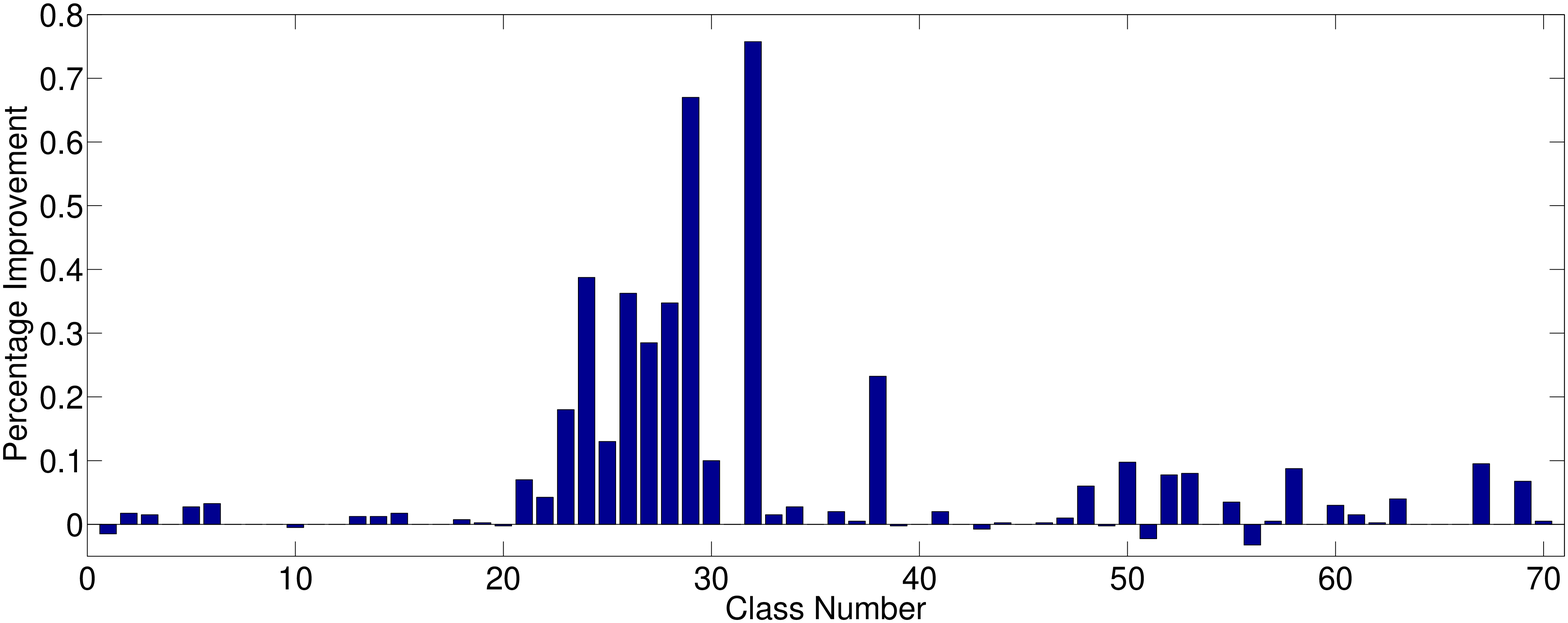}}

  \subfloat[]{\label{fig:together}\includegraphics[trim=3.5cm 0.3cm 0.8cm 0cm, clip=true, width=15cm, height=3.5cm]{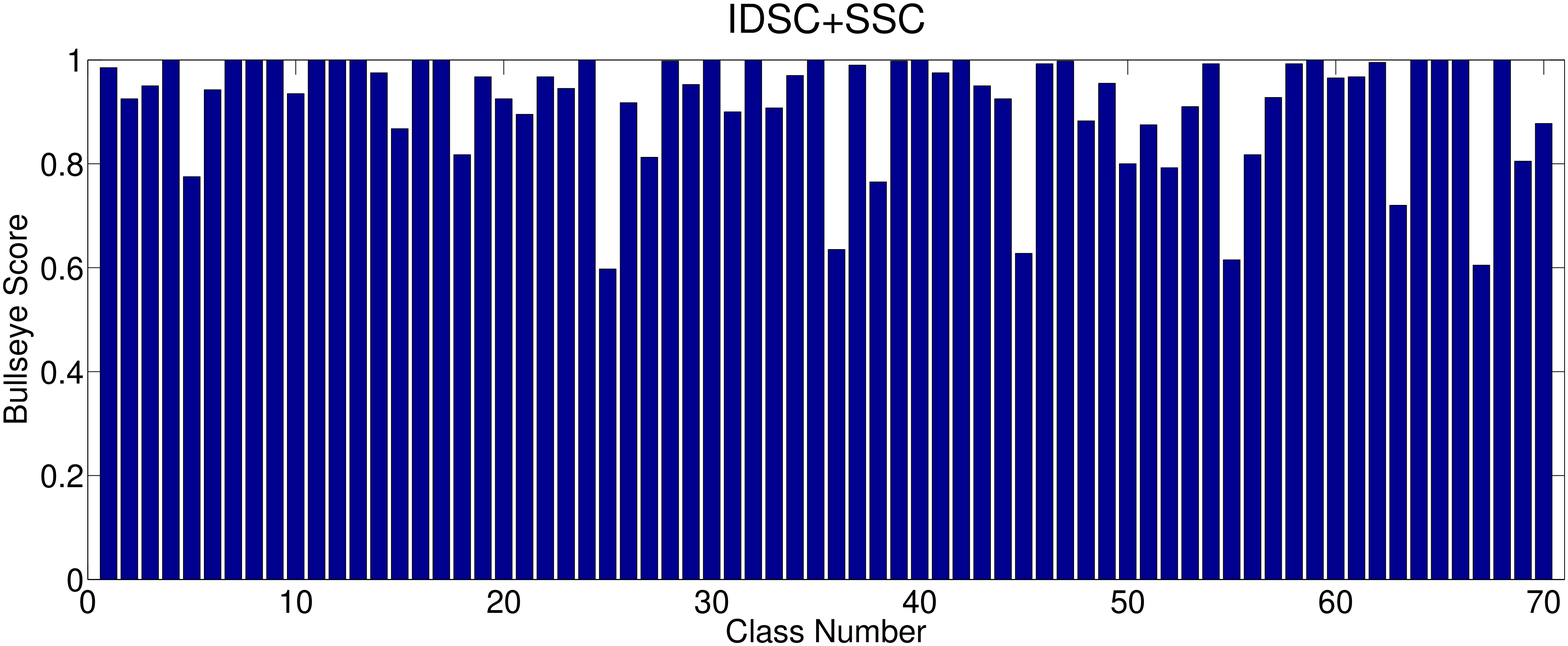}}

  \caption{(a) Subfigure shows the class-specific Bullseye score for IDSC (top) and SSC (bottom). We can see that SSC complements IDSC. SSC performs better than IDSC for classes 21 through 32, while IDSC performs better than SSC for some other classes. (b) Subfigure shows the percentage gain in Bullseye score for each class, when SSC is fused with IDSC, over IDSC alone. We can see a significant improvement in the Bullseye score for the classes 21 through 32, which correspond to classes with visually similar objects, but having many indents in their contours. (c) Class-specific Bullseye score for IDSC+SSC. The bar chart shows a much more evened out score among all the classes.}
\end{figure}

Figure \ref{fig:individual} shows class-specific Bullseye scores for both IDSC and SSC. We can see that the SSC performs better than IDSC for classes 21 through 32. These are the classes where there are a lot of indentations in the objects. Also, IDSC performs better than SSC in some other classes. These classes correspond to the the classes of articulating objects. Ex: Lizard, Octopus, etc. From the bar chart, we can see that SSC complements IDSC well. Figure \ref{fig:diffHist} shows the class-specific gain in Bullseye score when SSC is fused with IDSC, over IDSC alone. We can see a significant gain in the Bullseye score for a number of classes. Most of the classes that have a gain correspond to the classes where the objects have an overall visual similarity. Many objects in these classes have a number of indentations in their contours. Figure \ref{fig:together} shows the class-specific Bullseye score when IDSC is fused with SSC. We can see a much more evened out score among all the classes.

\begin{figure*}
  \centering

  \subfloat[IDSC]{\shortstack{\includegraphics[trim=0.1cm 0cm 0.12cm 0cm, clip=true, width=1.2cm, height=1.2cm]{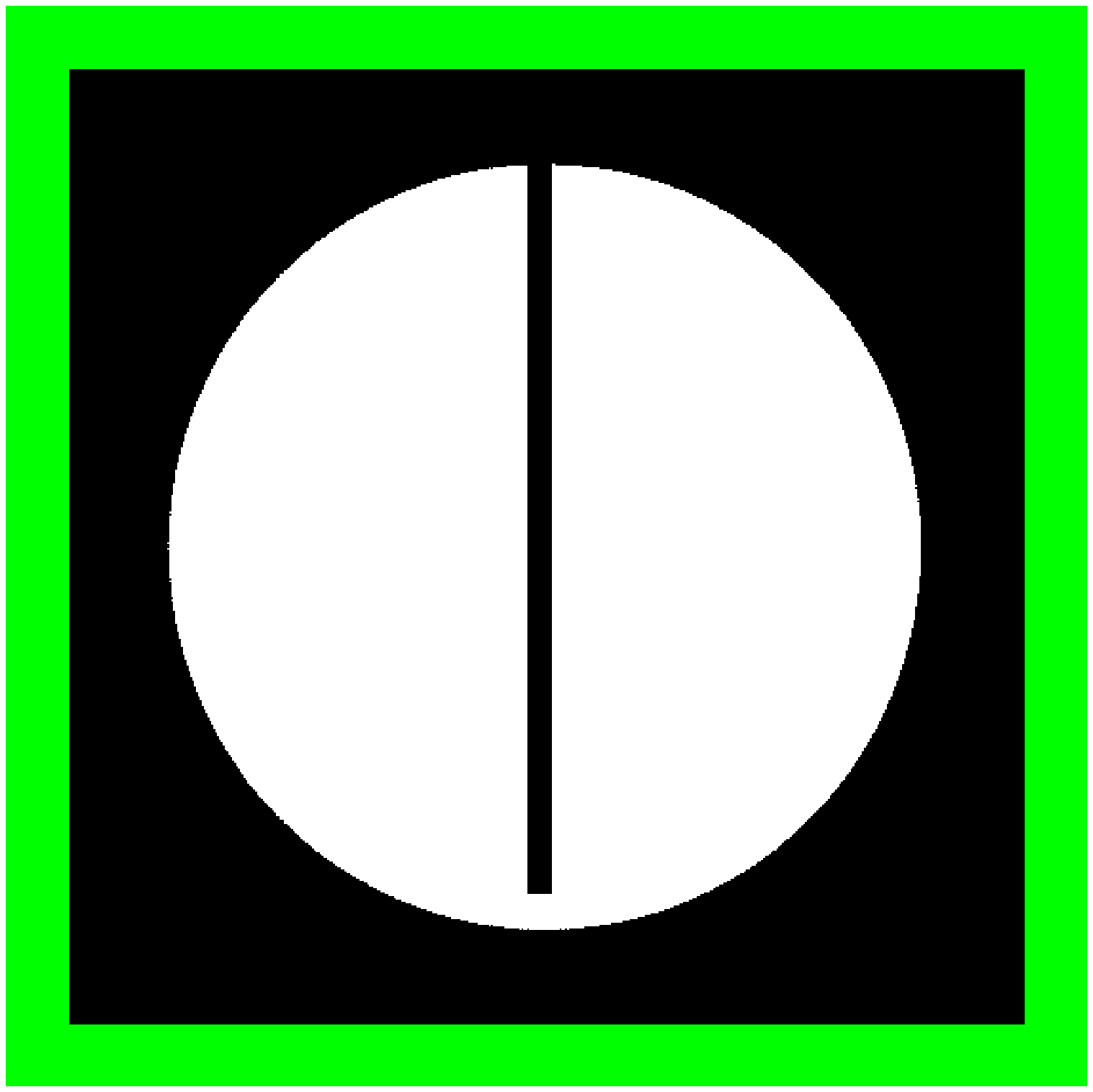}\hspace{-0.215em}
\includegraphics[trim=0.1cm 0cm 0.12cm 0cm, clip=true, width=1.2cm, height=1.2cm]{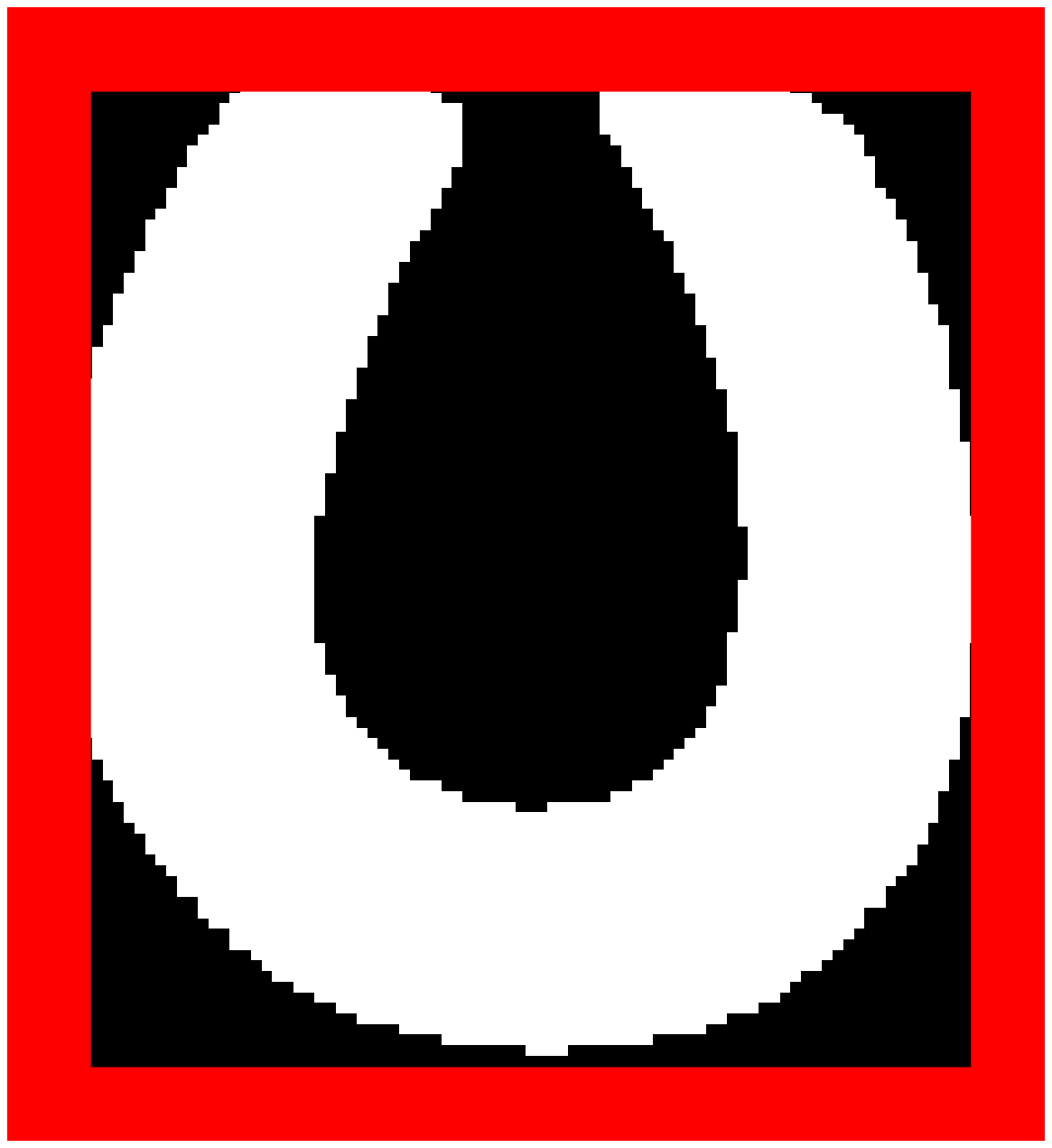}\hspace{-0.215em}
\includegraphics[trim=0.1cm 0cm 0.12cm 0cm, clip=true, width=1.2cm, height=1.2cm]{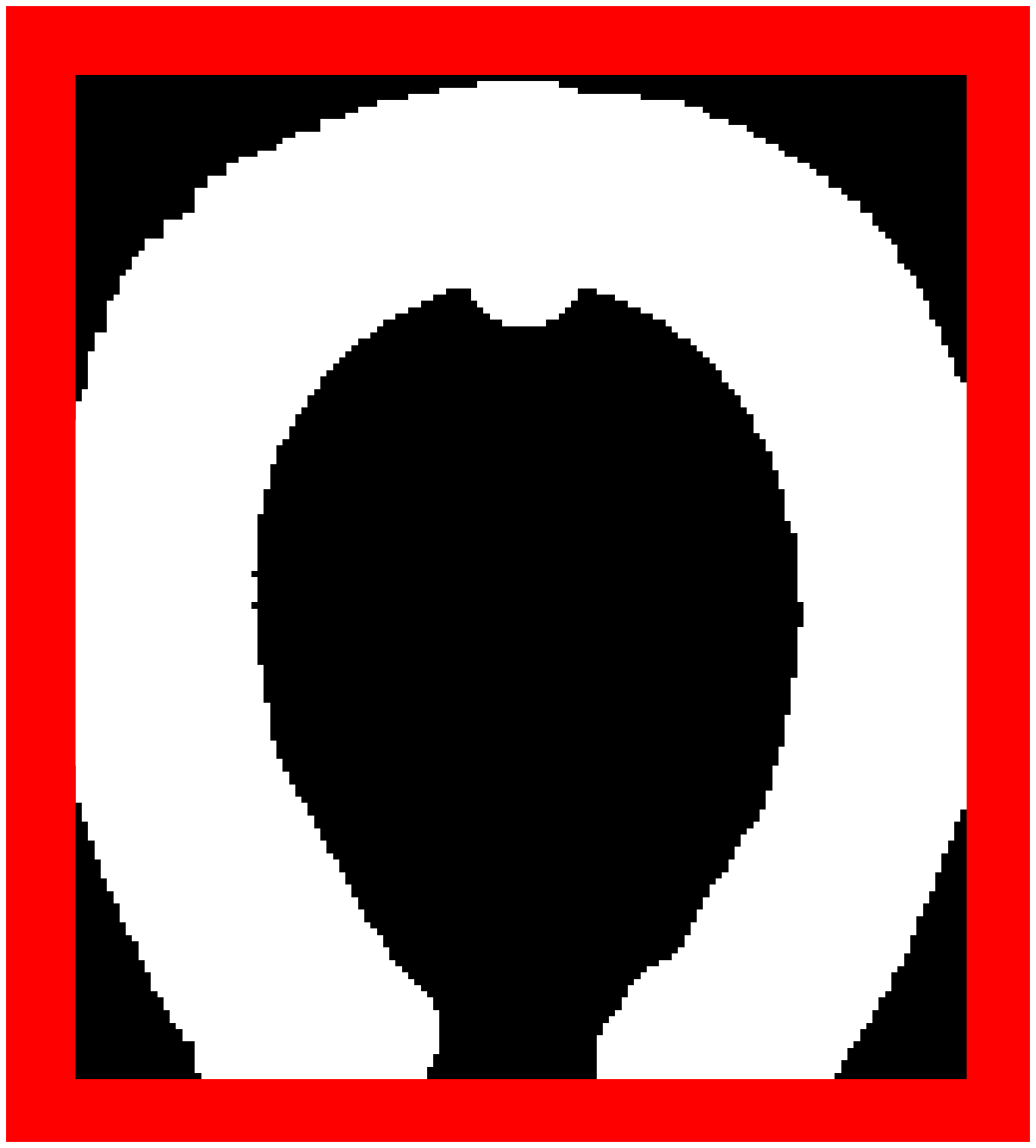}\hspace{-0.215em}
\includegraphics[trim=0.1cm 0cm 0.12cm 0cm, clip=true, width=1.2cm, height=1.2cm]{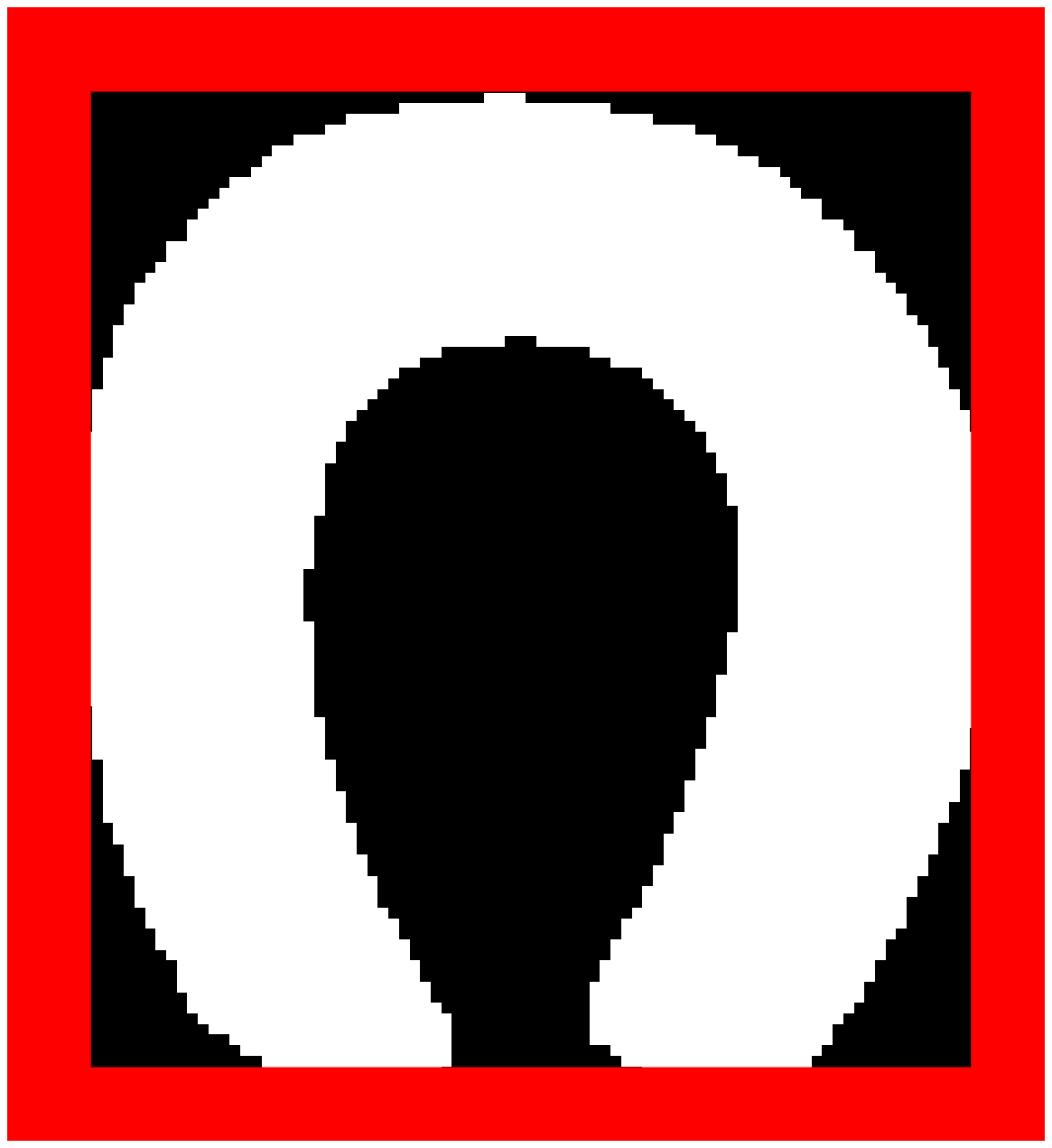}\hspace{-0.215em}
\includegraphics[trim=0.1cm 0cm 0.12cm 0cm, clip=true, width=1.2cm, height=1.2cm]{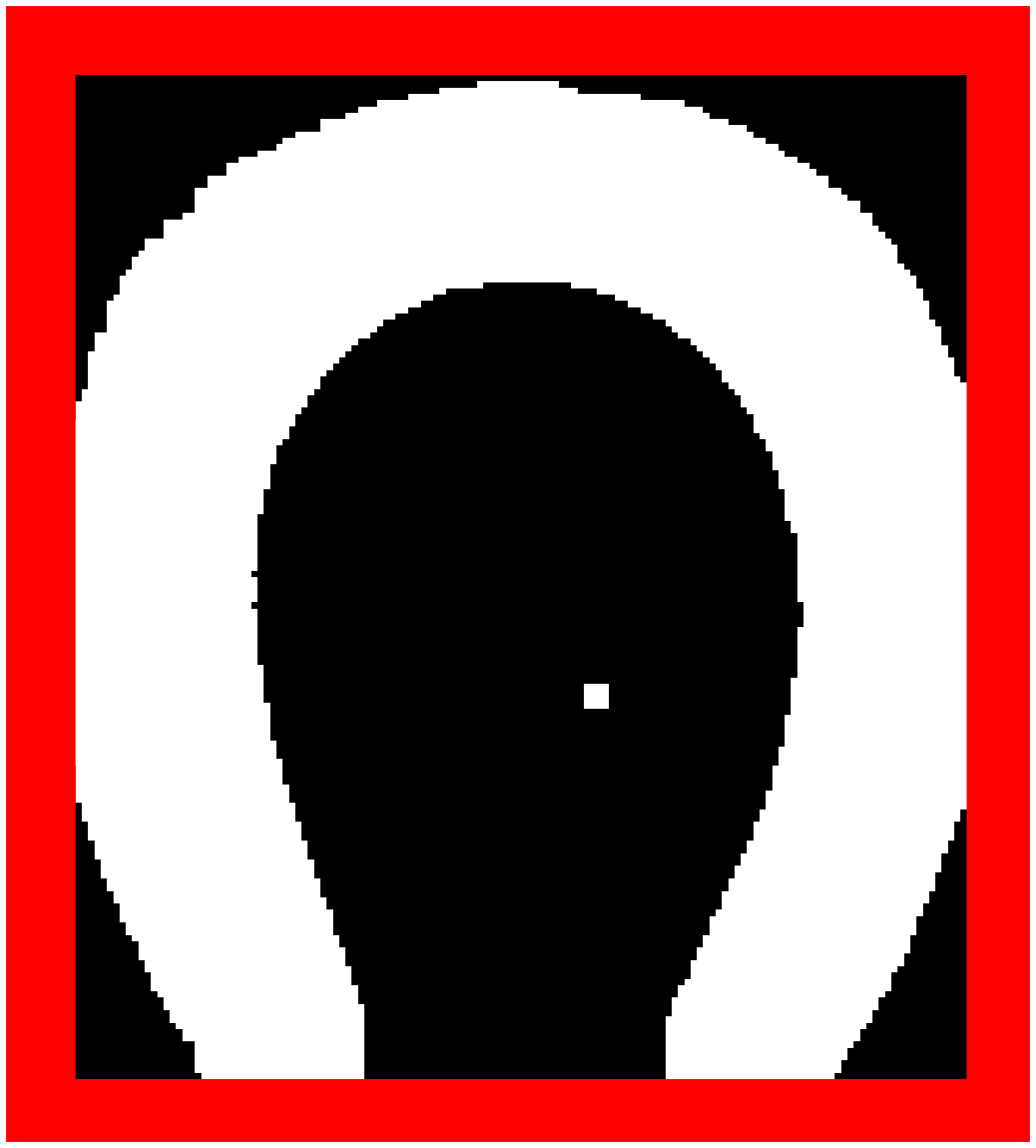}\hspace{-0.215em}
\includegraphics[trim=0.1cm 0cm 0.12cm 0cm, clip=true, width=1.2cm, height=1.2cm]{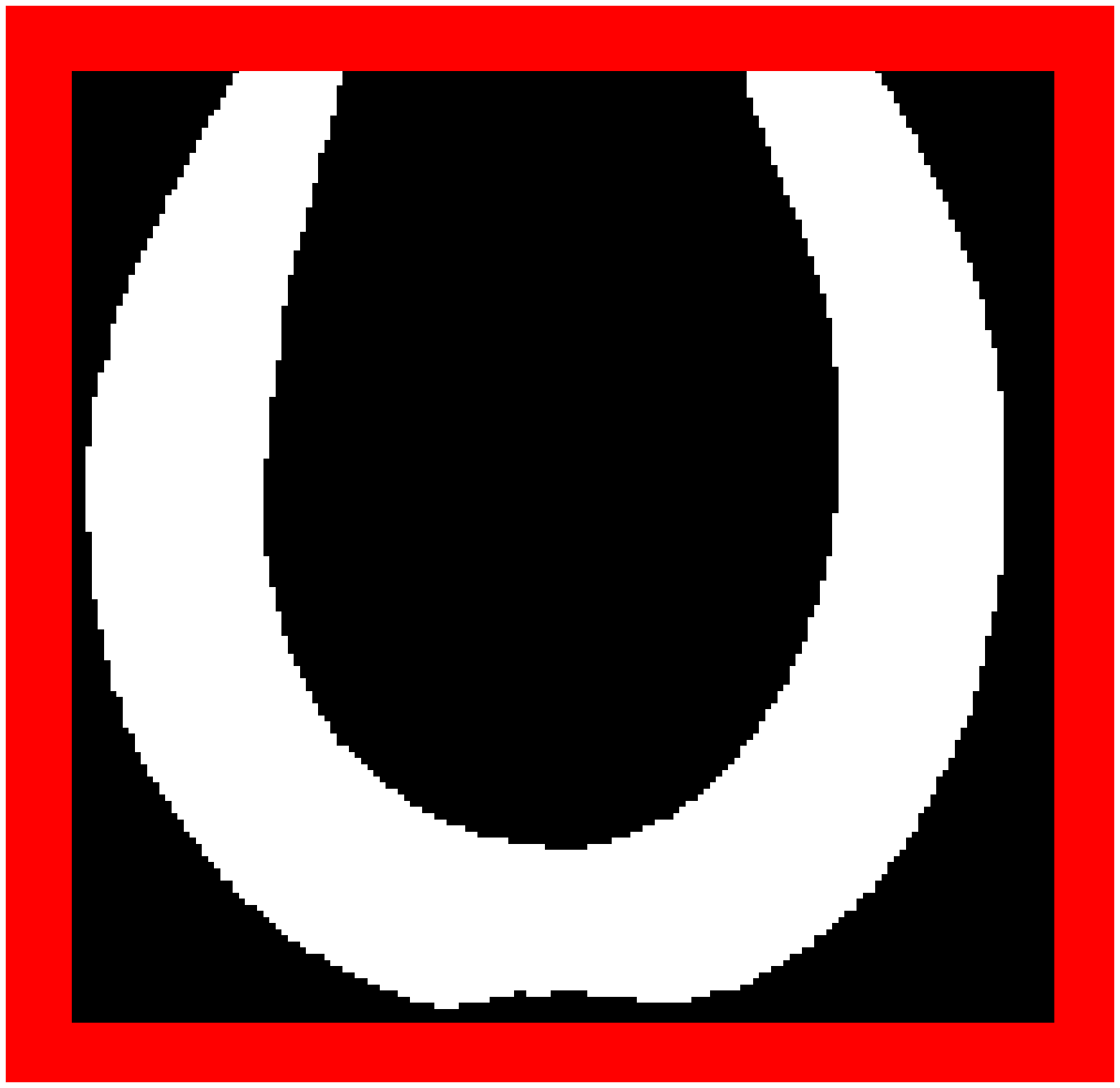}\hspace{-0.215em}
\includegraphics[trim=0.1cm 0cm 0.12cm 0cm, clip=true, width=1.2cm, height=1.2cm]{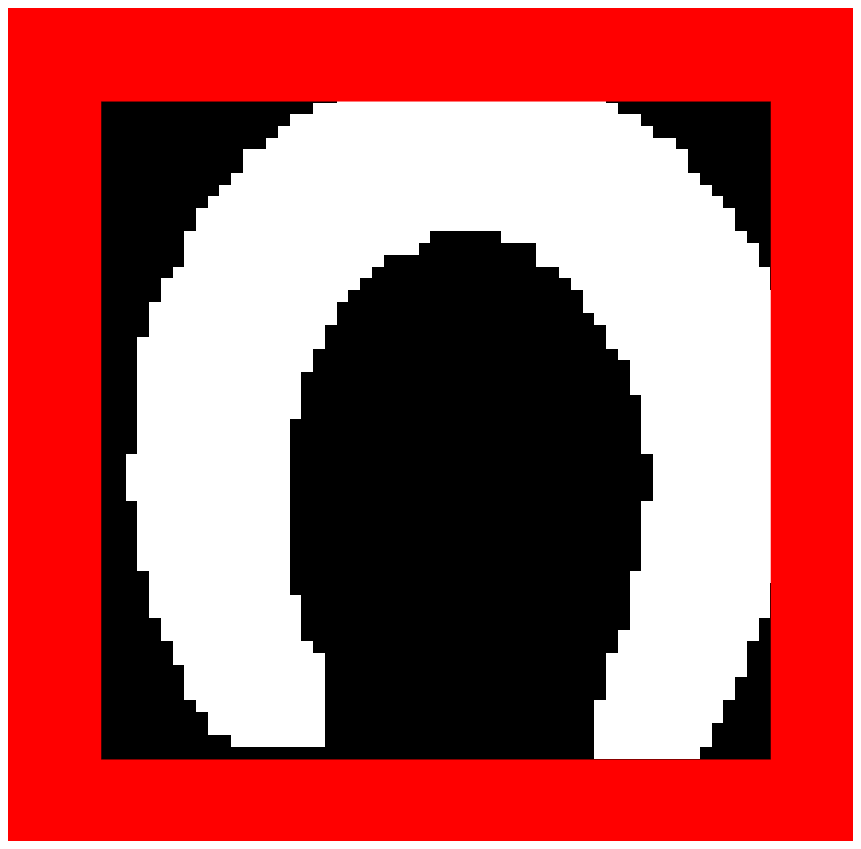}\hspace{-0.215em}
\includegraphics[trim=0.1cm 0cm 0.12cm 0cm, clip=true, width=1.2cm, height=1.2cm]{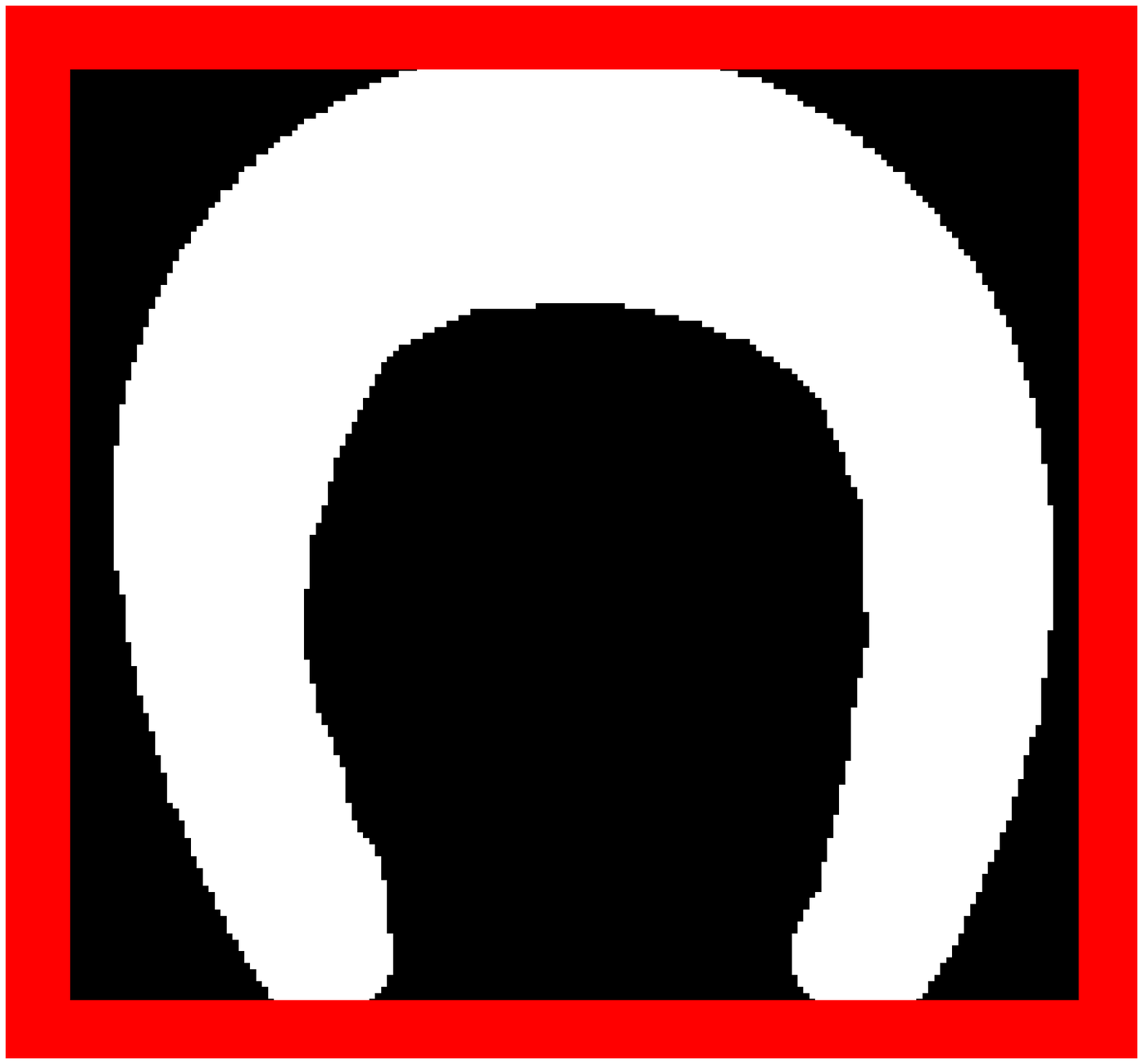}\hspace{-0.215em}
\includegraphics[trim=0.1cm 0cm 0.12cm 0cm, clip=true, width=1.2cm, height=1.2cm]{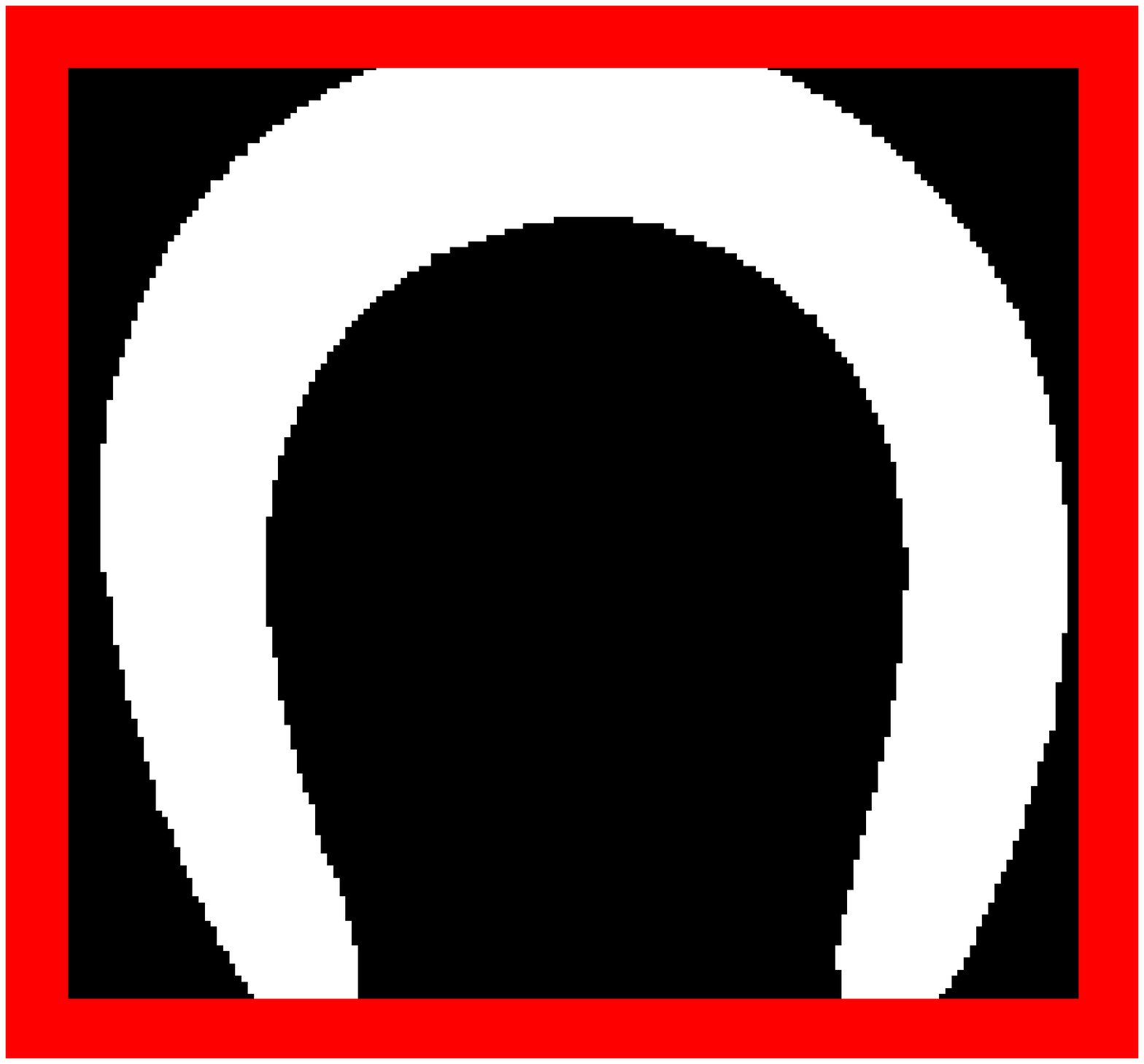}\hspace{-0.215em}
\includegraphics[trim=0.1cm 0cm 0.12cm 0cm, clip=true, width=1.2cm, height=1.2cm]{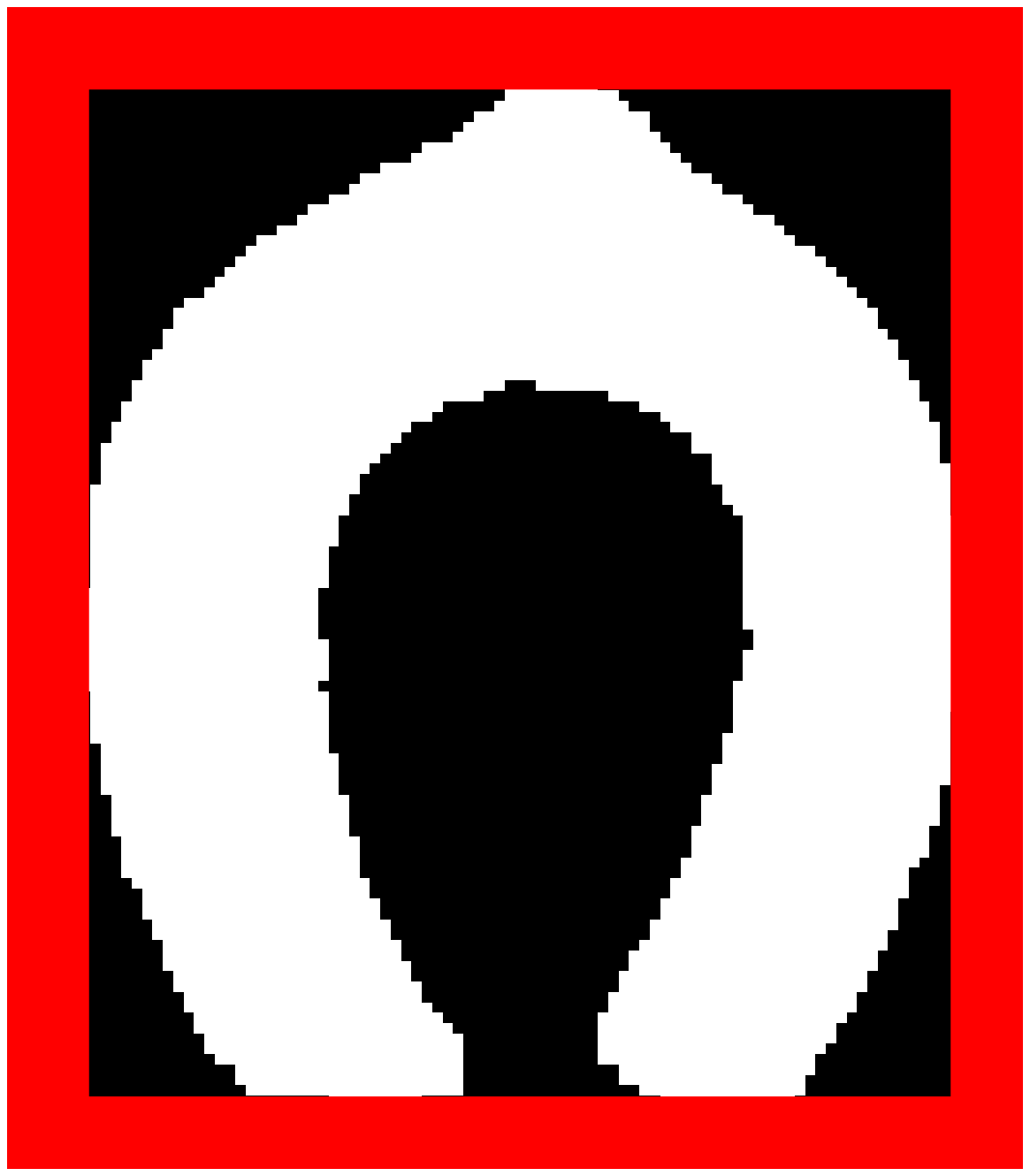}\hspace{-0.215em}\\
\includegraphics[trim=0.1cm 0cm 0.12cm 0cm, clip=true, width=1.2cm, height=1.2cm]{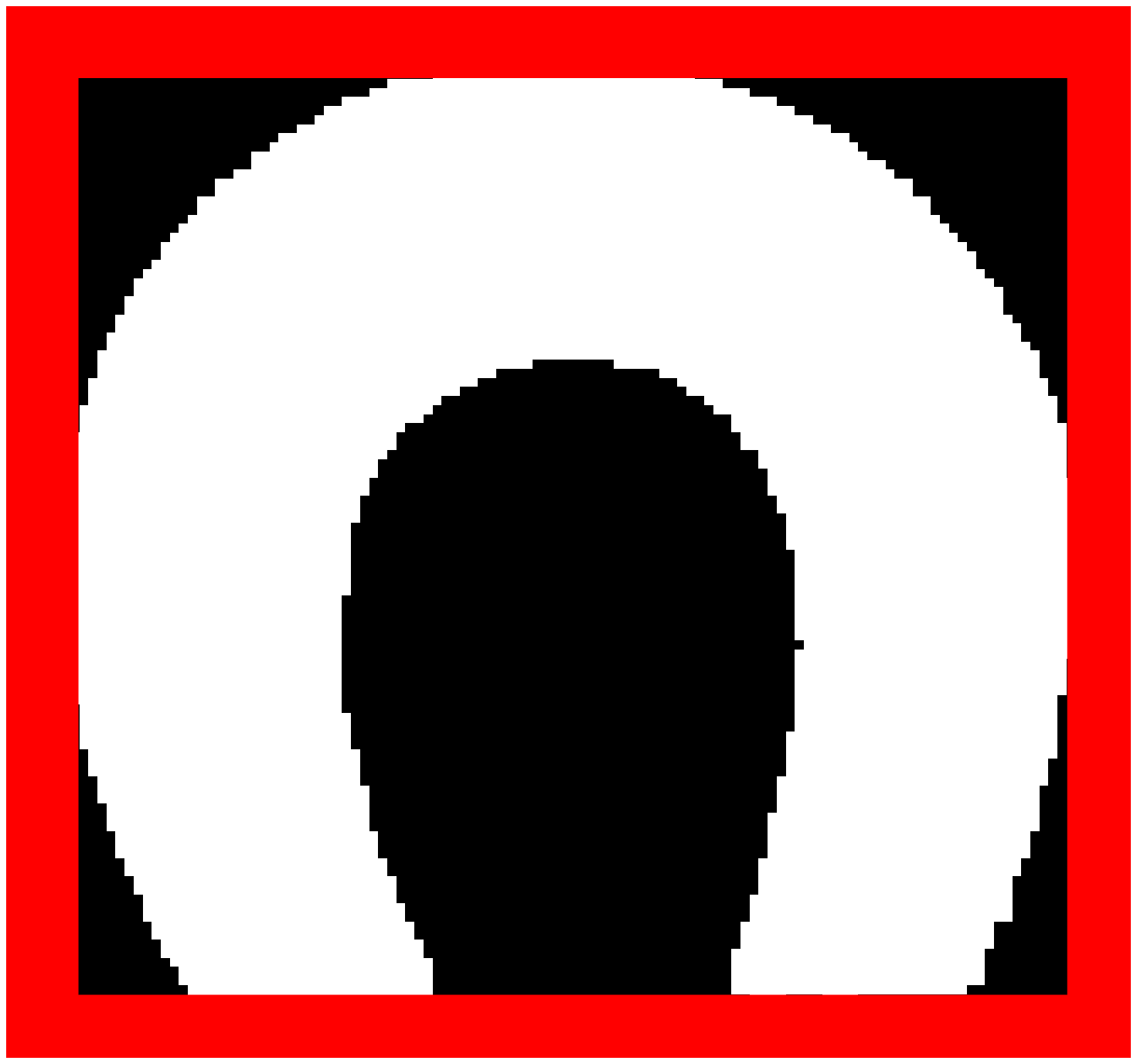}\hspace{-0.215em}
\includegraphics[trim=0.1cm 0cm 0.12cm 0cm, clip=true, width=1.2cm, height=1.2cm]{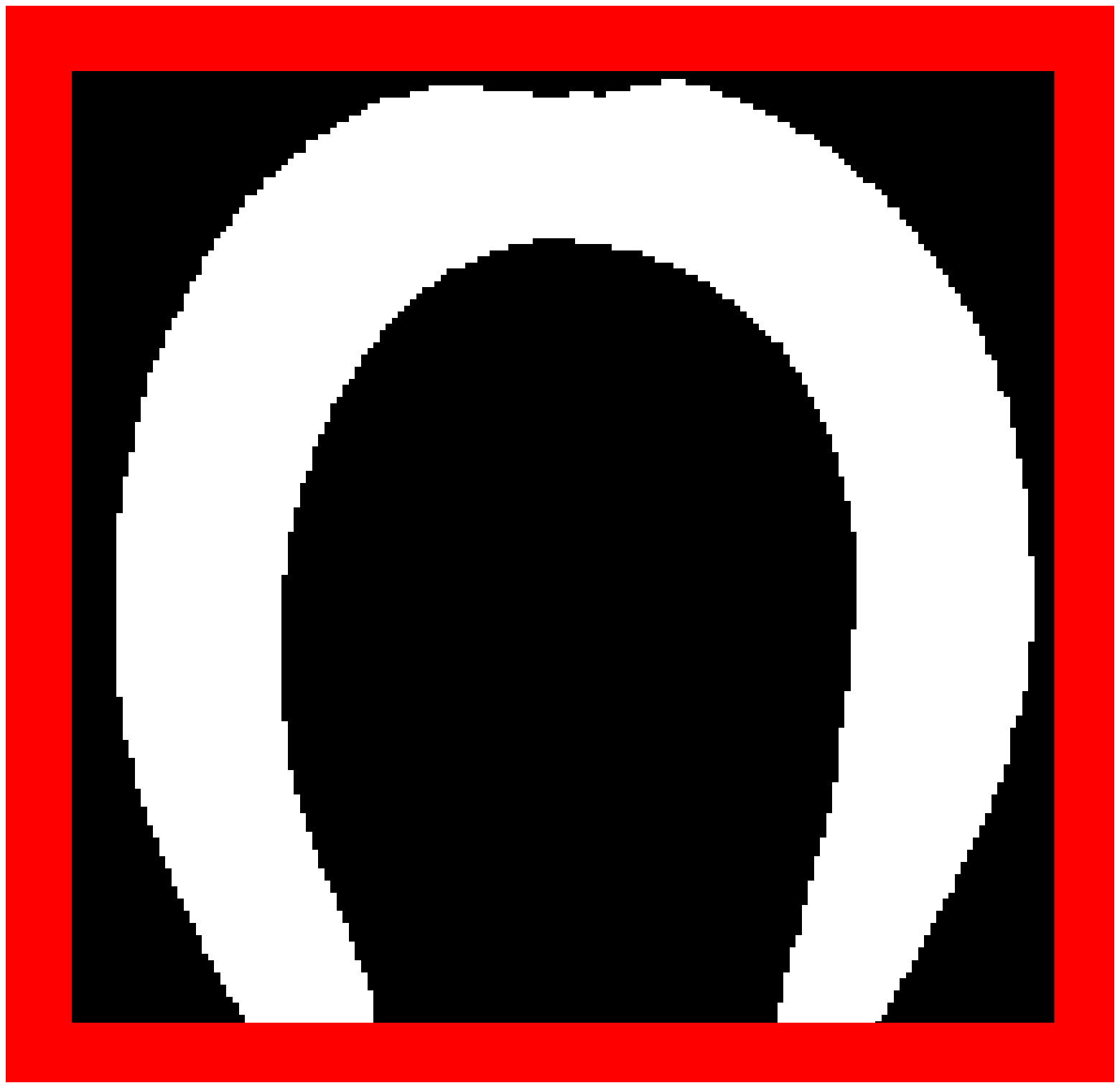}\hspace{-0.215em}
\includegraphics[trim=0.1cm 0cm 0.12cm 0cm, clip=true, width=1.2cm, height=1.2cm]{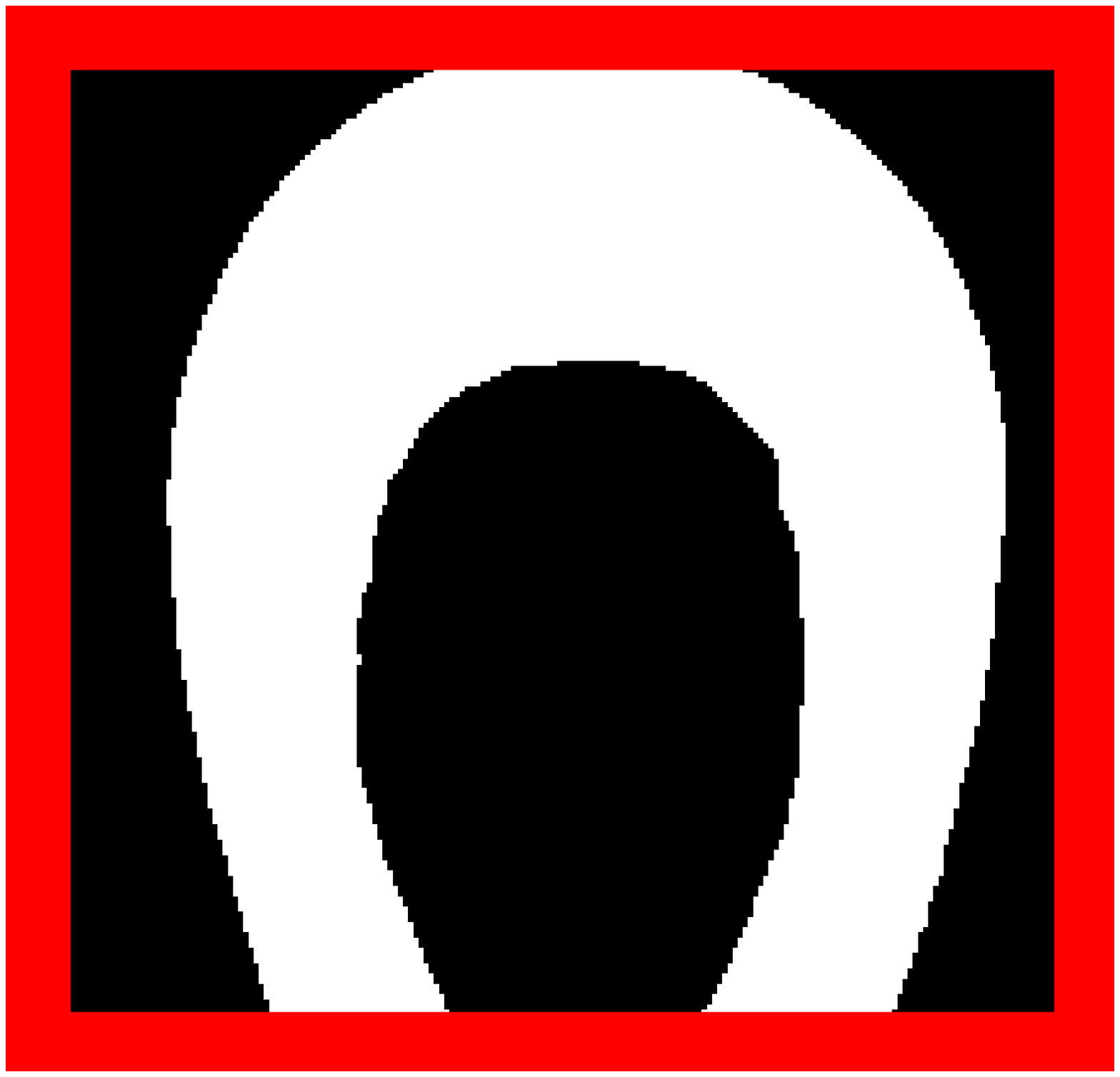}\hspace{-0.215em}
\includegraphics[trim=0.1cm 0cm 0.12cm 0cm, clip=true, width=1.2cm, height=1.2cm]{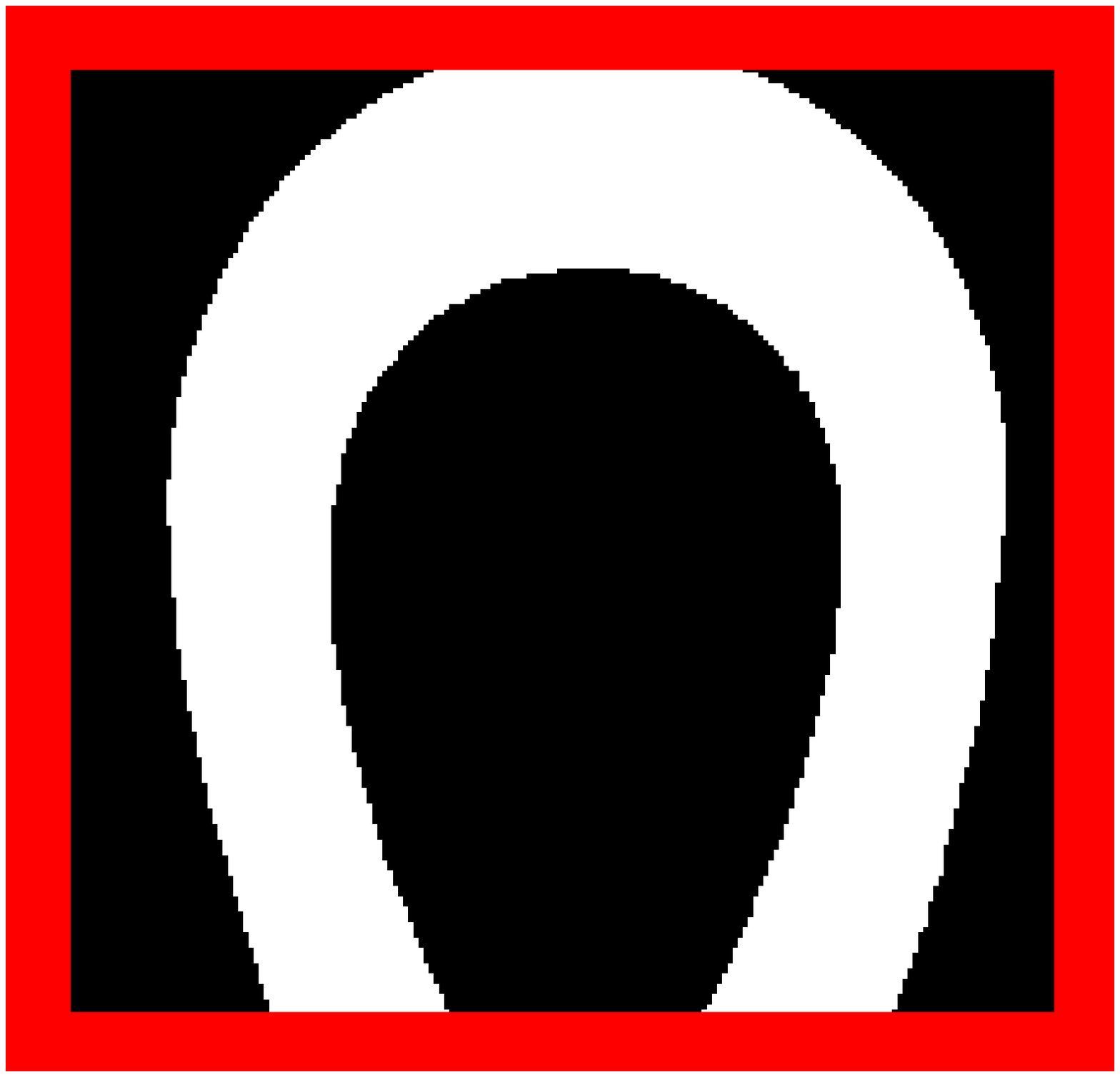}\hspace{-0.215em}
\includegraphics[trim=0.1cm 0cm 0.12cm 0cm, clip=true, width=1.2cm, height=1.2cm]{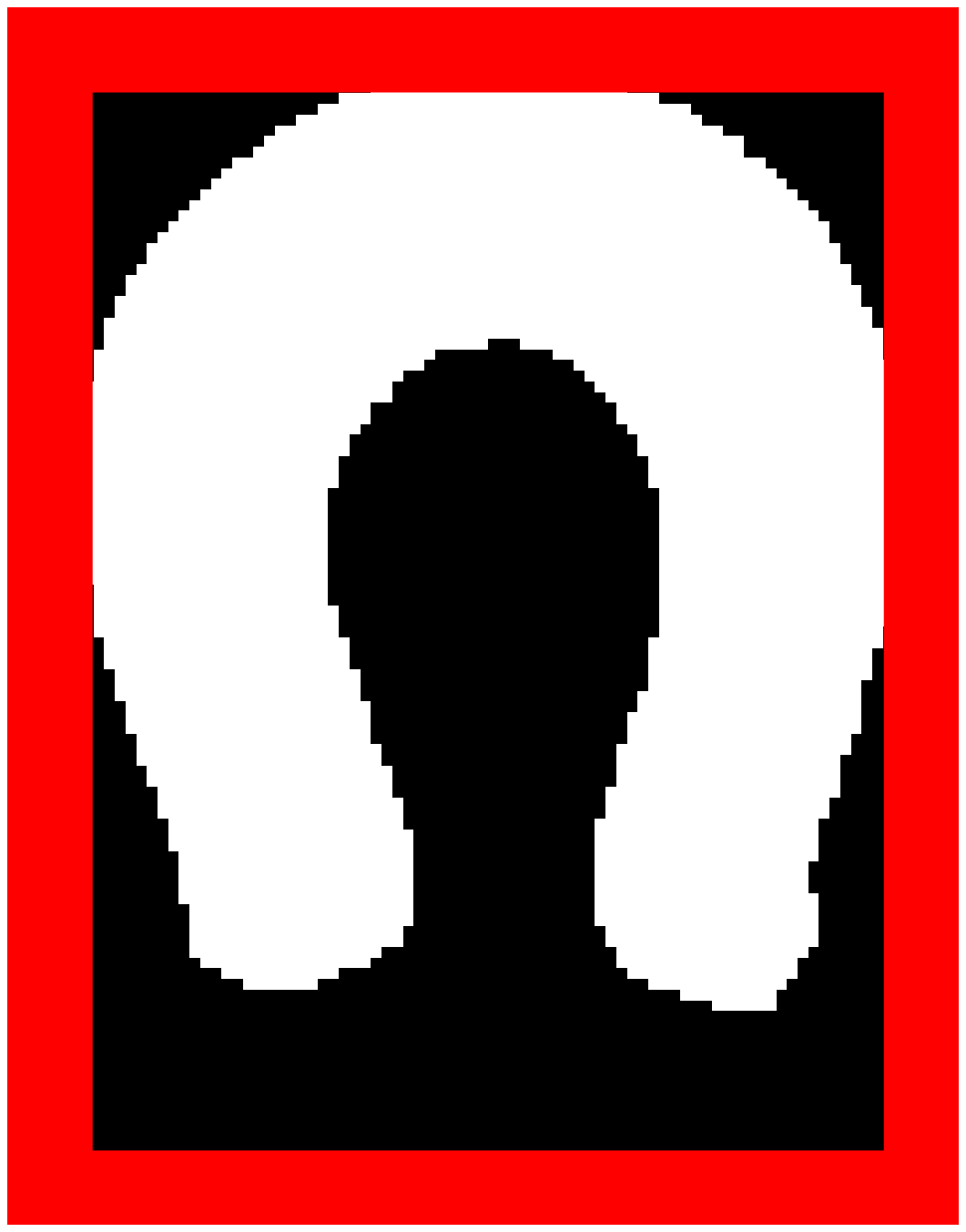}\hspace{-0.215em}
\includegraphics[trim=0.1cm 0cm 0.12cm 0cm, clip=true, width=1.2cm, height=1.2cm]{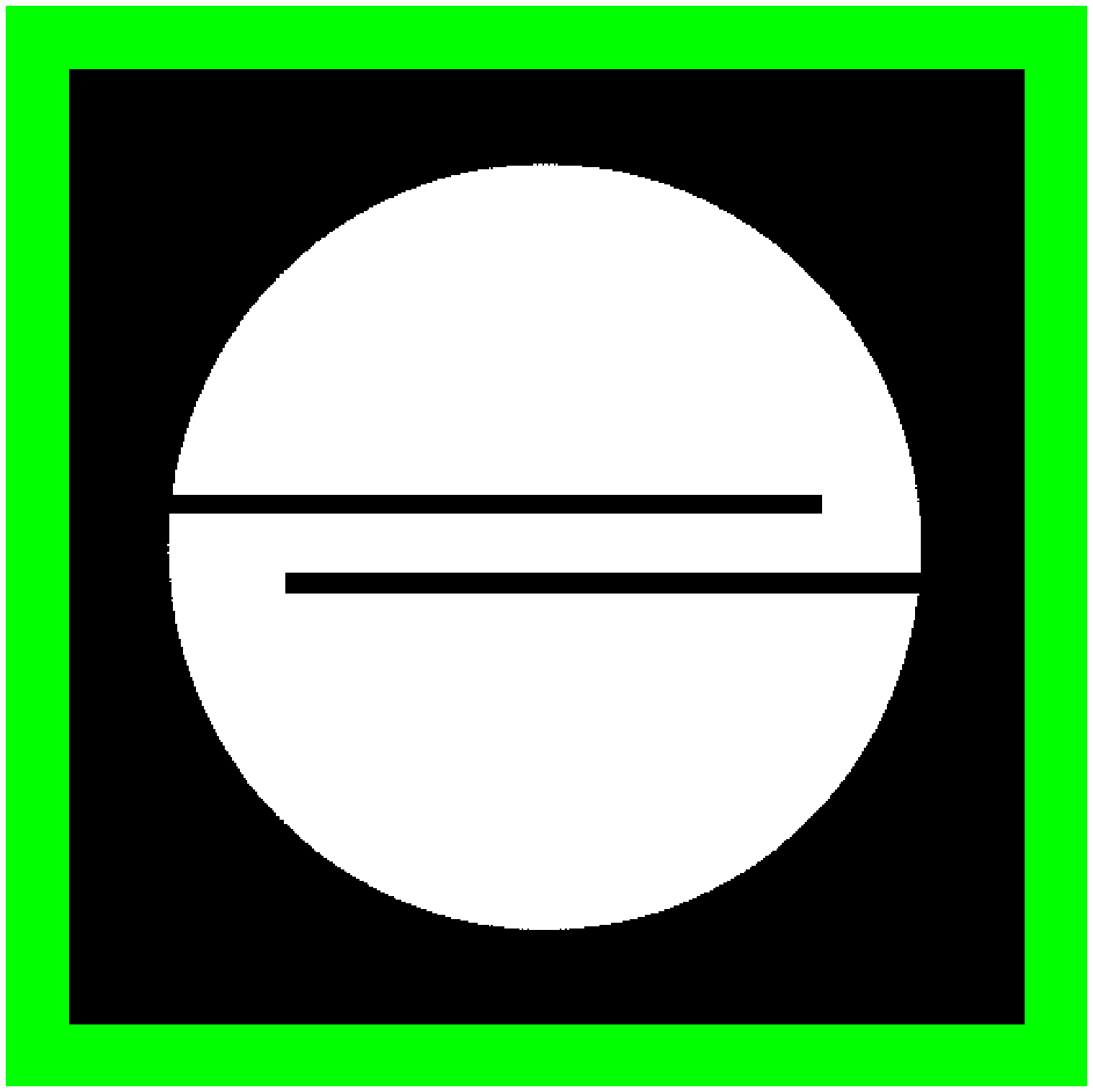}\hspace{-0.215em}
\includegraphics[trim=0.1cm 0cm 0.12cm 0cm, clip=true, width=1.2cm, height=1.2cm]{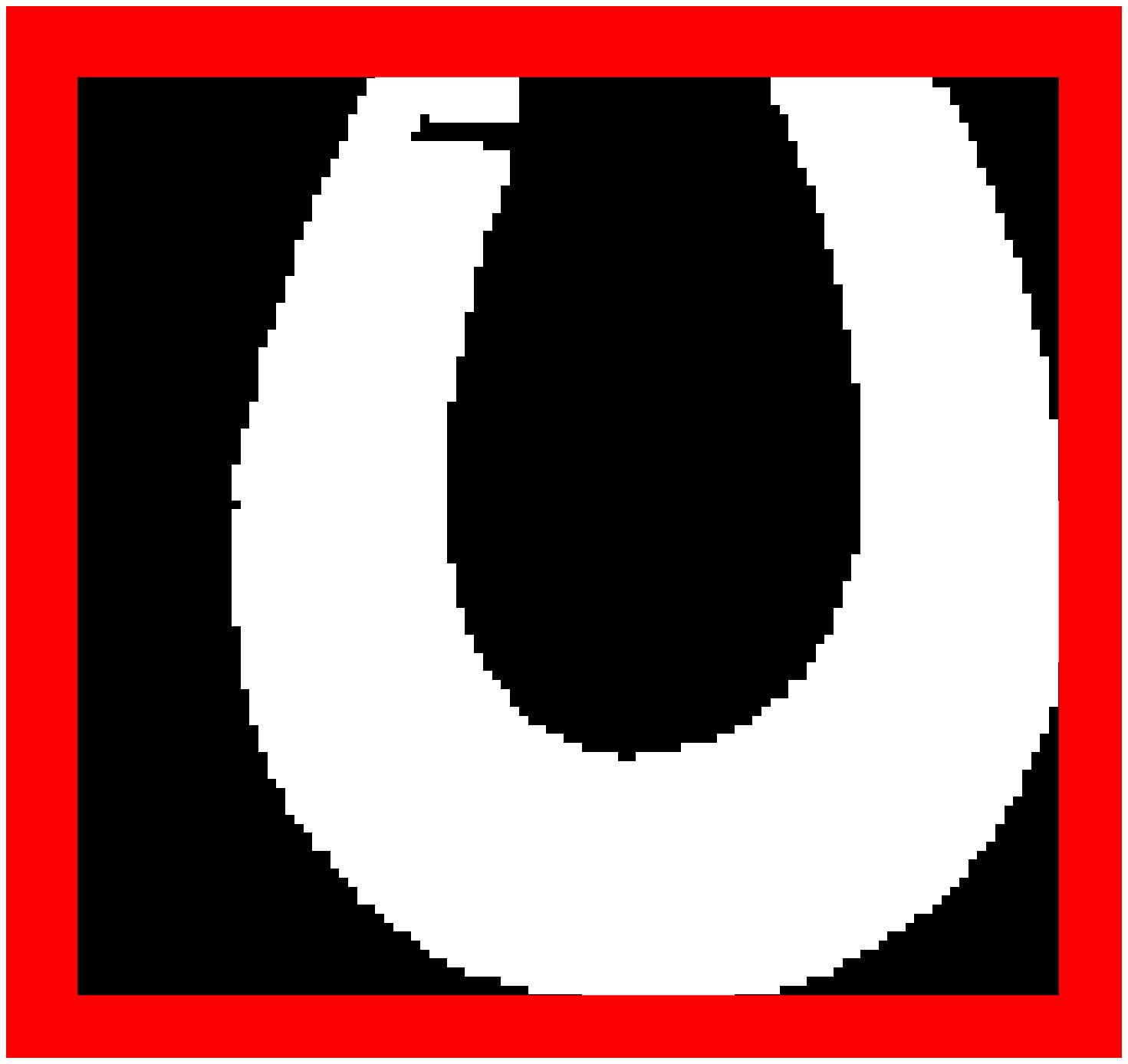}\hspace{-0.215em}
\includegraphics[trim=0.1cm 0cm 0.12cm 0cm, clip=true, width=1.2cm, height=1.2cm]{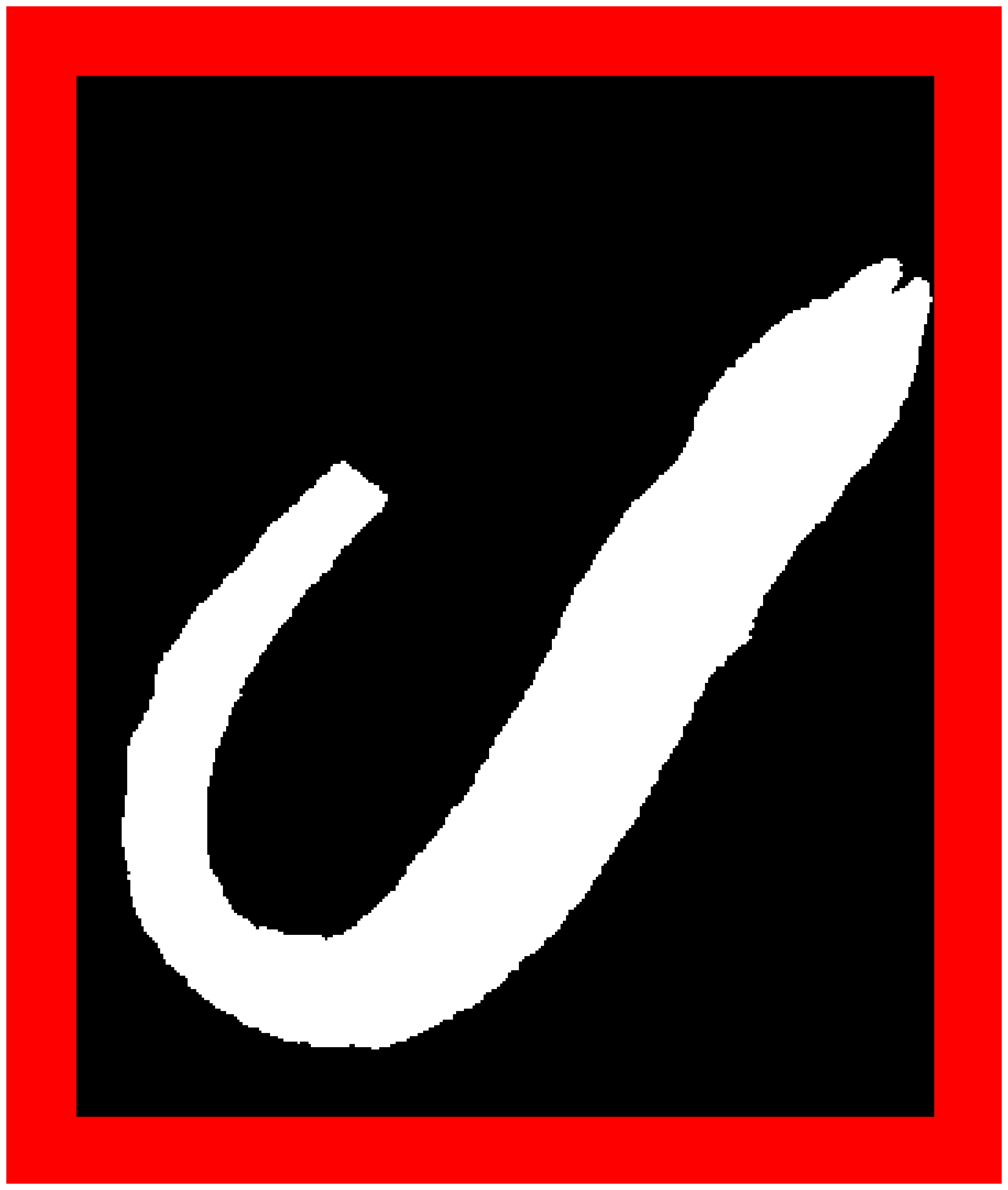}\hspace{-0.215em}
\includegraphics[trim=0.1cm 0cm 0.12cm 0cm, clip=true, width=1.2cm, height=1.2cm]{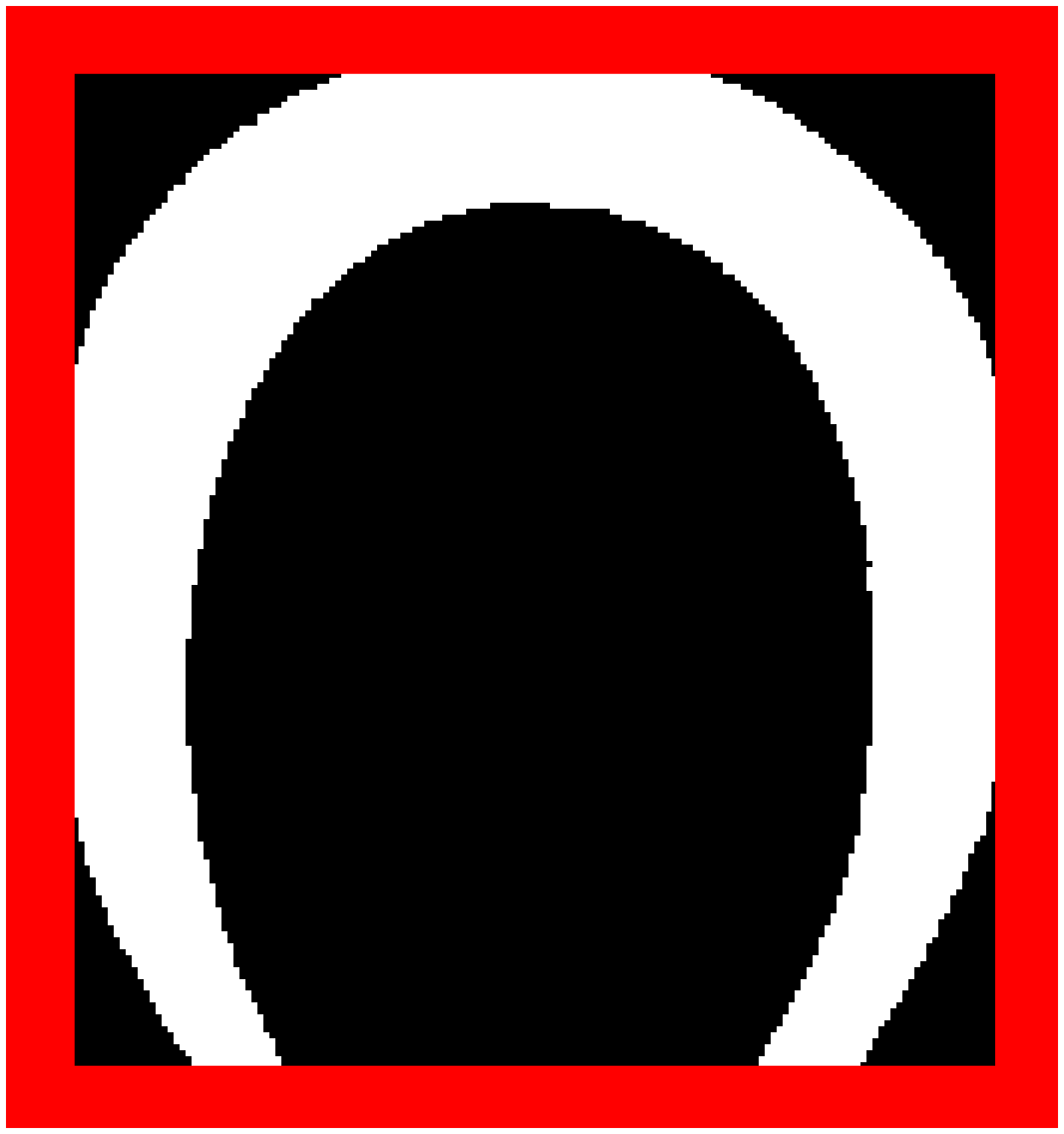}\hspace{-0.215em}
\includegraphics[trim=0.1cm 0cm 0.12cm 0cm, clip=true, width=1.2cm, height=1.2cm]{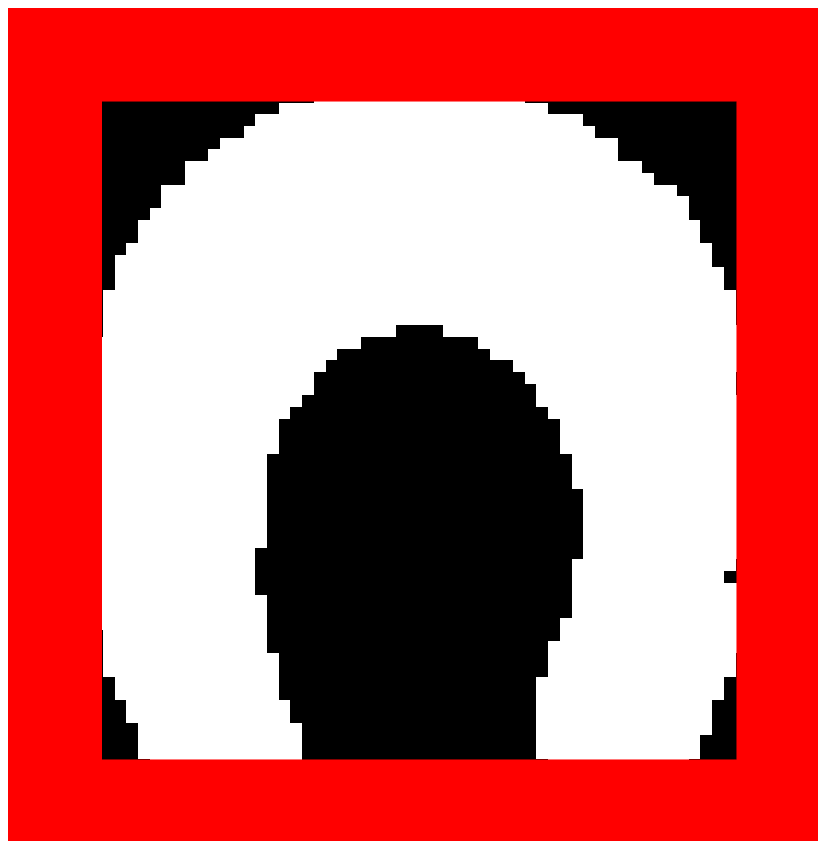}\hspace{-0.215em}\\
\includegraphics[trim=0.1cm 0cm 0.12cm 0cm, clip=true, width=1.2cm, height=1.2cm]{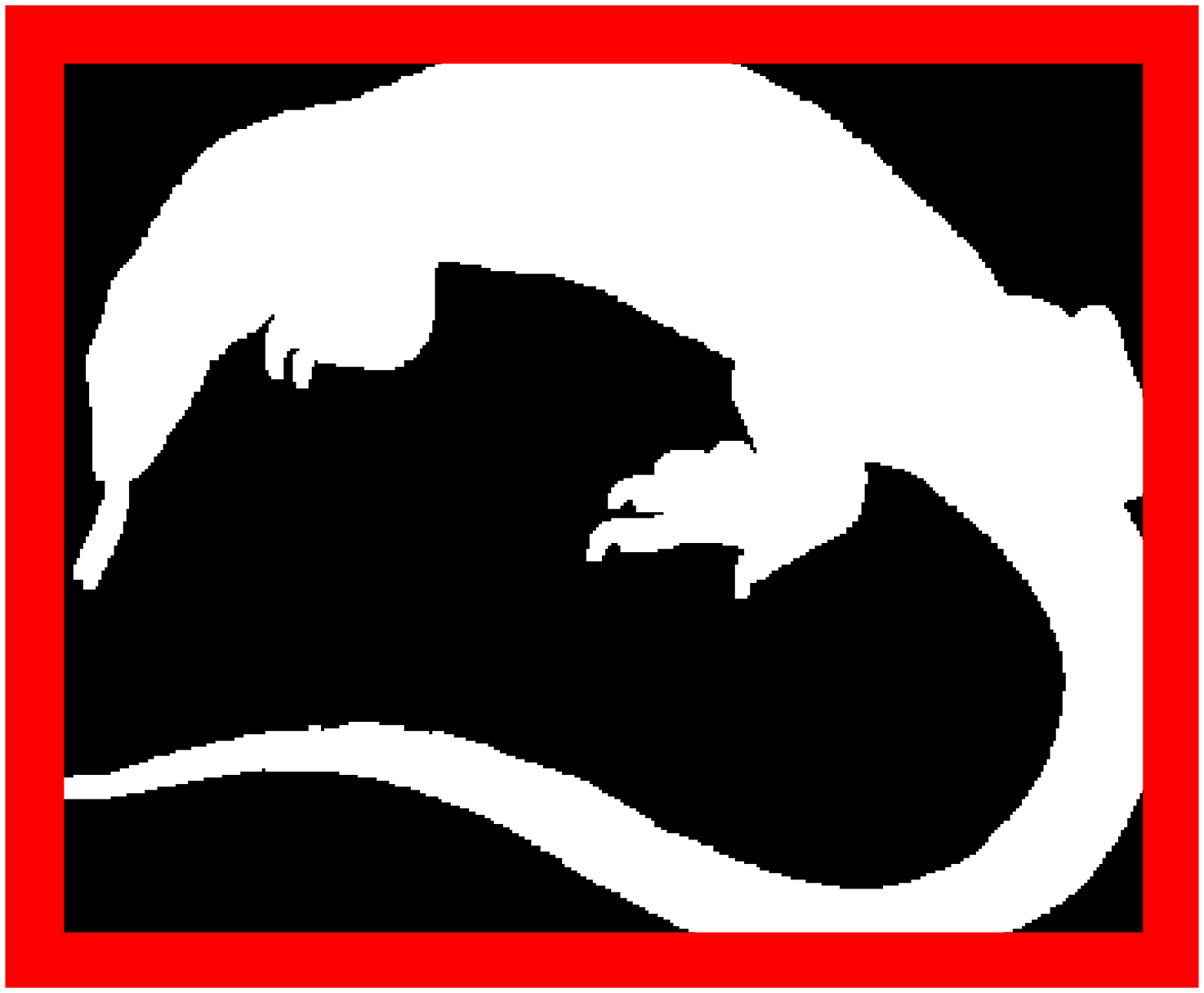}\hspace{-0.215em}
\includegraphics[trim=0.1cm 0cm 0.12cm 0cm, clip=true, width=1.2cm, height=1.2cm]{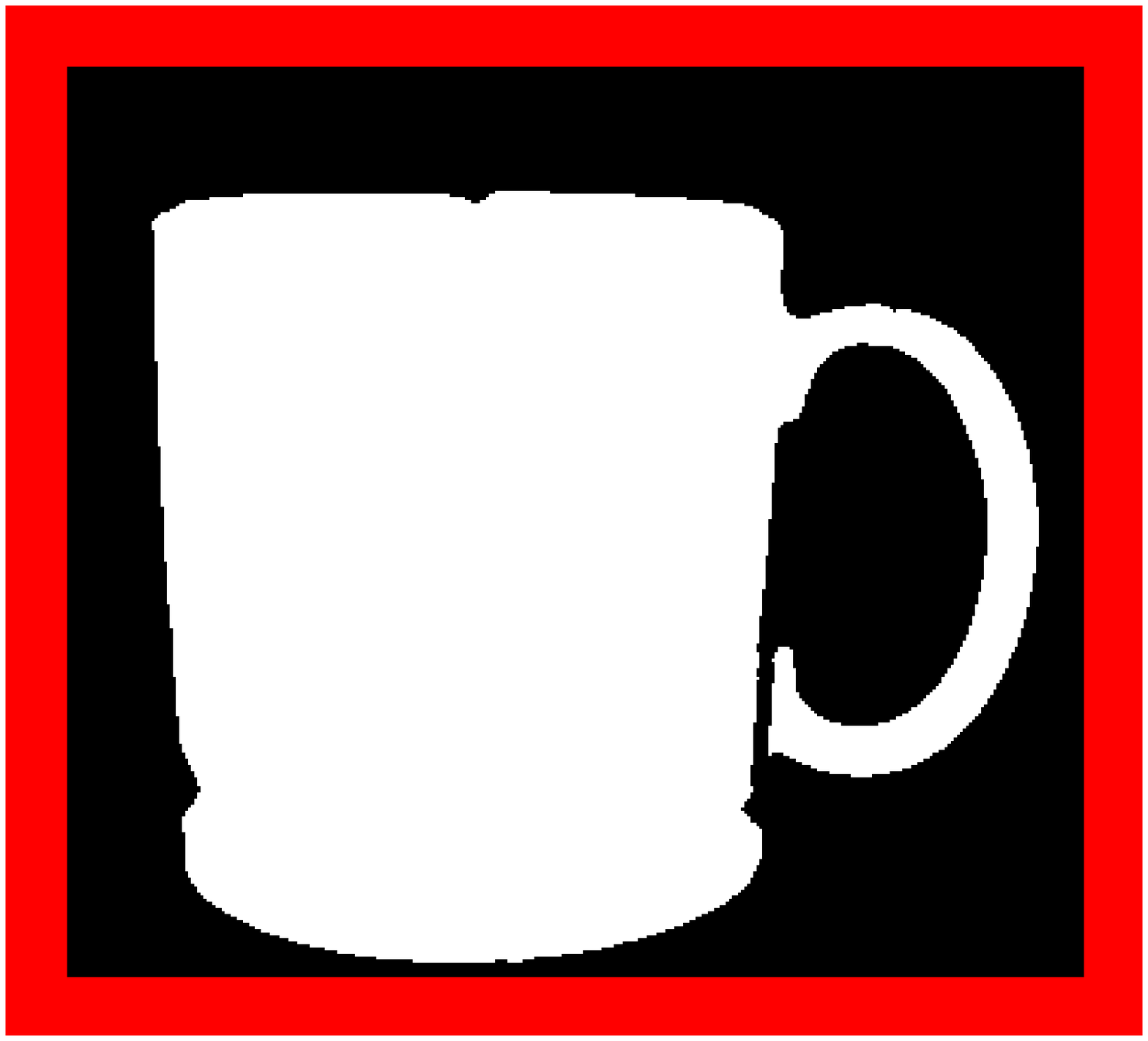}\hspace{-0.215em}
\includegraphics[trim=0.1cm 0cm 0.12cm 0cm, clip=true, width=1.2cm, height=1.2cm]{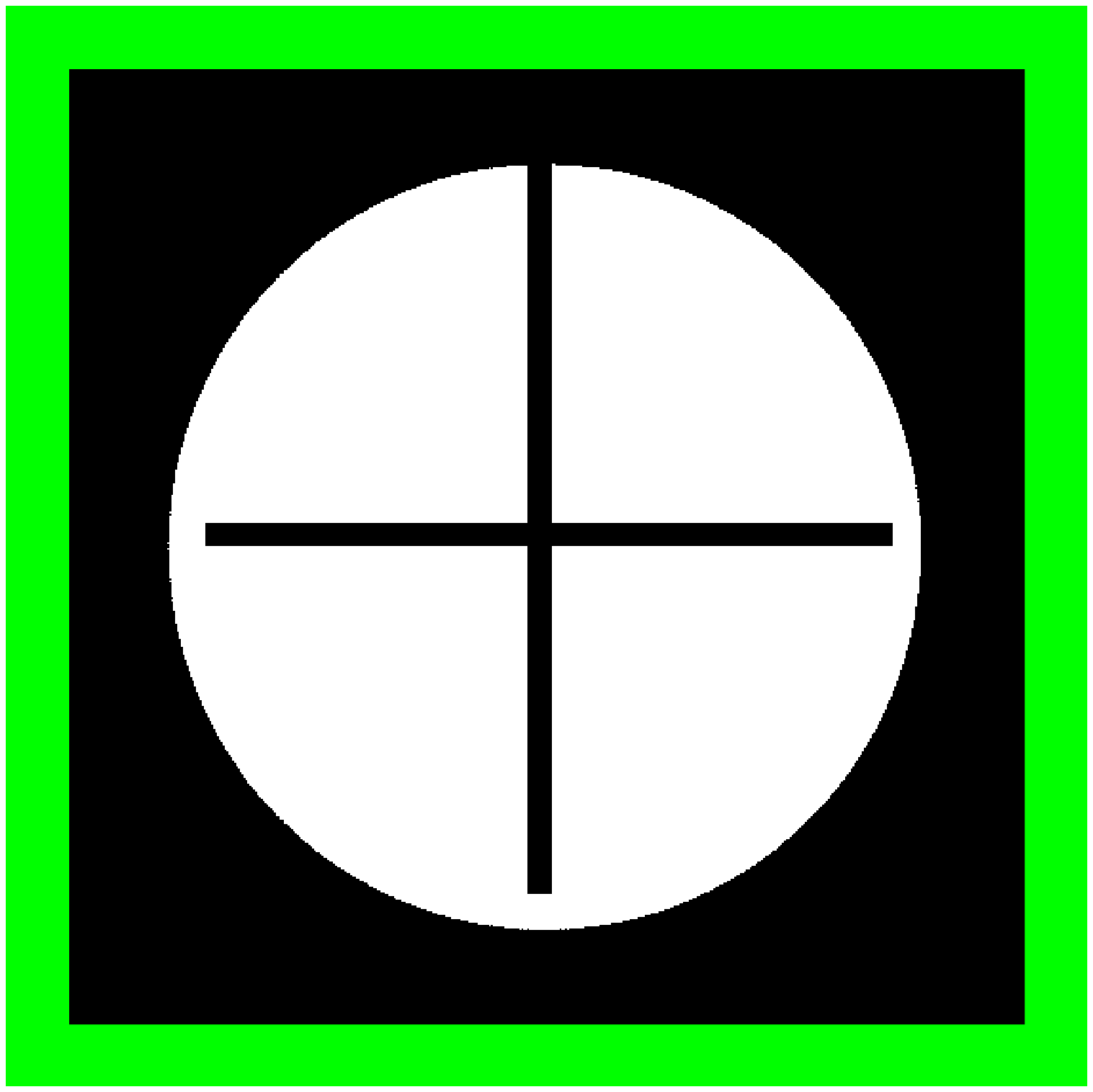}\hspace{-0.215em}
\includegraphics[trim=0.1cm 0cm 0.12cm 0cm, clip=true, width=1.2cm, height=1.2cm]{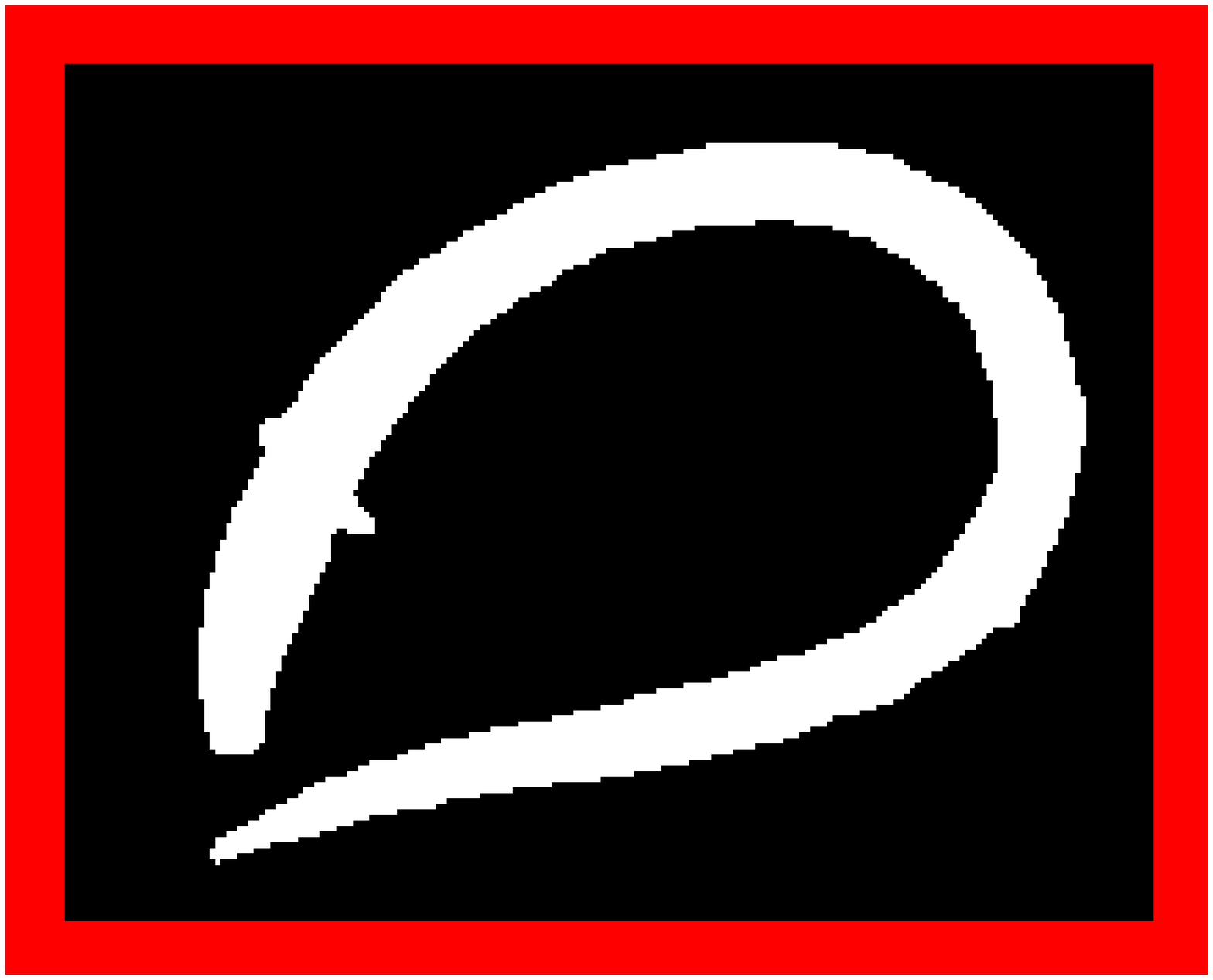}\hspace{-0.215em}
\includegraphics[trim=0.1cm 0cm 0.12cm 0cm, clip=true, width=1.2cm, height=1.2cm]{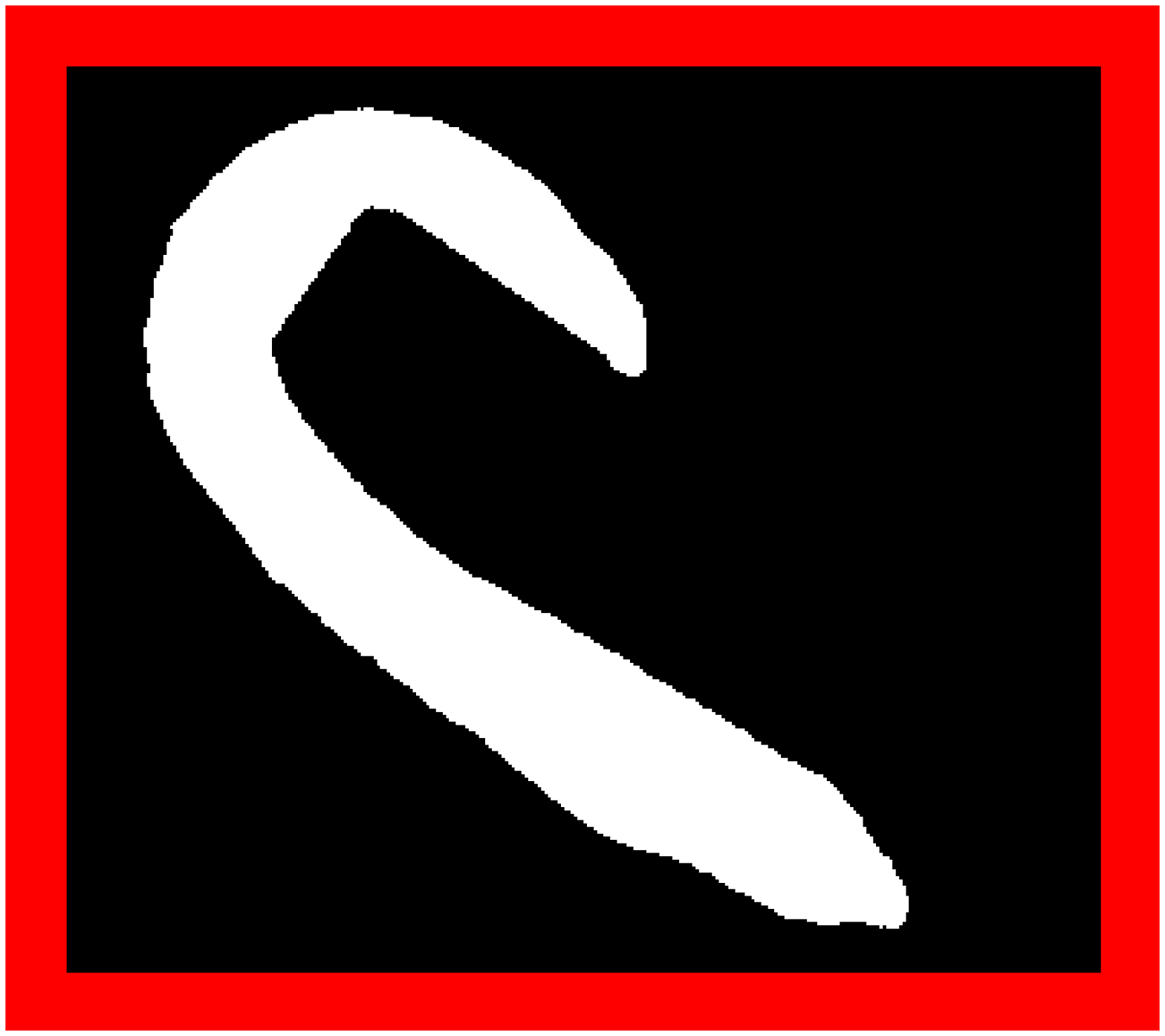}\hspace{-0.215em}
\includegraphics[trim=0.1cm 0cm 0.12cm 0cm, clip=true, width=1.2cm, height=1.2cm]{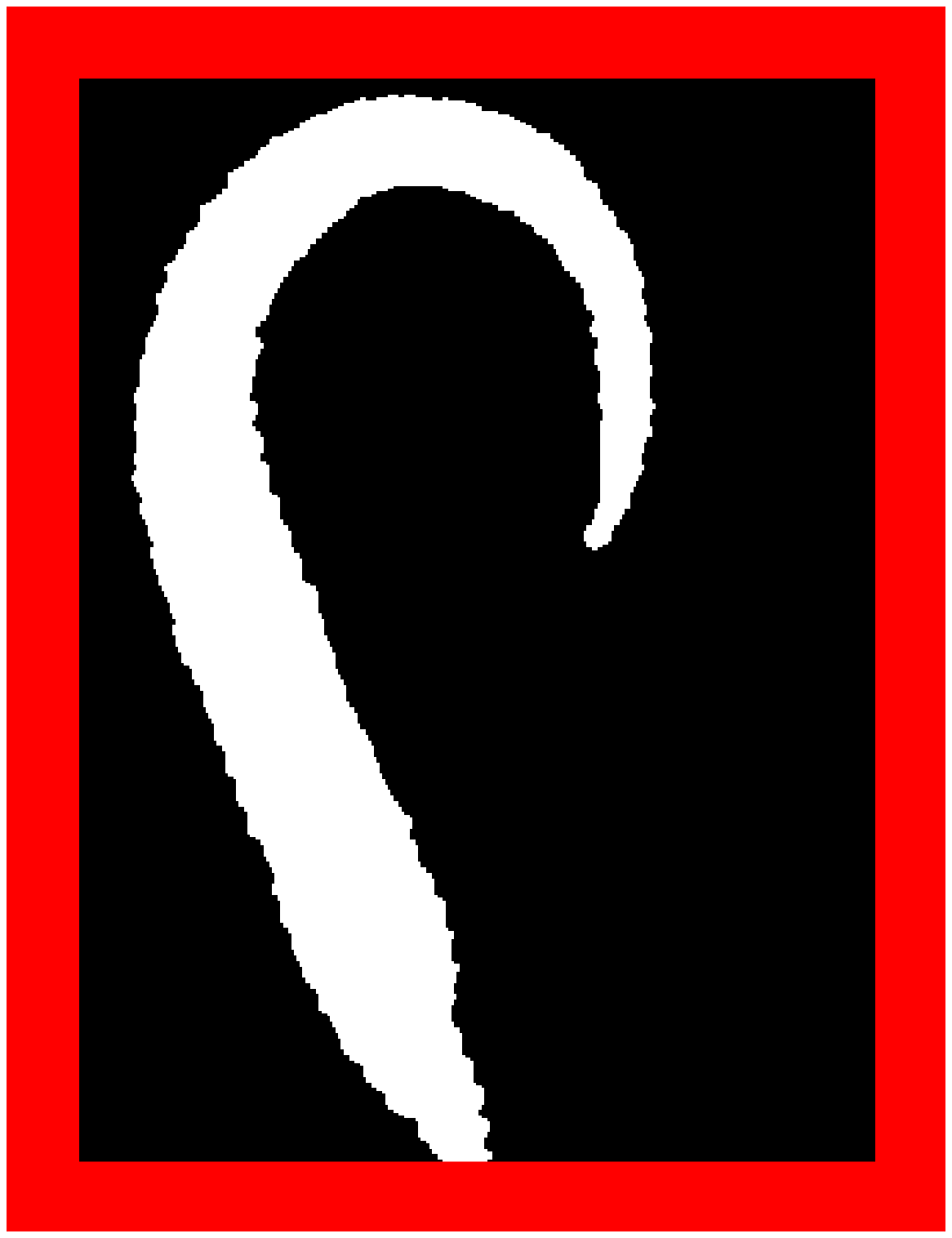}\hspace{-0.215em}
\includegraphics[trim=0.1cm 0cm 0.12cm 0cm, clip=true, width=1.2cm, height=1.2cm]{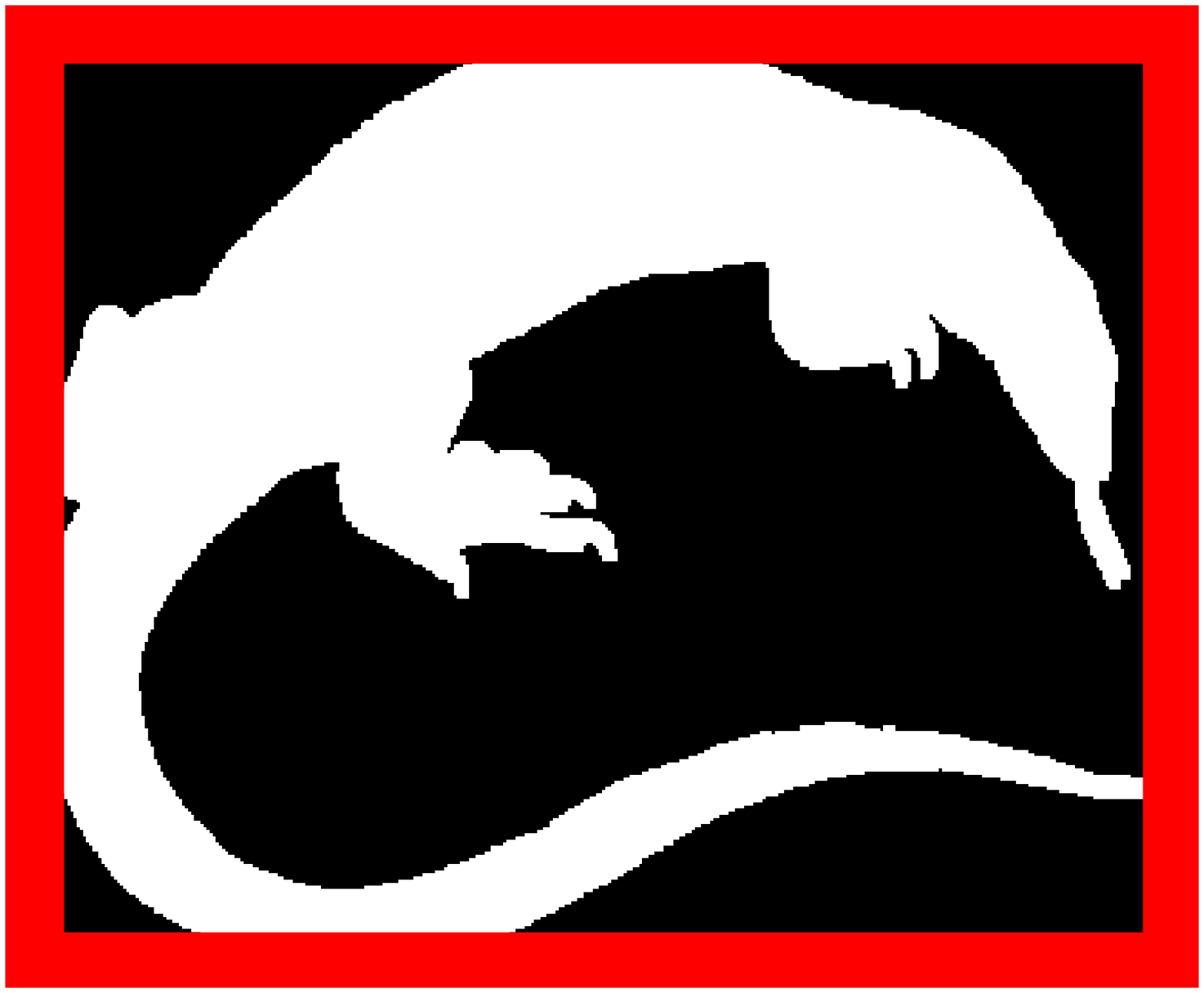}\hspace{-0.215em}
\includegraphics[trim=0.1cm 0cm 0.12cm 0cm, clip=true, width=1.2cm, height=1.2cm]{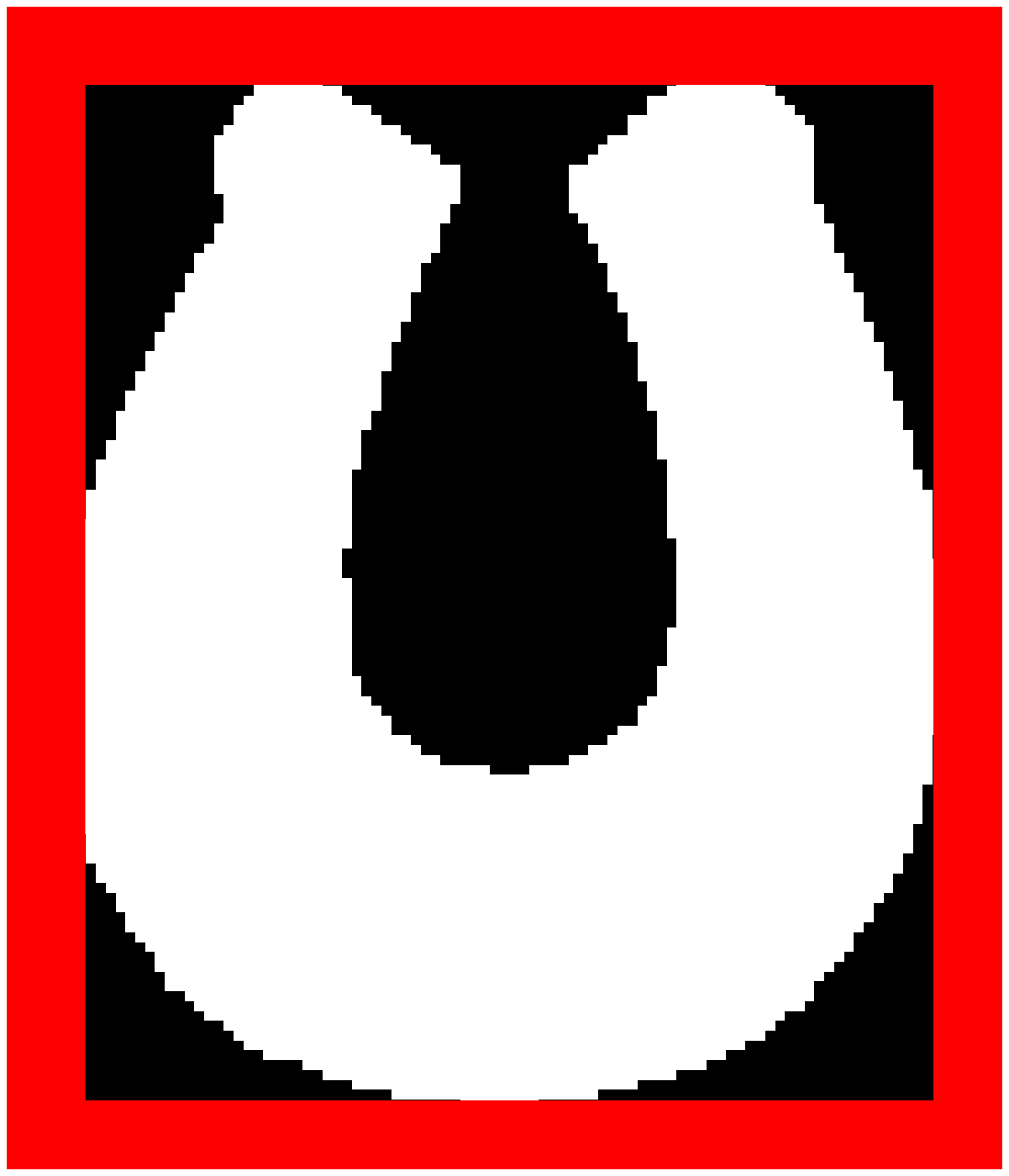}\hspace{-0.215em}
\includegraphics[trim=0.1cm 0cm 0.12cm 0cm, clip=true, width=1.2cm, height=1.2cm]{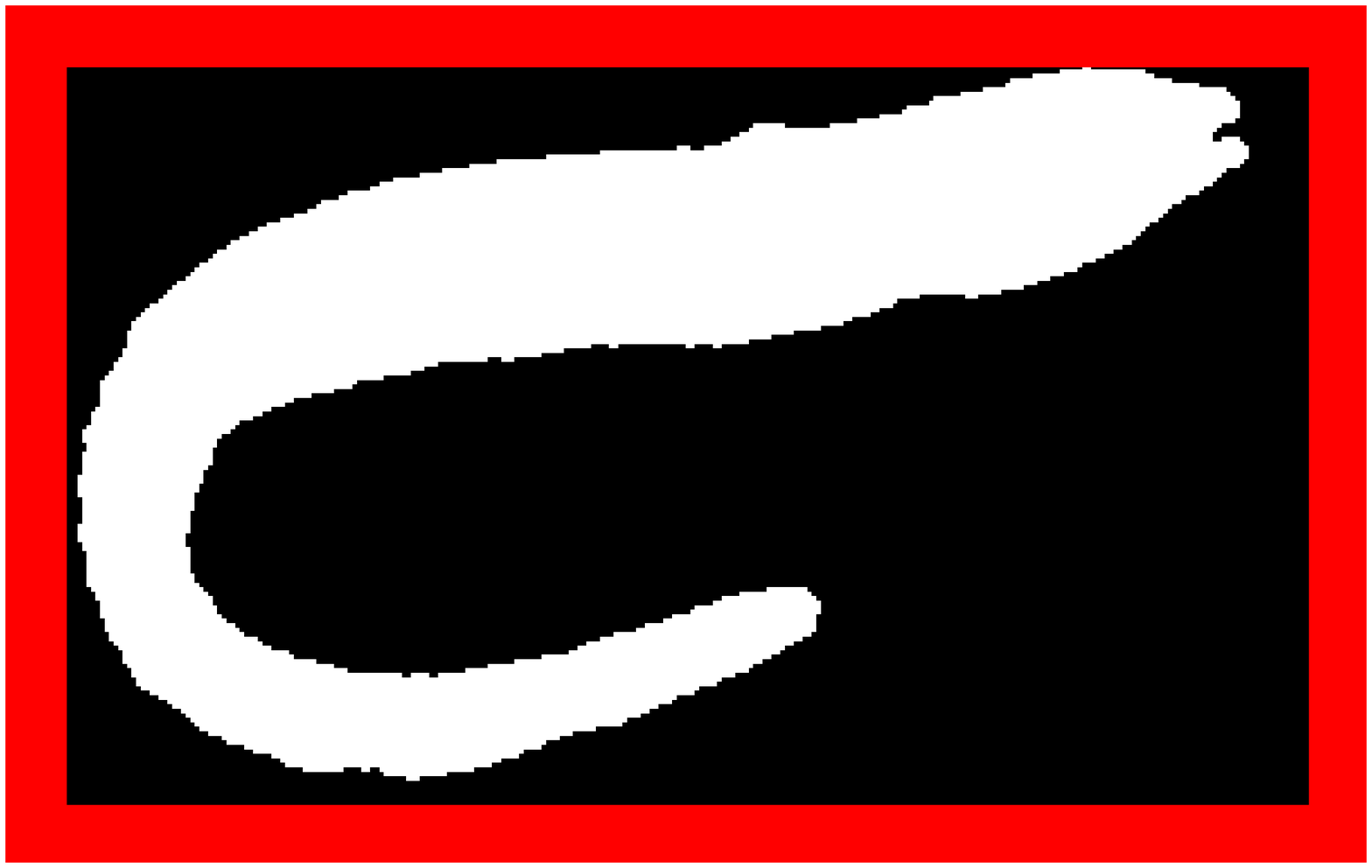}\hspace{-0.215em}
\includegraphics[trim=0.1cm 0cm 0.12cm 0cm, clip=true, width=1.2cm, height=1.2cm]{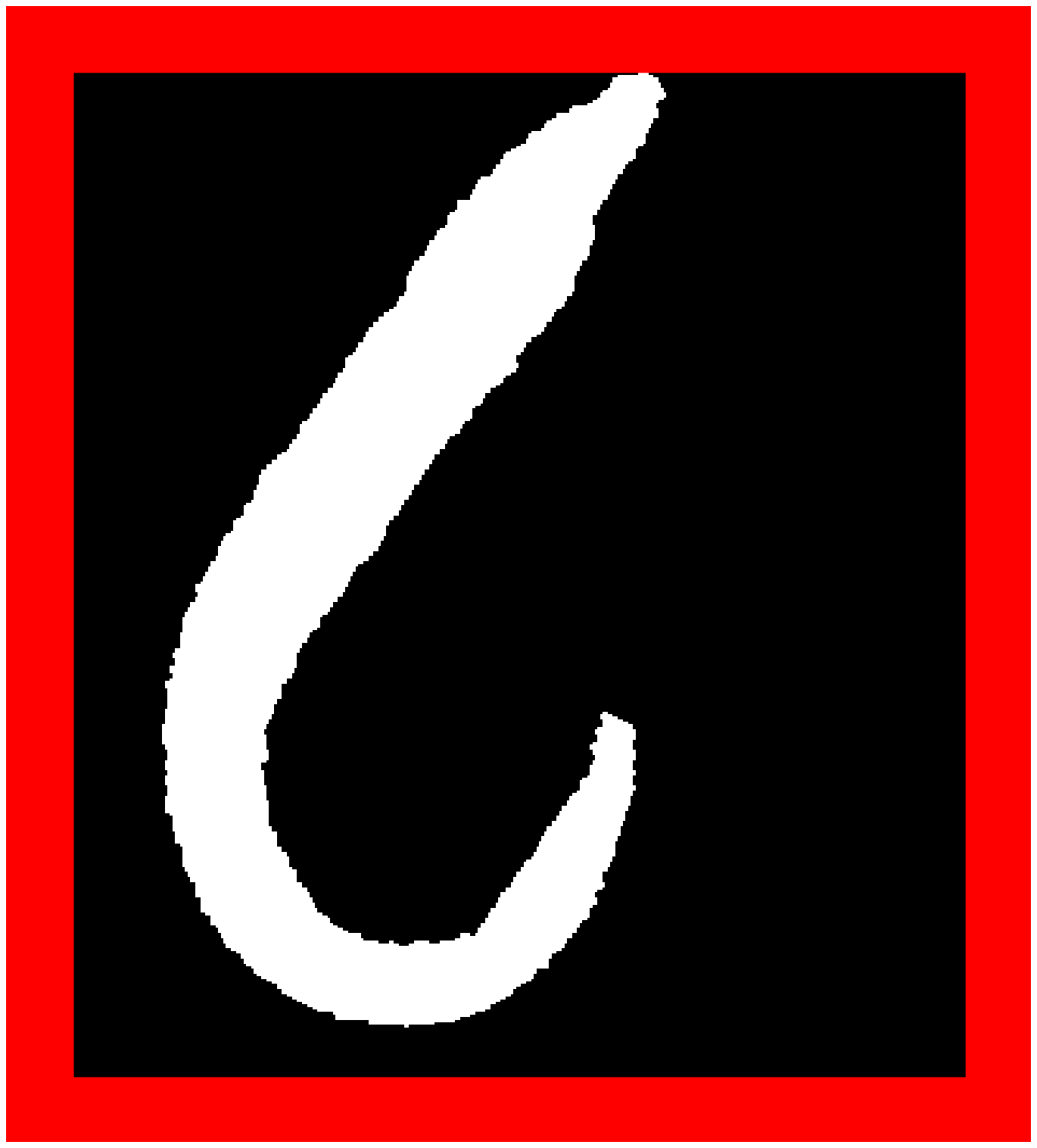}\hspace{-0.215em}\\
\includegraphics[trim=0.1cm 0cm 0.12cm 0cm, clip=true, width=1.2cm, height=1.2cm]{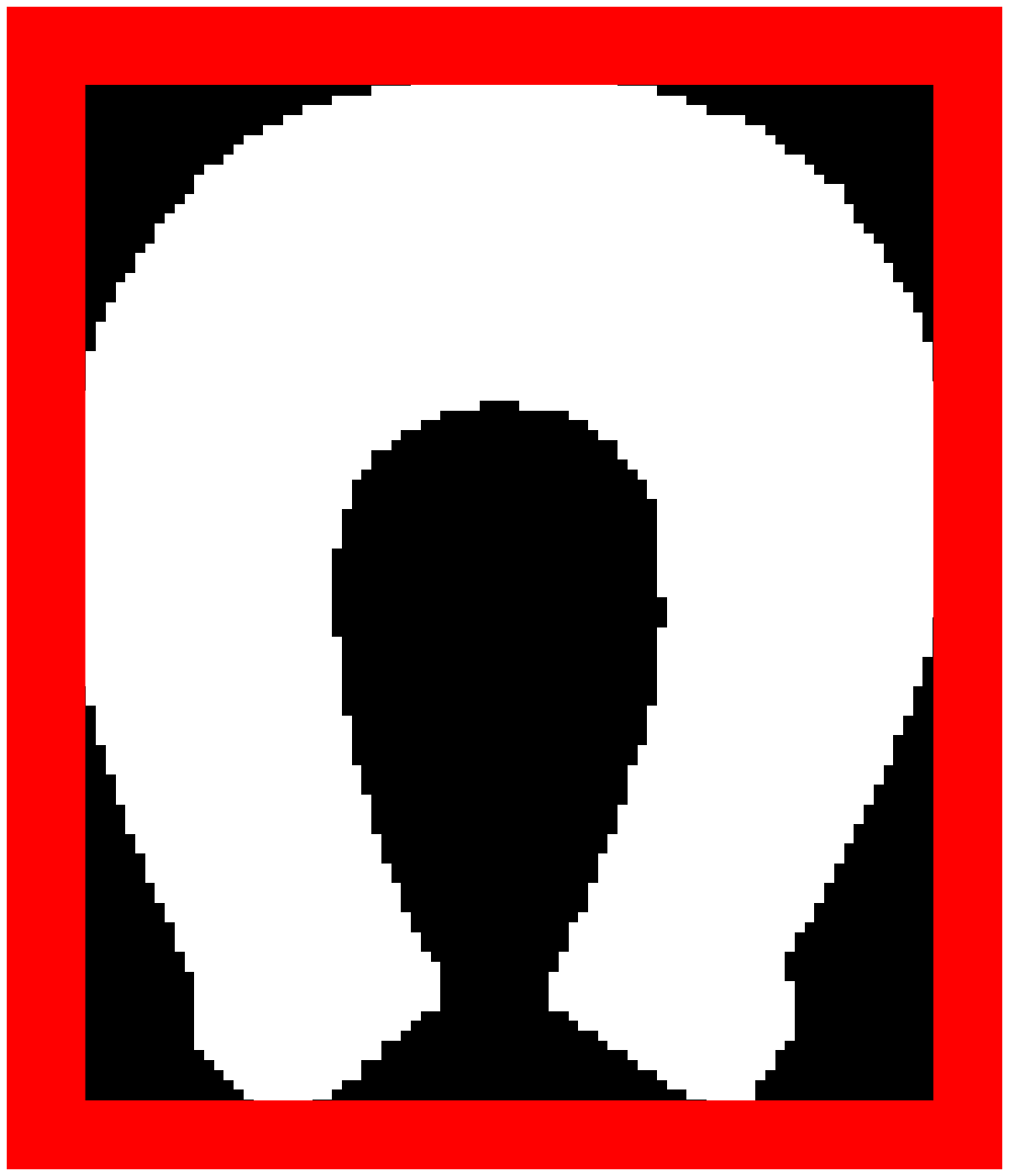}\hspace{-0.215em}
\includegraphics[trim=0.1cm 0cm 0.12cm 0cm, clip=true, width=1.2cm, height=1.2cm]{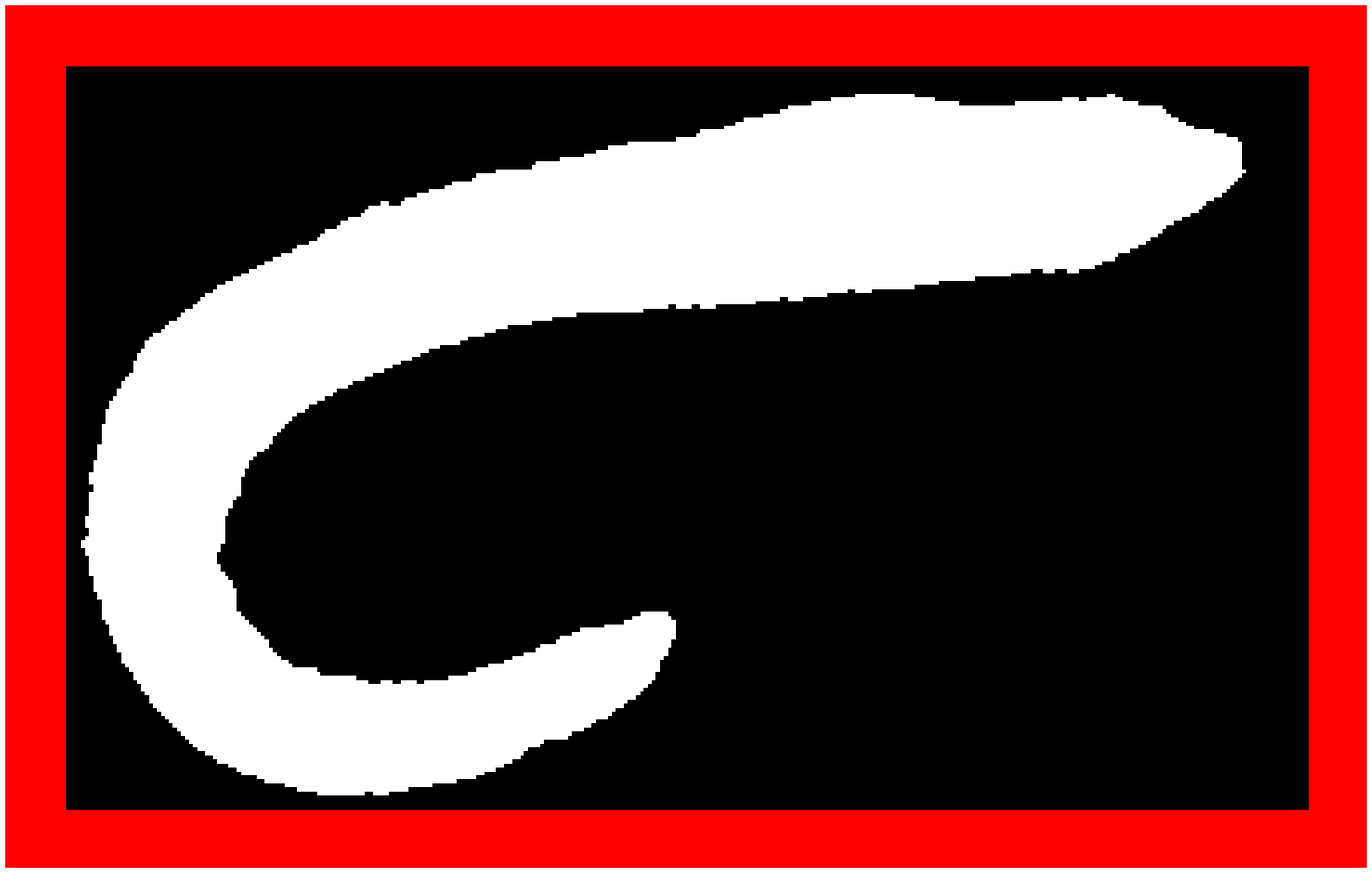}\hspace{-0.215em}
\includegraphics[trim=0.1cm 0cm 0.12cm 0cm, clip=true, width=1.2cm, height=1.2cm]{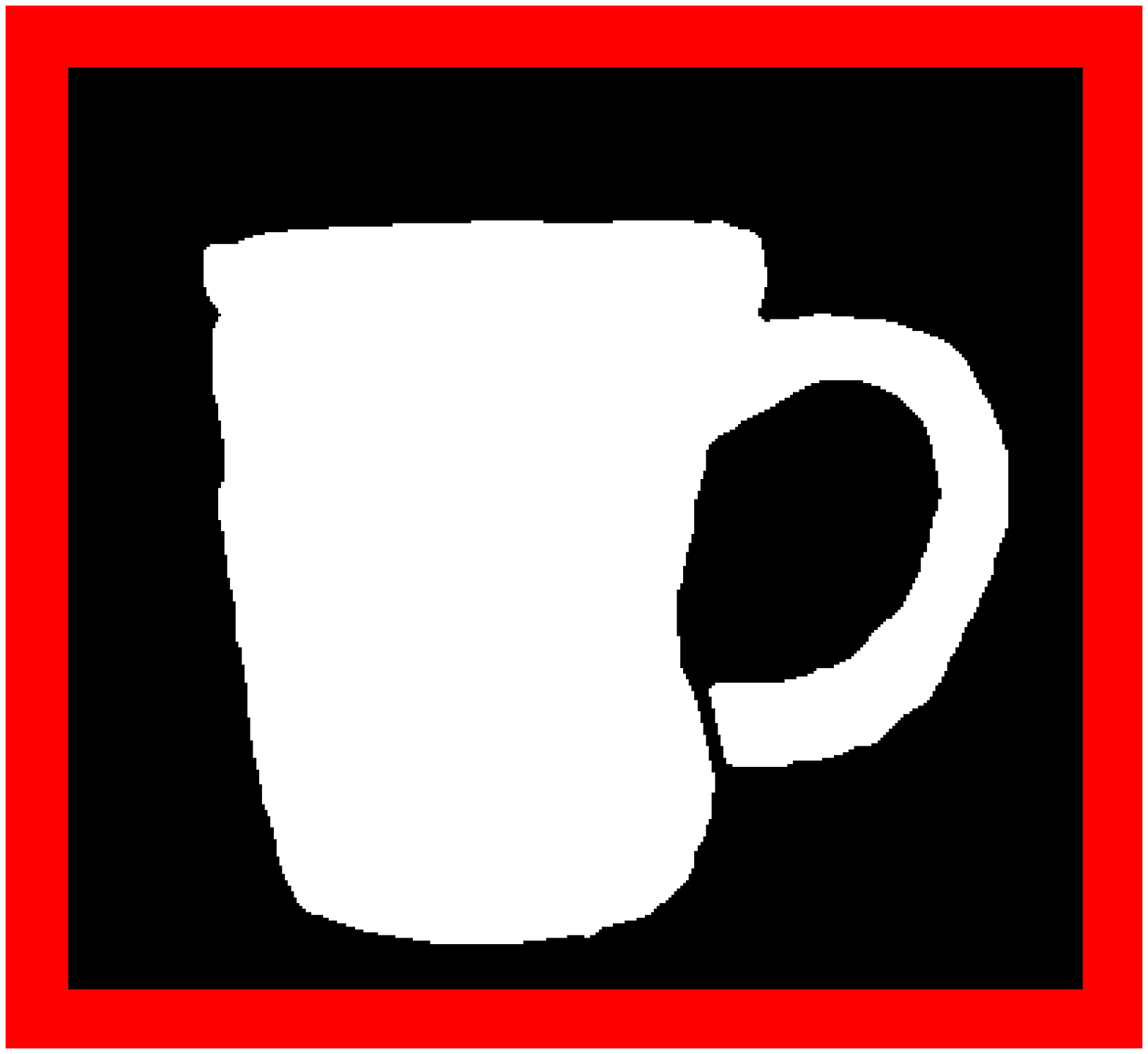}\hspace{-0.215em}
\includegraphics[trim=0.1cm 0cm 0.12cm 0cm, clip=true, width=1.2cm, height=1.2cm]{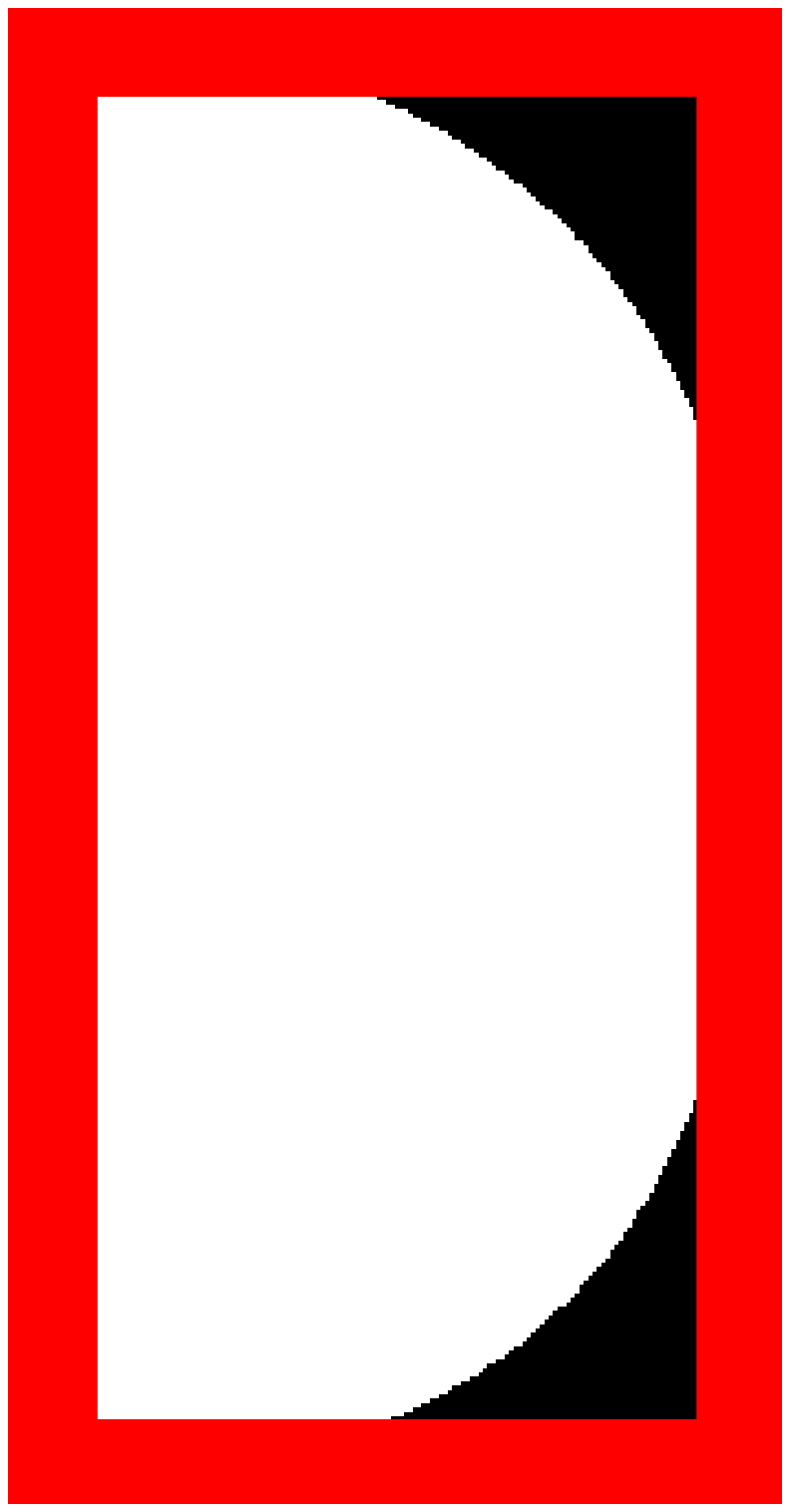}\hspace{-0.215em}
\includegraphics[trim=0.1cm 0cm 0.12cm 0cm, clip=true, width=1.2cm, height=1.2cm]{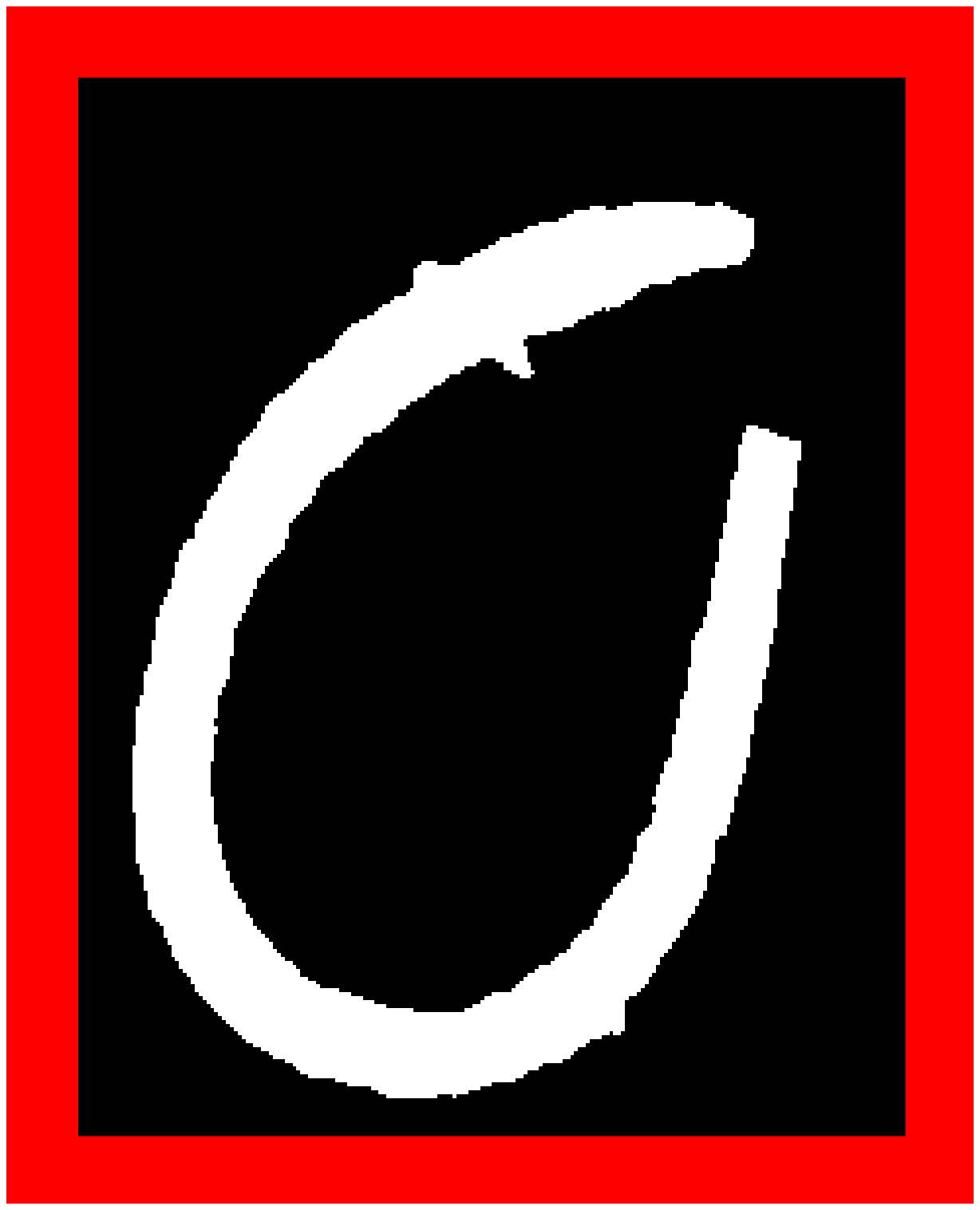}\hspace{-0.215em}
\includegraphics[trim=0.1cm 0cm 0.12cm 0cm, clip=true, width=1.2cm, height=1.2cm]{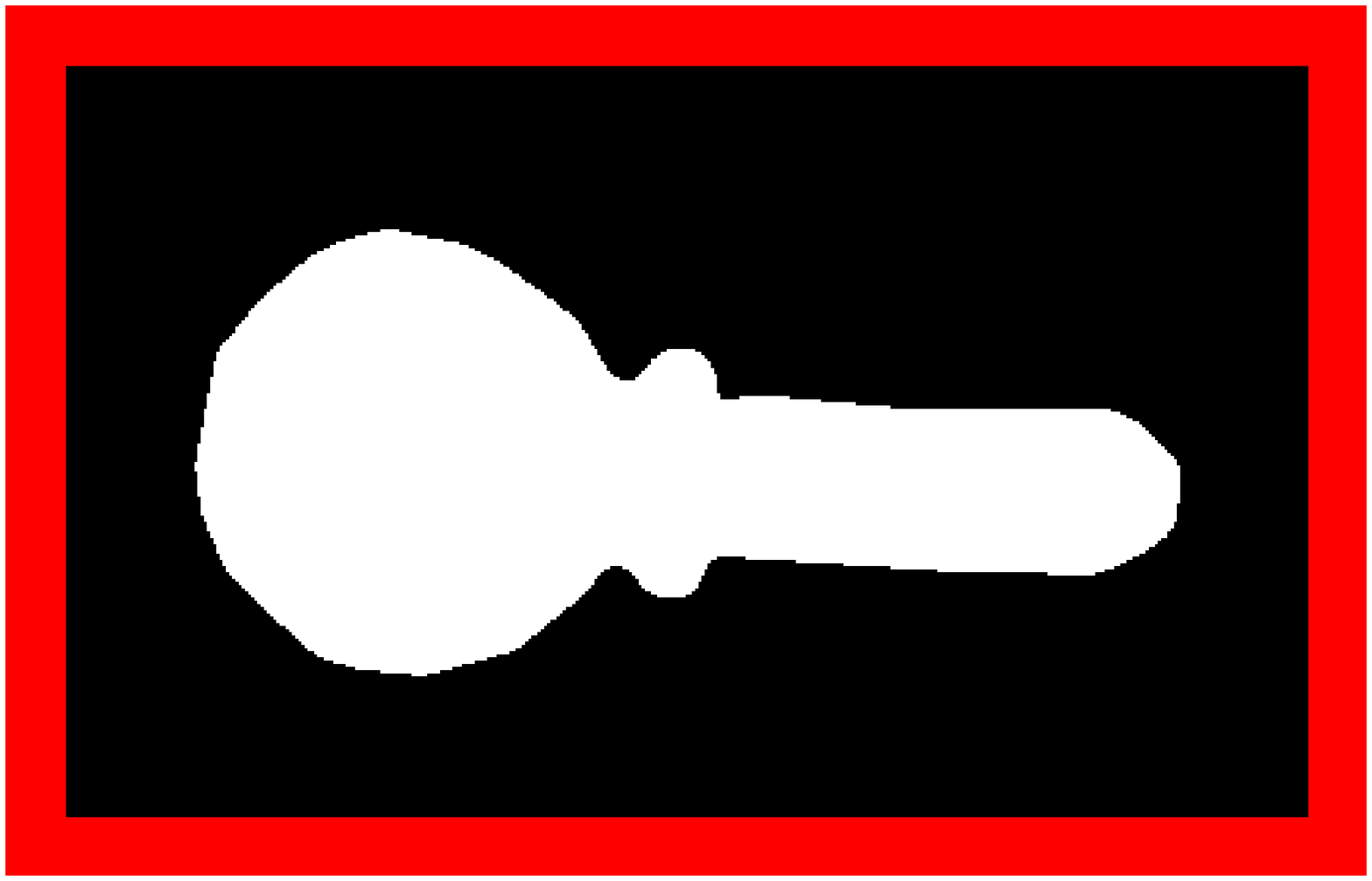}\hspace{-0.215em}
\includegraphics[trim=0.1cm 0cm 0.12cm 0cm, clip=true, width=1.2cm, height=1.2cm]{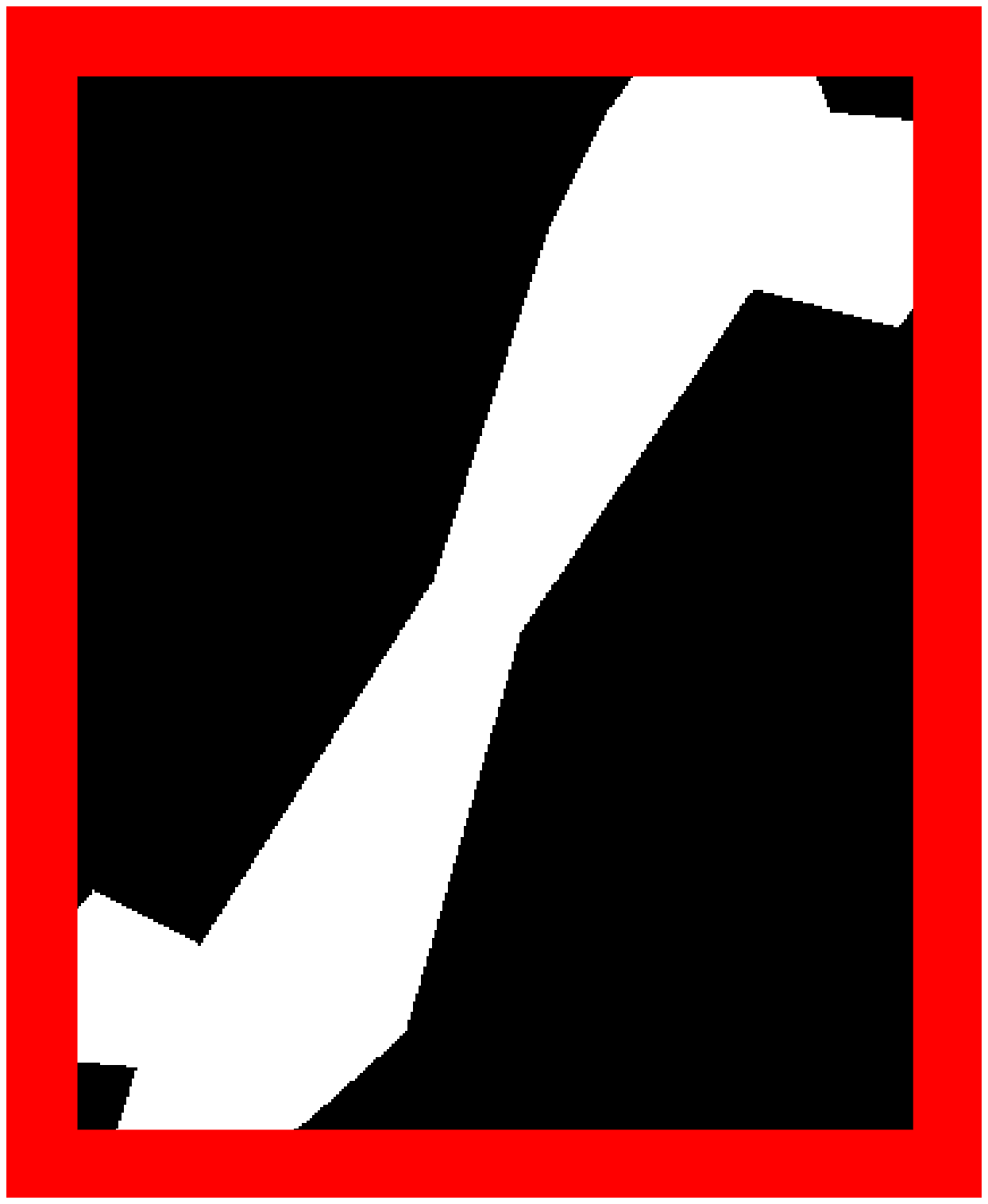}\hspace{-0.215em}
\includegraphics[trim=0.1cm 0cm 0.12cm 0cm, clip=true, width=1.2cm, height=1.2cm]{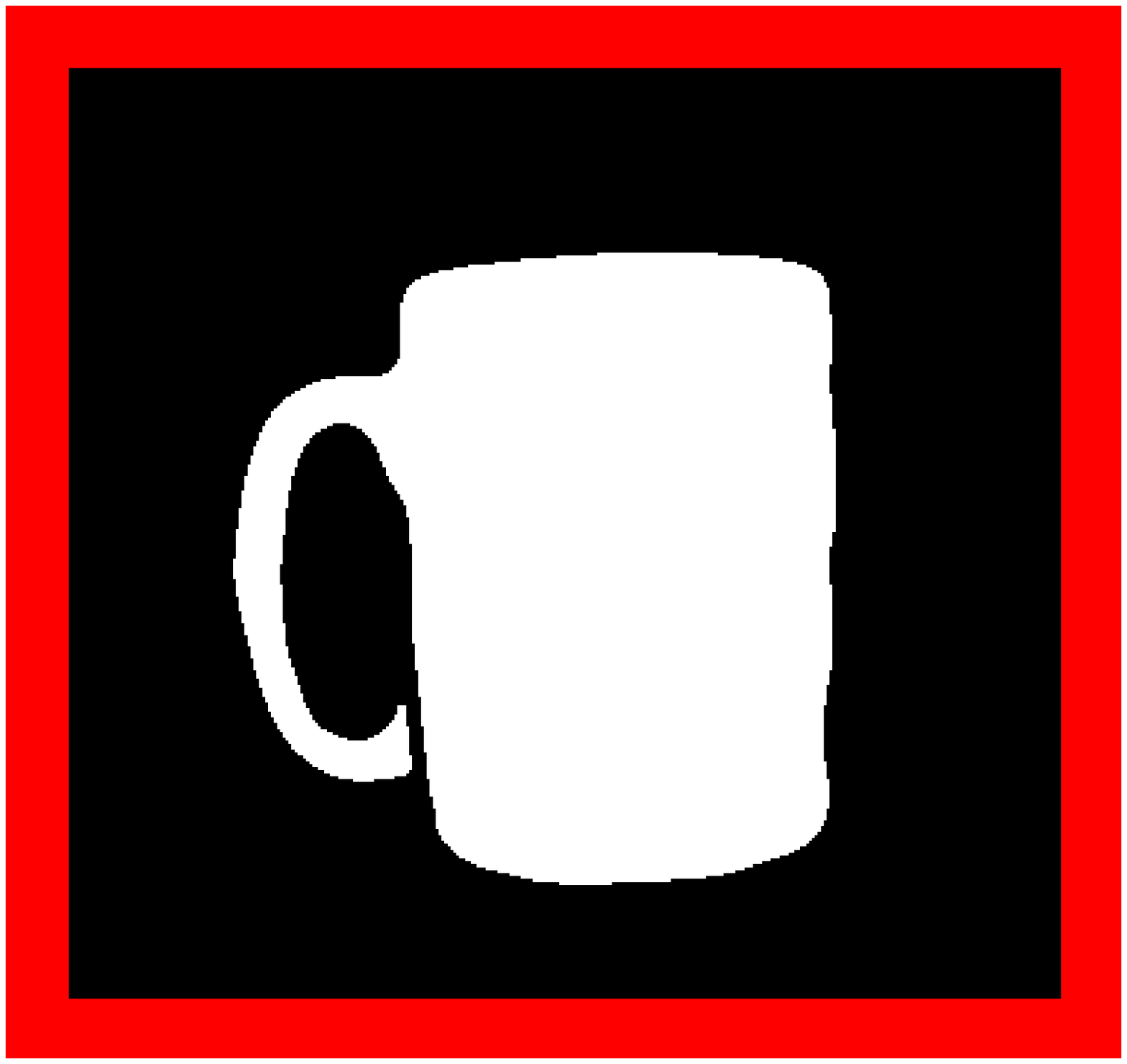}\hspace{-0.215em}
\includegraphics[trim=0.1cm 0cm 0.12cm 0cm, clip=true, width=1.2cm, height=1.2cm]{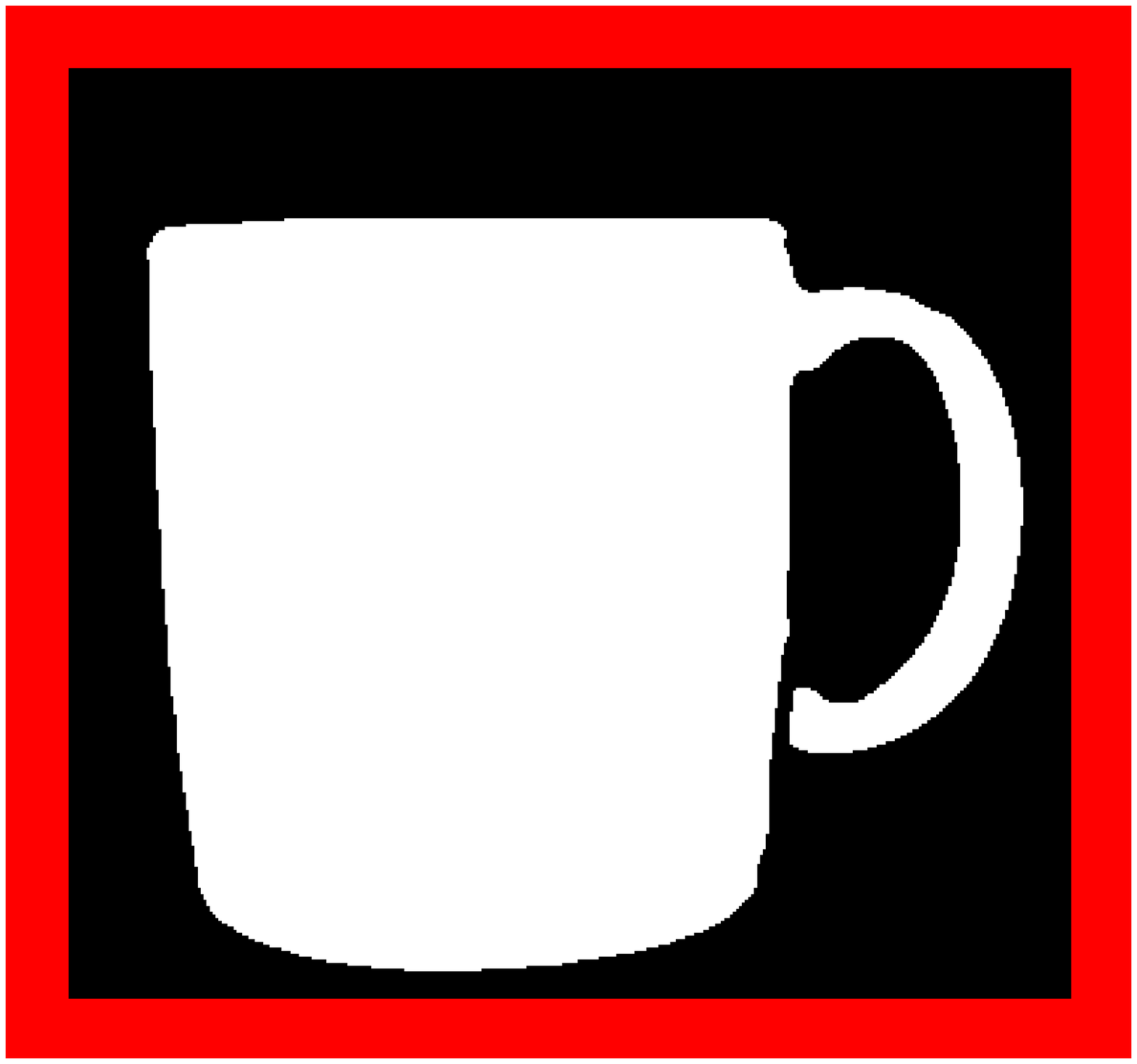}\hspace{-0.215em}
\includegraphics[trim=0.1cm 0cm 0.12cm 0cm, clip=true, width=1.2cm, height=1.2cm]{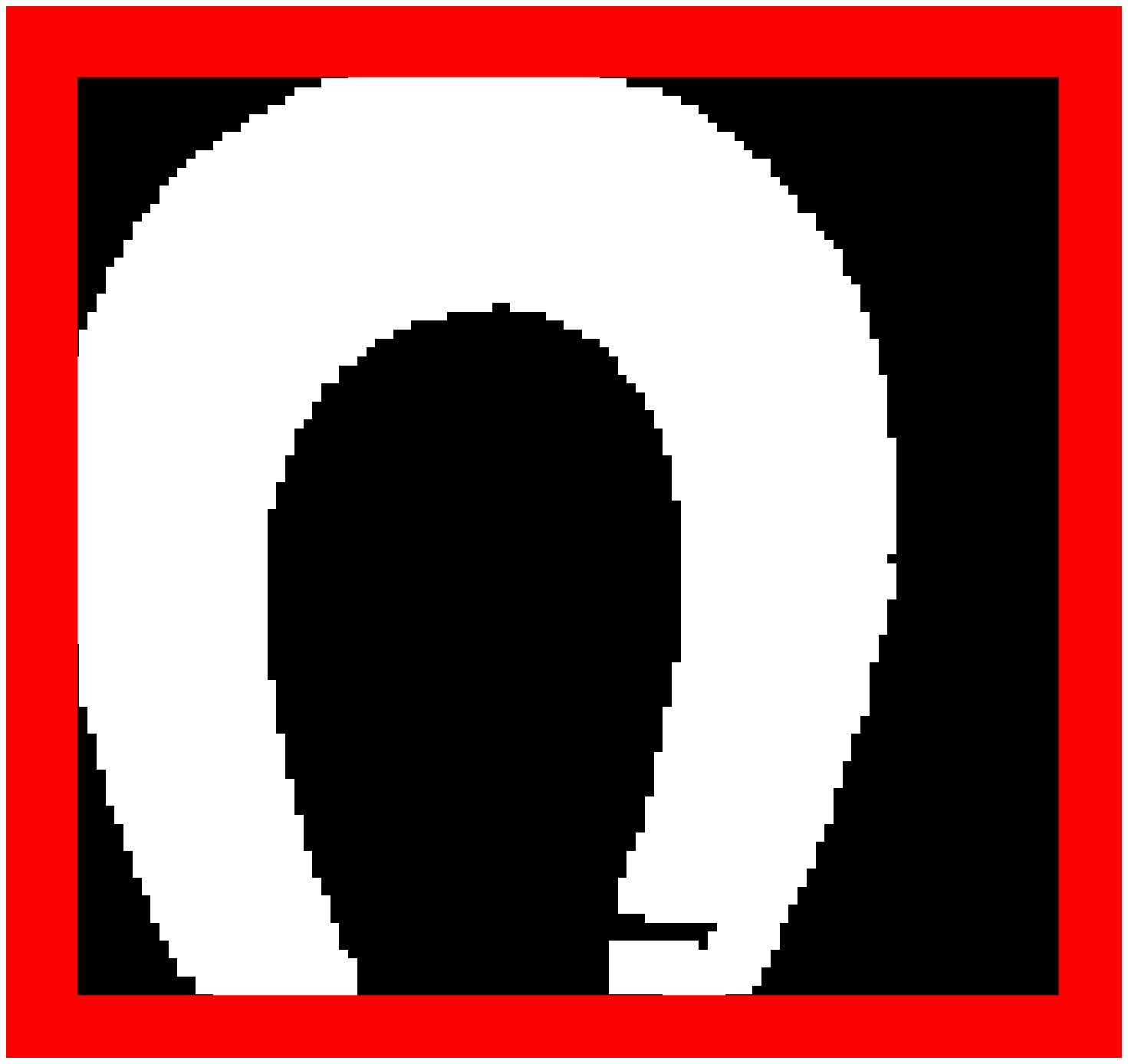}\hspace{-0.215em}}}

\subfloat[SSC]{\shortstack{\includegraphics[trim=0.1cm 0cm 0.12cm 0cm, clip=true, width=1.2cm, height=1.2cm]{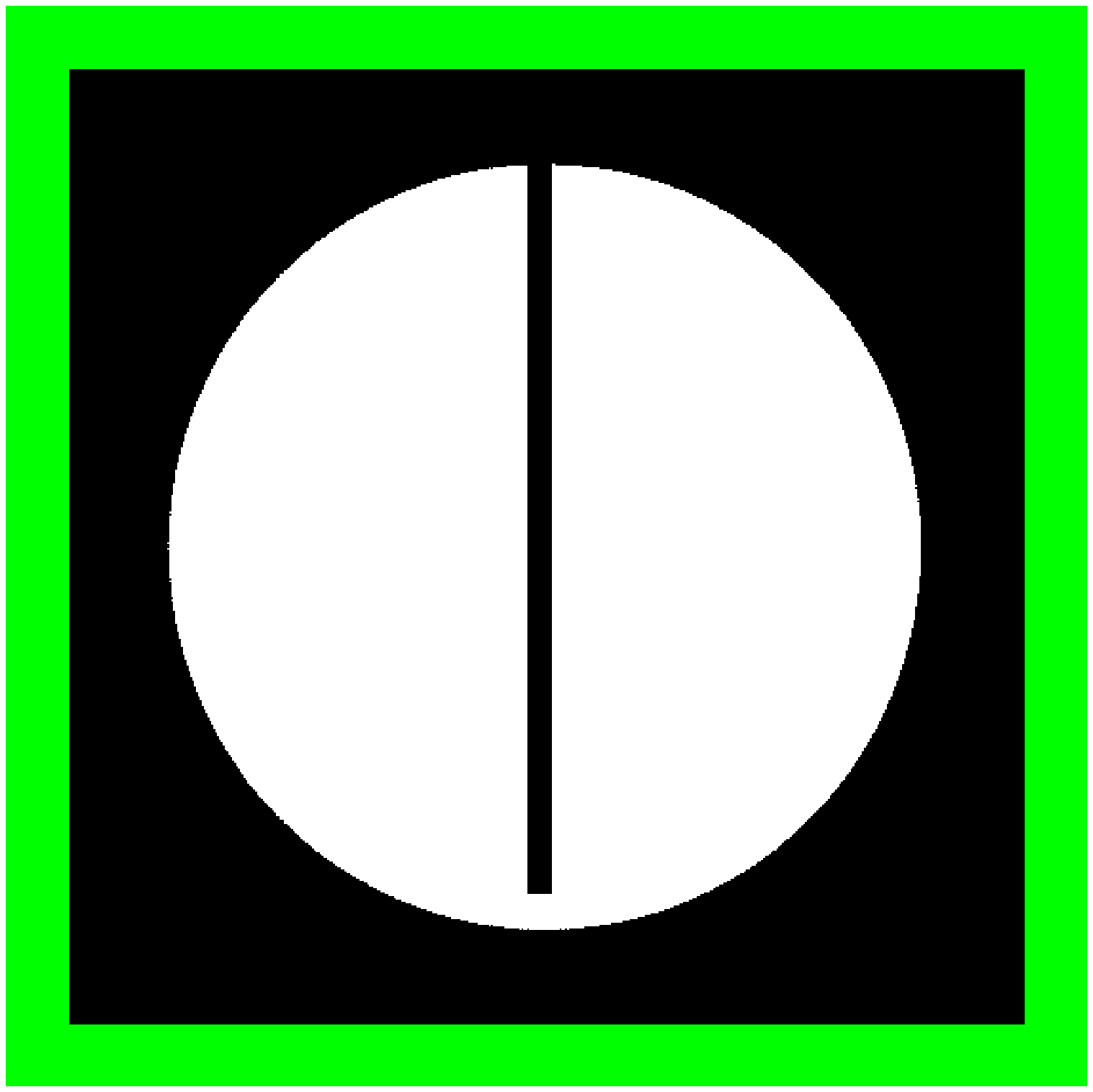}\hspace{-0.215em}
\includegraphics[trim=0.1cm 0cm 0.12cm 0cm, clip=true, width=1.2cm, height=1.2cm]{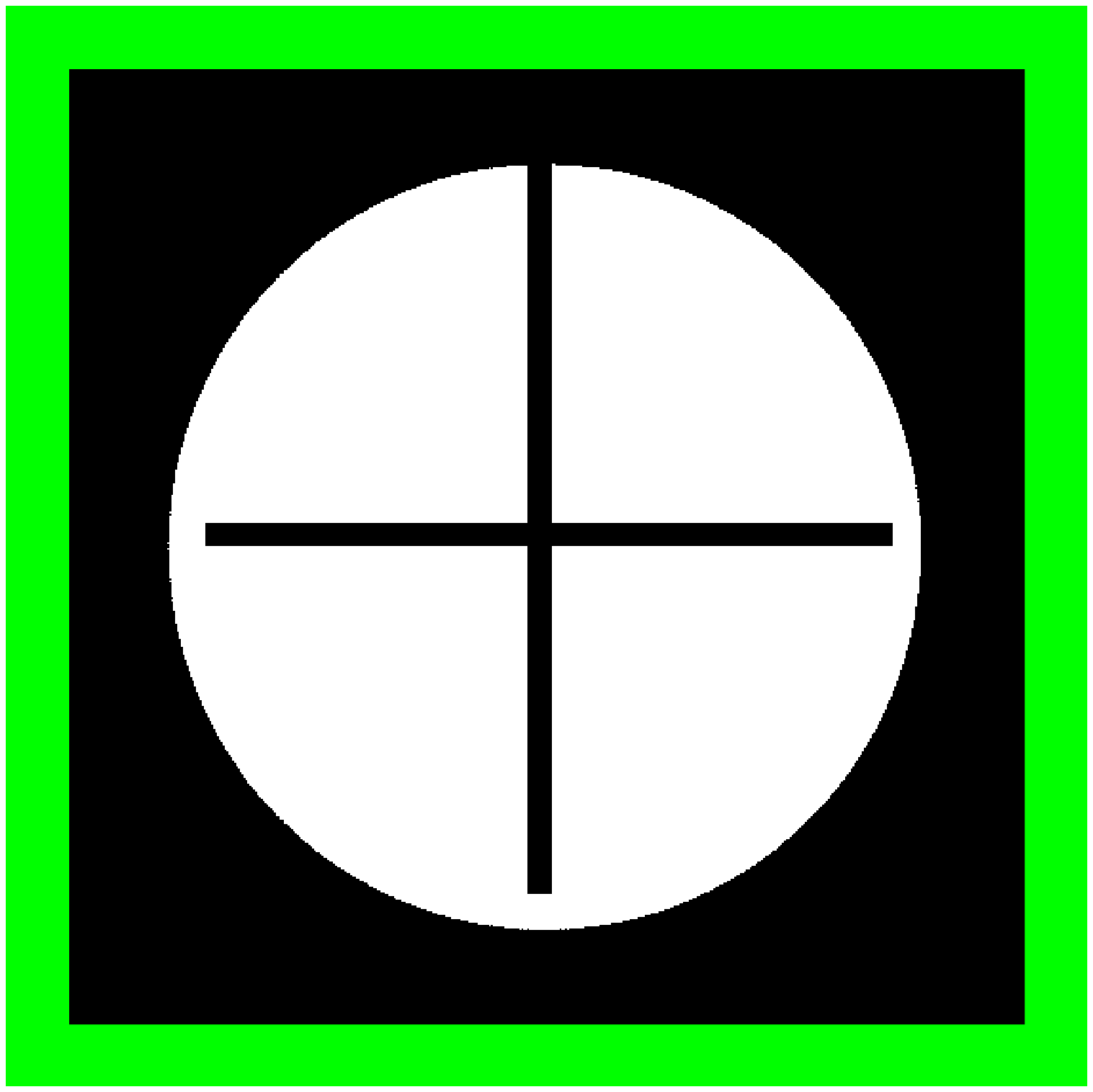}\hspace{-0.215em}
\includegraphics[trim=0.1cm 0cm 0.12cm 0cm, clip=true, width=1.2cm, height=1.2cm]{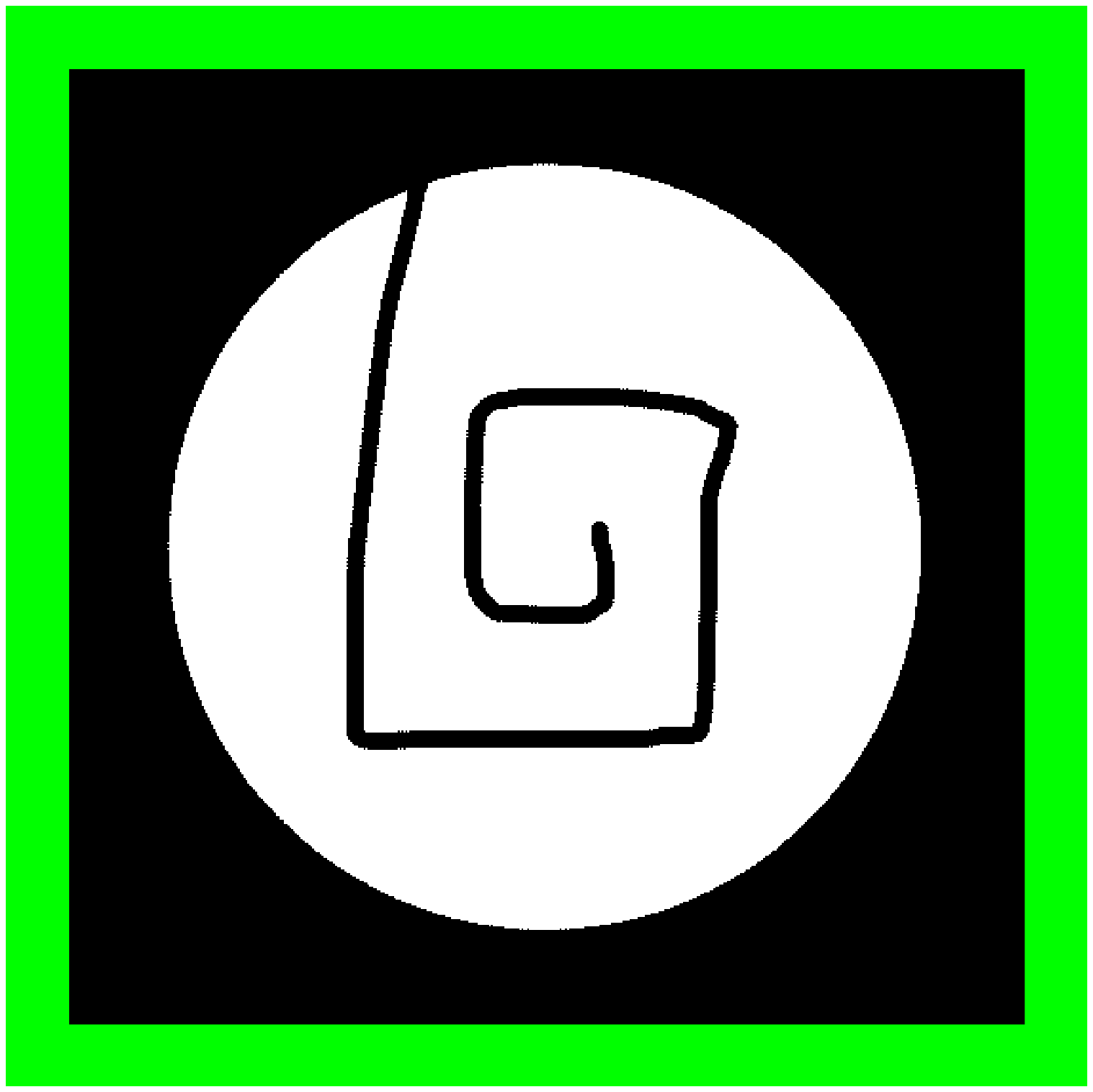}\hspace{-0.215em}
\includegraphics[trim=0.1cm 0cm 0.12cm 0cm, clip=true, width=1.2cm, height=1.2cm]{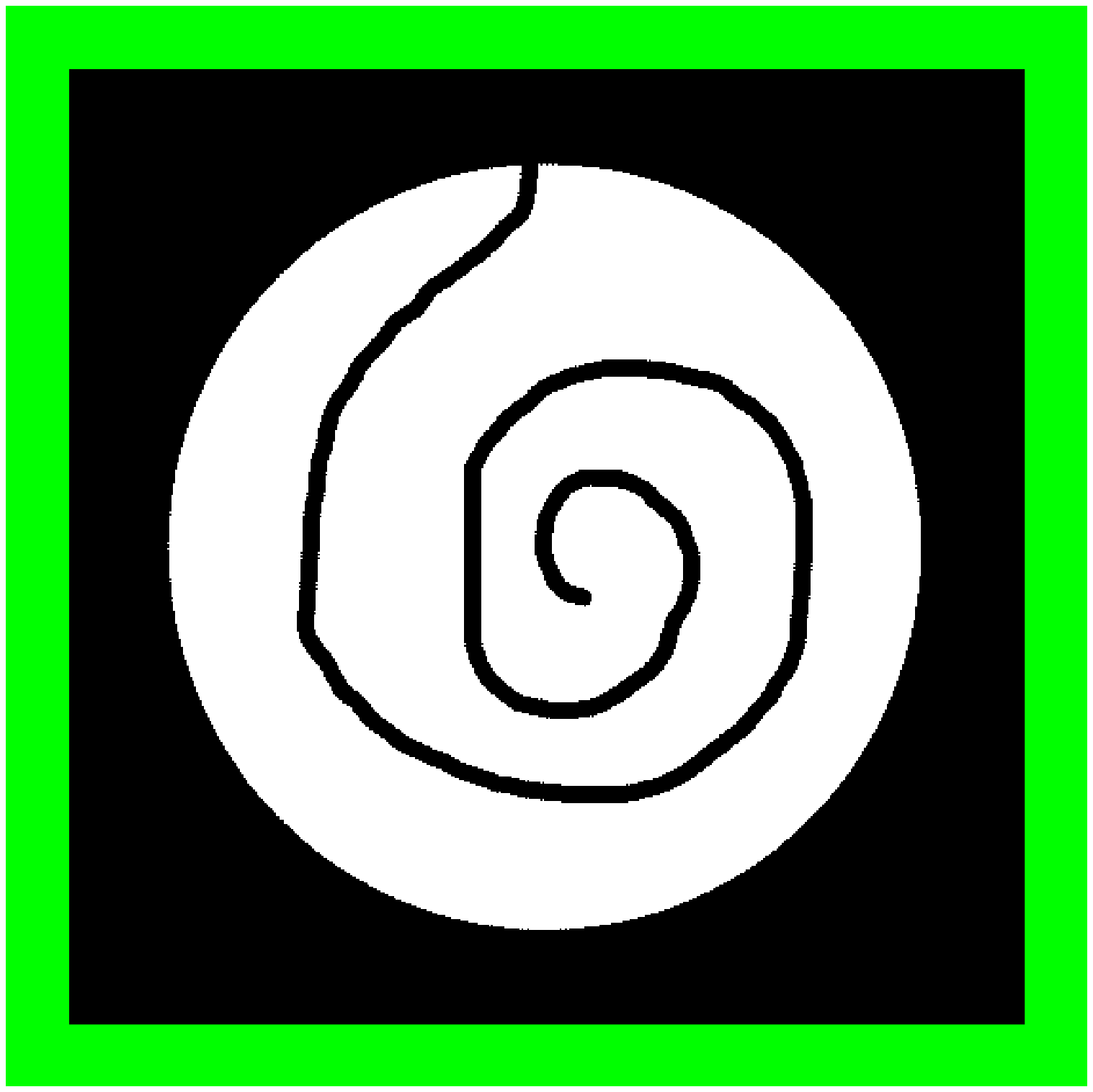}\hspace{-0.215em}
\includegraphics[trim=0.1cm 0cm 0.12cm 0cm, clip=true, width=1.2cm, height=1.2cm]{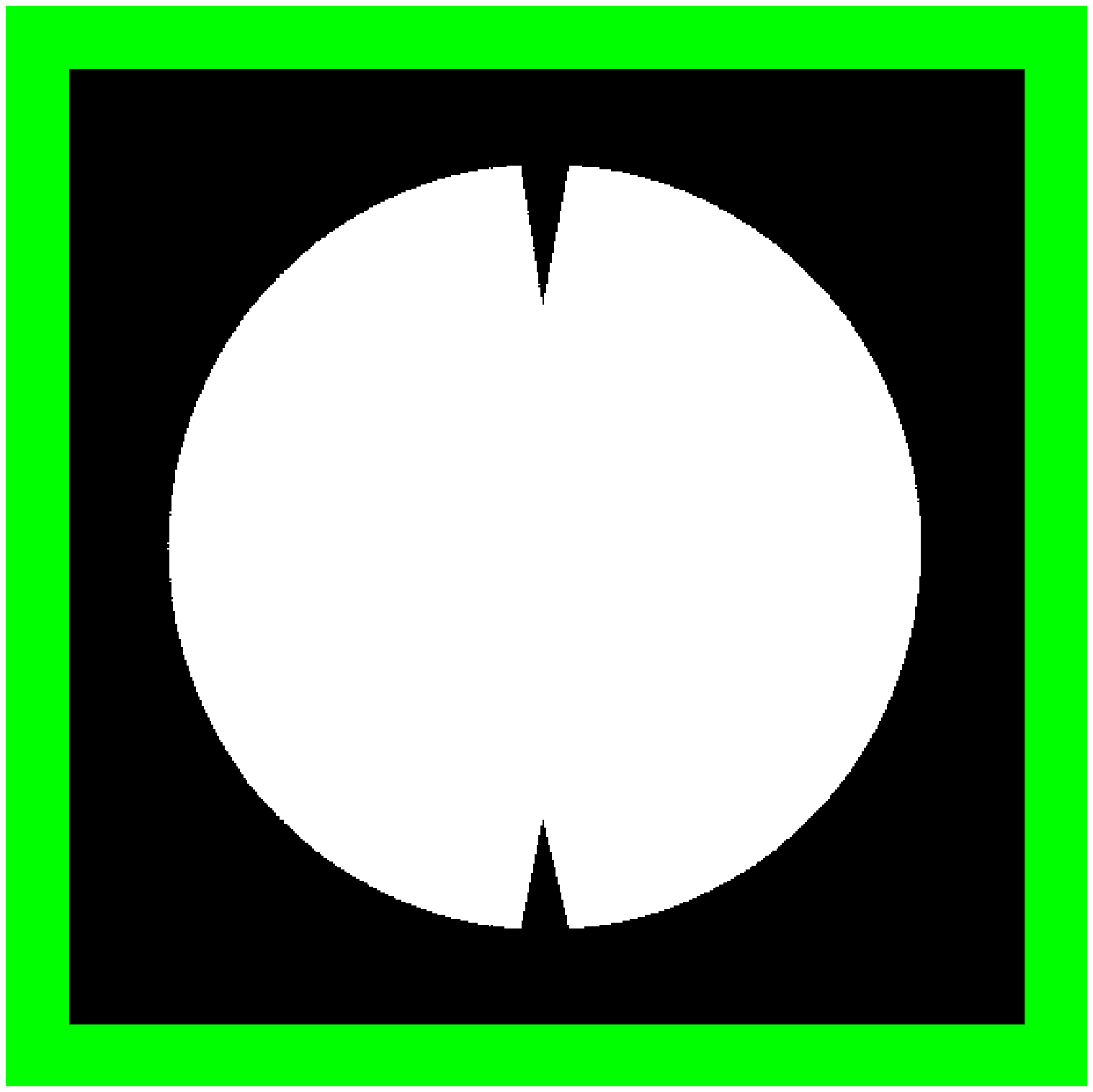}\hspace{-0.215em}
\includegraphics[trim=0.1cm 0cm 0.12cm 0cm, clip=true, width=1.2cm, height=1.2cm]{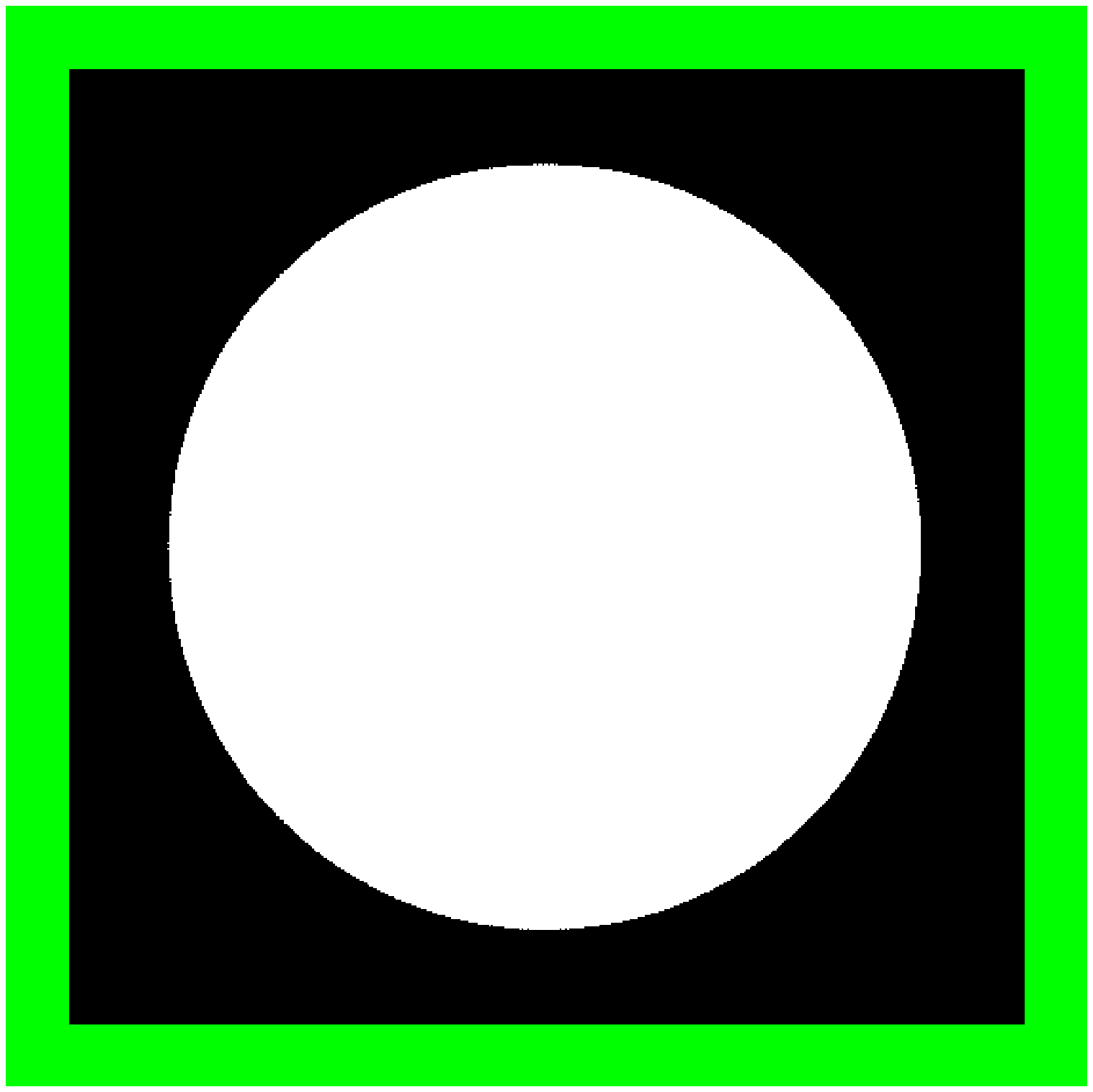}\hspace{-0.215em}
\includegraphics[trim=0.1cm 0cm 0.12cm 0cm, clip=true, width=1.2cm, height=1.2cm]{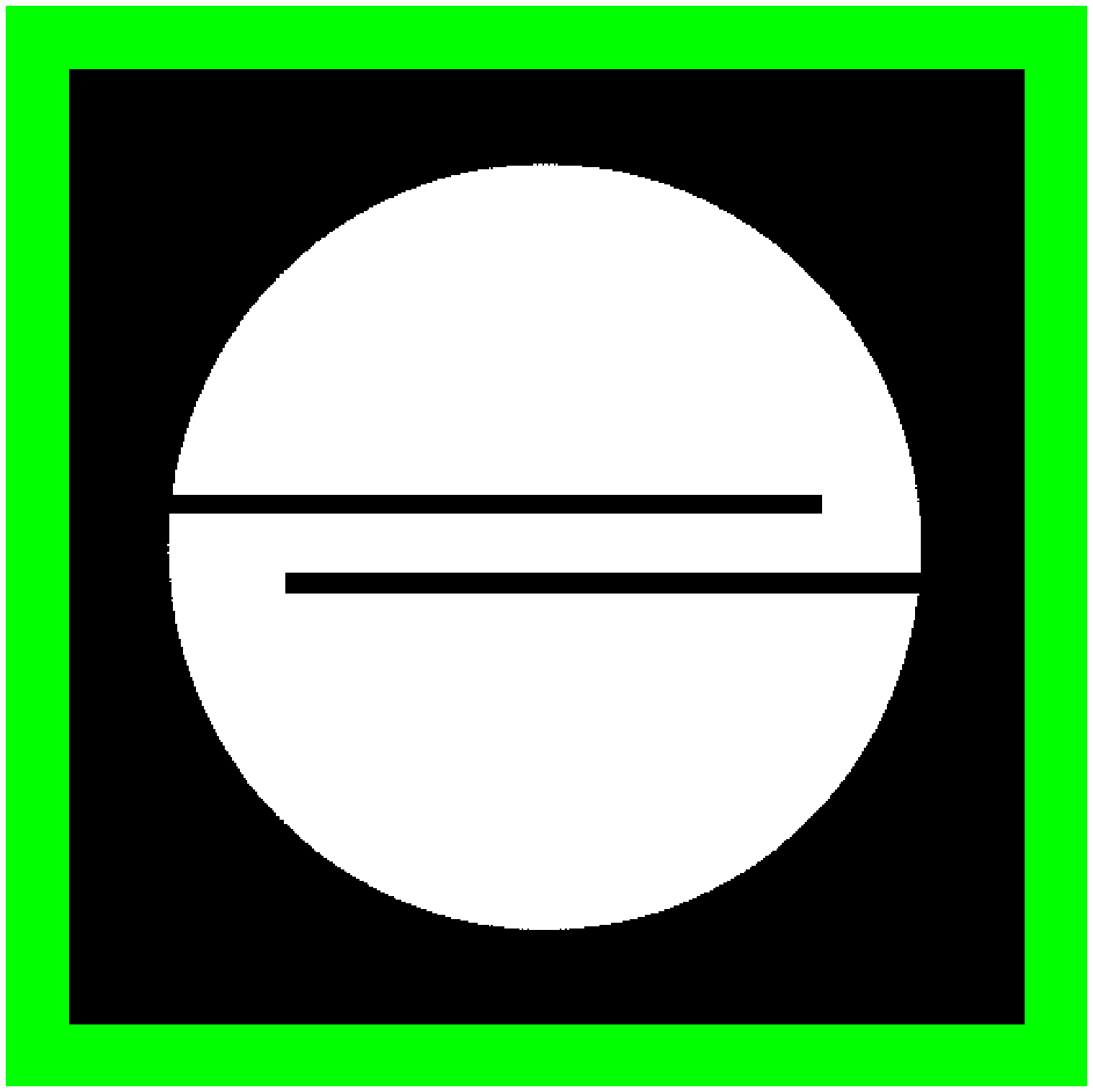}\hspace{-0.215em}
\includegraphics[trim=0.1cm 0cm 0.12cm 0cm, clip=true, width=1.2cm, height=1.2cm]{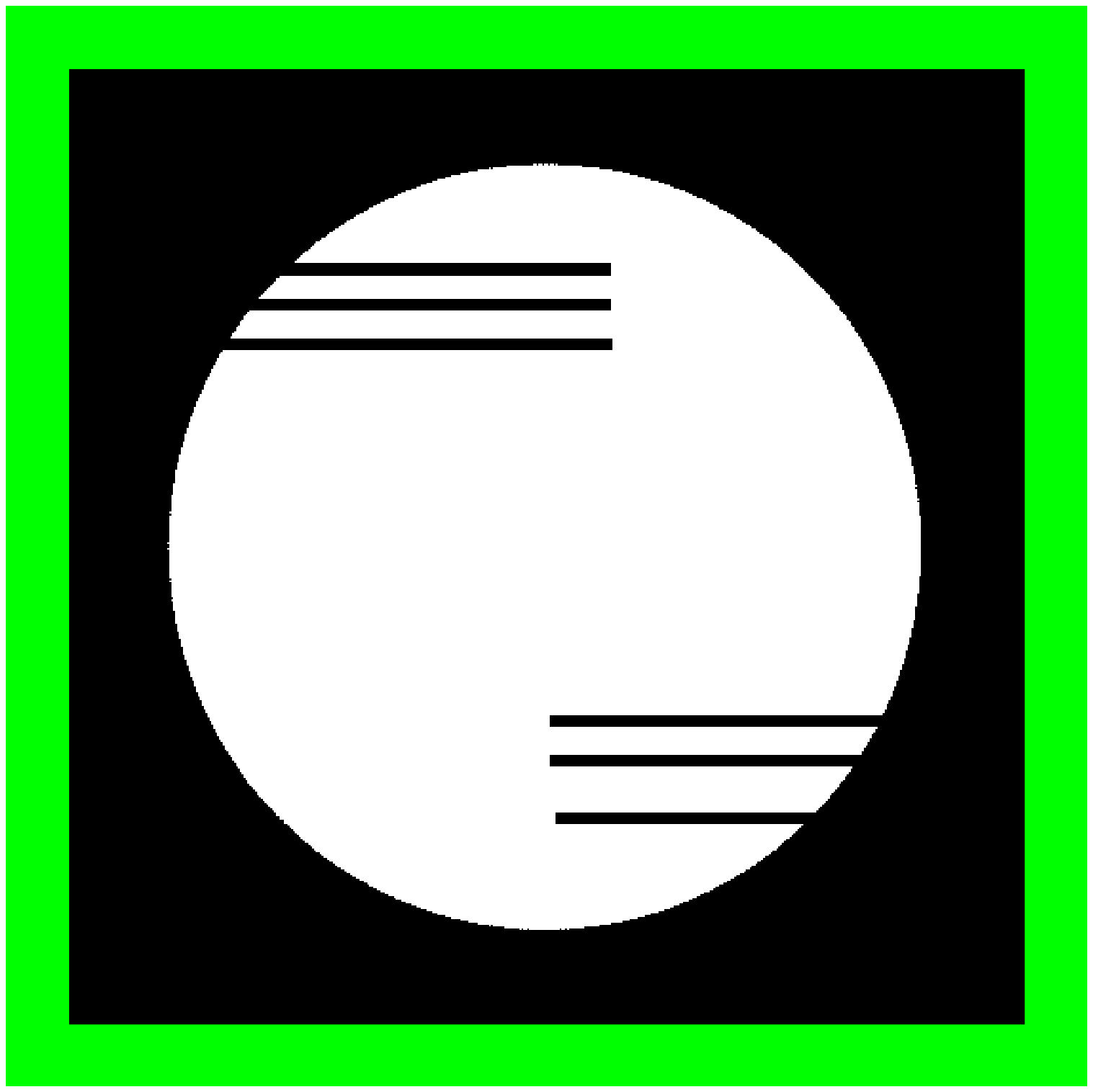}\hspace{-0.215em}
\includegraphics[trim=0.1cm 0cm 0.12cm 0cm, clip=true, width=1.2cm, height=1.2cm]{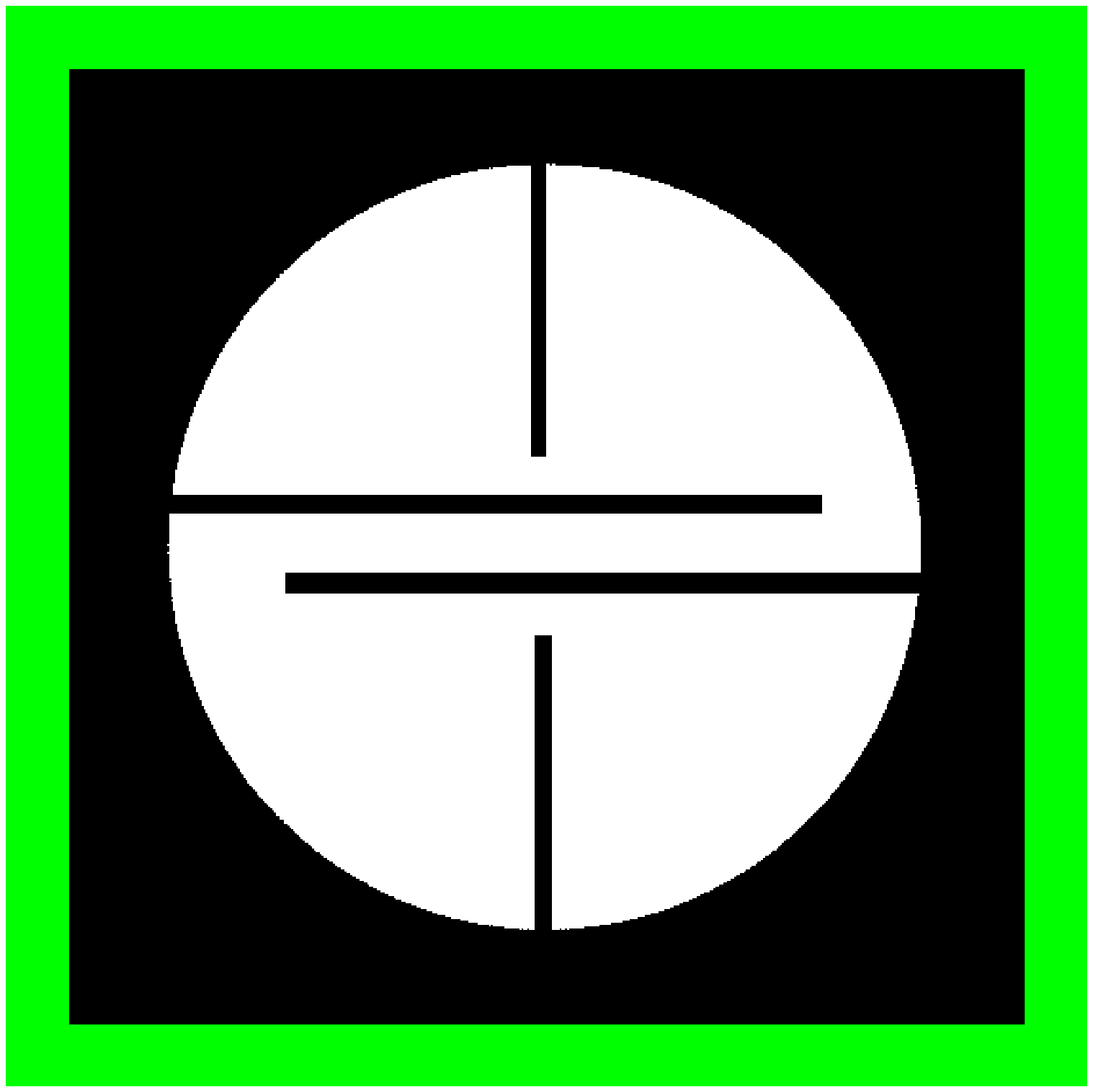}\hspace{-0.215em}
\includegraphics[trim=0.1cm 0cm 0.12cm 0cm, clip=true, width=1.2cm, height=1.2cm]{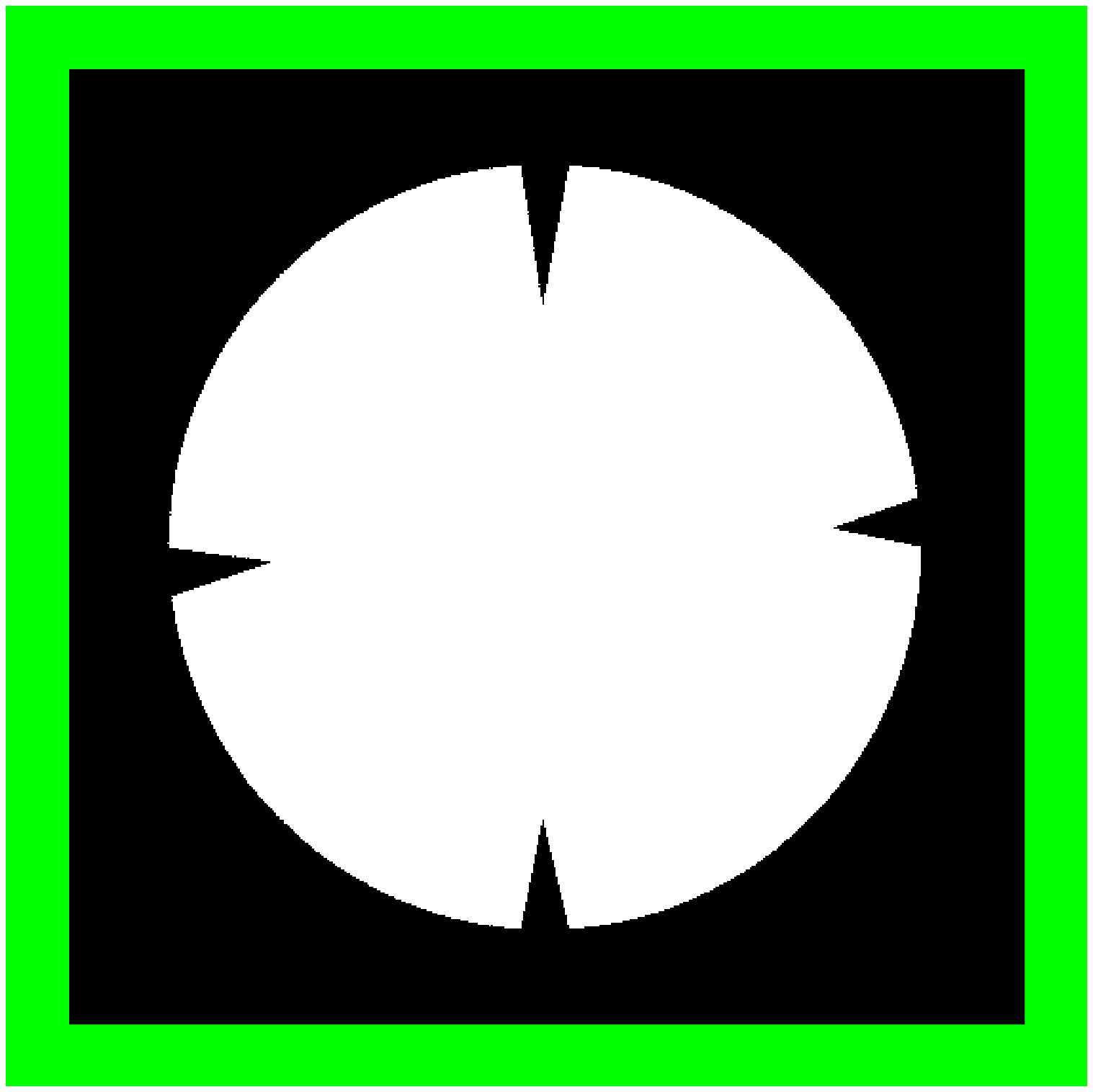}\hspace{-0.215em}\\
\includegraphics[trim=0.1cm 0cm 0.12cm 0cm, clip=true, width=1.2cm, height=1.2cm]{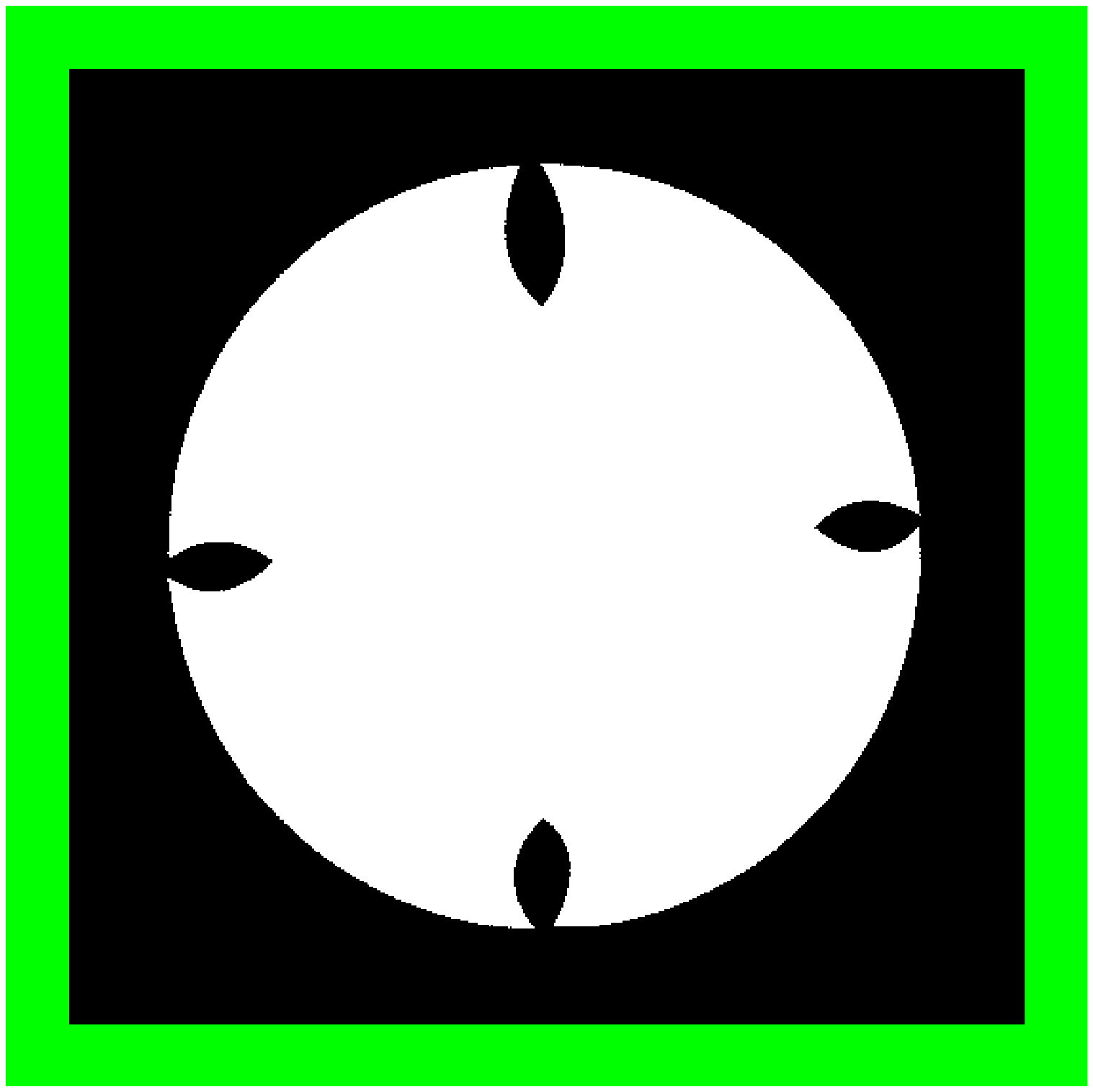}\hspace{-0.215em}
\includegraphics[trim=0.1cm 0cm 0.12cm 0cm, clip=true, width=1.2cm, height=1.2cm]{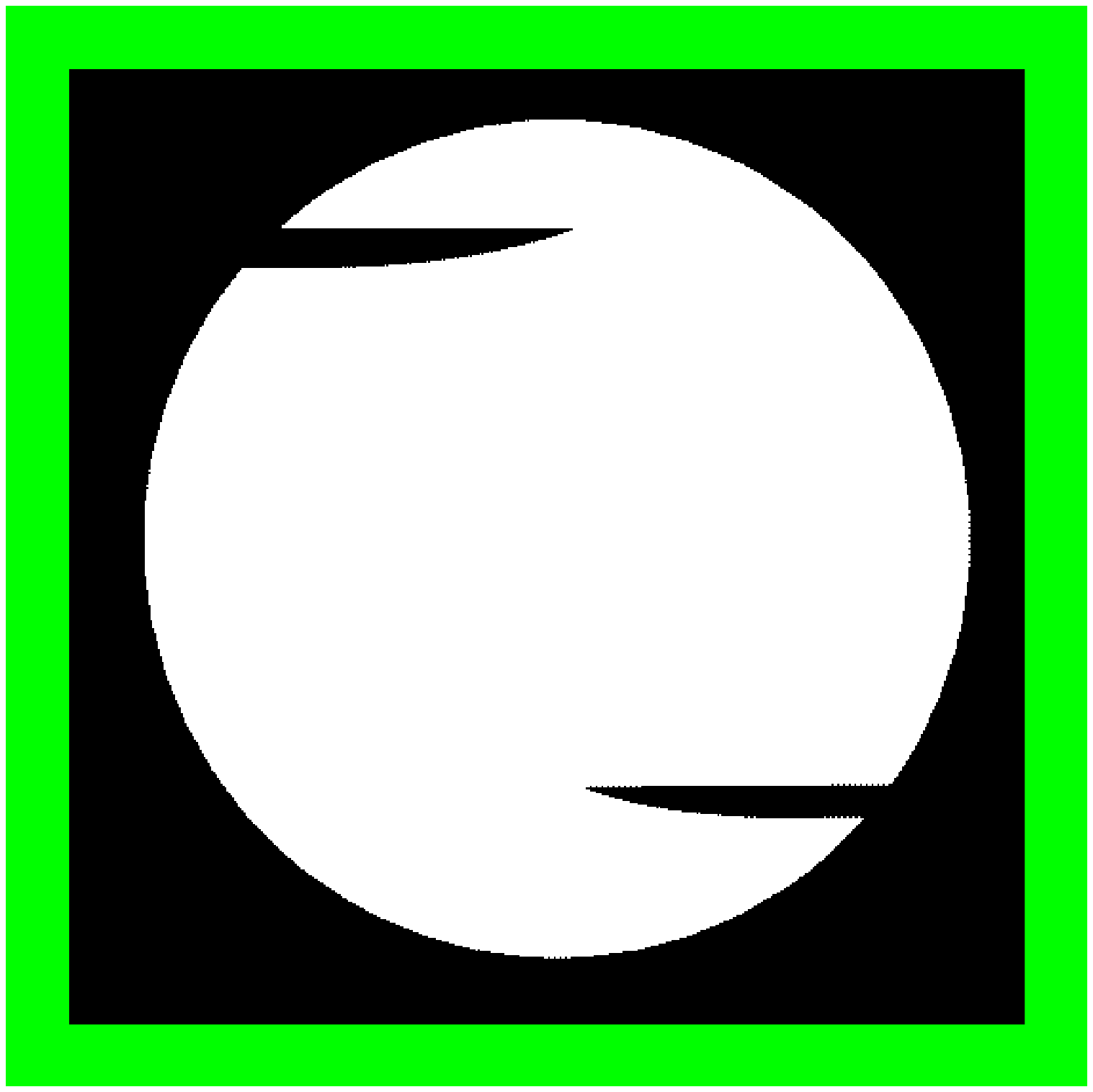}\hspace{-0.215em}
\includegraphics[trim=0.1cm 0cm 0.12cm 0cm, clip=true, width=1.2cm, height=1.2cm]{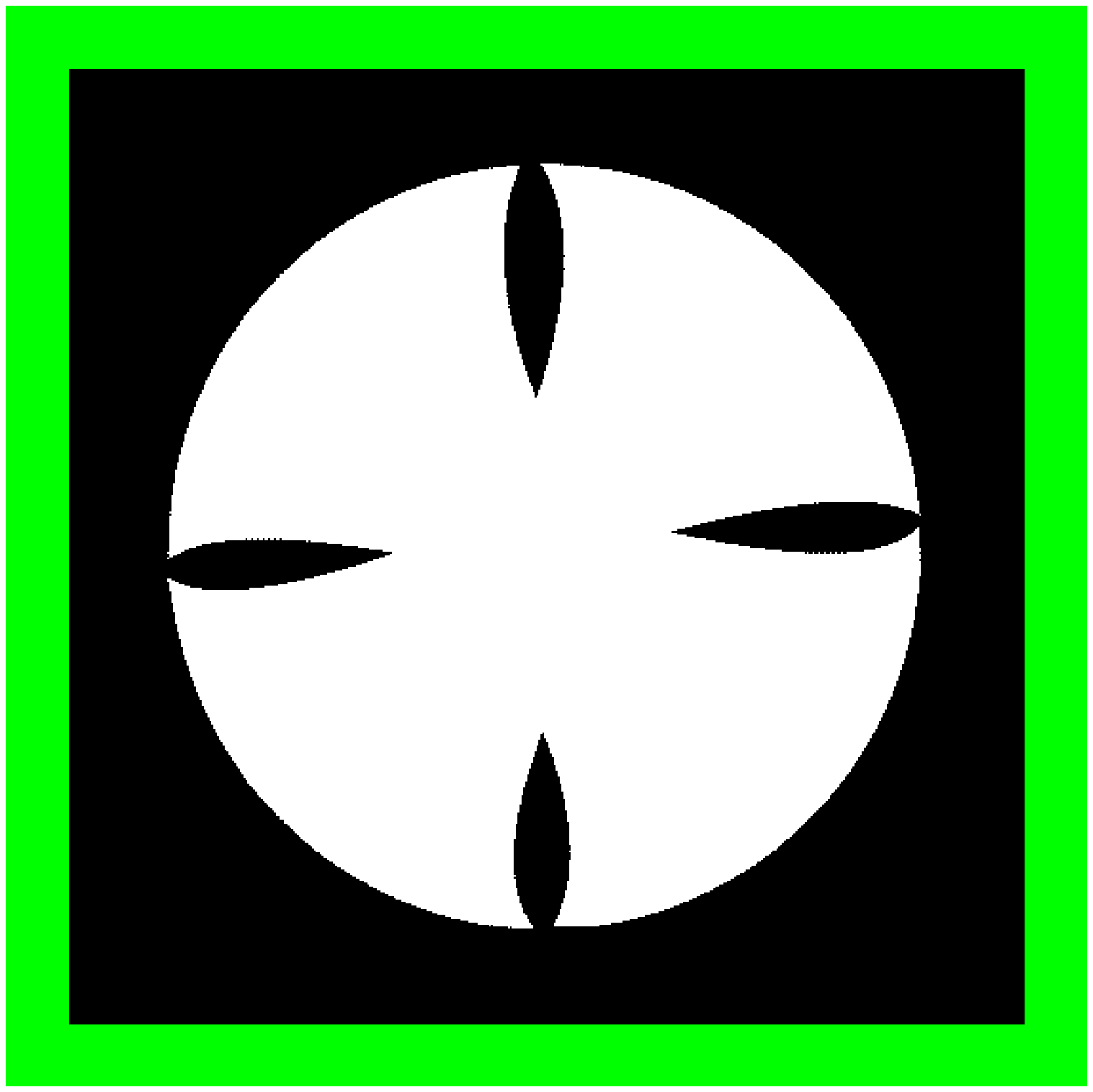}\hspace{-0.215em}
\includegraphics[trim=0.1cm 0cm 0.12cm 0cm, clip=true, width=1.2cm, height=1.2cm]{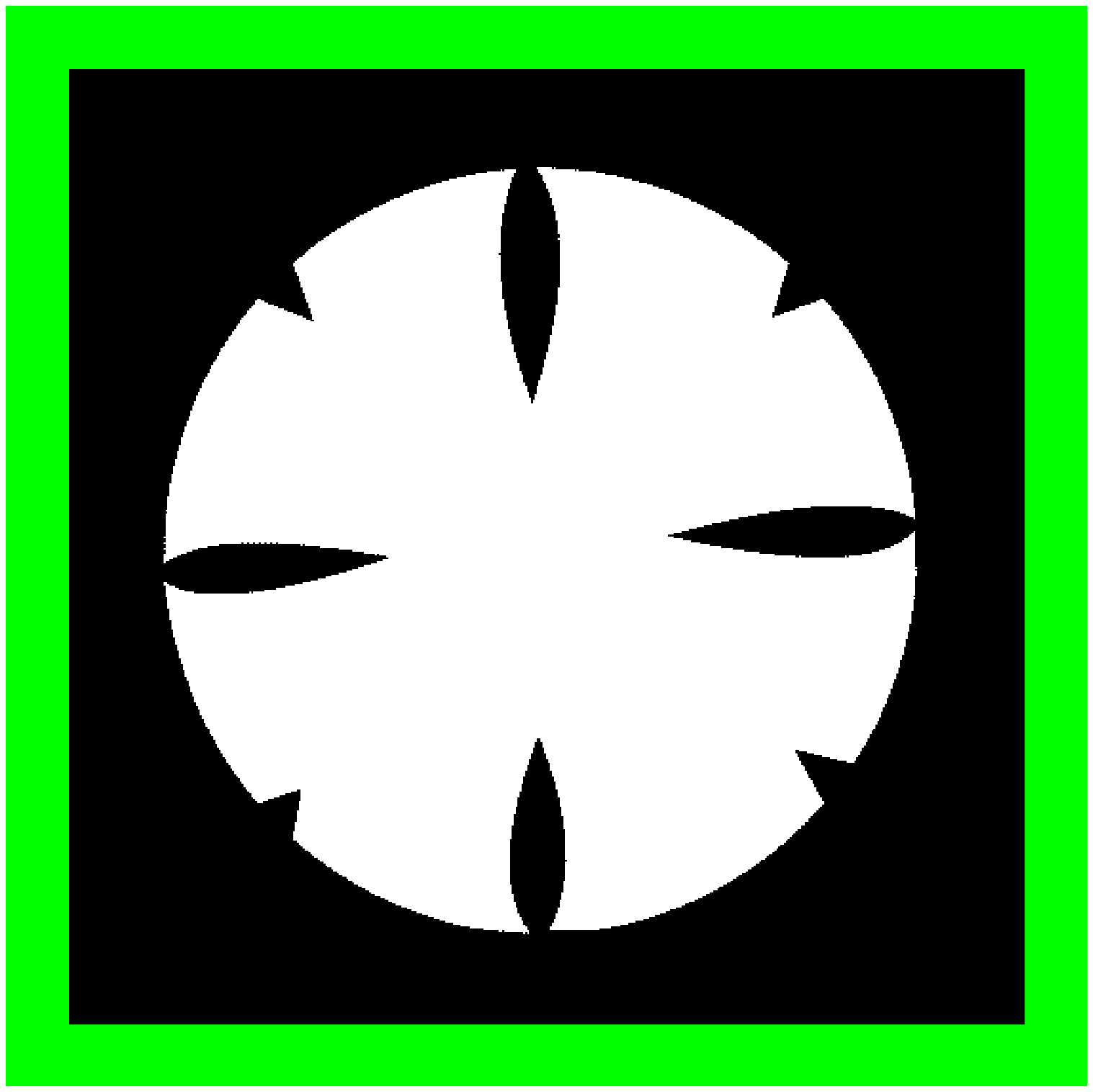}\hspace{-0.215em}
\includegraphics[trim=0.1cm 0cm 0.12cm 0cm, clip=true, width=1.2cm, height=1.2cm]{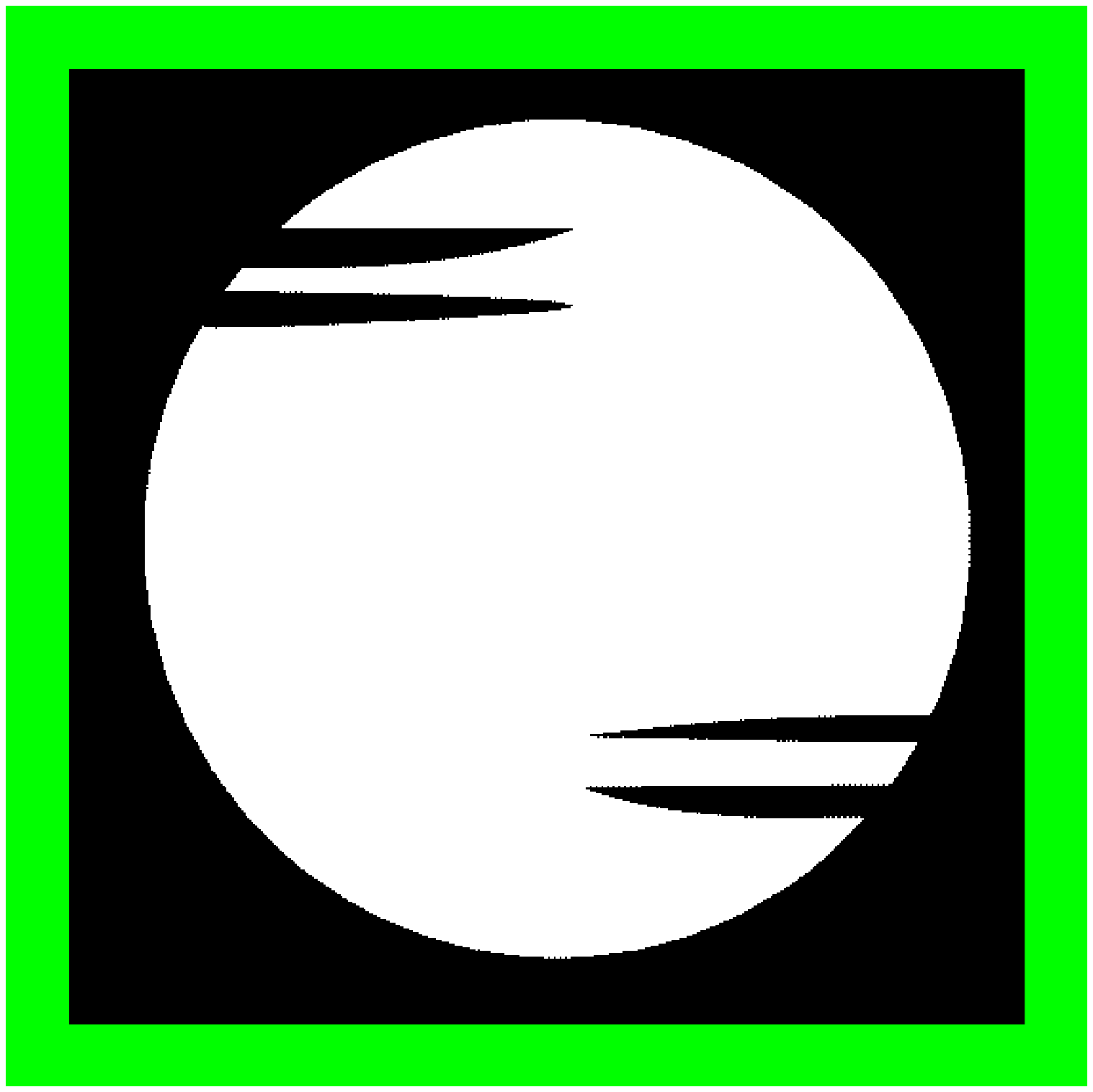}\hspace{-0.215em}
\includegraphics[trim=0.1cm 0cm 0.12cm 0cm, clip=true, width=1.2cm, height=1.2cm]{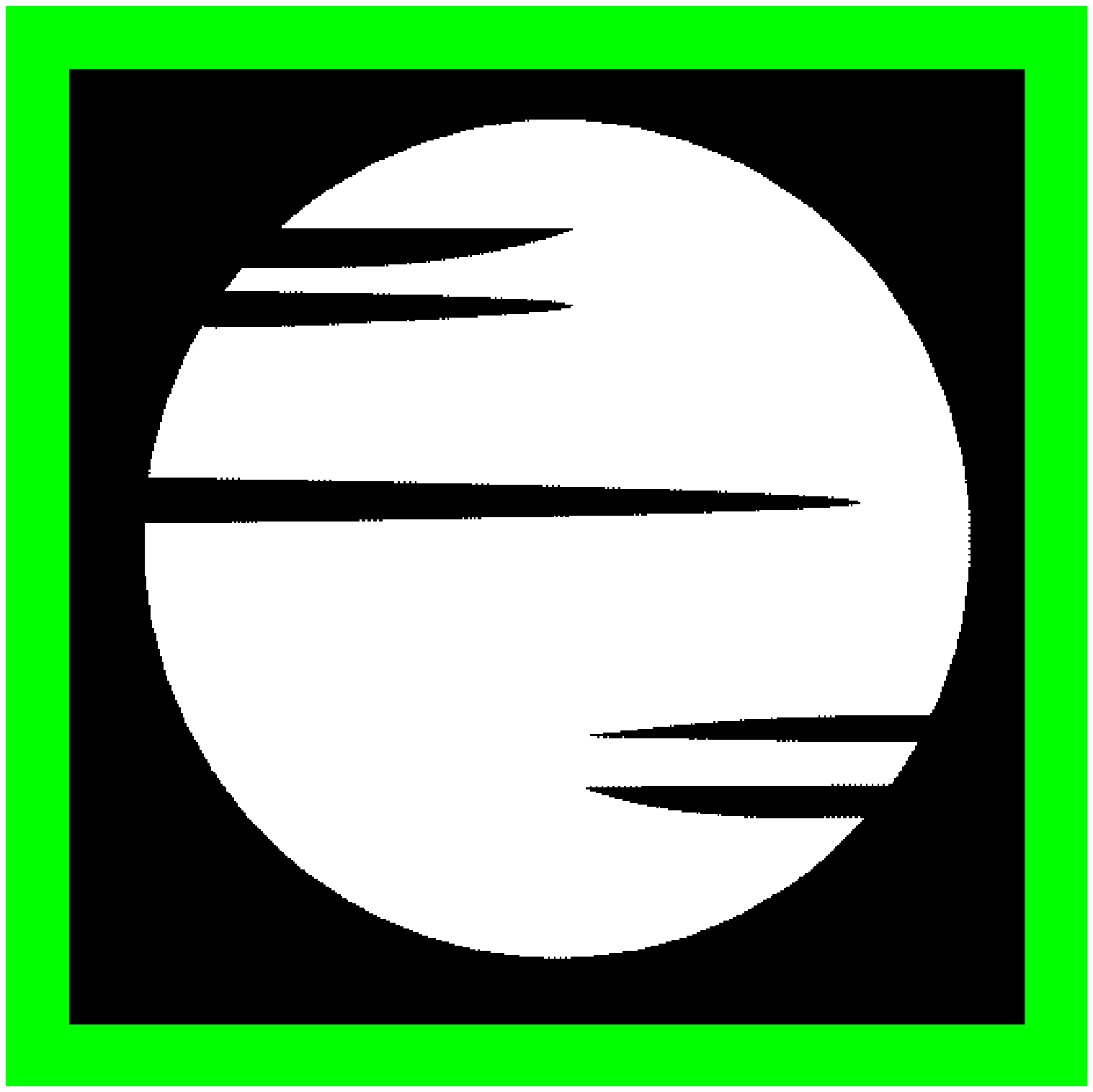}\hspace{-0.215em}
\includegraphics[trim=0.1cm 0cm 0.12cm 0cm, clip=true, width=1.2cm, height=1.2cm]{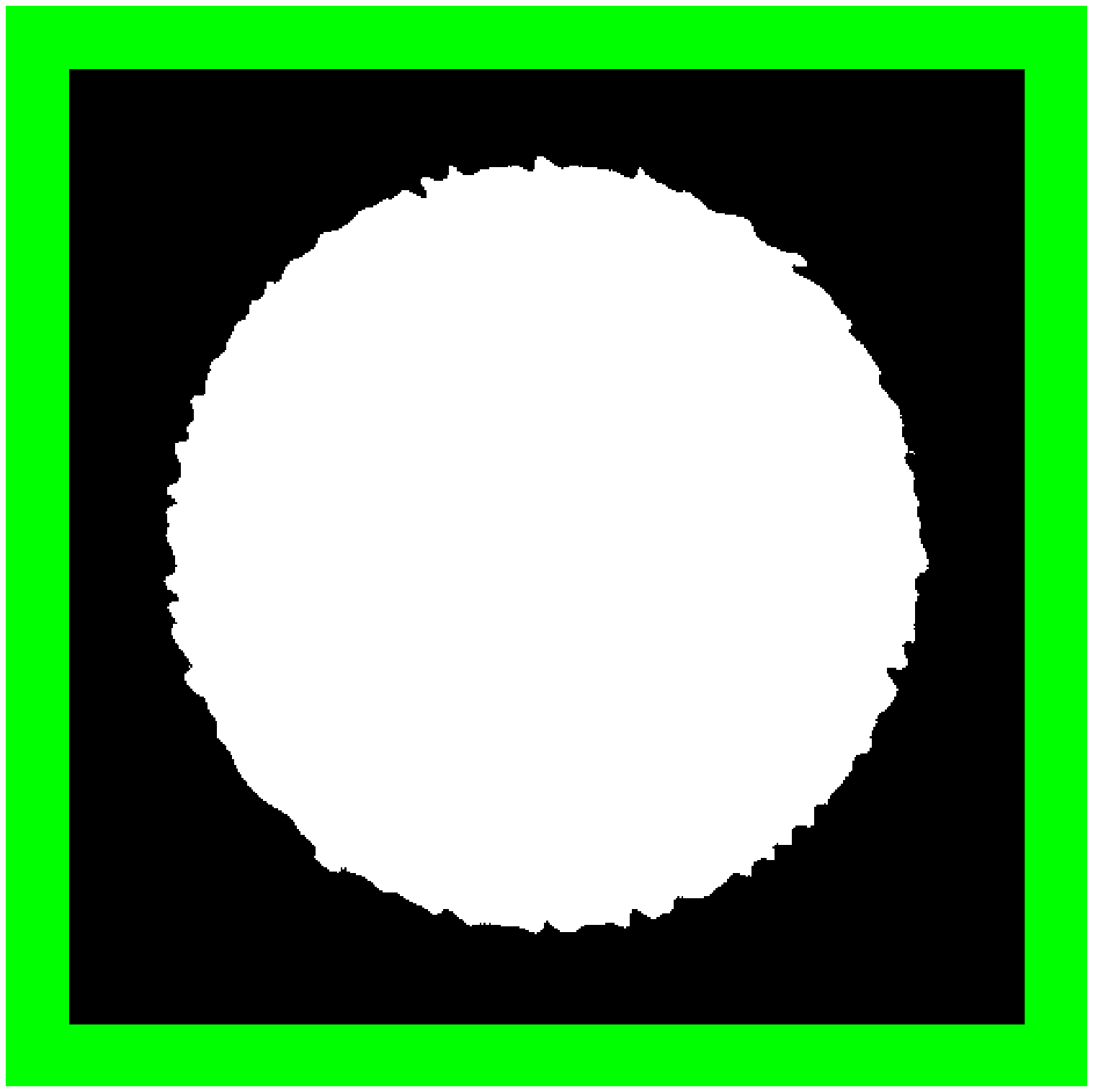}\hspace{-0.215em}
\includegraphics[trim=0.1cm 0cm 0.12cm 0cm, clip=true, width=1.2cm, height=1.2cm]{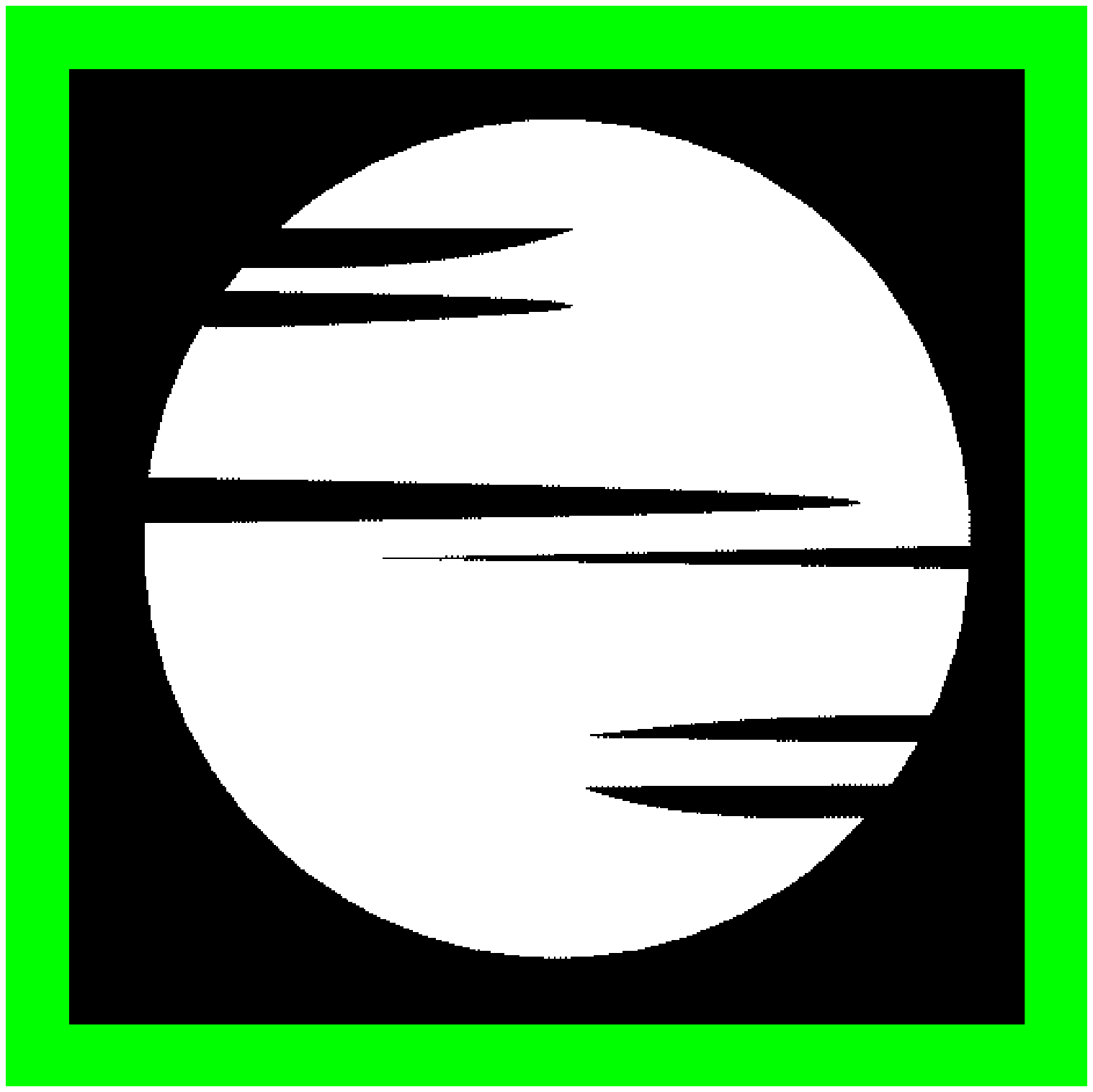}\hspace{-0.215em}
\includegraphics[trim=0.1cm 0cm 0.12cm 0cm, clip=true, width=1.2cm, height=1.2cm]{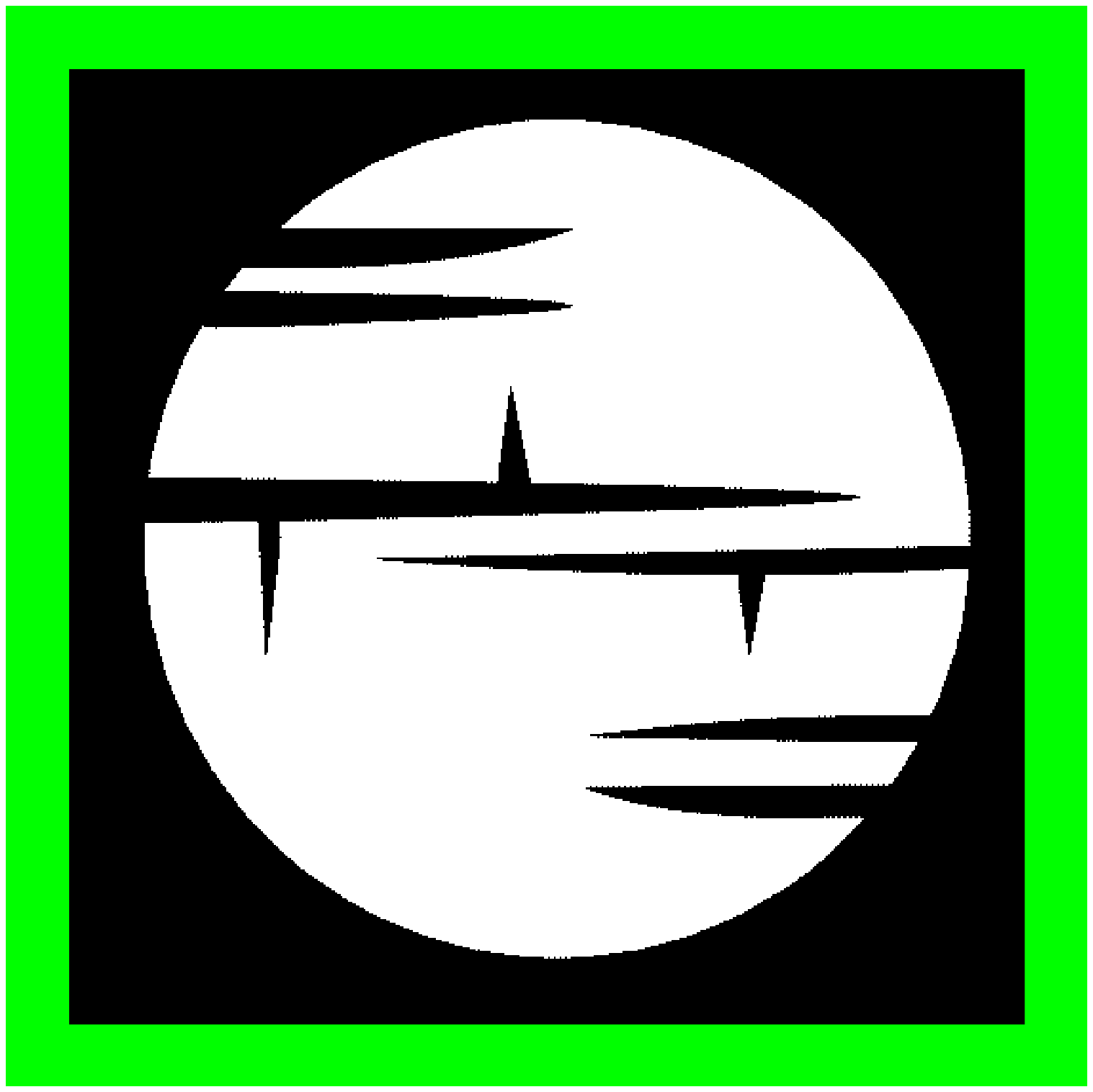}\hspace{-0.215em}
\includegraphics[trim=0.1cm 0cm 0.12cm 0cm, clip=true, width=1.2cm, height=1.2cm]{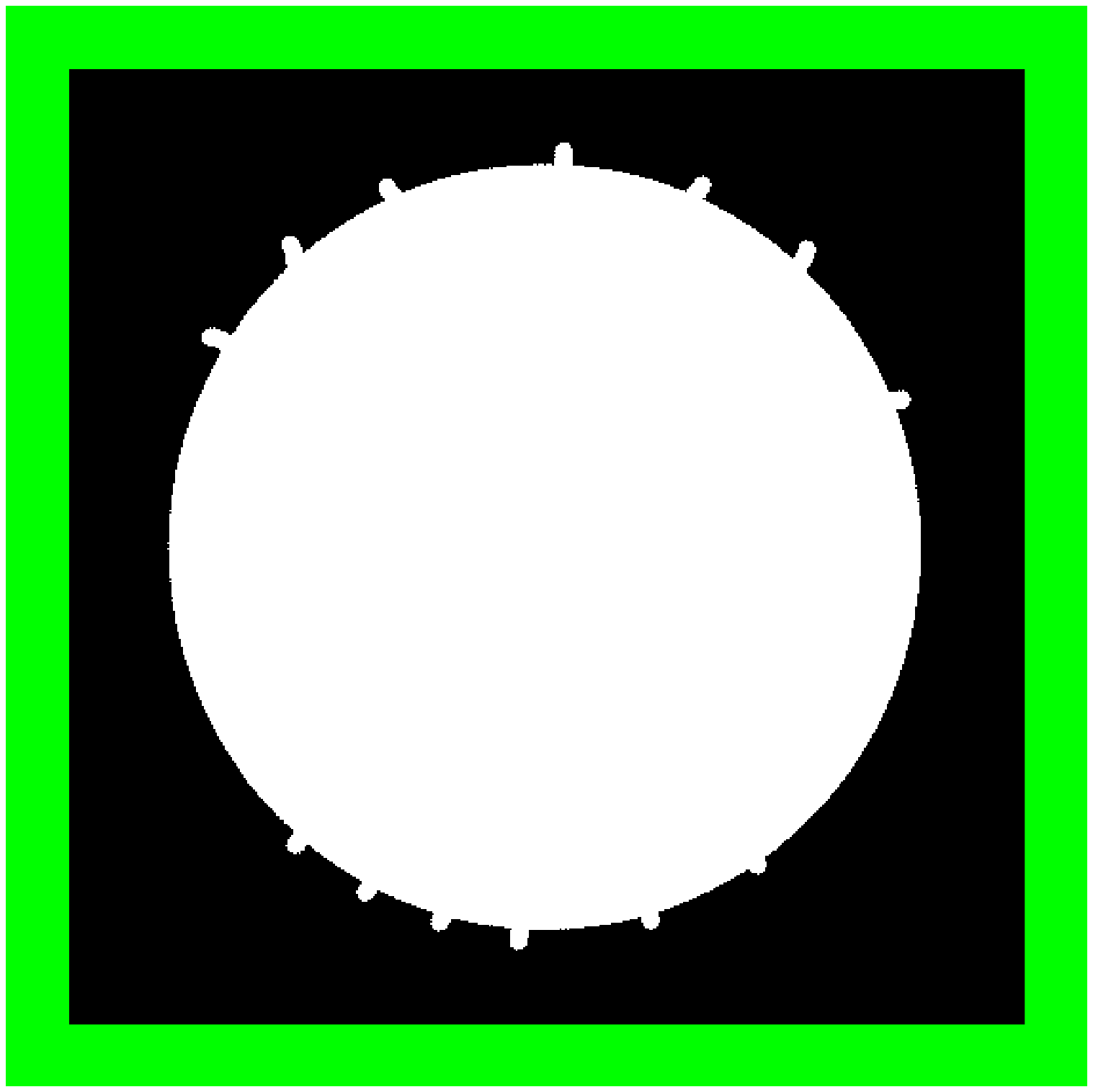}\hspace{-0.215em}\\
\includegraphics[trim=0.1cm 0cm 0.12cm 0cm, clip=true, width=1.2cm, height=1.2cm]{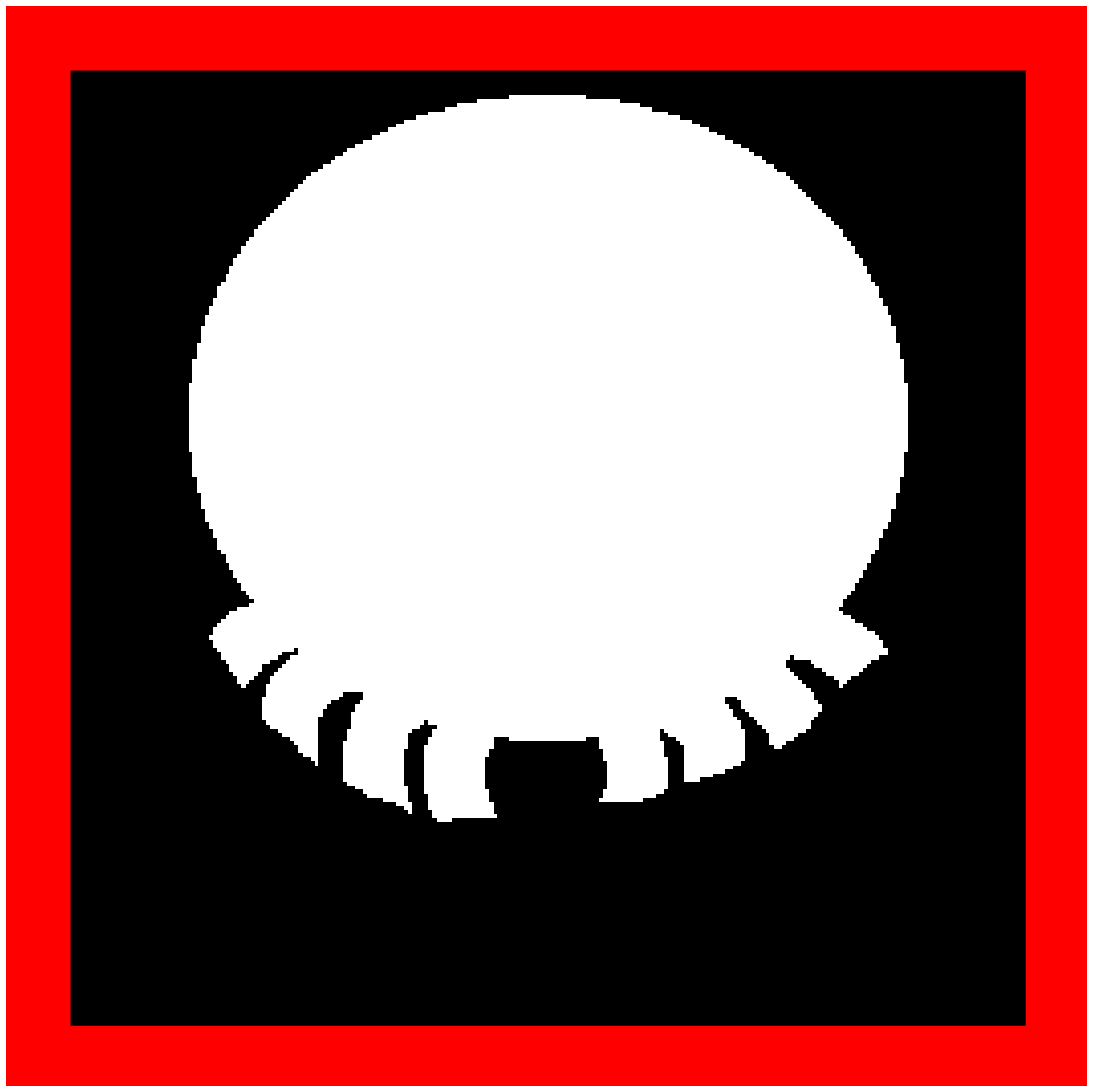}\hspace{-0.215em}
\includegraphics[trim=0.1cm 0cm 0.12cm 0cm, clip=true, width=1.2cm, height=1.2cm]{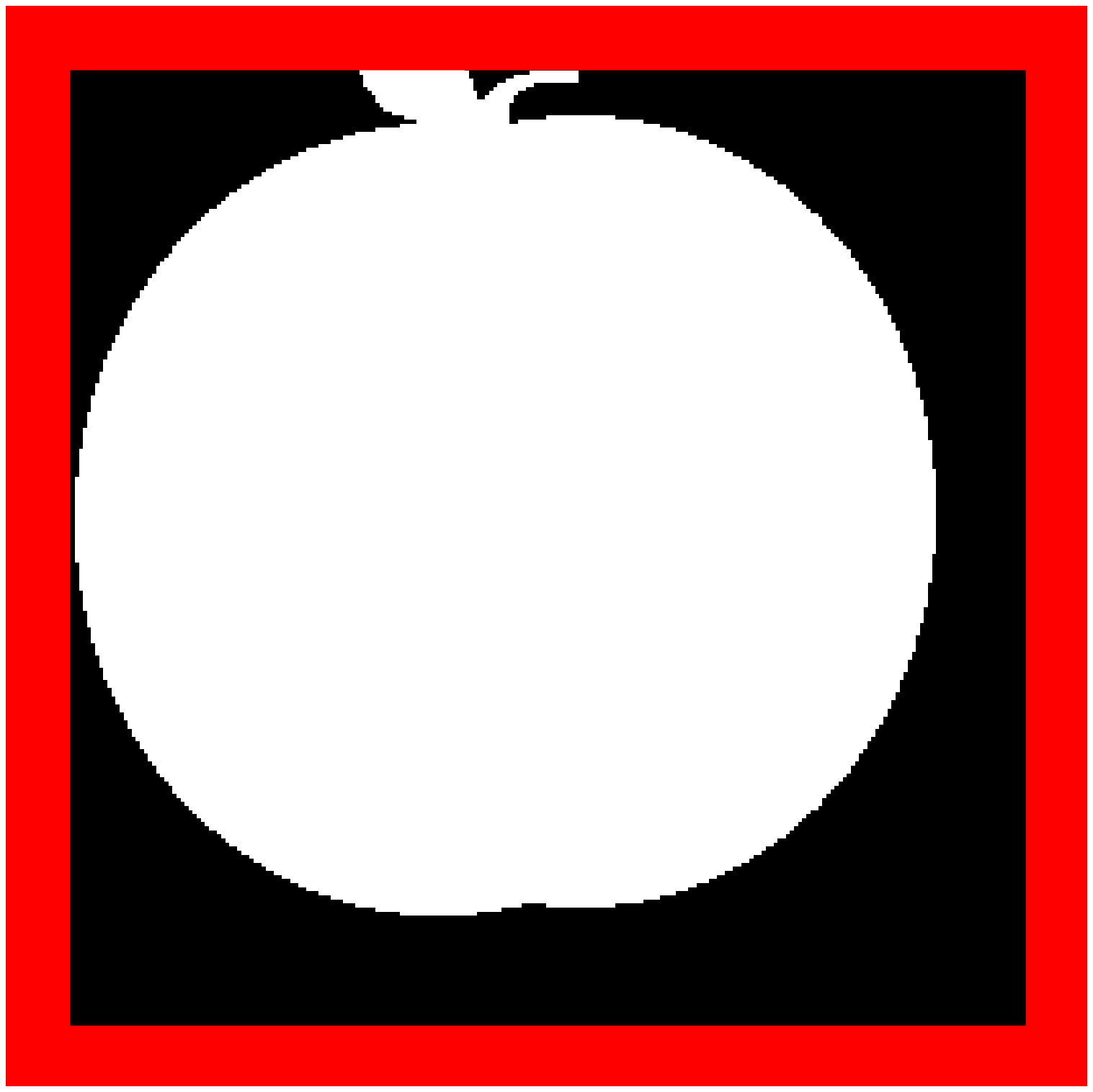}\hspace{-0.215em}
\includegraphics[trim=0.1cm 0cm 0.12cm 0cm, clip=true, width=1.2cm, height=1.2cm]{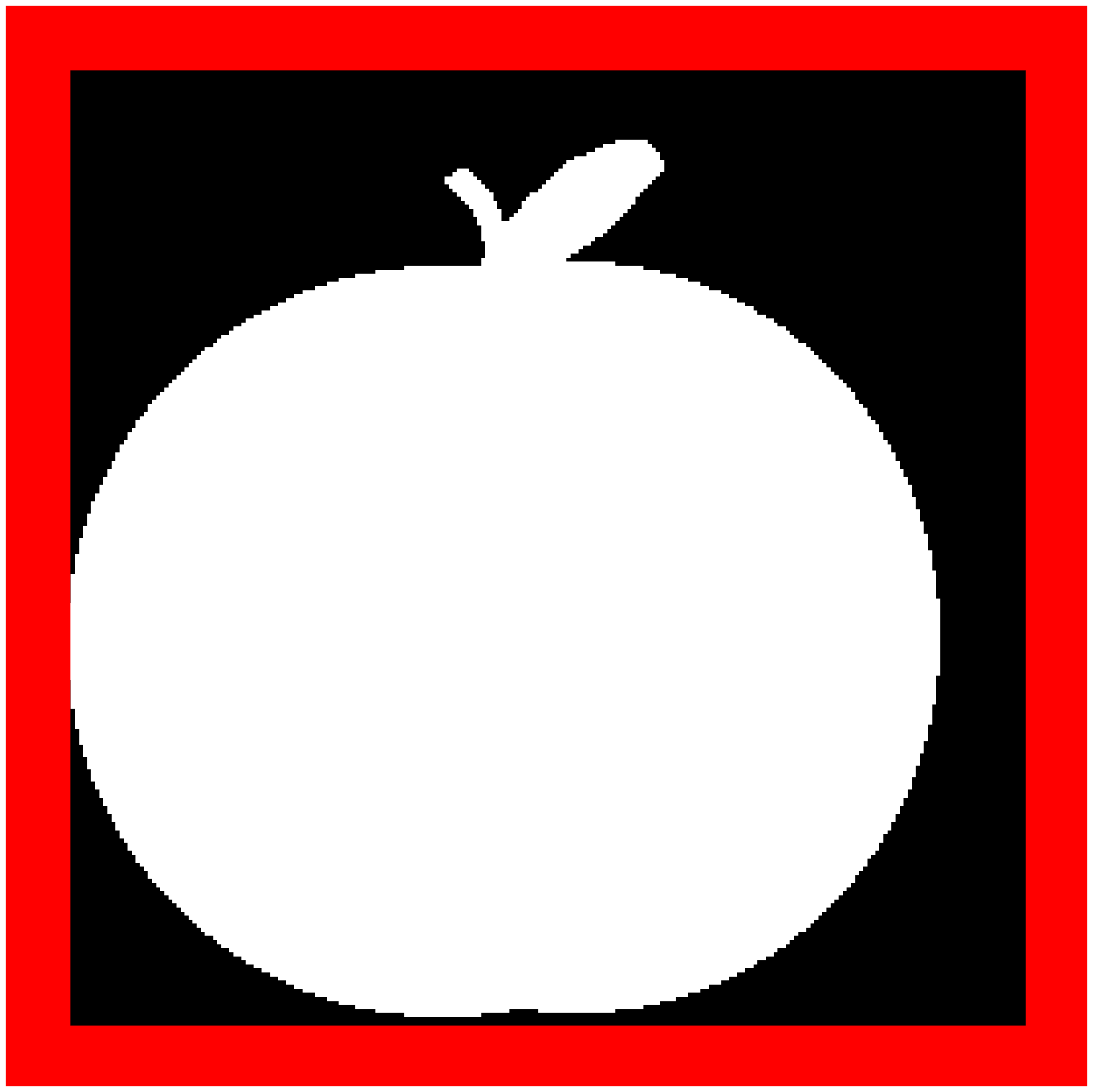}\hspace{-0.215em}
\includegraphics[trim=0.1cm 0cm 0.12cm 0cm, clip=true, width=1.2cm, height=1.2cm]{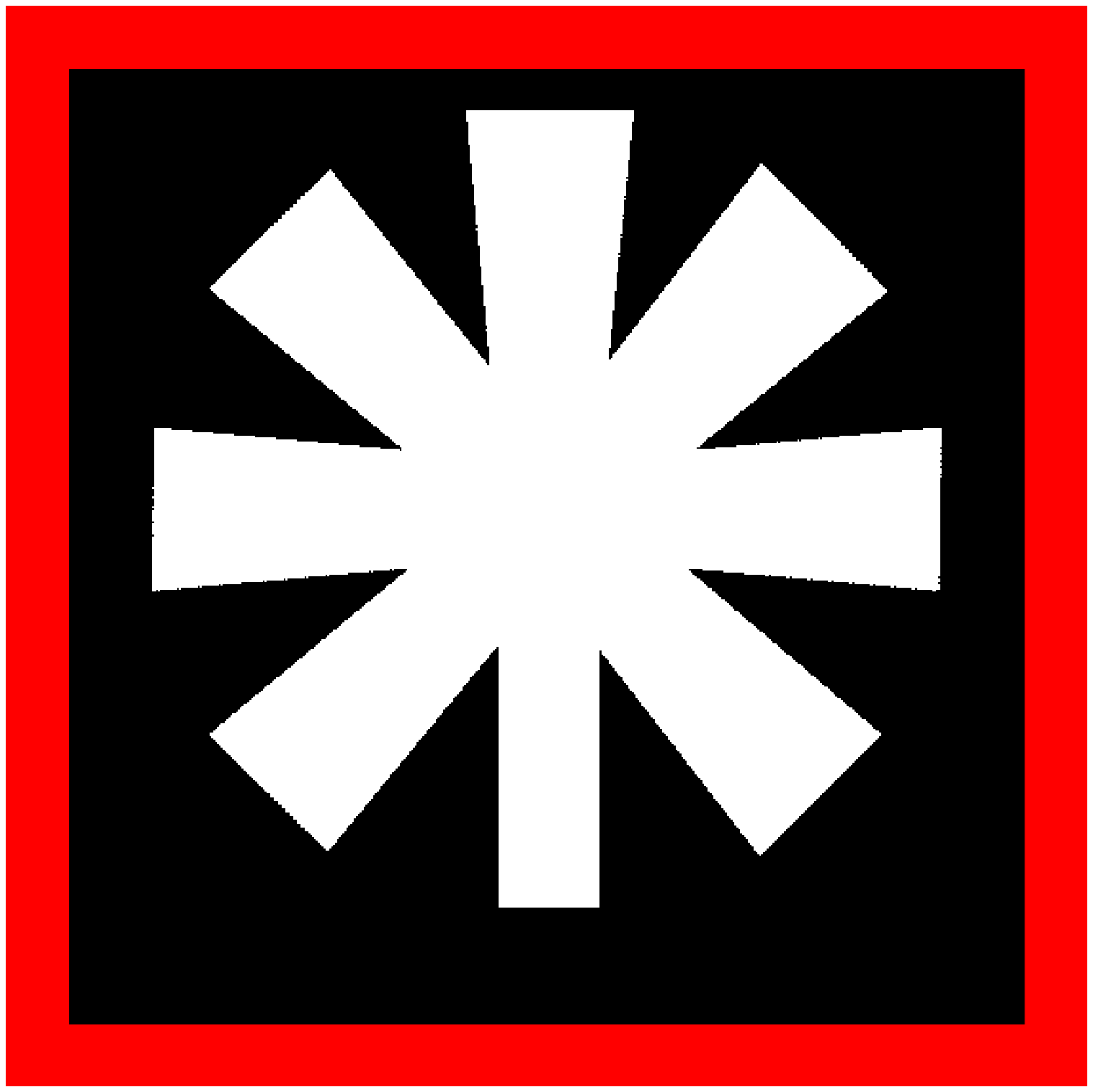}\hspace{-0.215em}
\includegraphics[trim=0.1cm 0cm 0.12cm 0cm, clip=true, width=1.2cm, height=1.2cm]{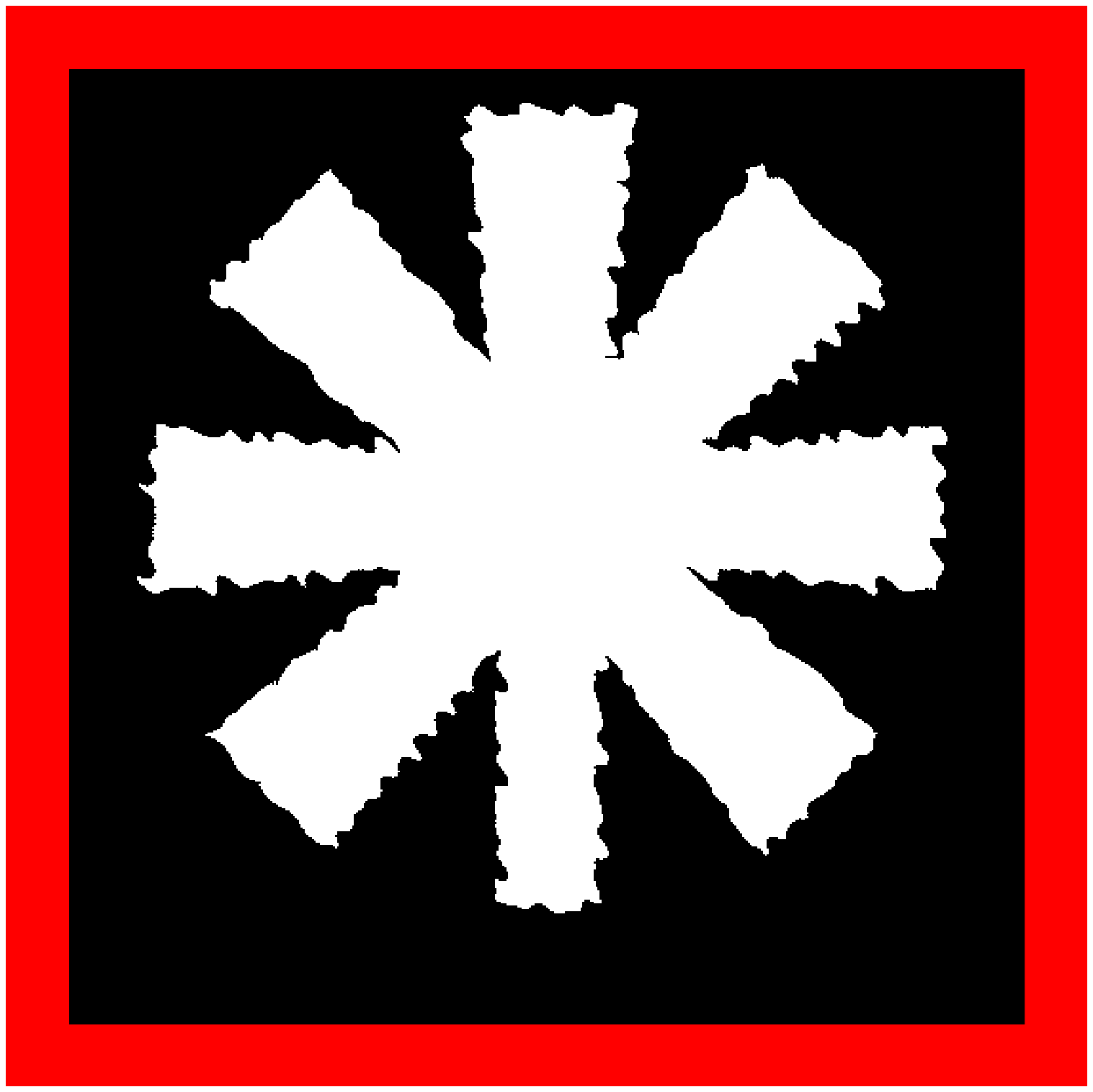}\hspace{-0.215em}
\includegraphics[trim=0.1cm 0cm 0.12cm 0cm, clip=true, width=1.2cm, height=1.2cm]{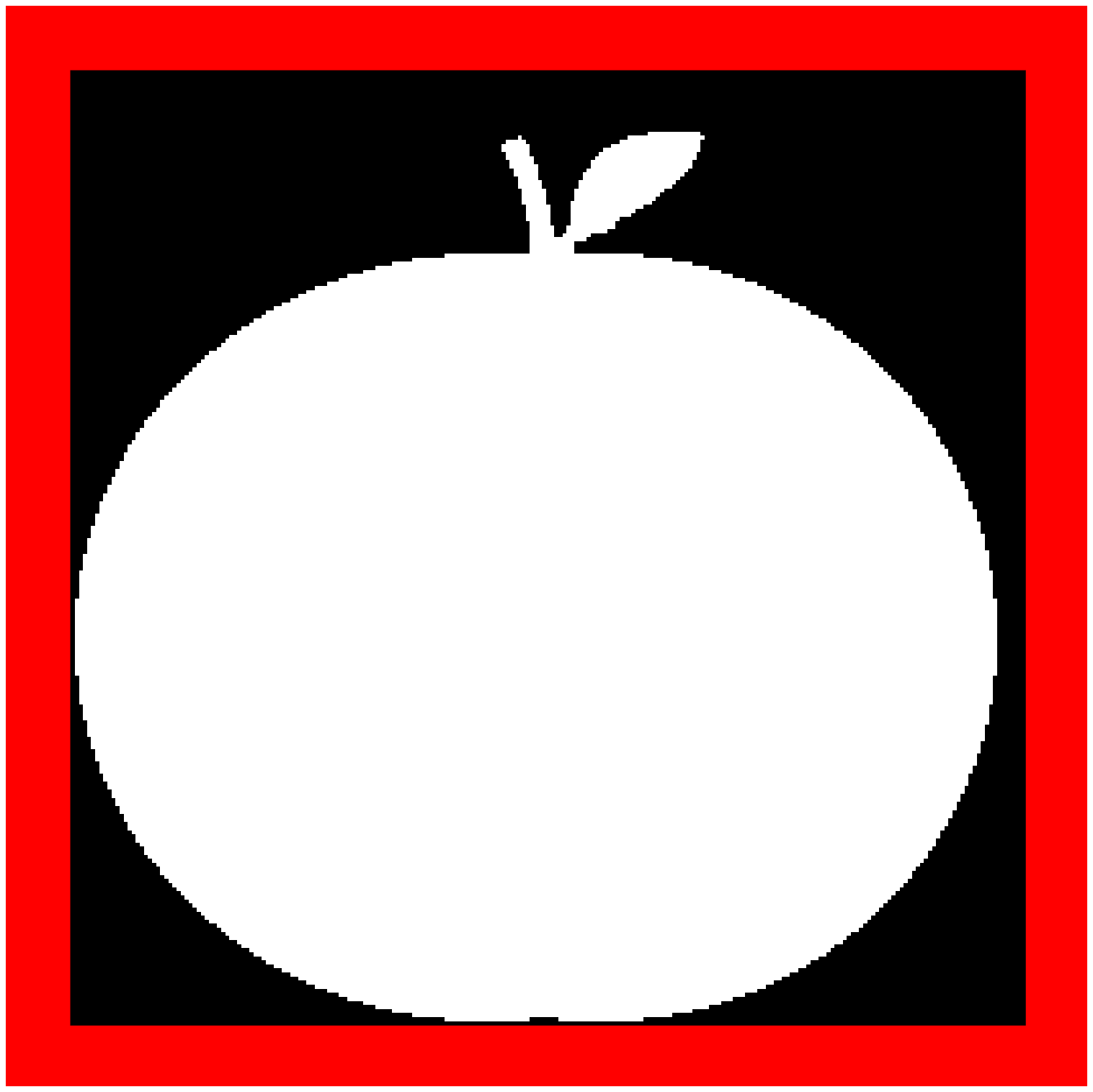}\hspace{-0.215em}
\includegraphics[trim=0.1cm 0cm 0.12cm 0cm, clip=true, width=1.2cm, height=1.2cm]{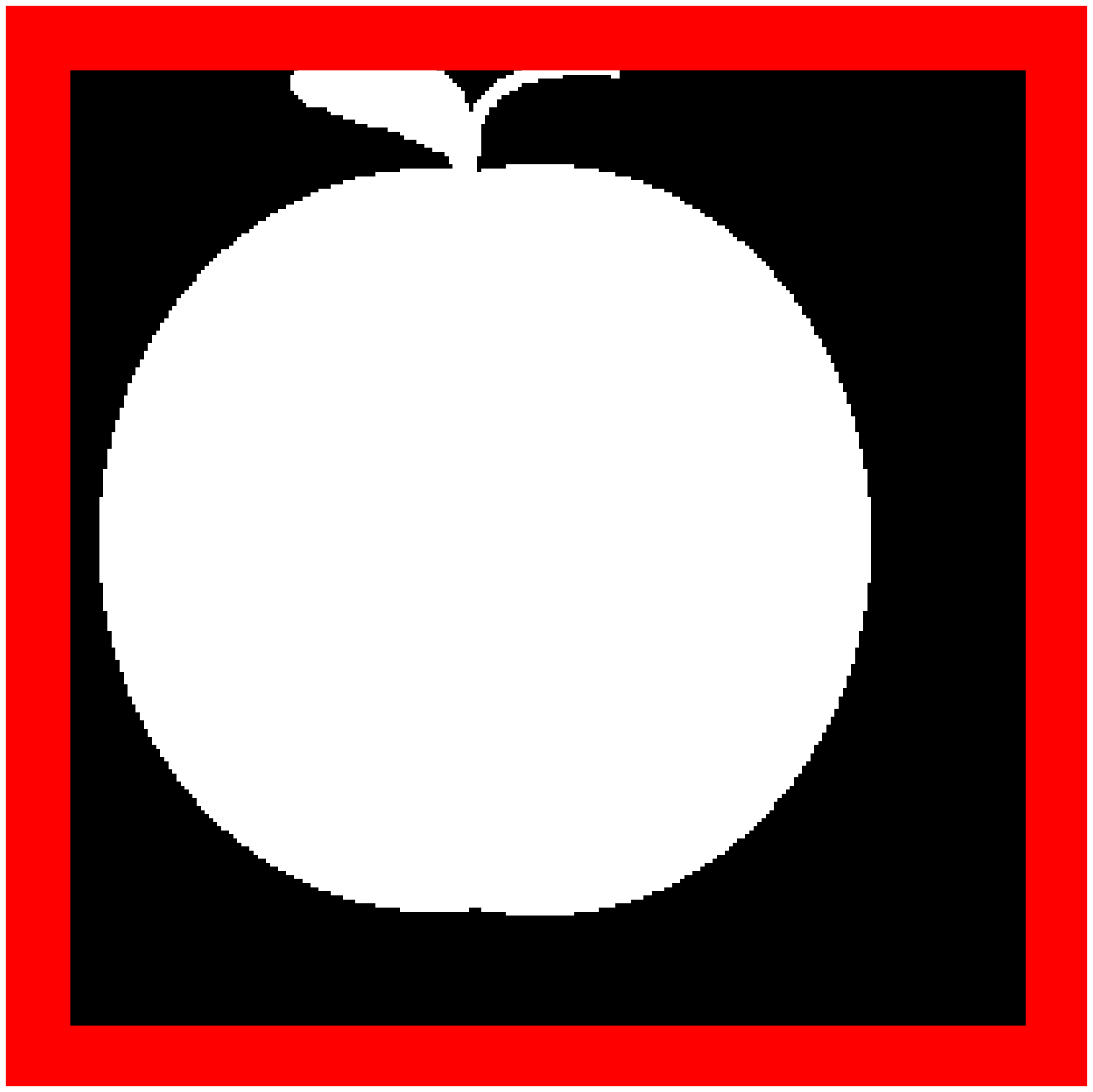}\hspace{-0.215em}
\includegraphics[trim=0.1cm 0cm 0.12cm 0cm, clip=true, width=1.2cm, height=1.2cm]{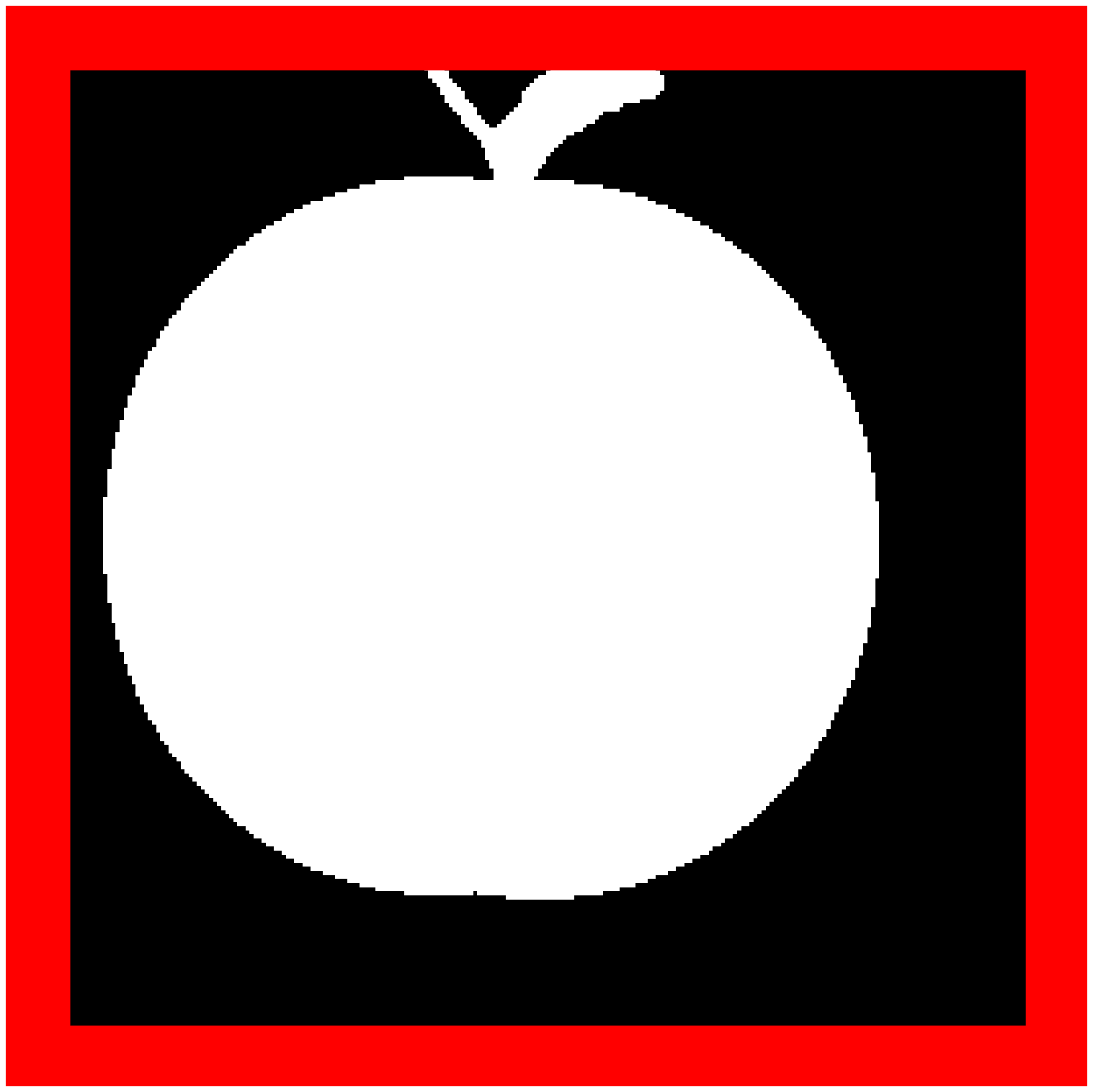}\hspace{-0.215em}
\includegraphics[trim=0.1cm 0cm 0.12cm 0cm, clip=true, width=1.2cm, height=1.2cm]{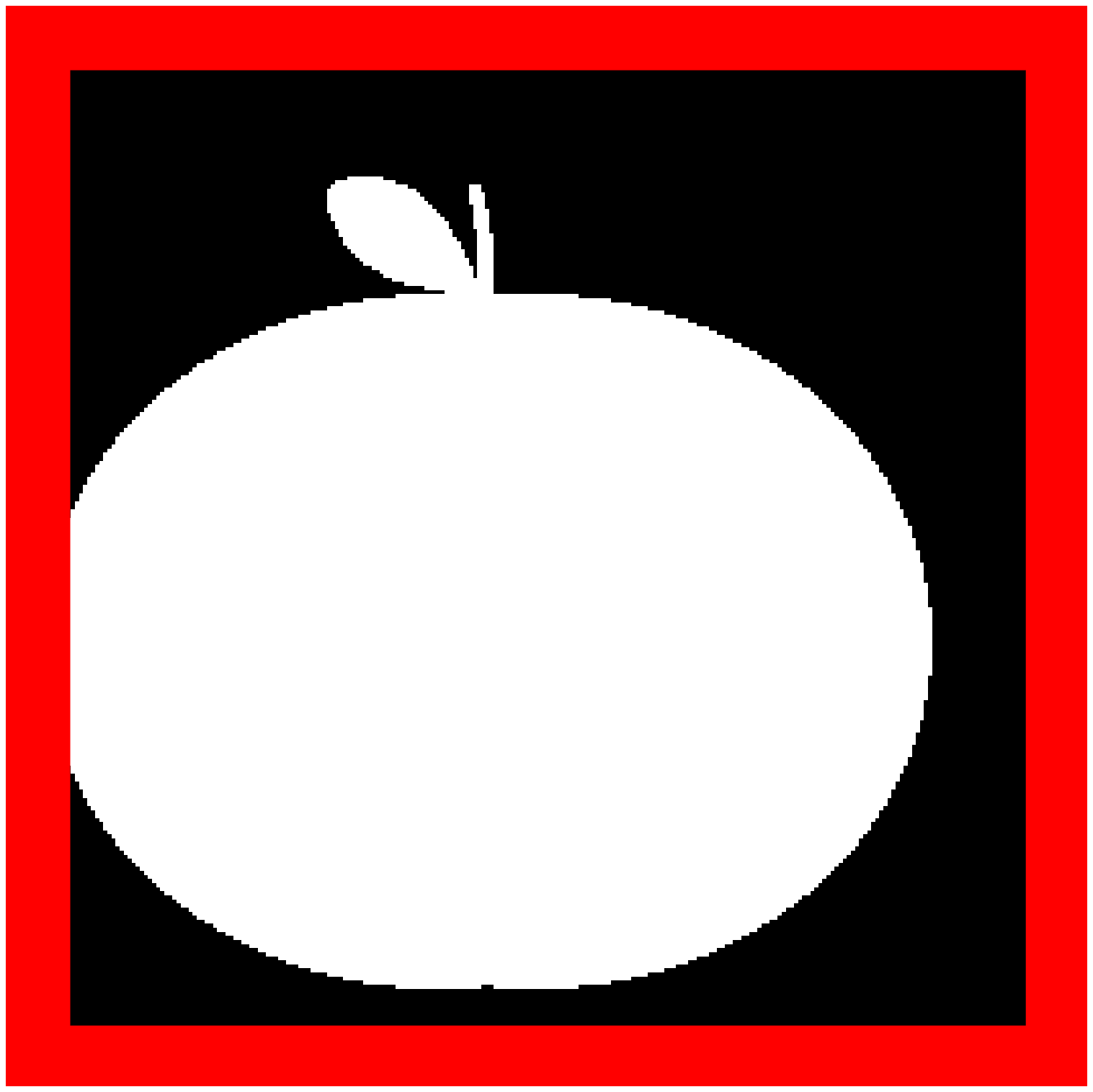}\hspace{-0.215em}
\includegraphics[trim=0.1cm 0cm 0.12cm 0cm, clip=true, width=1.2cm, height=1.2cm]{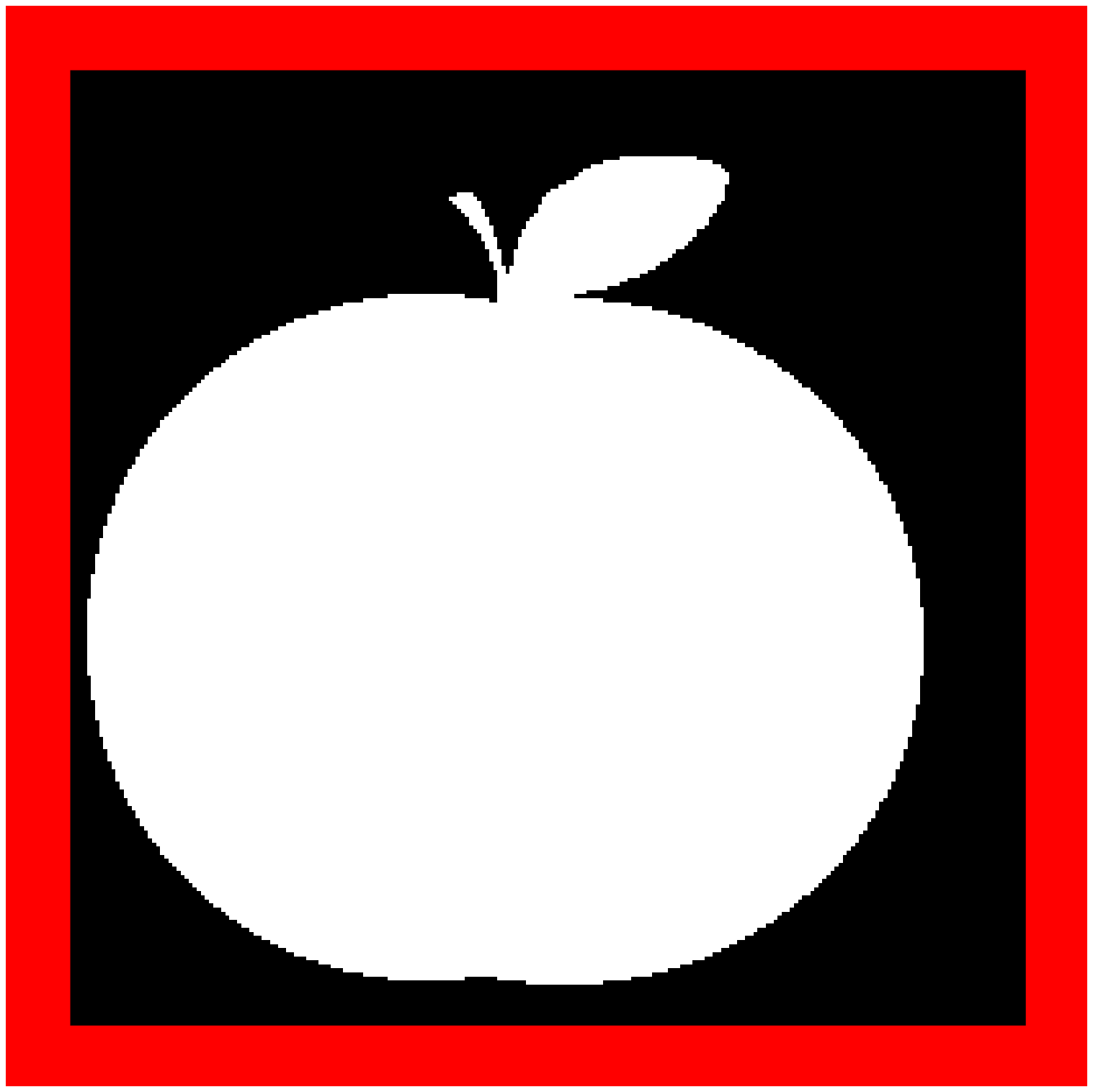}\hspace{-0.215em}\\
\includegraphics[trim=0.1cm 0cm 0.12cm 0cm, clip=true, width=1.2cm, height=1.2cm]{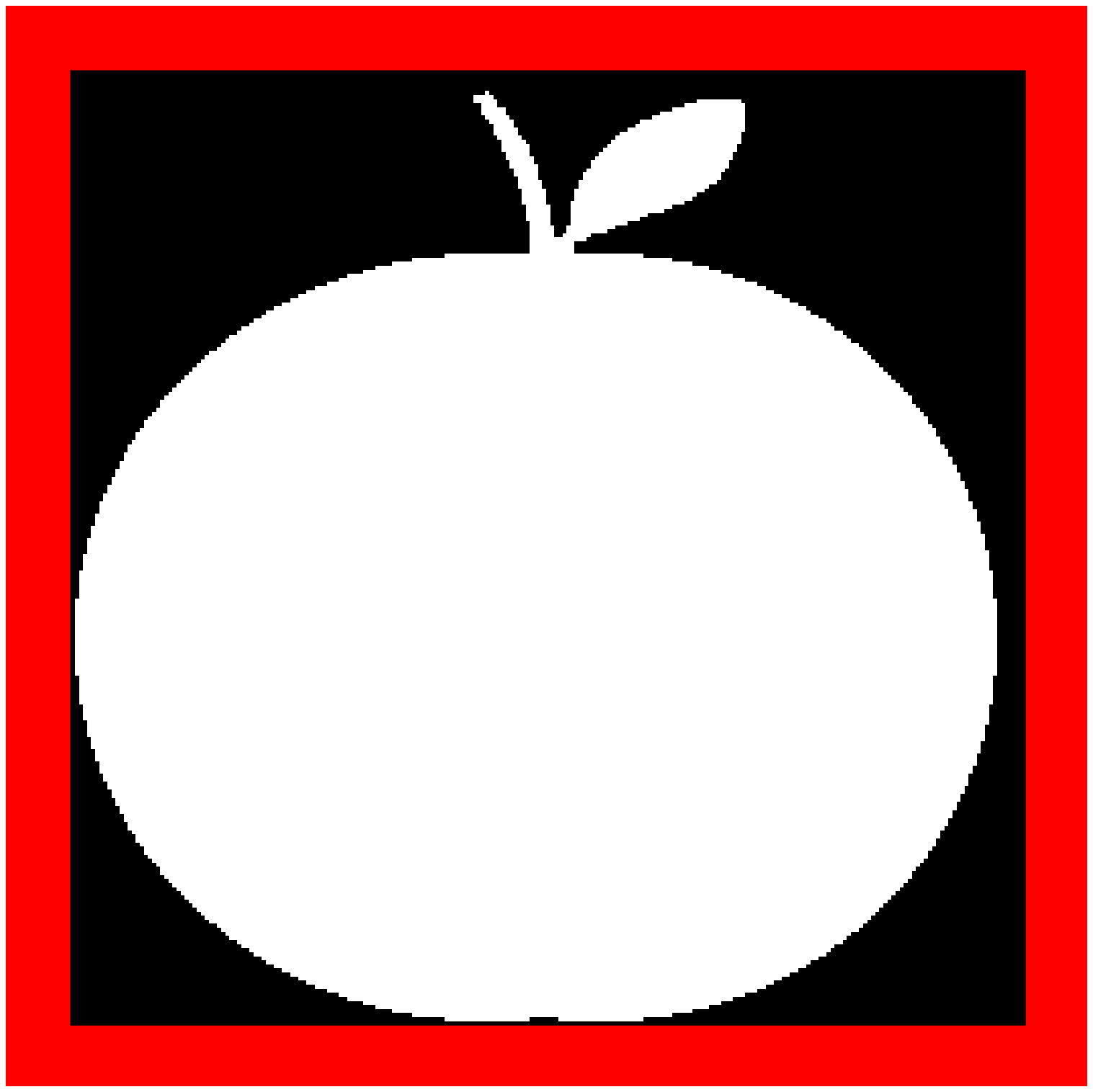}\hspace{-0.215em}
\includegraphics[trim=0.1cm 0cm 0.12cm 0cm, clip=true, width=1.2cm, height=1.2cm]{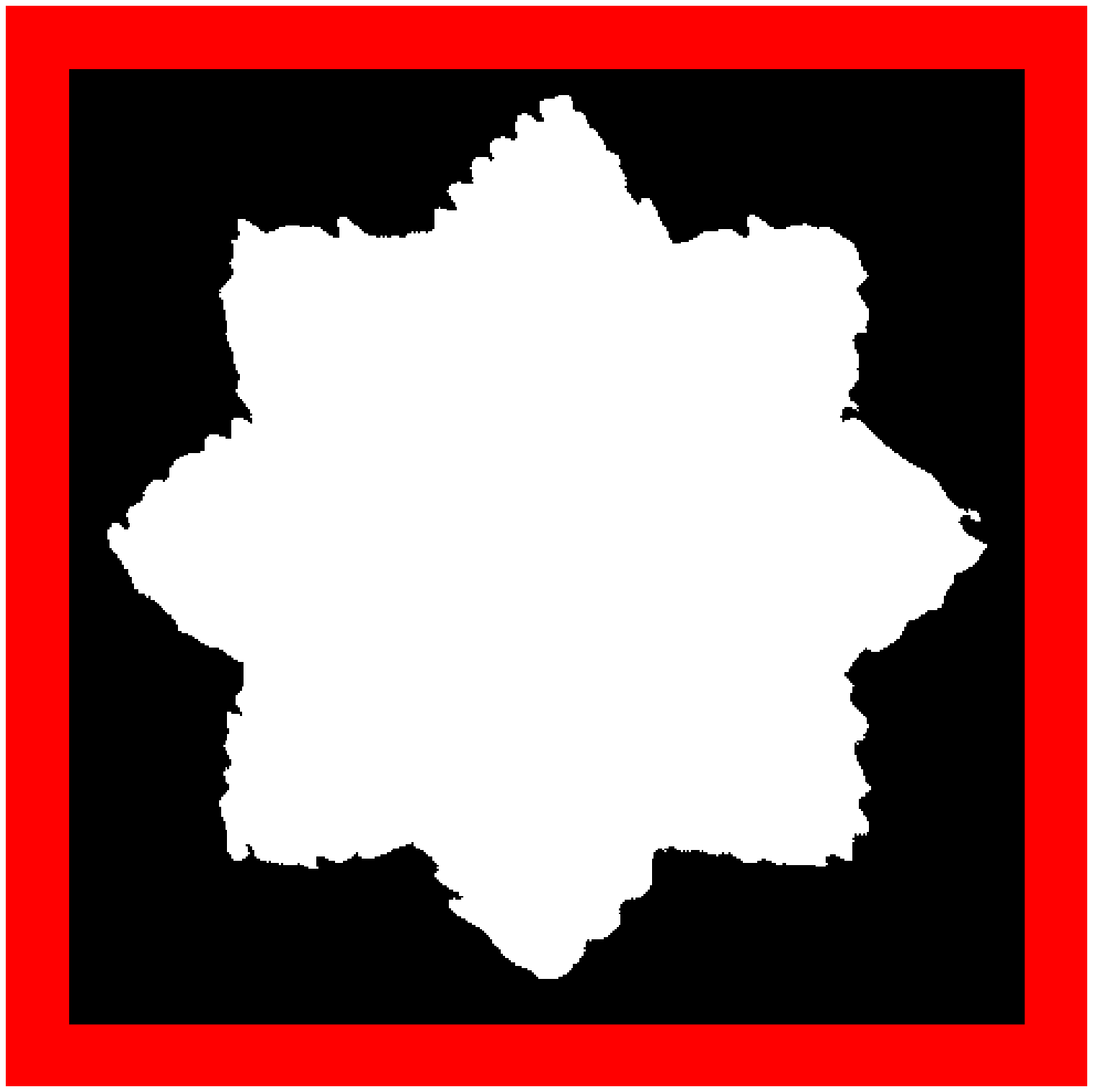}\hspace{-0.215em}
\includegraphics[trim=0.1cm 0cm 0.12cm 0cm, clip=true, width=1.2cm, height=1.2cm]{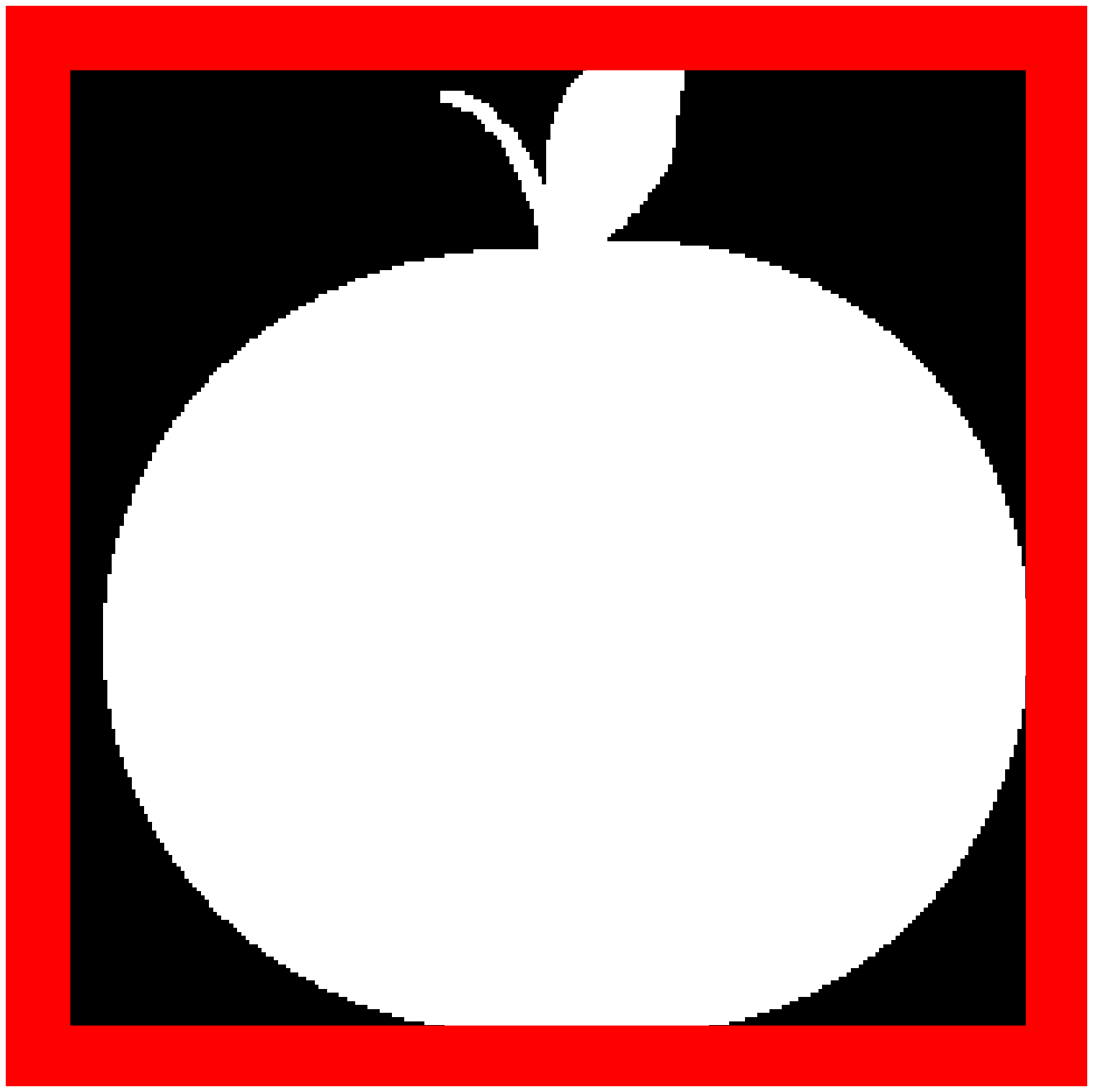}\hspace{-0.215em}
\includegraphics[trim=0.1cm 0cm 0.12cm 0cm, clip=true, width=1.2cm, height=1.2cm]{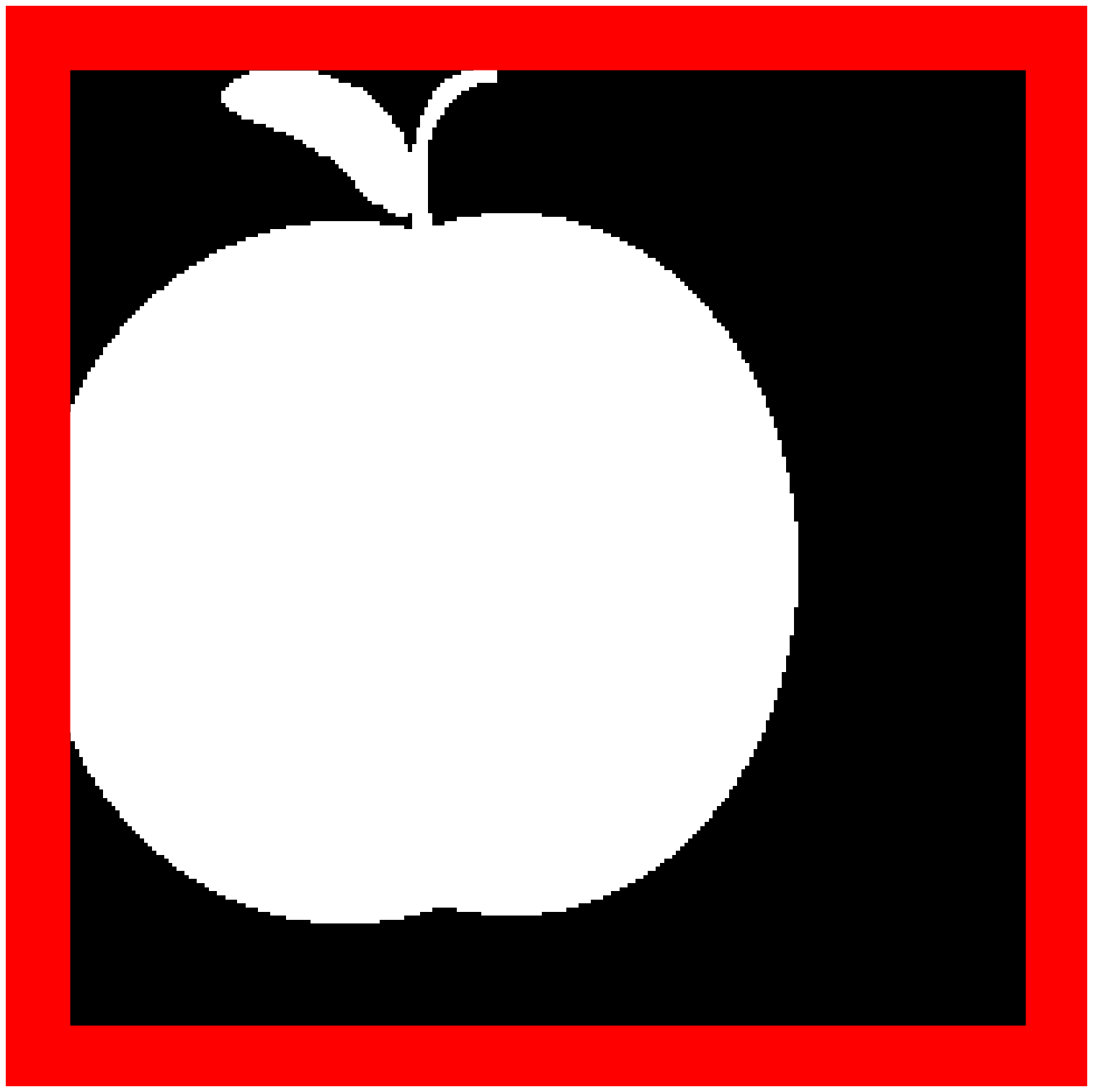}\hspace{-0.215em}
\includegraphics[trim=0.1cm 0cm 0.12cm 0cm, clip=true, width=1.2cm, height=1.2cm]{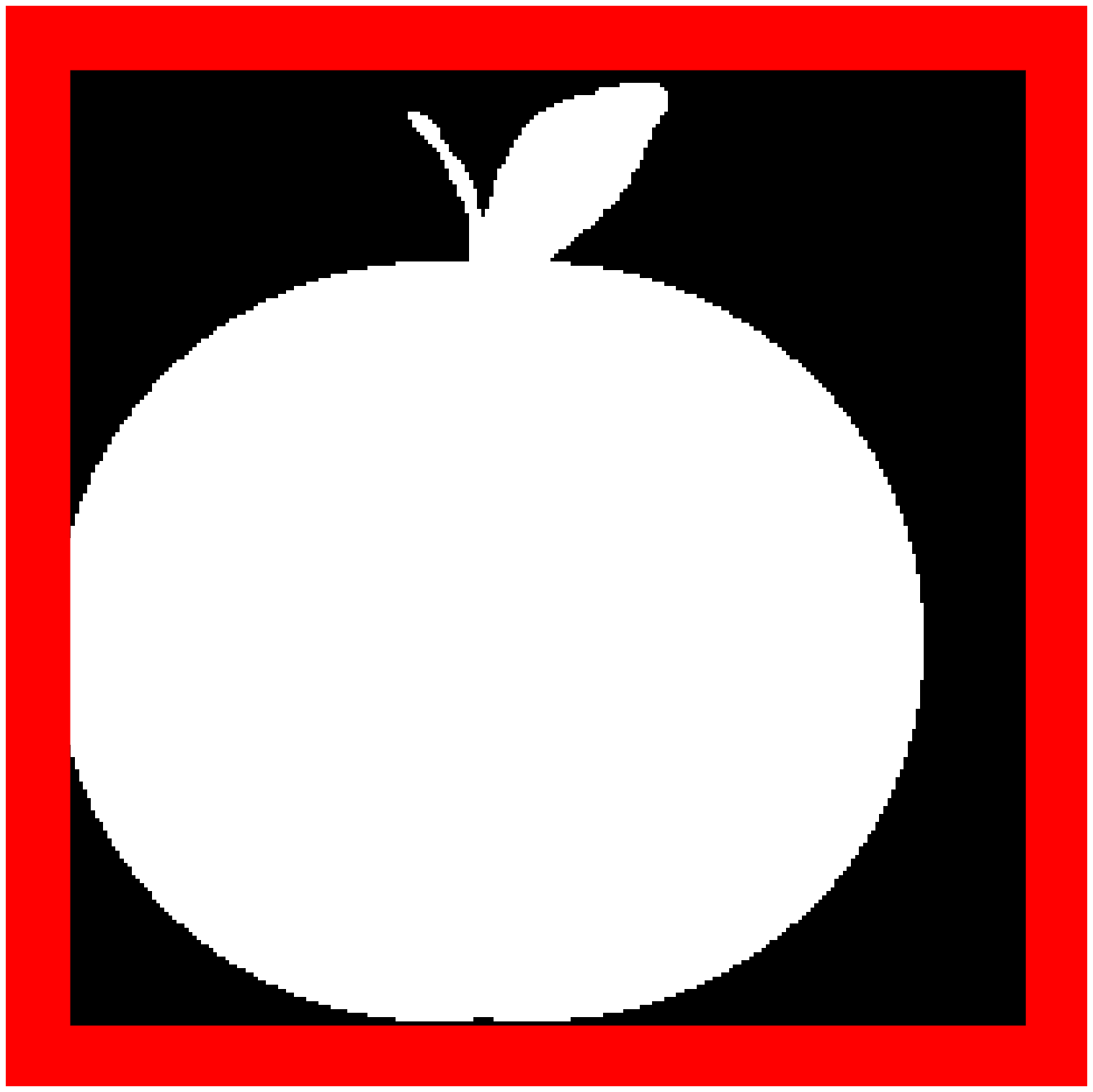}\hspace{-0.215em}
\includegraphics[trim=0.1cm 0cm 0.12cm 0cm, clip=true, width=1.2cm, height=1.2cm]{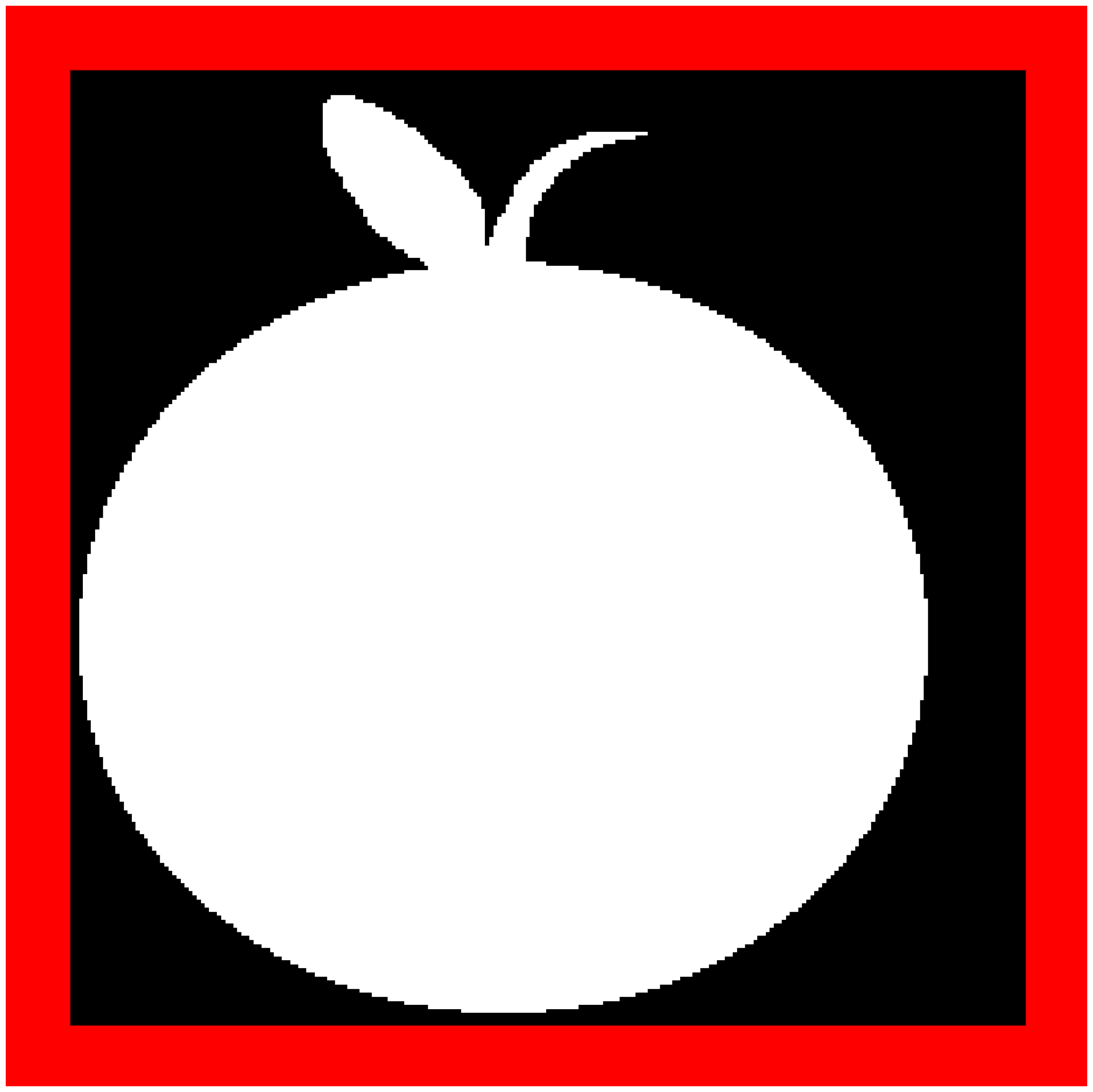}\hspace{-0.215em}
\includegraphics[trim=0.1cm 0cm 0.12cm 0cm, clip=true, width=1.2cm, height=1.2cm]{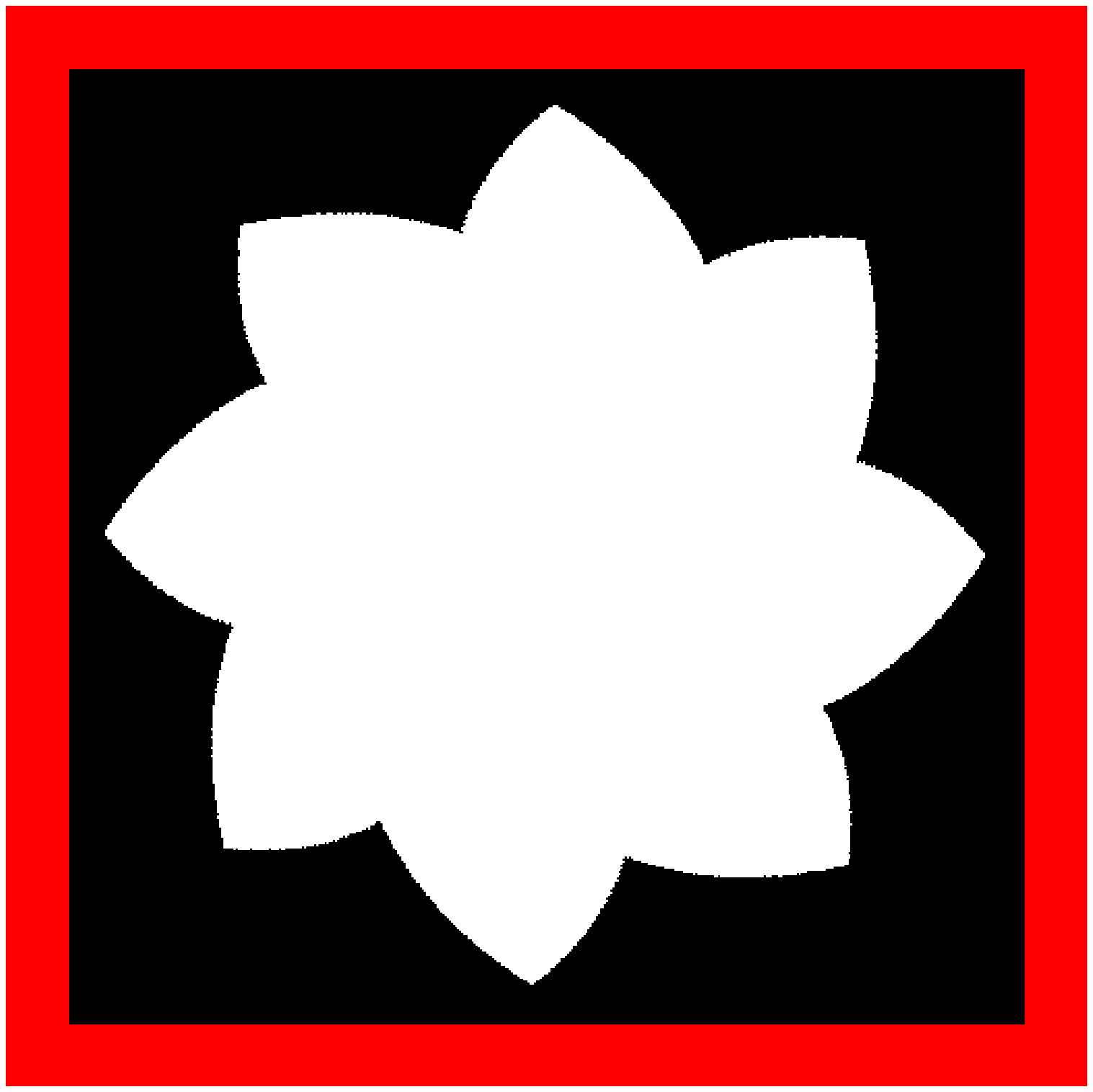}\hspace{-0.215em}
\includegraphics[trim=0.1cm 0cm 0.12cm 0cm, clip=true, width=1.2cm, height=1.2cm]{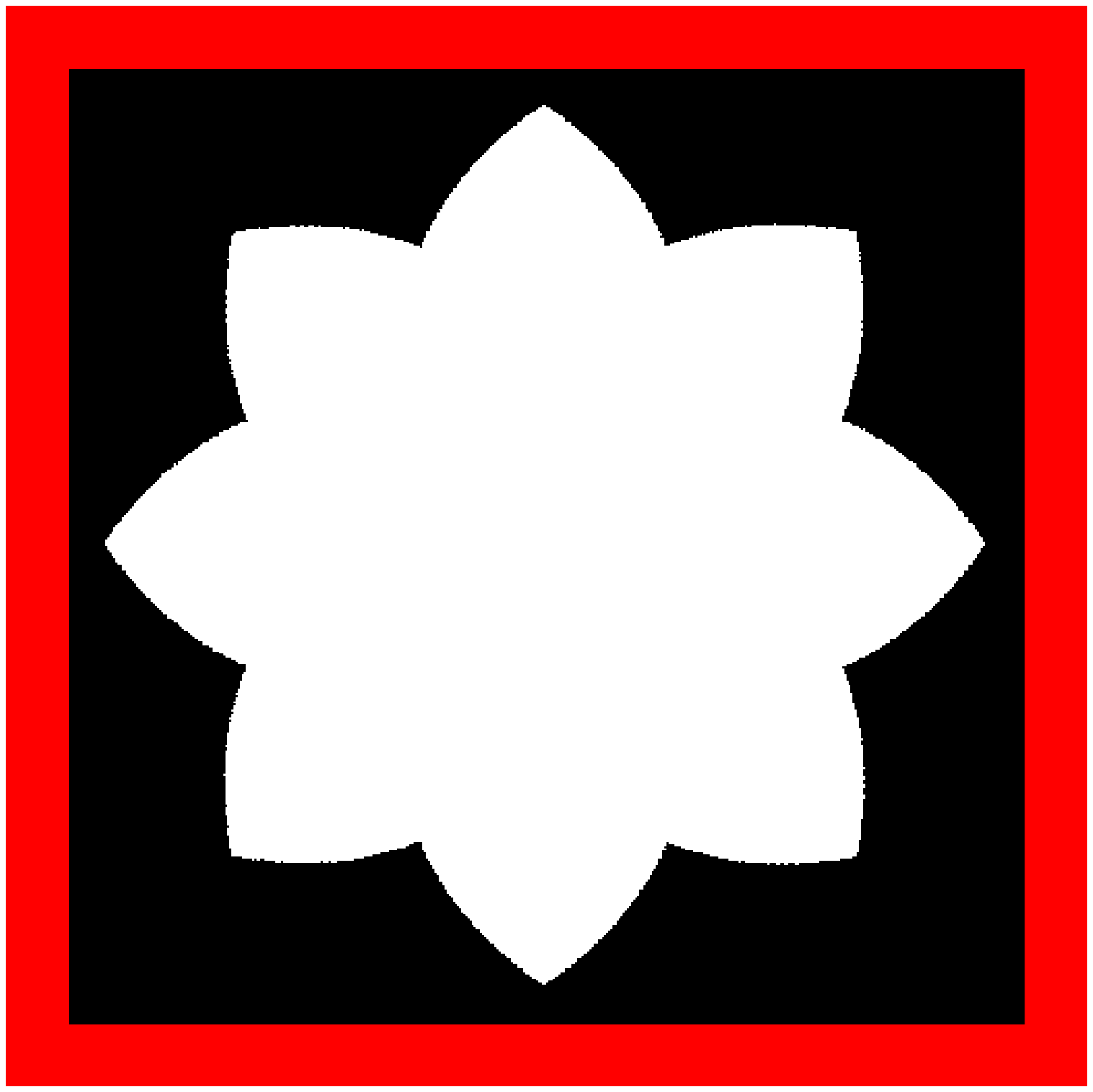}\hspace{-0.215em}
\includegraphics[trim=0.1cm 0cm 0.12cm 0cm, clip=true, width=1.2cm, height=1.2cm]{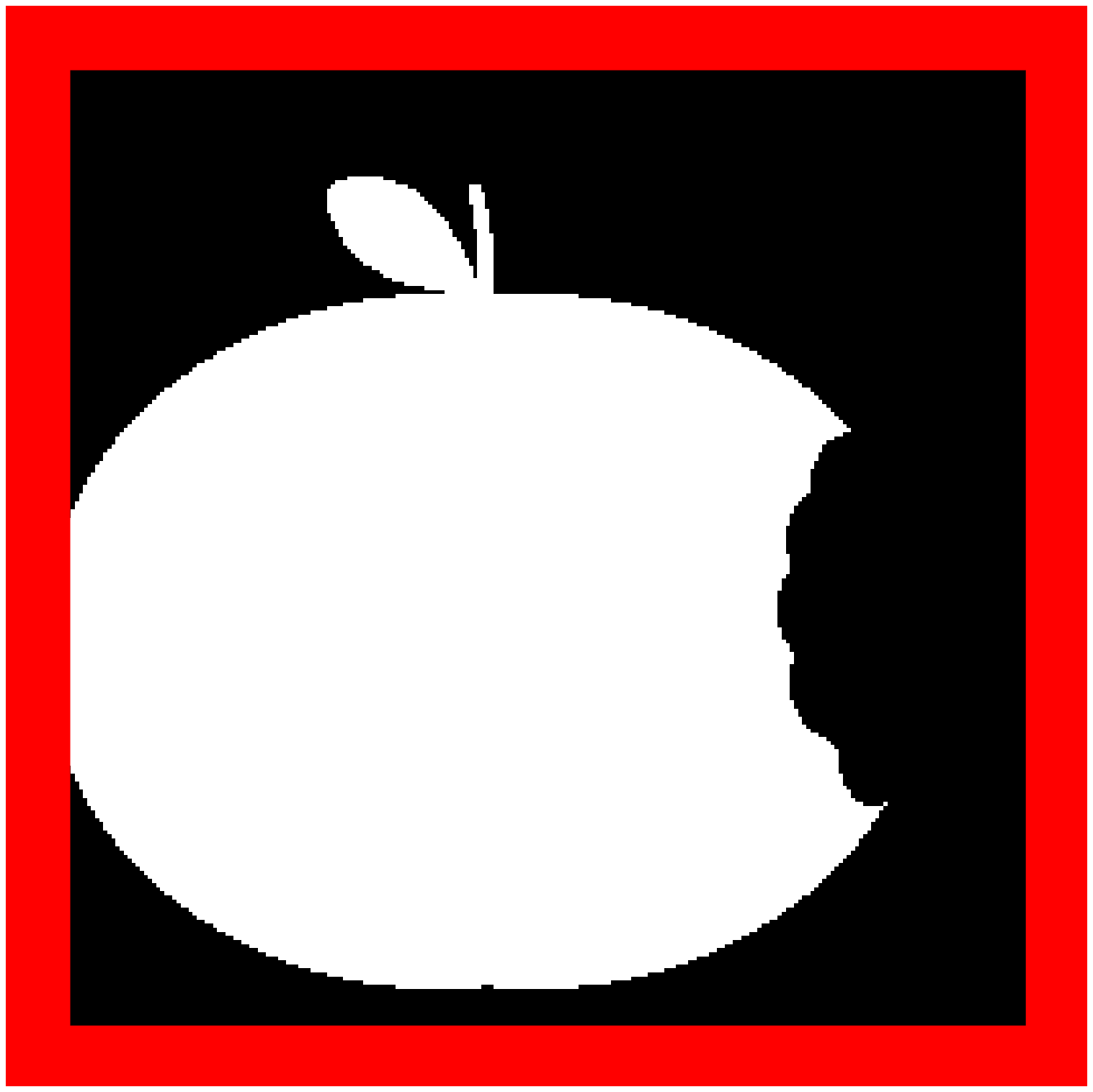}\hspace{-0.215em}
\includegraphics[trim=0.1cm 0cm 0.12cm 0cm, clip=true, width=1.2cm, height=1.2cm]{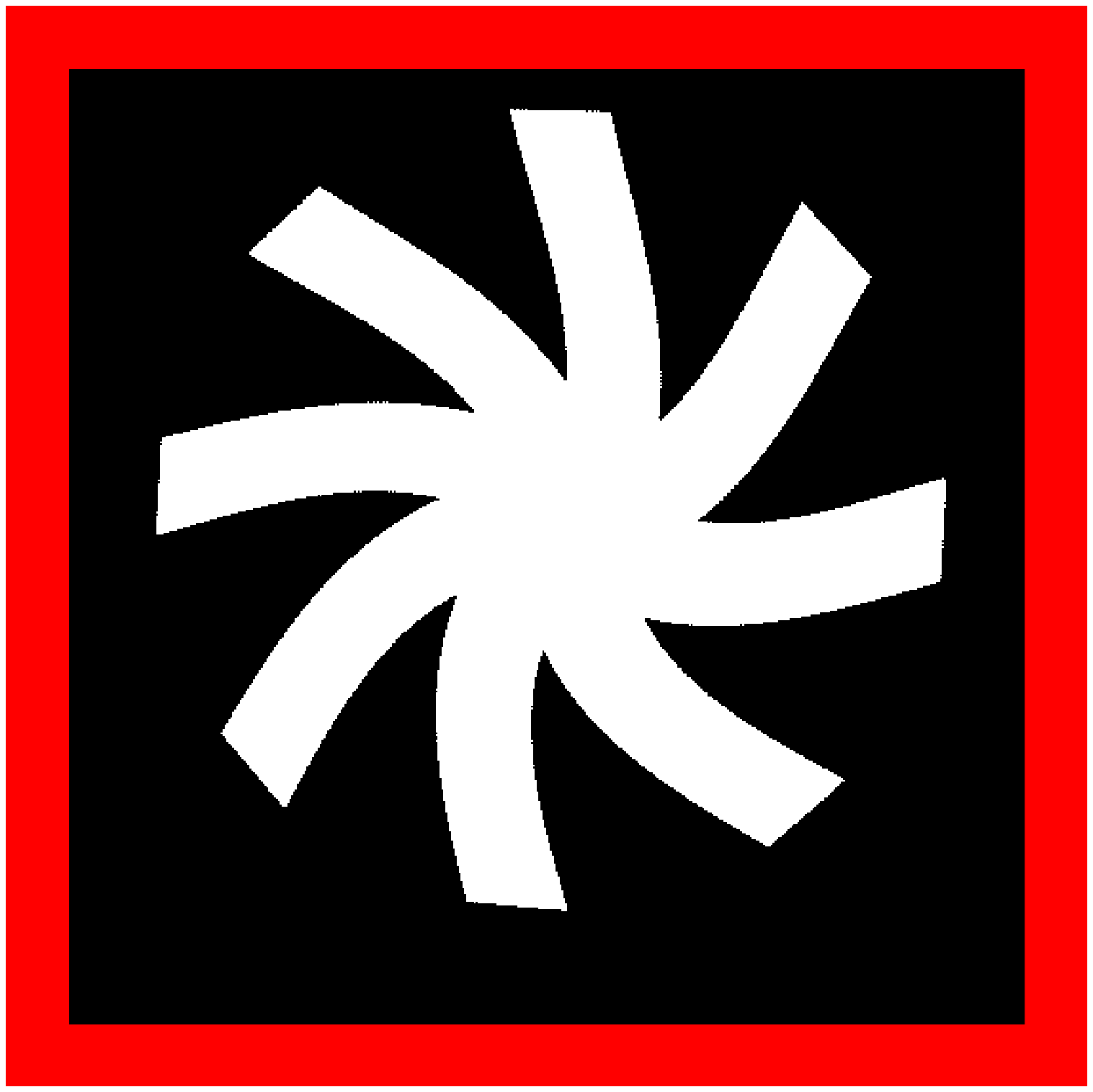}\hspace{-0.215em}}}
  \caption{[Best viewed in colour] (a) Retrieval example from IDSC. We can see that just 3 of the top 40 best matching objects belong to the same class. (b) Retrieval example from SSC. All 20 objects, from the same class as the query object, have been retrieved. Also noticeable is that all the 20 objects lie in the top-20 locations.}\label{fig:comparison}
\end{figure*}

Figure \ref{fig:comparison} shows a comparison of the retrieval results for an example object. The first object is the query object and the rest are its top-40 best matching objects. The objects with green bounding box are correct retrievals and the objects with red bounding box are incorrect retrievals. As can be seen from the figure, IDSC retrieves just $3$ correct objects, while SSC retrieves all $20$ objects belonging to the same class as the query object. Moreover, while using SSC, all of the $20$ objects lie in the top-20 locations.

In this paper, we tackle the case where objects have minor and major indents in their contours.
Our method works even when there are breaks in contours, such as the character `M' shown in Figure \ref{fig:similar}.
Fusing our method with the costs of IDSC means that SSC will take over whenever IDSC performs poorly, and vice-versa.
So, in the cases of major protrusions from the shape's boundary, the cost from IDSC will take over.
We show below, from experiments on the Kimia database, that fusing the two costs has very minimal negative effect on the overall results.
To correctly match shapes with major protrusions, one might employ the strand removal method from \cite{temlyakov2010two}, and fuse a third cost in Eq. \ref{eq:fusion}, as done by the authors of the same.
No one method can correctly match all types of objects. This is why recent works (see Section \ref{sec:related}) have adopted to fusing two or more costs.
The results that we show in Table \ref{tab:bullseye} are obtained by fusing just two costs. The Bullseye score will increase further if a third cost (from strand removal), or even more complementary costs \cite{ling2010balancing}, are fused together.

We use IDSC as the base algorithm since its code, and matrix, are easily available. From Table \ref{tab:bullseye}, we can see that \cite{gopalan2010articulation} produces the best Bullseye score without manifold learning. However, as mentioned in Section \ref{sec:related}, \cite{gopalan2010articulation} decomposes the object into multiple convex parts and performs affine normalization of the individual parts. Doing so would cause unwanted partitioning of the objects such as those shown in Figures \ref{fig:similar} and \ref{fig:device6}. We would expect a high cost of matching when the top-left object in Figure \ref{fig:similar} is matched with second object in the same figure, if we used the method in \cite{gopalan2010articulation}. We believe that if SSC was combined with \cite{gopalan2010articulation}, it would improve its Bullseye score similar to how it currently improves IDSC's Bullseye score. The method of \cite{gopalan2010articulation} is not designed for shapes with indentations (such as Figure \ref{fig:device6}), which our method is suited for, and therefore combining the two methods should produce a better overall cost as the two methods are otherwise compatible. SSC being complementary to IDSC, would also be complementary to the articulation invariant representation used in \cite{gopalan2010articulation}. Thus combining SSC with \cite{gopalan2010articulation} would help us get state-of-the-art results on the MPEG7 database, before manifold learning.

We also calculated the percentage of correct retrievals among the top-20 locations for IDSC and IDSC+SSC.
When IDSC is used alone, it provides a correct retrieval percentage of 76.96\%, while IDSC+SSC gives a correct retrieval percentage of 83.78\%.
We also calculated the average first position of a wrongly classified shape.
For each shape, we find the location of the first wrongly classified shape, and take the average of this location over all shapes.
The average first position of a wrongly classified shape, over all shapes, for IDSC, was found to be 14.43, and for IDSC+SSC, it was found to be 16.1371.
Since there are 20 objects in each class, the best average first position of a wrongly classified shape is 21. So, the closer this number is to 21, the better.
We can see that IDSC makes mistakes much earlier in the retrieval ordering when compared to IDSC+SSC.

In Section \ref{sec:related}, we mentioned certain works that improved the retrieval results by allowing similar shapes to influence the pair-wise scores \cite{yang2009locally, bai2010learning, kontschieder2010beyond, yang2008improving}. These methods try to learn the underlying shape manifold structure and thus learn a better geodesic distance between two shapes. These methods take the pair-wise similarity matrix and perform diffusion on it. In order to end up with a better similarity matrix, it would be ideal if we had a good similarity matrix to start off. A good similarity matrix does not necessarily mean a matrix that produces a good Bullseye score. A good similarity matrix is one that has a low cost for similar shapes and a high cost for dissimilar shapes. This means that the costs between similar shapes, to begin with, are much more closer to the true geodesic distance on the shape manifold. Our similarity matrix does have such properties.

\begin{figure}
  \centering
  \includegraphics[trim=4.5cm 0.5cm 2cm 1cm, clip=true, width=14.5cm, height=7cm]{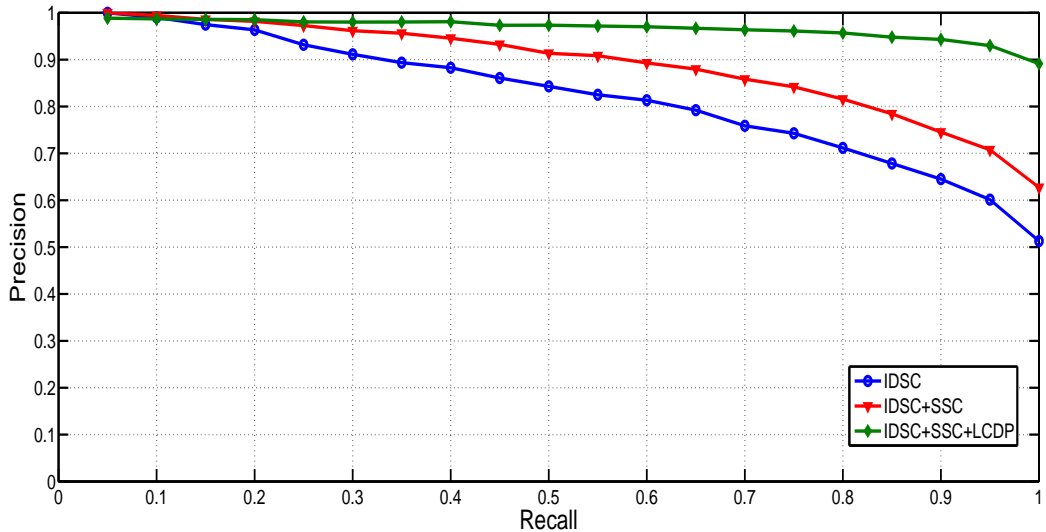}\\
  \caption{[Best Viewed in Color] Precision-Recall curves for IDSC, IDSC+SSC, and IDSC+SSC+LCDP. We can see that the IDSC+SSC curve is clearly above the IDSC curve for all recalls.}\label{fig:prplot}
  \vspace{-3ex}
\end{figure}

We use the Locally Constrained Diffusion Process (LCDP) \cite{yang2009locally} to perform the diffusion on our augmented matrix. The use of LCDP increases the Bullseye score of IDSC+SSC from $91.83\%$ to $98.85\%$. The improvement of the Bullseye score to close to $100\%$ shows that the matrix we started off with had good pair-wise similarity scores. The state-of-the-art results of $99.99\%$ shown in Table \ref{tab:bullseye} were achieved when diffusion was performed on the matrix from \cite{gopalan2010articulation}, which has a higher Bullseye score though it is not perceptually motivated. Moreover, the diffusion was performed using a Tensor Product Graph (TPG) affinity learning procedure \cite{yang2012affinity}, which uses higher order relations between shapes. We, on the other hand, show results that were obtained by performing diffusion using LCDP, which uses just single-order relations between shapes, and on a matrix that obtained by augmenting the IDSC matrix. We use LCDP
because it facilitates comparison with other techniques that also use LCDP \cite{ling2010balancing, temlyakov2010two, Hu20123222, yang2009locally}, and, in addition, its source code is available.  However, we can use TPG for learning the shape manifold as well.

In Figure \ref{fig:prplot}, we plot the precision-recall curves, as in \cite{baseski2009dissimilarity}. The curve compares the precision of IDSC, with that of IDSC+SSC, over various recalls. From the figure, we can clearly see that IDSC+SSC has a better precision than IDSC alone, over all recalls. We also plot the curve for IDSC+SSC+LCDP.

We use IDSC as a base technique, and LCDP as the diffusion technique, since the code for both the algorithms is easily available. We stress that the costs obtained from our algorithm can be fused with the costs from any other algorithm. Finally, we tested our descriptor primarily on the MPEG7 database as it is one of the most challenging shape databases. Other databases such as Kimia database, Natural Silhouette Database, ETH-80 Shape Database, etc, are composed of relatively simple shapes, and have much lesser number of objects compared to the MPEG7 database. Moreover, it is only the MPEG7 database that has the perceptually similar shape classes with vastly different contour properties. We do, however, compare the average first position of a wrongly classified shape for the Kimia dataset 2 \cite{sebastian2004recognition}. When IDSC is used alone, the average first position of a wrongly classified shape is 11.5455, while that for IDSC+SSC is 11.3939; 12 being the ideal score, as there are 11 objects in each class of the Kimia database. This shows that even though the Kimia database does not have objects with intrusions in its contour (in fact, it has objects with major protrusions), fusing SSC with IDSC has very minimal negative effect on the overall results.

\section{Conclusions and Future Work}
\label{sec:conclusion}
In this paper, we identified certain problems that traditional contour-based shape matching techniques face while performing shape matching. We showed that the shape interiors play an important role in object recognition, and proposed a perceptually motivated variant of the well-known Shape Context descriptor, which captures the shape properties in their entirety. We showed the benefits of modelling the interior properties of the shape using \emph{Dense Points}. We proposed a new way for sampling from within a shape boundary, in order to capture the interior properties of the shape. We then listed out the advantages of using the convex hull of a shape to select the landmark points. We also showed how augmenting traditional shape-matching techniques with the costs from our SSC descriptor can significantly improve the retrieval rates.

As for future research directions, we feel a need for the construction of a database that consists of shapes that are identified by humans based on the Gestalt properties of the Human Visual System. This is an area of research that has not been explored well in the community. We encounter such objects multiple times, in our daily lives. There are many instances where we find characters written using the ``stencil font". Most road markings use stencil font for conveying messages (final example of Figure \ref{fig:similar}). Also, the logos of many companies are based on stencil font.

We hope that this work of ours would motivate other researchers in the community to take the area of shape matching to the next level, by coming up with other perceptually motivated techniques.





\bibliographystyle{elsarticle-num}
\bibliography{full,references}







\end{document}